\documentclass[pdflatex,sn-mathphys]{sn-jnl}

\usepackage{diagbox}
\usepackage{cleveref}
\usepackage{algorithm}
\usepackage[section]{placeins}
\usepackage{subfigure}
\usepackage{float}
\usepackage{multirow}
\usepackage{graphicx}
\usepackage{caption}

\jyear{2022}%

\theoremstyle{thmstyleone}%

\theoremstyle{thmstyletwo}%

\theoremstyle{thmstylethree}%

\DeclareUnicodeCharacter{2061}{}

\begin{document}

\title{The Deep Learning model of Higher-Lower-Order Cognition, Memory, and Affection- More General Than KAN}

\author{\normalsize Jun-Bo Tao$^{1,**}$, Bai-Qing Sun$^{2,**}$, Wei-Dong Zhu$^{3}$, Shi-You Qu$^{4,**}$, Jia-Qiang Li$^5$, Guo-Qi Li$^{6,**}$, Yan-Yan Wang$^7$, Ling-Kun Chen$^8$, Chong Wu$^{9,*}$, Yu Xiong$^{10,*}$, and Jiaxuan Zhou$^{11}$ \\
\\
\small ${}^1$Ph. D. candidate, School of Management, Harbin Institute of Technology, Harbin, Heilongjiang 150001, China. Email: ulanara@hotmail.com\\
\small ${}^2$Ph. D., School of Management, Harbin Institute of Technology, Harbin, Heilongjiang 150001, China. Email: baiqingsun@hit.edu.cn\\
\small ${}^3$Ph. D., Department of Mechanical Engineering University of Maryland, Baltimore County, Baltimore, MD 21250, U.S.A. Email: wzhu@umbc.edu\\
\small ${}^4$Ph. D., School of Management, Harbin Institute of Technology, Harbin, Heilongjiang 150001, China. Email: qushiyou@hit.edu.cn\\
\small ${}^{5}$Ph. D.,Institute No.25 of the Second Academy China Aerospace Science and\\
\small Industry Corporation Limited, Beijing 100048, China. Email: lijiaqiang1988926@126.com\\
\small ${}^{6a}$ Institute of Automation, Chinese Academy of Science, Beijing, 100190, China. Email: guoqi.li@ia.ac.cn\\
\small ${}^{6b}$ University of Chinese Academy of Science, Beijing, 100049, China\\
\small ${}^{7}$Ph. D., Faculty of Humanities and Social Sciences, Harbin Institute of Technology, Harbin, Heilongjiang 150001, China. Email: wyy@hit.edu.cn\\
\small ${}^{8a}$ Ph. D., College of Civil Science and Engineering, Yangzhou University, Yangzhou,\\
\small ${}^{8b}$ Jiangsu 225127, China; Email: lingkunchen08@hotmail.com\\
\small ${}^{8c}$ Department of Civil and Environmental Engineering, University of California, Los Angeles, CA 90095, USA\\
\small ${}$School of Civil Engineering, Southwest Jiaotong University, Chengdu, 610031 Sichuan, China\\
\small ${}^9$Ph. D., School of Management, Harbin Institute of Technology, Harbin, Heilongjiang 150001, China. Email: nnem2022@outlook.com\\
\small ${}^{10}$Ph. D., Surrey Business School, University of Surrey, Guildford, Surrey GU2 7XH, United Kingdom. Email: y.xiong@surrey.ac.uk\\
\small ${}^{11}$School of Management, Harbin Institute of Technology, Harbin, Heilongjiang 150001, China. Email: 1191000315@stu.hit.edu.cn\\
\small ${}^*$ Correspondence and requests for materials should be addressed to Chong Wu (email: nnem2022@outlook.com) and Yu Xiong(email: y.xiong@surrey.ac.uk)\\
\small ${}^{**}$These authors contributed to the work equally and should be regarded as co-first authors.}


\maketitle

In brief: We try to comprehend the workings of brains through artificial intelligence, quantum mechanics, and brain dynamics and gave the Artificial Multiple cortexes Heart-Brain model. The brain functions as a Deep Learning model from input to output, that is, from the $n$th external cortex to the internal hippocampus. The mechanism of our proposed Neural Network (NN) is very well in line with the results of the latest brain plasticity study, in which researchers found that as a synapse strengthens, neighboring synapses automatically weaken themselves to compensate \cite{bib14}. Nevertheless, our neural network treats synapses as a group of intelligent agents to achieve synaptic strength rebalance. And Dr. Luo's team has put forward that competition regarding synapse formation for dendritic morphogenesis is crucial. This is also reflected in the fact that PNN simulates synapse formation \cite{bib15}. The paper's analysis suggests a retrograde mechanism, which was reflected in PNN through retrograde retrieval memory \cite{bib16}. The decline of neurogenesis was considered throughout aging. And PNN simulates synapse formation and decrease the number of neurons at the same time \cite{bib17}. However, the controversy claimed that human hippocampal neurogenesis persists throughout aging, we suggested that it may form a new and longer circuit in the late stages of iteration \cite{bib18}. We try to conduct research on the mechanism of failure in brain plasticity by model at the closure of critical period in details by contrasting with studies before \cite{bib19}. By simulations, the PNN gave the two cases with and without negative cortex memories persistence. And the paper proposed $\alpha$7-nAChR-dependent astrocytic responsiveness is an integral part of the cellular substrate underlying memory persistence by impairing fear memory\cite{bib20}. The effect of astrocytes phagocytose synapses  made local synaptic effective range remain in an appropriate length at critical period\cite{bib21}. Both negative and positive memories can be reflected in brain cortex, and its simulations fits very well with experiments in the relationship of human intelligence and cortical thickness, as well as individual differences in brain\cite{bib33}. PNN also considered the memory engram cells that strengthen synaptic strength\cite{bib22}. And paper found that 1064-nm Transcranial photobiomodulation (tPBM) applied to the right prefrontal cortex (PFC) improves visual working memory capacity\cite{bib34}, the effect of PNN memory structure may be the same with smaller losses in signals. Kerskens et. al. suggests consciousness is non-classical \cite{bib36}, so PNN introduces exacted memory brain plasticity by quantum computing in retrograde mechanism, non-classical short-term memory brain plasticity affected by heart frequency, is quantum entanglement, in extracted relatively good or inferior memories brains plasticity will exhibit exponential decay of wave function for a while because of barriers. The research group propose a thalamo-cortical circuit that gates memory consolidation, and suggest the selection and stabilization of hippocampal memories into longer-term cortical storage\cite{bib37}. Researchers found that efficient information transmission of the brain relies on turbulence\cite{bib38}. So short-term memory turns long-term memory due to the laminar-turbulent transition.

The Joris de Wit's group and Bart De Stroope's group paper suggest that impaired MCH-system function contributes to aberrant excitatory drive and sleep defects, which can compromise hippocampus-dependent functions \cite{bib40}. And energy loss of hippocampus might change the gradient direction of cognitive search.

Through the PNN tests, we found that the astrocytic cortex memory persistence factor, like the astrocytes phagocytose synapses factor, produces the effect of reducing the local accumulation of synapses. Considering both negative and positive cortex memories persistence help activate synapse length changes. In the calculation to update the synaptic effective range, the PNN in which only the effect of astrocytes phagocytose synapses is considered regardless of the gradient update, proves simple and effective.

Positive or negative working memory or short-term memory brain plasticity is quantum, and exhibits exponential decay of wave function for a while, produced in the hippocampus. And exponential decay because of barriers, and barriers may relate to astrocytes. Working memory brain plasticity goes through brain from hippocampus to cortices by directional derivative. The strong short-term memory brain plasticity turns to long-term memory means maximum of directional derivatives, that is gradient. So long-term memory means gradient of short-term memory brain plasticity. Similarly, memory flows to the $n$th external cortex may be the $n$th derivative of brain plasticity.

~\\
Summary: Beyond the consideration of the basic connection, this research focuses on a new Neutral Network (NN) model, referred as PNN (Plasticity Neural Network) in which the current and memory effective range between synapses and neurons plastic changes with iterations, are both taken into account at the same time, specifically, the current and memory synaptic effective range weights.

In addition, this mechanism of synaptic strength rebalance is consistent with the findings in the recent research on brain plasticity. Regarding the importance of this mechanism \cite{bib14}, our Neural Network is, it treats synapses as a group of intelligent agents to achieve synaptic strength rebalance. Dr. Luo's team has mentioned that the competition regarding synapse formation for dendritic morphogenesis is important, also reflected PNN simulates synapse formation\cite{bib15}. We try to examine the mechanism of failure in brain plasticity by model at the closure of critical period in details by contrasting with studies before\cite{bib19}.

The PNN model is not just modified on the architecture of NN based on current gradient informational synapse formation and brain plasticity, but also the memory negative and positive gradient informational synapse formation and memory brain plasticity at critical period. By simulations, the PNN gave the two cases with and without negative cortex memories persistence.  And the paper proposed $\alpha$7-nAChR-dependent astrocytic responsiveness is an integral part of the cellular substrate underlying memory persistence by impairing fear memory\cite{bib20}. In addition, memory brain plasticity involves the plus or minus disturbance-astrocytes phagocytose synapses, through which the synaptic homeostasis is achieved\cite{bib21}. The effect of astrocyte made local synaptic effective range remain in an appropriate length at critical period\cite{bib21}. Both negative and positive memories can be reflected in brain cortex, and its simulations fits very well with experiments in the relationship of human intelligence and cortical thickness, as well as individual differences in brain\cite{bib33}. The PNN also considered the memory engram cells that strengthened synaptic strength\cite{bib22}. And paper found that 1064-nm Transcranial photobiomodulation (tPBM) applied to the right prefrontal cortex (PFC) improves visual working memory capacity \cite{bib34}, the effect of PNN memory structure may be the same with smaller losses in signals.

We can regard PNN as a special Recurrent Neural Network (RNN), each input variable corresponds to neurons shared connection weights over a period of time interval; the weights are updated by the MSE loss of the activation function of output within this synaptic effective range and connection. The synaptic effective range is reflected in the time interval, for which the real value of the simulation would lead to gradient-based change in the synaptic effective range by Back Propagation. The Back Propagation of PNN includes shared connection weights and synaptic effective range weights. A simulation of PNN was conducted on the cognitive processes in the brain from infancy to senile phase, which is interpreted by the observation of decreasing the synapse population and increasing the minimum of synaptic effective range in PNN evolution of synapse formation, hence the loss of diversity and plasticity. This explanation is similar to the Dr. Luo and his colleagues' research, and on PNN which shows that the synapse formation in a certain extent may inhibit dendrite. And we also analyzed the different brain plasticity by Tables at critical or the closure of critical period. When ORPNN contains both astrocytic synapse formation cortex memory persistence factor and astrocytes phagocytose synapses factor will have better results in correlation coefficients of Tables respectively at critical period.

The cosine filter is used in the tests for comparing the result of the three cases, namely gradient optimized synaptic effective range (ORPNN), random synaptic effective range (RRPNN), and constant synaptic effective range (CRPNN). The result indicates that the ORPNN runs the best. Through the PNN tests, we found that the astrocytic cortex memory persistence factor, like the  astrocytes phagocytose synapses factor, produces the effect of reducing the local accumulation of synapses. Considering both negative and positive cortex memories persistence yields better results than considering only positive cortex memory persistence. In the calculation to update the synaptic effective range, the PNN in which only the effect of astrocytes phagocytose synapses is considered regardless of the gradient update, proves simple and effective.

\section{Introduction}\label{sec1}

Researchers in the Deep Learning community have long regarded simulation of the human brain as an important means of advancing Neural Networks (NNs). In machine learning and computational neuroscience, Artificial Neural Networks (ANNs) are used as mathematical and computational models that mimic the structure and function of biological Neural Networks. Estimations and approximations conducted by the functions have achieved great success over the recent decades. The more commonly used Neural Networks models today include Perceptron in 1957 \cite{bib1}, Hopfield network in 1982 \cite{bib2}, Boltzmann in 1983 \cite{bib3}, Back Propagation (BP) in 1986 \cite{bib4},  Convolutional Neural Networks (CNN) in 1989 (from Neocognitron \cite{bib5} to CNN \cite{bib6}), Spiking Neural Networks (SNN) in 1997 \cite{bib7}, Long Short-Term Memory (LSTM) in 1997 \cite{bib8}, Deep Belief Network (DBN) in 2006 \cite{bib9}, Deep Neural Networks (DNN) in 2012 \cite{bib10}, Deep Forest (DF) in 2019 \cite{bib11} and Deep Residual Learning (ResNet) in 2016 \cite{bib12}. 

Human brains can be highly flexible due to their plasticity. Neural activity can induce the strengthening or weakening (potentiation or depression) of synapses, leading to plasticity. The brain plasticity therefore refers to the connections of synapses and neurons, and to establish new connections as a result of learning and experiencing, thus exerting impacts on the behaviors of individuals. The synapses are the roles between neurons permitting the transmission of signals or stimulus. Therefore, our new NN model considers not only the synaptic connections, but also how the synaptic effective range at current and history moments changes plastically with the iteration, particularly at critical period, namely, in addition to the shared weights of the current and memory synaptic connections, it also considers the weights of the current and memory synaptic effective ranges and gradient information at critical period.

Researchers presented the \emph{One Hundred Year Study on Artificial Intelligence (AI100) 2021 Study Panel Report}, in which one question concerns how much progress we have made in UNDERSTANDING the PRINCIPLES of HUMAN INTELLIGENCE. First, a fundamental principle of cognitive neuroscience is that individual characteristics such as working memory and executive control are crucial for domain-independent intelligence, which determines an individual's performance on all cognitive tasks regardless of mode or subject. A second hypothesis gaining traction is that higher-ability persons enjoy more efficient patterns of brain connectivity—higher-level brain regions in the parietal-prefrontal cortex. The third concept is more revolutionary, which suggests that the neural correlates of intelligence are dispersed throughout the brain. According to this theory, human intelligence is fundamentally flexible, constantly updating past information and making new predictions \cite{bib13}. The new NN proposed in this thesis is grounded on the first and third notion, taking into consideration the current and past brain plasticity and current gradient informational and memory positive and negative gradient informational synapse formation at critical period and closure of critical period.

Scientists have unveiled for the first time how this balance is achieved in synapses: Professor Mriganka Sur likens this behavior to a massive school of fish in the sea. Immediately when the lead fish changes direction, other fish will follow suit, presenting a delicate marine "dance". “Collective behaviors of complex systems always have simple rules. When one synapse goes up, within 50 micrometers there is a decrease in the strength of other synapses using a well-defined molecular mechanism.” the scientists stated \cite{bib14}. The lead behavior of the school of fish is well embodied in our PNN. However, our Neural Network is no longer confined to the simple physical concept of synaptic strength strength rebalance - it treats synapses as a group of intelligent agents to achieve synaptic strength rebalance. 

Findings largely consistent with that of Dr. Luo's team in PNN tests are obtained in our study, that is, the synapse formation will inhibit dendrites generation to a certain extent \cite{bib15}, with the reason shown in our simulation results being that synaptic growth leads to a reduction in changes in synaptic effective range, which disrupts diversity and plasticity of brain was reflected in our simulations.

The Alcino J Silva's team suggested that memory ensembles recruit presynaptic neurons during learning by a retrograde mechanism \cite{bib16}. The retrograde mechanism reflected in memory retrieval process at upstream brain regions of PNN.

The Alvarez-Buylla's lab proposed that recruitment of young neurons in the primate hippocampus decreases rapidly during the first years of life, and that neurogenesis in the dentate gyrus does not continue, or is greatly rare, in adult humans \cite{bib17}. The PNN gave the simulation results of decrease of neurons based on synapse formation.

But Boldrini et al suggested healthy older subjects display preserved neurogenesis. It is possible that ongoing hippocampal neurogenesis sustains human-specific cognitive function throughout life \cite{bib18}. And PNN's guess decline in neurogenesis may be repaired by a new and longer circuit in late iteration.

The finding has significant implications for the learning and further development of NN. Moreover, the mechanism of our NN is basically consistent with that of the latest brain plasticity study, in which researchers found that as a synapse strengthens, neighboring synapses automatically weaken themselves to compensate \cite{bib14}. In tests, possible explanations of dendrite morphogenesis are derived, which demonstrate that dendrite generation, to a certain extent, is curbed by synapse formation \cite{bib15}. Unconventional astrocyte connexin signaling hinders expression of extracellular matrix-degrading enzyme matrix metalloproteinase 9 (MMP9) through RhoA-guanosine triphosphatase activation for controlling critical period closure. The astrocyte impacts on brain plasticity and synapse formation is an important mechanism of our Neural Network at critical period, and failures by the closure of critical period result in neurodevelopmental disorders \cite{bib19}. In the model at critical period, the hypothesis is the best previous brain plasticity affects current brain plasticity and the best previous synapse formation, relatively good and relatively inferior synapse formation affects current synapse formation. While their experimental research features a combination of cutting-edge imaging and genetic tools, our study lays more emphasis on the model, derivation and simulation of a new NN. At the same time, the current and memory brain plasticity-the synaptic effective range are taken into consideration. Furthermore, the architecture of new NN is based on current gradient informational, and memory positive and negative gradient informational synapse formation for cortex memory persistence-long-term memory. By simulations, the PNN presented the two cases with and without negative cortex memories persistence. And the paper proposed $\alpha$7-nAChR-dependent astrocytic responsiveness is an integral part of the cellular substrate underlying memory persistence by impairing fear memory\cite{bib20}. In addition, memory brain plasticity involves the plus or minus disturbance- astrocytes phagocytose synapses, through which the dynamic synaptic strength balance is achieved \cite{bib21}. The effect of astrocyte made local synaptic effective range remain in an appropriate length at critical period \cite{bib21}. Our new NN may give more inspirations to experiments in the relationship of human intelligence and cortical thickness, as well as individual differences in cerebral cortex \cite{bib33}. By using learning-dependent cell labeling, Dr. Susumu Tonagawa and his team identified an increase of synaptic strength and dendritic spine density specifically in consolidated memory engram cells. For improving synaptic strength in PNN, the memory retrieval process by memory engram cells includes memory negative and positive gradient information, and memory engram cells are helpful to the memory retrieval process \cite{bib22}. In this study, researchers found that 1064-nm tPBM applied to the right prefrontal cortex improves visual working memory capacity \cite{bib34}, and PNN has same effect for improving memory capacity by memory structure of relatively good or inferior memory. The Dr. Christian Kerskens and David López Pérez of Neuroscience's findings suggest that they may have witnessed entanglement mediated by consciousness-related brain functions, because consciousness-related or electrophysiological signals are unknown in nuclear magnetic resonance (NMR). Those brain functions must then operate non-classically, which would mean that consciousness is non-classical \cite{bib36}, so PNN introduces brain plasticity of extracted relatively good or inferior memories brains plasticity, which reflected consciousness by quantum computing. Priya Rajasethupathy research group developed longitudinal simultaneous imaging in hippocampus, thalamus, and cortex of mouse brain, and propose a thalamo-cortical circuit that gates memory consolidation, and suggested a mechanism suitable for the selection and stabilization of hippocampal memories into longer-term cortical storage \cite{bib37}. In simulations of PNN, it converts hippocampal non-classical exacted memories to long-term classical memories, maximum of directional derivatives is gradient, and barriers will lead to exponential decay of wave function.

Researchers used magnetoencephalography (MEG) technology to analyze turbulence throughout the brain, and the study found that efficient information transmission of the brain relies on turbulence, through which the phenomenon of turbulence in rapid neuronal brain dynamics can be directly measured \cite{bib38}.

This paper's work suggests a model in which impaired MCH-dependent synaptic function in CA1 and perturbed Rapid Eye Movement (REM) sleep synergistically compromise neuronal homeostasis, contributing to aberrant neuronal activity in CA1\cite{bib40}. It is also indirectly proved our possible mechanism of Alzheimer's disease.

Our newly proposed NN is named the Plasticity Neural Network (PNN). Synaptic competition causes the enhancement of signal-stimulated synapses to be reflected in an increase in synaptic effective range, while the weakening of peripherally stimulated synapses is reflected in a shortening of synaptic effective range. The long-term memory of brain architecture's gradient at the upstream brain regions, the short-term memory of brain architecture at the downstream brain regions.

The following papers are academic frontiers in neuroscience and Artificial Neural Networks which we endeavor to make connections with our PNN.

Based on the mechanism of flow stability, Dr. Hua-Shu Dou proposed the Energy Gradient Theory. The maximum of directional derivatives means short-term memory turns to long-term memory might be modified by a critical angel, and the angle 0 turns to $\alpha_c$ \cite{bib39}.

Their research shows that SynCAM 1 actively limits cortical plasticity in the mature brain. Plasticity tapers off when the brain matures and the conclusion is sufficiently substantiated by visual input in adult animal models \cite{bib35}.

The Self-Back Propagation (SBP) phenomenon, first discovered in hippocampal neurons \cite{bib23,bib27}, involves cross-layer Back Propagation (BP) of Long-Term Potentiation (LTP)and Long-Term Depression (LTD ) from output to input synapses of a neuron to enable the strengthening or weakening of synaptic connections. Other forms of nonlocal spreads of LTP and LTD in the pre- and post-synaptic neurons have already been researched extensively \cite{bib23,bib27,bib24}. The SBP phenomenon induces a form of nonlocal activity-dependent synaptic plasticity that may endow developing neural circuits with the capacity to modify the weights of input synapses on a neuron in accordance with the status of its output synapses \cite{bib24}. The existence of SBP was demonstrated in developing retinotectal circuits in vivo \cite{bib25,bib26}. These papers endeavor to shed light on plasticity rules involving activity-dependent modification of synapses to obtain synaptic activity \cite{bib25,bib26,bib27,bib23,bib24}. 

Bertens and Lee proposed an Evolvable Neural Unit (ENU) that can evolve individual somatic and synaptic compartment models of neurons in a scalable manner and try to solve a T-maze environment task \cite{bib28}.

Wang and Sun stated that in a unidirectional RNN, the linking with the emotion regions and the somatic motor cortex includes three basic units: input units arriving from the emotion regions, only one hidden unit composed of self-feedback connected medial prefrontal cortex neurons, and output units located at the somatic motor cortex \cite{bib29}. If they used PNN regardless of RNN may give more insights in their research. Although PNN studies have focused on the prefrontal cortex and emotion regions.

In the simulations of PNN in brain plasticity at critical period, we had an idea to modify the synapses for repairing brain plasticity, and later we found this paper to build and remodel of synapses \cite{bib30}.

Instead of conducting research simply on the basis of Self-Back Propagation phenomenon, the PNN we propose in this thesis focuses more on plasticity and strength rebalance as a result of its tapering off -the brain from infancy to senile phase during critical period and the closure of critical period. The model also delves into the competition of brain plasticity which introduces a novel form of current gradient informational, and memory positive and negative gradient informational synapse formation means long-term memory, as well as current and memory brain plasticity means short-term memory at critical period. The strengthening and weakening of synaptic connections help to curb and boost current and history synaptic effective ranges, resulting in an increased accuracy of PNN.

The research, based on Genetic Algorithm or Particle Swarm Optimization in search of the connection weights for each time interval of the dynamic problems, and on PNN to find the weights of the connection and the effective range, takes into account the specificity of the new coronavirus variant, whose viral strain is highly variable. The viral load of the Delta (or Omicron) strain is 1,260 times higher than the prevalent strain of last year, thus doubling the infectious rate of last year's original strain. The major difference between PNN and RNN is that the former has a formula to estimate the parameters (shared connection weights) for their respective parameters' ranges (synaptic effective range weights) at critical period. So PNN is based on the practical dynamic problem in an effort to obtain the new NN. As for the selection of the appropriate mechanism, we chose the hypotheses and possible explanations of synaptic strength rebalance, competition and effect of astrocyte \cite{bib19,bib14,bib15,bib16,bib17,bib18,bib20,bib21,bib22,bib33,bib34,bib36,bib37,bib38,bib40}. In this way, not only the architecture of NN is transformed by current gradient information, but also memory positive and negative gradient informational synapse formation and memory brain plasticity are taken into account. The memory gradient information needs to consider astrocytic synapse formation cortex memory persistence factor (including both negative and positive memories). By simulations, the PNN presented the two cases with and without negative cortex memories persistence. And the paper proposed $\alpha$7-nAChR-dependent astrocytic responsiveness is an integral part of the cellular substrate underlying memory persistence by impairing fear memory\cite{bib20}. In addition, memory brain plasticity involves the astrocytes phagocytose synapses \cite{bib21}. And it also comparing with the simulations and experiments in the relationship of human intelligence and cortical thickness, as well as individual differences in brain \cite{bib33}. In PNN, the memory retrieval process includes memory negative and positive gradient information\cite{bib22}. And PNN has same effect for improving memory capacity by memory structure of salient point and concave point \cite{bib34}.

After representing the synaptic plasticity competition in Formula \eqref{eq2}, strength rebalance in Formula \eqref{eq3}, current gradient informational and memory positive and negative gradient informational synapse formation at critical period in Formula \eqref{eq2}, current and memory brain plasticity at critical period in Formula \eqref{eq4}, cosine filter is used in tests to verify the PNN and compare the results of the following three situations: the case dubbed as CRPNN, in which the synaptic effective range for various neurons remains unchanged; the RRPNN case, in which synaptic effective range for each neuron generates in a random manner; and the last one is the ORPNN case, in which optimization of synaptic effective range connects neurons with iterations. The findings also show that synapse formation will, to some extent, inhabit the tests of PNN. And we also analyzed the different brain plasticity by Tables at critical period or the closure of critical period. The effect of astrocytic synapse formation memory cortex persistence factor and astrocytes phagocytose synapses factor in Tables.

\section{Method}\label{sec2}

We developed PNN under the premise that the sum of synaptic effective ranges of these neurons' connections remains a constant value for the rebalance of synaptic strength. The size of RNN training sets has been determined by sum of synaptic effective ranges ${\rm l_{max}}$. This hypothesis applies to both RNN as well as BP. However, synapse formation will lead to convergence and the loss of plasticity, therefore diminishing the cognitive abilities of PNN. For RNN, the input variables correspond to the different types of neurons' within a time interval, and each time interval corresponds to a connection weight. And the weight is updated by the MSE loss of the activation function of output within this synaptic effective range. The synaptic effective range is reflected in the time interval, and alters with the gradient by simulating real value- training sets with iterations.

Formula \eqref{eq1} and \eqref{eq2} are both obtained through the reasoning of MSE loss function gradient. The activation function for the amount of change in the output result is processed through the tanh function.

Formula \eqref{eq1} updates the shared connection weight of the $k$th synapse of the $m$th input neuron, $f()$ represents the output of different variables and synapses, $m$ represents the label of the input variable, $k$ represents label of synapses connecting two neurons, and the range of values of time interval $t$ is $[n_3(m,k)\ n_3(m,k+1)-1]$. $h(t)$ indicates actual output, and $f(t)$ represents desired output. $n_3(m,k)$ and $n_3(m,k+1)$ depict the location of the $m$th input variable and $k$th synaptic initiation and termination points. $ \eta = 1/1\rm E3 \times (\rm j_{max}-j)/\rm j_{max}$, namely, the learning rate is reflected in formulas \eqref{eq1} and \eqref{eq2}. Considering RNN, $f(t)=f(t-1) + \tanh [\Delta f(t-1)]$, the derivative of $f(t)-h(t)$ is $1-\Delta f(t-1)^2$.

\begin{equation}
\begin{split}
g_{w}{(m,k)}=\frac{\partial E}{\partial w(m,k)} \times \eta = [f(t)-h(t)] \times [1-\Delta f(t-1)^2] \times \Delta f(t-1)^{'}_{w(m,k)} \times \eta \\
\centering =[f(t)-h(t)] \times [1-\Delta f(t-1)^2] \times \Delta f(t-1)^{'}_{w(m,k)} /1 \rm E3 \times({\rm j_{max}}-j)/{\rm j_{max}}  \label{eq1}
\end{split}
\end{equation}

\begin{equation}
w{(m,k)}=w{(m,k)}-g_{w}{(m,k)} \nonumber
\end{equation}

The synaptic effective range weights of the $m$th input variable and the $k$th synapse is updated by Formula \eqref{eq2}. The relationship between the synaptic location and the effective range satisfies the equation $n_3(m, k+1)-n_3(m,k)=n_1(m,k)$, in which $n_1(m,k)$ represents the synaptic effective range. When the synaptic position of the lead neurons is changed, these neurons, much like the heads in a massive school of fish, also exerts an impact on the postsynaptic neurons. Formula \eqref{eq2} considers current gradient informational and memory positive and negative gradient informational synapse formation updated by the gradient of the best previous synaptic effective range weight $g_r (m,k)_{best}$, or gradient of relatively inferior synaptic effective range weight $g_r (m,k)_{worse}$ or relatively good synaptic effective range weight $g_r (m,k)_{better}$, so that architecture of PNN bears current and memory gradient information at critical period. With reference to Formula \eqref{eq3}, Formula \eqref{eq2} $n_1(m,k)^{'}_{r(m,k)}=[\frac{-r(m,k)}{sum(r(m,:))^2}+\frac{1}{sum(r(m,:))}] \times (\rm {l_{max}}- \rm {k_{max}} \times \rm {l_{min}})$. $MW_{i, i=1,2,3}$ is memory weight. The memory weight of $g_r (m,k) _{worse}$ is relatively small for jumping out of the local optimum, the proportion of people's negative emotions should not be too large.

\begin{equation}
\begin{aligned}
g_{r}{(m,k)}=&\frac{\partial E}{\partial r(m,k)}\times \eta\\
=&[f(t)-h(t)]\times[1-\Delta f(t-1)^{2}] \times \Delta f(t-1)^{'}_{n1(m,k)} \times{n_1(m,k)^{'}_{r(m,k)}}\\
&\times \eta\\
=&[f(t)-h(t)] \times[1-\Delta f(t-1)^{2}] \times \Delta f(t-1)^{'}_{n1(m,k)}\\
&\times[\frac{-r(m,k)}{\rm sum(r(m,:))^2}+\frac{1}{\rm sum(r(m,:))}]\times({\rm l_{max}}-{\rm k_{max}}{\rm l_{min}}) /1{\rm E}3 \\ 
&\times(j_{\rm{max}}-j)/{j_{max}}
\end{aligned}
\label{eq2}
\end{equation}

\begin{equation}
r{(m,k)}=r{(m,k)}-g_{r}{(m,k)} \nonumber
\end{equation}

\begin{equation}
or\ r(m,k)=r(m,k)-g_{r}{(m,k)}+[g_{r}{(m,k)}_{best}-g_{r}{(m,k)}]\times M \nonumber
\end{equation}

\begin{equation}
\begin{aligned}
or\ r(m,k)=& r(m,k)-g_{r}{(m,k)}\\
& +[MW_1 \times g_{r}{(m,k)}_{better}+MW_2 \times g_{r}{(m,k)}_{worse}-g_{r}{(m,k)}]\\
&\times M \ MW_1=0.6666, MW_2=0.3333 \nonumber
\end{aligned}
\end{equation}

\begin{equation}
\begin{aligned}
or\ r(m,k)=&r(m,k)-g_{r}{(m,k)}\\
& +[MW_1 \times g_{r}{(m,k)}_{best} + MW_2 \times g_{r}{(m,k)}_{worse} \\
& + MW_3 \times g_{r}{(m,k)}_{better} -g_{r}{(m,k)}]\times M  \\
& MW_1=0.5, MW_2=0.25, MW_3=0.25\nonumber
\end{aligned}
\end{equation}

\begin{equation}
\begin{aligned}
r(m,k)=&r(m,k)-[h_r {m,k}]^{-1} \times g_{r}{(m,k)}\\
& +[MW_1 \times [h_{r}{(m,k)}_{best}]^{-1} \times g_r{(m,k)}_{best} \\
& + MW_2 \times [h_{r}{(m,k)}_{worse}]^{-1} \times g_r{(m,k)}_{worse} \\
& + MW_2 \times [h_{r}{(m,k)}_{better}]^{-1} \times g_r{(m,k)}_{better} \\
& -[h_{r}{(m,k)}]^{-1} \times g_{r}{(m,k)}] \times M , \\
& MW_1=0.5, MW_2=0.25, MW_3=0.25\nonumber
\end{aligned}
\end{equation}

Formula \eqref{eq3} serves to update the range of connecting neurons at the $k$th synapse of the $m$th input variable, and normalization of result is achieved through $r(m,k)/sum(r(m,:))$ in Formula \eqref{eq3}. The result of  $r{(m,k)}=r{(m,k)}-g_{r}{(m,k)}+[0.6666\times g_{r}{(m,k)}_{better}+0.3333\times g_{r}{(m,k)}_{worse}-g_{r}{(m,k)}]\times M$ may be better than $r(m,k)=r(m,k)-g_r (m,k)$ in simulation, because negative and positive memories reflecting the diversity. If Formula \eqref{eq2} just has current gradient information, $r(m,k)=r(m,k)-g_r (m,k)$, else if Formula \eqref{eq2} includes current and memory gradient information, $r(m,k)=r(m,k)-g_{r}{(m,k)}+[0.6666 \times g_{r}{(m,k)}_{better}+0.3333 \times g_{r}{(m,k)}_{worse}-g_{r}{(m,k)}]\times M$ needs to consider forgotten memory in synapse formation, cortex memory persistence of astrocytic function gradually lost by aging. The Gradient Descent Method to update the synaptic effective range weights by long-term memory means $g_r (m,k)_{better}$ or $g_r (m,k)_{worse}$. Strong short-term memory was consolidated and became long-term memory. Short-term memory brain plasticity goes through brain from hippocampus to cortices by directional derivative. The strong short-term memory brain plasticity turns to long-term memory means maximum of directional derivatives, and maximum of directional derivatives is gradient. At upstream brain regions, considering the second-order Taylor approximation at $r_0$ point, $f=f_{r0}+g_{r0} \times (r-r_0)+0.5 \times h_{r0} \times (r-r_0)^2$, we also get the derivative formula $g=g_{r0}+h_{r0} \times (r-r_0)$, through formula $g=0$ we get $r(m,k)=r(m,k)-[h_r(m,k)]^{-1} g_r(m,k)$, the same goes for the first-order Taylor approximation of the downstream brain regions.

The derivation of cortex memory persistence factor is as follows, may happen in memory engram cells of different cortices: The update of $r(m,k)=r(m,k)-g_{r}(m,k)$ is based on the Gradient Descent Method. From $r(m,k)=r(m,k)-g_{r} (m,k)_{best}$, we can obtain $r(m,k)_{j+1}=r(m,k)_{j}-g_{r} (m,k)_{best}$, $r(m,k)_{j+1}$ is the best previous synaptic effective range weight, and $g_{r} (m,k)_{best}$ is the gradient of the best previous synaptic effective range weight. Time reversal from $j+1$ to $j$ to return to in situ to maintain memory can also be written as $r(m,k)_j=r(m,k)_{j+1}+g_{r} (m,k)_{best}$, after which, by adding $r(m,k)_{j+1}=r(m,k)_{j}-g_{r} (m,k)$ and $r(m,k)_{j}=r(m,k)_{j+1}+g_{r} (m,k)_{best}$, we get $r(m,k)=r(m,k)-0.5 \times g_{r} (m,k)+0.5 \times g_{r} (m,k)_{best}$ .Considering the flexibility of learning rate, it can also be written as $r(m,k)=r(m,k)-g_{r} (m,k)+[g_{r} (m,k)_{best}-g_{r} (m,k)] \times M$. If we take both positive and negative memories into account, introducing the gradient of the best previous synaptic effective range weight $g_{r} (m,k)_{best}$, gradient of relatively inferior synaptic effective range weight $g_{r} (m,k)_{worse}$ and gradient of relatively good synaptic effective range weight $g_{r} (m,k)_{better}$, therefore the equation can be modified as follows: $r(m,k)=r(m,k)-g_{r}{(m,k)}+[0.6666 \times g_{r}{(m,k)}_{better}+0.3333 \times g_{r}{(m,k)}_{worse}-g_{r}{(m,k)}]\times M$ or $r(m,k)=r(m,k)-g_{r}{(m,k)}+[0.5 \times g_{r}{(m,k)}_{best}+0.25 \times g_{r}{(m,k)}_{worse}+0.25 \times g_{r}{(m,k)}_{better}-g_{r}{(m,k)}]\times M$. As indicated in Fig.\ref{fig20}(b), the best previous solution appears at time point $j+$1, with the smallest Loss. And the next generation demonstrates an upward trend, featuring a relatively inferior solution with an increase in Loss which appears at time point $j+$3. The generation demonstrates a downward trend, featuring a relatively good solution with a decrease in Loss which appears at time point $j+$5. The relatively good and inferior gradient information in synapse formation of retrograde circuit means memory engram cells in long-term memory.

In derivation of formula \eqref{eq2}, the $[MW_1 \times g_{r}{(m,k)}_{best}+MW_2 \times g_{r}{(m,k)}_{worse}+MW_3 \times g_{r}{(m,k)}_{better}-g_{r}{(m,k)}]$ of $r(m,k)=r(m,k)-g_{r}{(m,k)}+[MW_1 \times g_{r}{(m,k)}_{best}+MW_2 \times g_{r}{(m,k)}_{worse}+MW_3 \times g_{r}{(m,k)}_{better}-g_{r}{(m,k)}]\times M$ illustrated that recruitment of presynaptic neurons is triggered by downstream neurons through a retrograde mechanism\cite{bib16}, so the long-term memory gradients $MW_1 \times g_{r}{(m,k)}_{best}$,$MW_2 \times g_{r}{(m,k)}_{worse}$,$MW_3 \times g_{r}{(m,k)}_{better}$ are positive, and the current gradient -$g_{r}{(m,k)}$ is negative.

We hypothesize that the hippocampus stores memory brain architecture as $r$, this $r$ may be affected by quantum entanglement from the heart frequency and the brain architecture. If $r$ is transmitted from the hippocampus to the first cortex of the brain, memory produces a rate of change $\frac{d r}{d n_1}$, and the second cortex conducts memory through the rate of change to take $\frac{d^2 r}{dn_1 dn_2}$, and the rate of change of this cortex continues to conduct to achieve the flow of memory, ignoring the difference between different cortices, then the $n$th cortex is the $n$th derivative $\frac{d^n r}{d{n_1}^n}$ of brain plasticity. This is like the Chain Rule in Deep Learning. Deep Learning from input to output, that is, from the $n$th cortex on the outside to the hippocampus on the inside. The turbulent movement of the mnemonic logarithmic spiral spreading from one cortex to another is only a loss of energy, but the memory engrams are still approximate.

Turbulence occurs from the hippocampus to the first cortex, memory may be a first-order derivative of brain plasticity, means gradient or Jacobian matricx. Turbulence occurs from the first cortex to the second cortex, memory may be the first derivative of the brain plasticity Jacobian matricx, that is, second derivative of the brain plasticity, the Heisen matrix. $h_r(m,k)$ stands for the Heisen matrix. At this time, the Gradient Descent Method that updates synaptic effective range is changed into Newton Method. Formula \eqref{eq2} can be further modified to take into account the case of the Heisen matrix. Alzheimer's disease may first be positive about in the $g_r(m,k)$ of the first cortex, followed by positive gradient in the $[{h_r(m,k)}]^{-1} \times {g_r(m,k)}$ of the second cortex, and positive and negative changes in these derivatives lead to disturbances in synaptic excitation and inhibition. It is shown in Fig. \ref{fig20}(h). Similarly, memory flows to the $n$th external cortex may be the $n$th derivative of brain plasticity. We suspect that curly hair indicates that the brain has strong energy, and energy can be transmitted from the hippocampus to the epidermis of the brain at a critical angle of the cortex, so people with curly hair are not prone to Alzheimer's disease.

Back Propagation from the hippocampus to different cortexes, it requires higher-order optimization to process simple signals, indicates that the external part of the brain needs higher-order optimization, it is also possible to reduce the complexity of the calculation. During Forward Propagation from different cortexes to the hippocampus, complex signals require more negative memories to escape from the local optimal solution. In other words, more negative memories are necessary within the brain.

If we take the logarithmic spiral of the partial mnemonic brain architecture $r_m(m,k)=ae^{b\theta}$, that is, the memory may be two-dimensional logarithmic spiral in a certain cortex, only having the angular strain $\theta$, and the second cortex mnemonic architecture is $\frac{d r_m(m,k)}{d \theta}=abe^{b\theta} $, because turbulence diffuses the $n$th cortex mnemonic architecture $\frac{d^{n-1} r_m(m,k)}{d \theta^{n-1}}=ab^{n-1}e^{b\theta} $, and take $b<1$ means that the memory gradually weakens transmission to the upstream brain regions, the logarithmic spiral $n-1$st derivative is the memory engram shapes are approximate, but the image quality of the memory engram is compressed, $a$ is related to the synaptic effective range weight, $b$ is related to the synaptic connection and effective range weights. The mnemonic architecture of the downstream first cortex is $r_m$, and the mnemonic architecture of the upstream $n$th cortex may be approximately the $n-1$st derivative of $r_m$.

The formula for memory engram is given and the $n-1$st derivative of $r_m$. The following situations are met:

\begin{itemize}
        \item[1] The memory engram flows by turbulence from the downstream brain regions to the upstream brain regions, and the shape of the memory engram remains unchanged, because the resistance along the way and the pipe diameter gradually decreases, and the image quality of the memory engram will be compressed because of the derivative of memory engram.
        \item[2] The memory engram moves from the upstream brain regions to the downstream brain regions, and the shape of the memory engram remains unchanged, but there is no the resistance along the way and the pipe diameter gradually increases, and the memory engram image quality will not be compressed.
        \item[3] If the large middle blood vessels and large middle ducts in the downstream brain regions are destroyed, the analysis of reverse turbulence makes reversal of the situation 1 and 2. Situation 1: Memory engram does not lose image quality, and it will be better to obtain memory engram, but hallucinations may occur due to downstream information redundancy; Situation 2: Turbulence occurs across the cortex and memory engram loses image quality from upstream to downstream.
        \item[4] Our memory engram may be a two-dimensional logarithmic spiral in the brain cortex, which is derived from only one angle strain variable for turbulent movement across brain regions.
\end{itemize}

The angle $\psi$ between the logarithmic spiral tangent and the radius, which can be calculated that, $\psi=\tan^{-1}\frac{r_m(m,k)}{\frac{d r_m(m,k)}{d \theta_1}}=\cot^{-1}\frac{\frac{d r_m(m,k)}{d \theta_1}}{r_m(m,k)}=\cot^{-1}b$, if $b$ is too small, $\cot^{-1}b=\frac{\pi}{2}$, the logarithmic spiral is close to the circle, and the smooth surface weakens the synaptic connection ability. The critical angle of turbulence $\alpha$ can be understood in this way, only considering the angular motion of the logarithmic spiral, tangential strain of logarithmic spiral along the spiral $\theta_1=\frac{\pi}{2}-\psi$ and tangential strain of logarithmic spiral angular motion $\theta_2$. When the they become the force and counter-force, $\theta_1+\theta_2=0$, which can be calculated that, $\alpha=-\tan^{-1}b=\tan^{-1}\frac{\frac{d r_m(m,k)}{d \theta_2}}{r_m(m,k)}=\theta_2$. $\theta_2$ should be at least greater than the action of $\theta_1$, which can be turn to turbulence, and the nonlinear movement of the logarithmic spiral means the strain of the brain, which from the hippocampus to the prefrontal cortex, that is, the strain of the brain and the strain of memory may be an equal and opposite reaction, the angle changes in space-time are $\tan^{-1}b$, so the geometry of the longitudinal section of the brain is like to a spiral, with the beginning or O point of the spiral is the prefrontal cortex and ending in the hippocampus. if you do not consider different brain regions, only suppose the mechanical property of the isotropy of the brain, then the brain longitudinal section is more like the spiral. And the value of the critical angle is related to the mnemonic connection weight, and the life span of the logarithmic spiral is related to the mnemonic effective range and connection weights. See Fig. \ref{fig20}(i) and \ref{fig20}(j).

$\theta_1=\tan^{-1}b$ critical angle increase hinders the transmission of information between brain regions. The synapse formation increases the synaptic effective range weight associated with $a$, it has nothing to do with the critical angle, which increases the life span of the logarithmic spiral. And the loss of immune cells because of the reverse turbulence, which leads to neuroinflammation and increases the shared connection and effective range weights associated with $b$, which in turn increases the critical angle. Both the enlargement of $a$ and $b$ may impair cognition.

The element $g_{r} (m,k)_{worst,i}$ or $g_{r} (m,k)_{better,i}$ of history set $WORS=\{{g_{r} (m,k)_{worst,i} i\in N^*}\}$ or $BETT=\{{g_{r} (m,k)_{better,i}  i\in N^*}\}$ also can be considered in the formula \eqref{eq2}, and latest $g_{r} (m,k)_{worst}$ or $g_{r} (m,k)_{better}$ could be replaced by later $g_{r} (m,k)_{worse,i}$ or $g_{r} (m,k)_{better,i}$. The gradient information of later element $g_{r} (m,k)_{worse,i}$ or $g_{r} (m,k)_{better,i}$ is stored in memory engram cells of cortexes, namely larger $i$ of $g_{r} (m,k)_{worst,i}$  or $g_{r} (m,k)_{better,i}$ will be selected with upper probabilities.

Why formula \eqref{eq2} chooses the elements of history set $WORS$ or $BETT$?

Based on optimality principle, the relatively good gradient information can help to search optimal solution, and the relatively inferior gradient information can help to improve diversity of searching optimal solution. The location of relatively inferior gradient information is salient point, the directions of search both are downward and become more efficient. The location of relatively good gradient information is concave point, the directions of search both are upward and improve the efficiency.

But memory-forgotten means that the signals from input to output units dissipate with resistances in brain and do not activate the relevant gradient information, and these resistances in brain increase by aging.

In the model at critical period, the hypothesis is the best previous synapse formation and relatively inferior synapse formation affects current synapse formation. Bad memory of synapse formation $M=exp(-7\times j /{\rm j_{max}} ) \times rand$ or good memory $M=exp(-5\times j /{\rm j_{max}} ) \times rand$ can be called the astrocytic synapse formation cortex memory persistence factor. $j/{\rm j_{max}}$ illustrates the decreasing weights of factor we present at critical period in $M$.

\begin{equation}
n_{1}{(m,k)}=\frac{r{(m,k)}}{sum(r(m,:))}\times ({\rm l_{max}}-{\rm k_{max}}{\rm l_{min}}) + {\rm l_{min}} \label{eq3}
\end{equation}

For a certain input variable, the neuron has a total of ${\rm k_{max}}$ synapses,${\rm l_{min}}$ shows that the synaptic range change enjoys a certain degree of plasticity, and the minimum synaptic effective range is ${\rm l_{min}}$. $g_{r} (m,k)$ controls ${\rm l_{min}}$ with iterations and represents the elasticity or plasticity of synapses, and if the value of ${\rm l_{min}}$ is overly small, the change of  $n_1 (m,k)$ will be too elastic and may make the test results deteriorate. ${\rm k_{max}}$ indicates the synapse population involved in the neurons from input to output units, and ${\rm l_{max}}$ represents the sum of ranges of time series and is a constant. For the signal of a neuron (i.e., an input variable), when one synapse strengthens, there must be surrounding synapses that are weakened. In the case when ${\rm l_{max}}={\rm k_{max}} \times {\rm l_{min}}$, i.e., the case in which the synaptic effective range is constant, the $n_1 (m,k)={\rm l_{min}}$ is also a constant value. That is, for ${\rm l_{min}}={\rm l_{max}}/{\rm k_{max}}$, constant-range synapses also satisfy Formula \eqref{eq2} and Formula \eqref{eq3}. Formula \eqref{eq2} can be named the synaptic competitive formula, and Formula \eqref{eq3} can be dubbed as the synaptic strength rebalance formula.

\begin{equation}
\begin{aligned}
r{(m,k)}&=r{(m,k)}-[r{(m,k)}_{better}-r{(m,k)}]\times rand + P\\
or\ r{(m,k)}&=r{(m,k)}\\
&+[MW_1 \times r{(m,k)}_{best}+MW_2 \times r{(m,k)}_{worse}\\
&+MW_3 \times r{(m,k)}_{better}-r{(m,k)}] \times rand + P\\
&MW_1=0.5, MW_2=0.25, MW_3=0.25\\
or\  R&=r(m,k)_{best}+rand \times [r(m,k)_{best}-r(m,k)_{better}]\\
r(m,k)_{temp, better}&={\rm sign}(rand-0.5) \times \beta \times \\
&\lvert r(m,k)_{best}+rand \times [r(m,k)_{best}-r(m,k)_{better}]  \\ & -\sum_{i = j-LEN+1}^{j} {\frac{r(m,k)_{i,better}}{LEN}}\rvert \times \ln(\frac{1}{rand})\\
r(m,k)_{temp, worse}&={\rm sign}(rand-0.5) \times \beta \times \\
&\lvert r(m,k)_{best}+rand \times [r(m,k)_{best}-r(m,k)_{better}]  \\ & -\sum_{i = j-LEN+1}^{j} {\frac{r(m,k)_{i,worse}}{LEN}}\rvert \times \ln(\frac{1}{rand})\\
Q_P&=(rand > 0.5) \times r(m,k)_{temp, better} \\
Q_N&=(rand > 0.5) \times r(m,k)_{temp, worse}\\
P=&{\rm sign}(rand-0.5)\times rand \times ({\rm j_{max}}-j)/{\rm j_{max}}\\
r(m,k)=r(m,k)+rand \times &[MW_1 \times R + MW_2 \times Q_N + MW_3 \times Q_P-r(m,k)]+P\\
        &MW_1 = 0.5, MW_2 = 0.25, MW_3 = 0.25
\end{aligned}
\label{eq4}
\end{equation}


\begin{figure}[H]%
\centering
\subfigure[\label{a}]{
\includegraphics[width=0.9\textwidth]{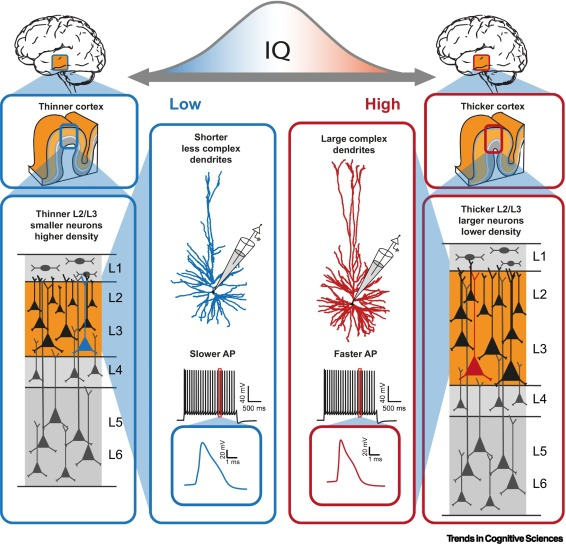}}
\subfigure[\label{b}]{
\includegraphics[width=0.9\textwidth]{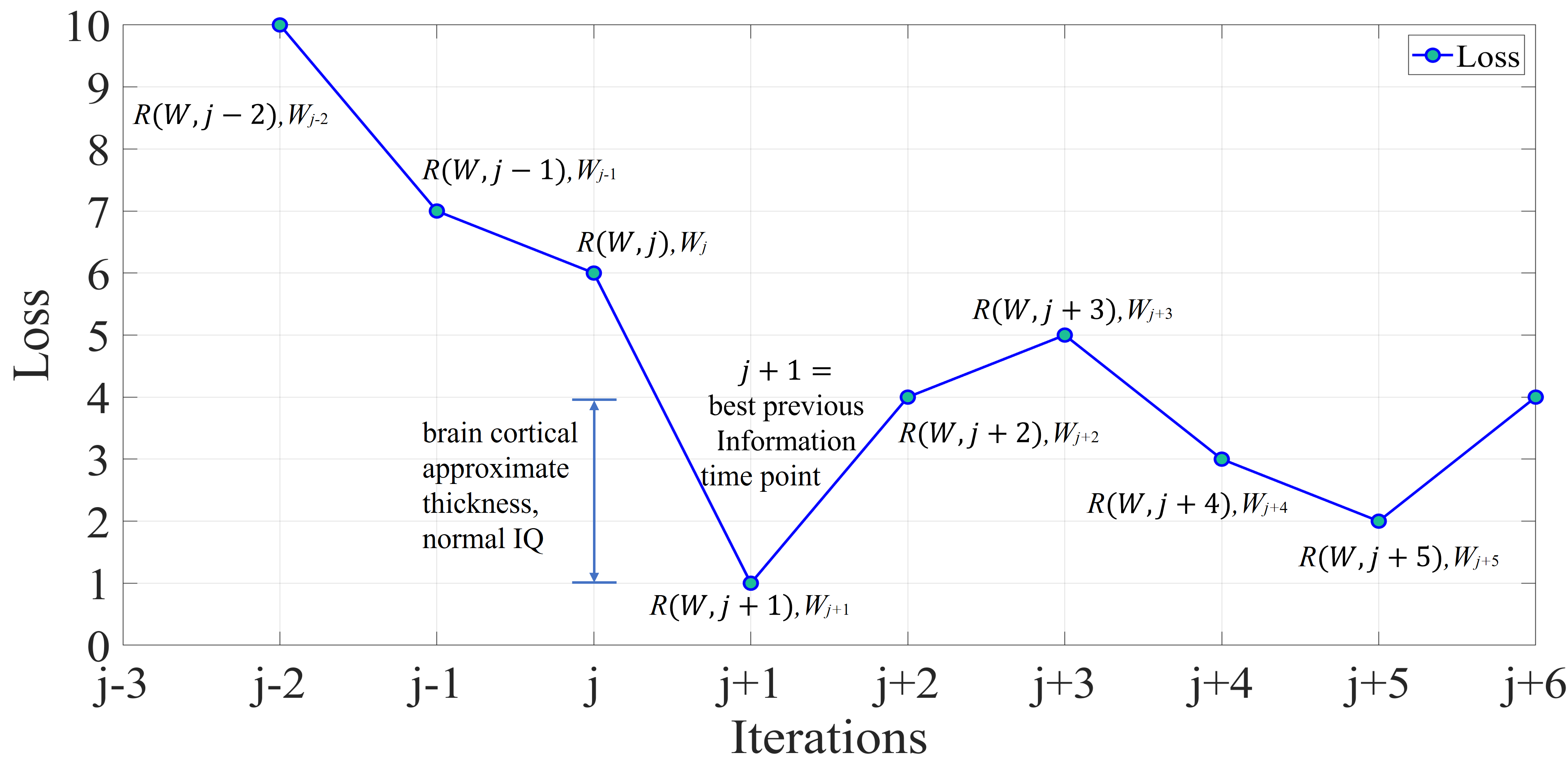}}
\end{figure}
\begin{figure}[H]%
\centering
\subfigure[\label{c}]{
\includegraphics[width=0.9\textwidth]{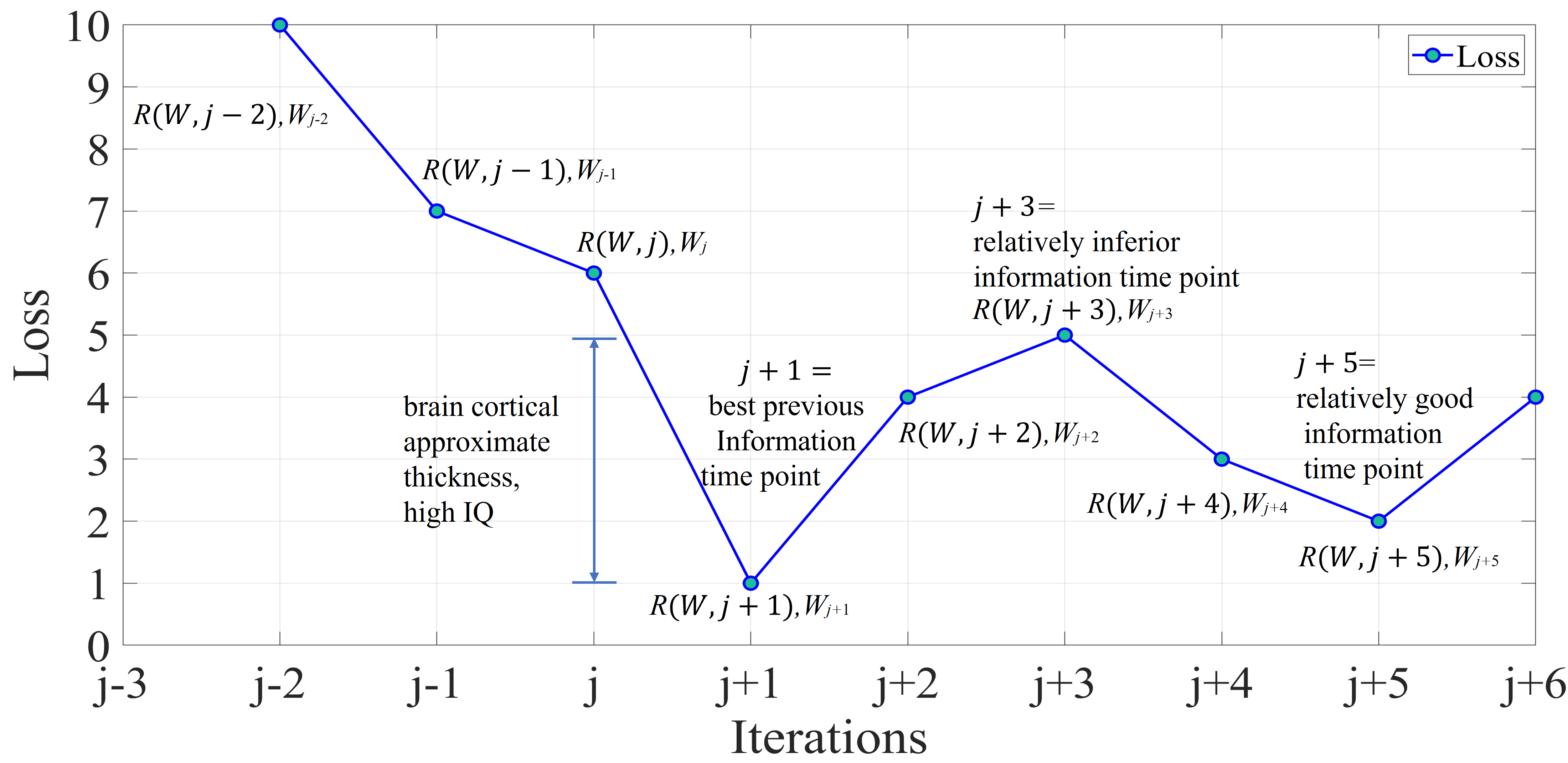}}
\subfigure[\label{d}]{
\includegraphics[width=0.9\textwidth]{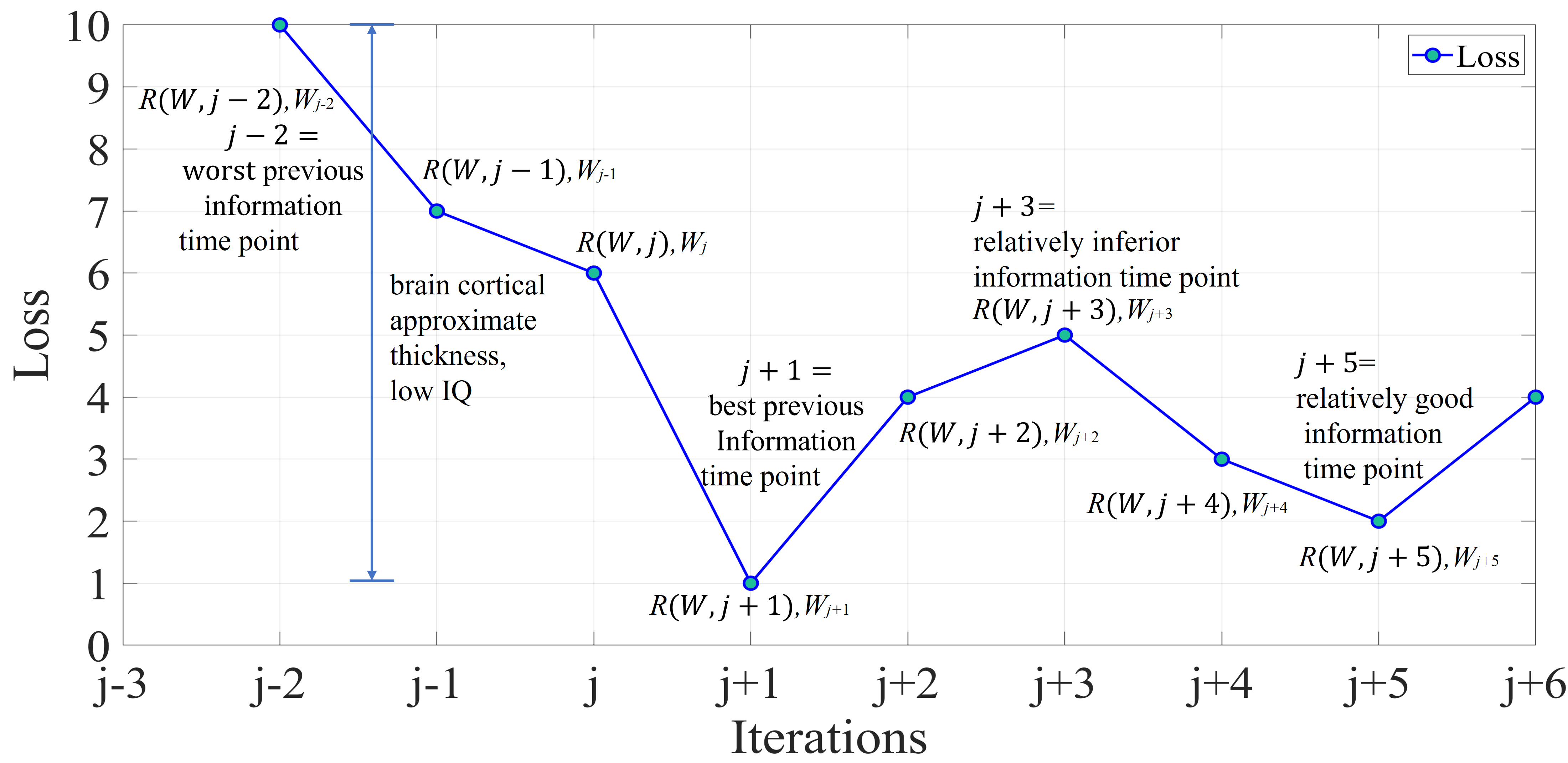}}
\subfigure[\label{e}]{
\includegraphics[width=0.9\textwidth]{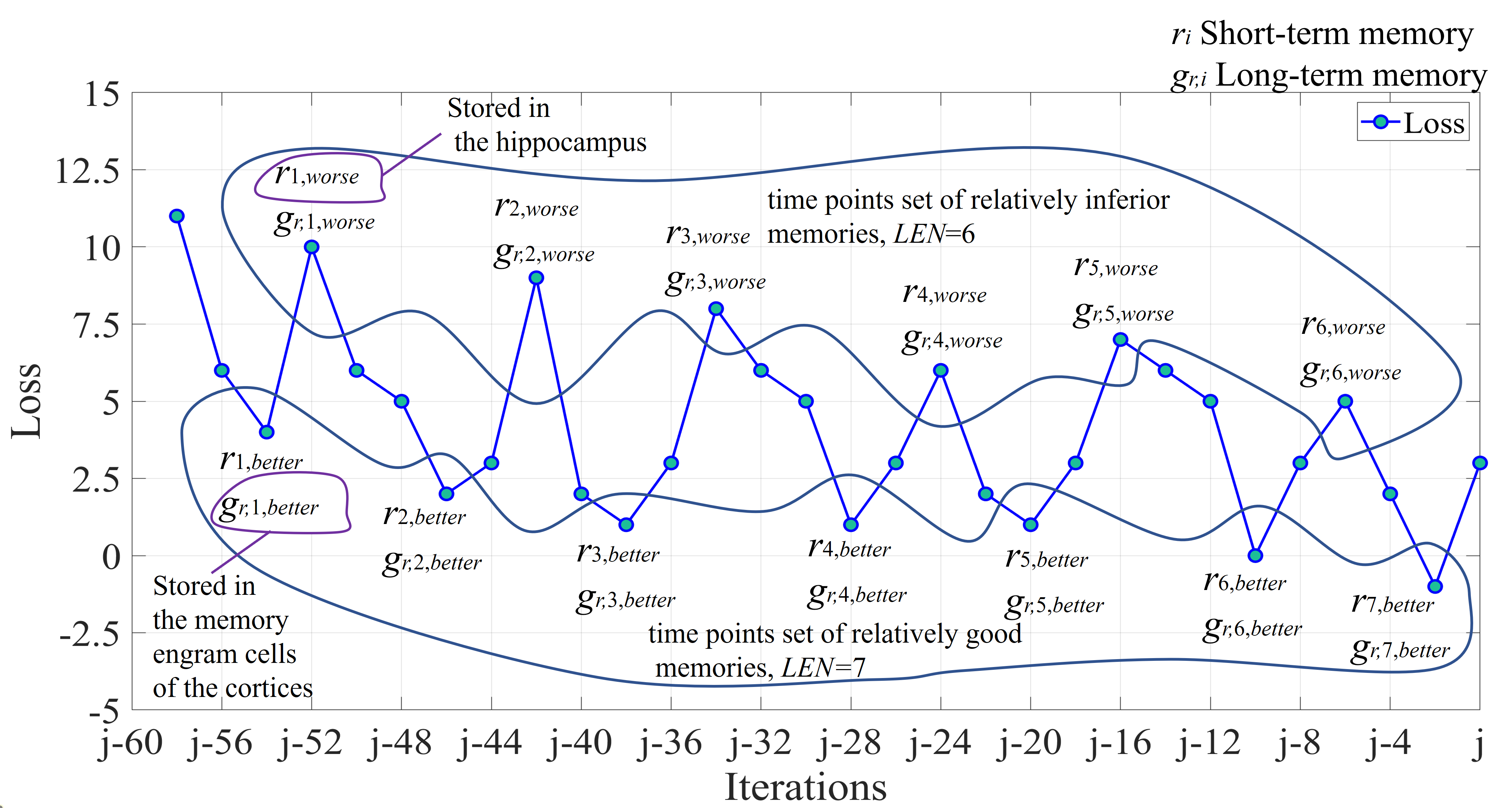}}
\end{figure}
\begin{figure}[H]%
\centering
\subfigure[\label{f}]{
\includegraphics[width=0.9\textwidth]{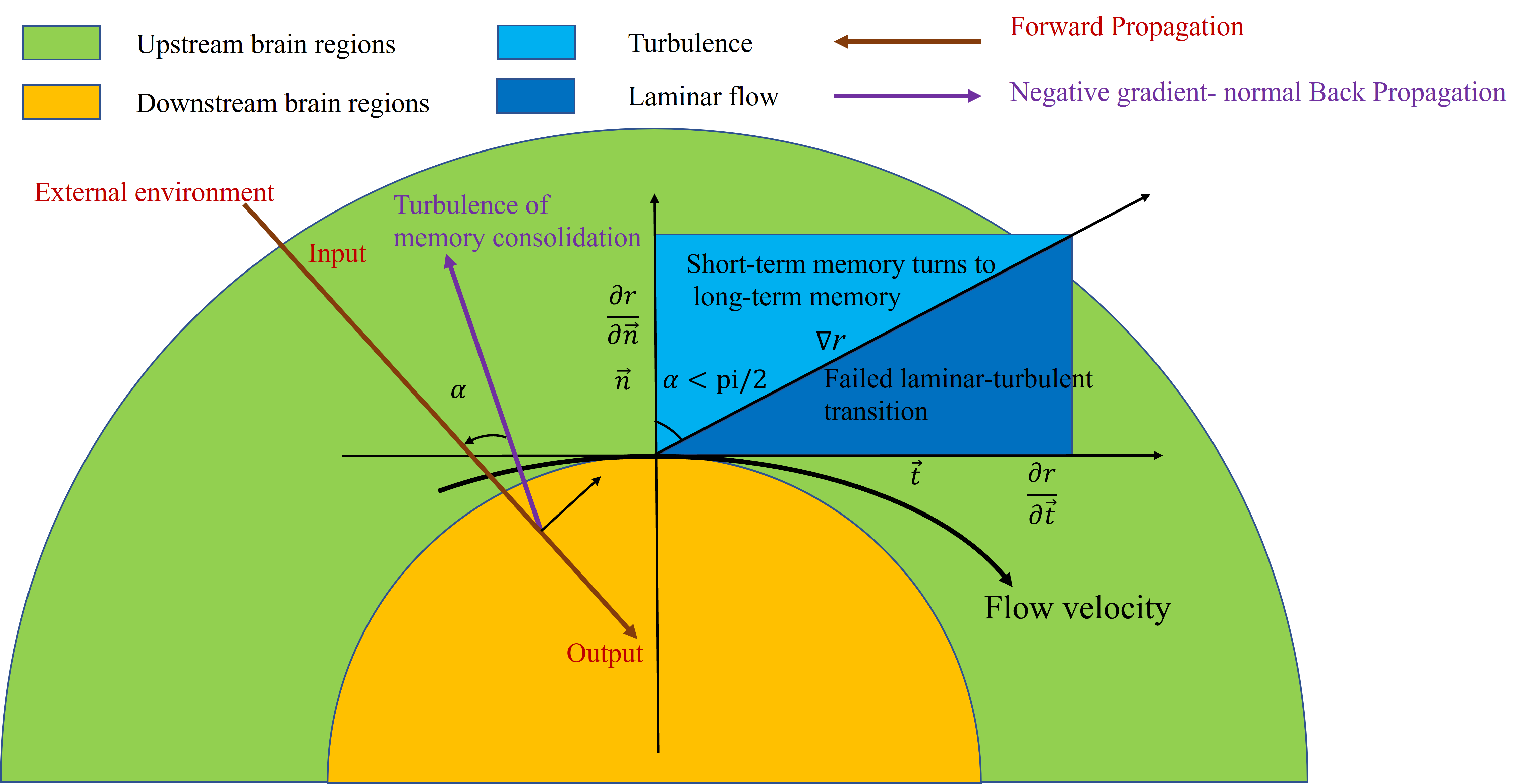}}
\subfigure[\label{g}]{
\includegraphics[width=0.9\textwidth]{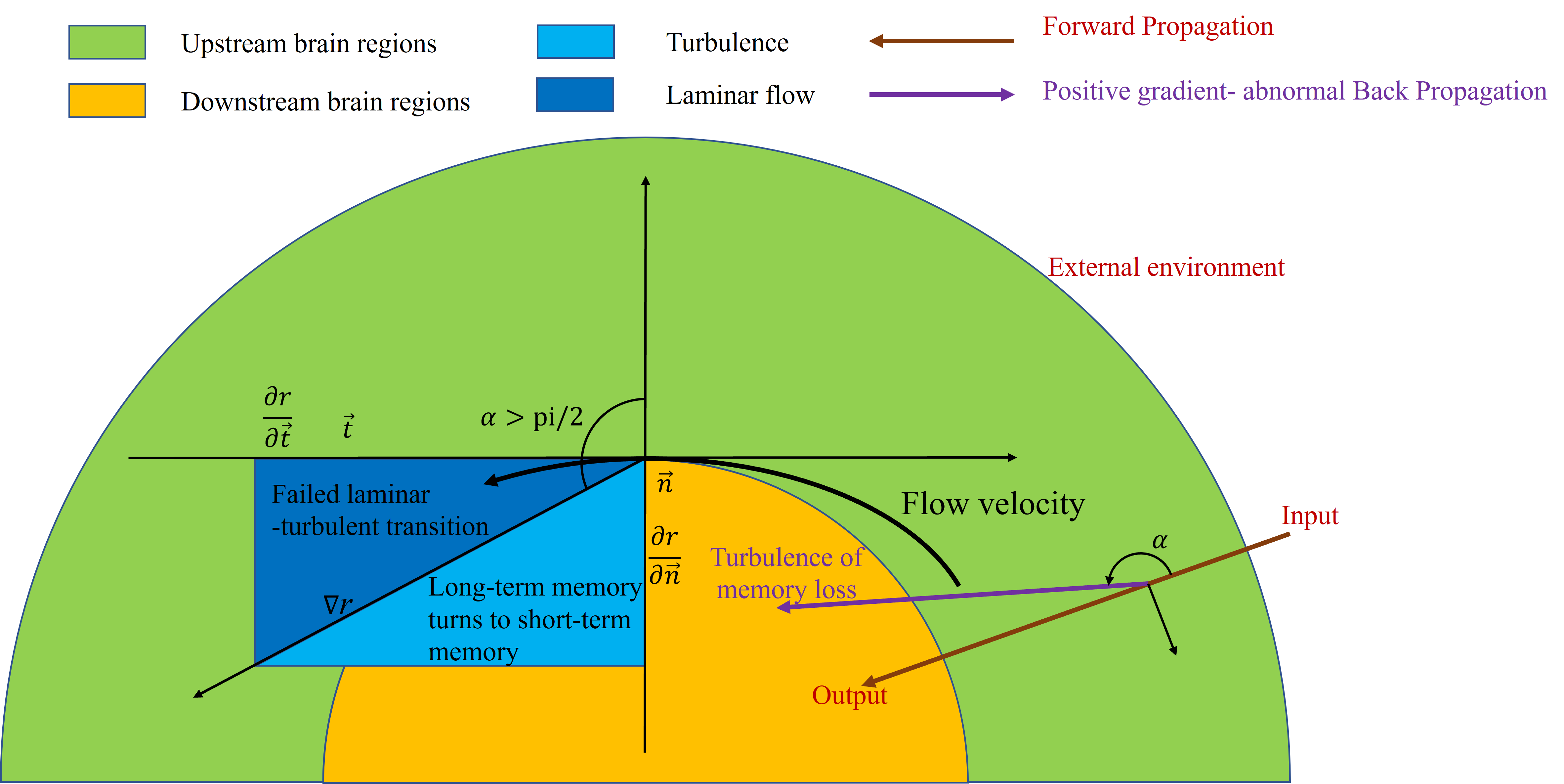}}
\includegraphics[width=0.9\textwidth]{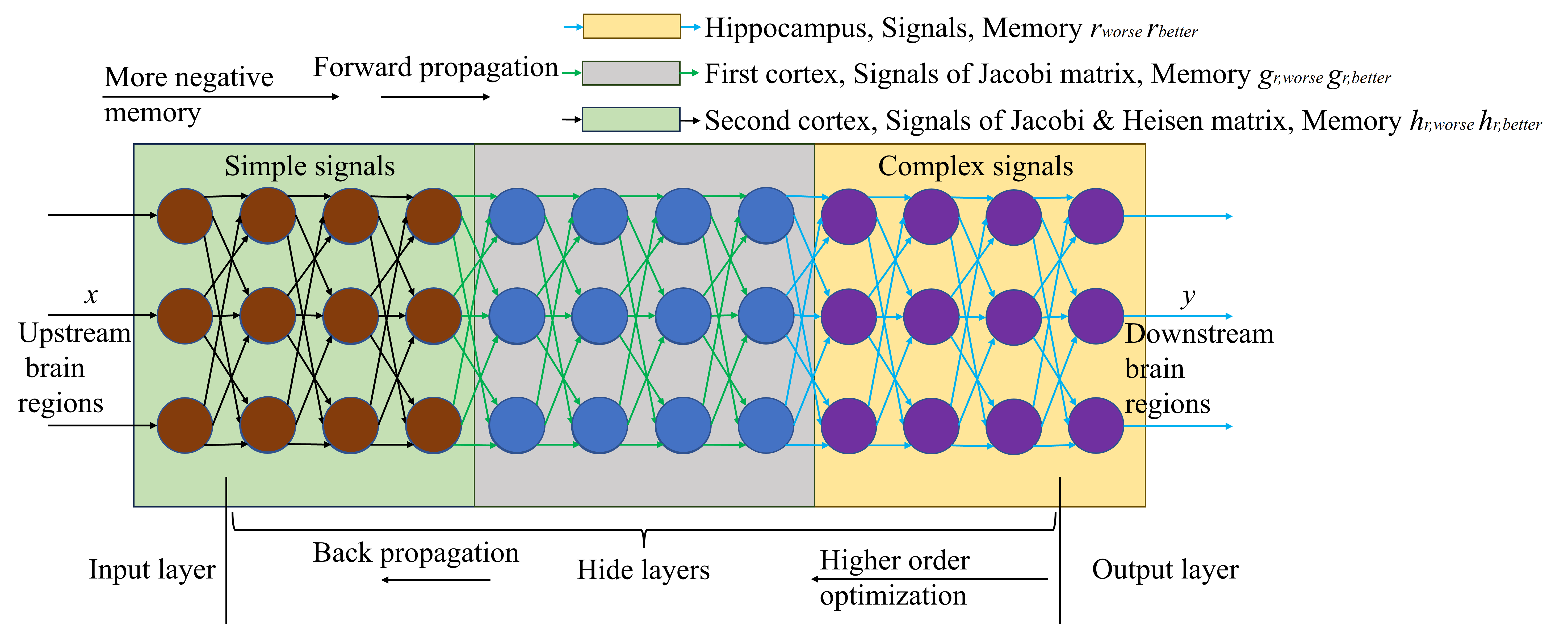}
\end{figure}
\begin{figure}[H]%
\centering
\subfigure[\label{h}]{
\includegraphics[width=0.9\textwidth]{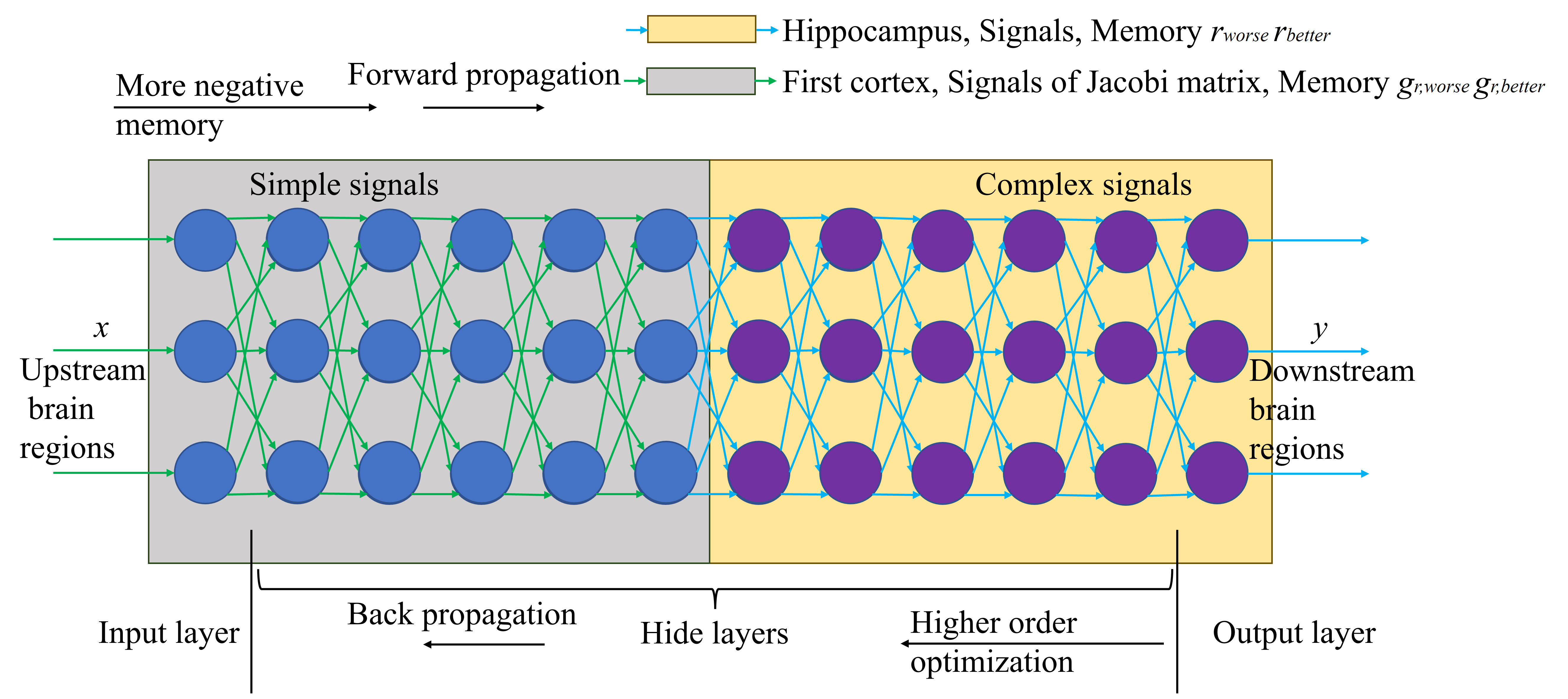}}
\subfigure[\label{i}]{
\includegraphics[width=0.9\textwidth]{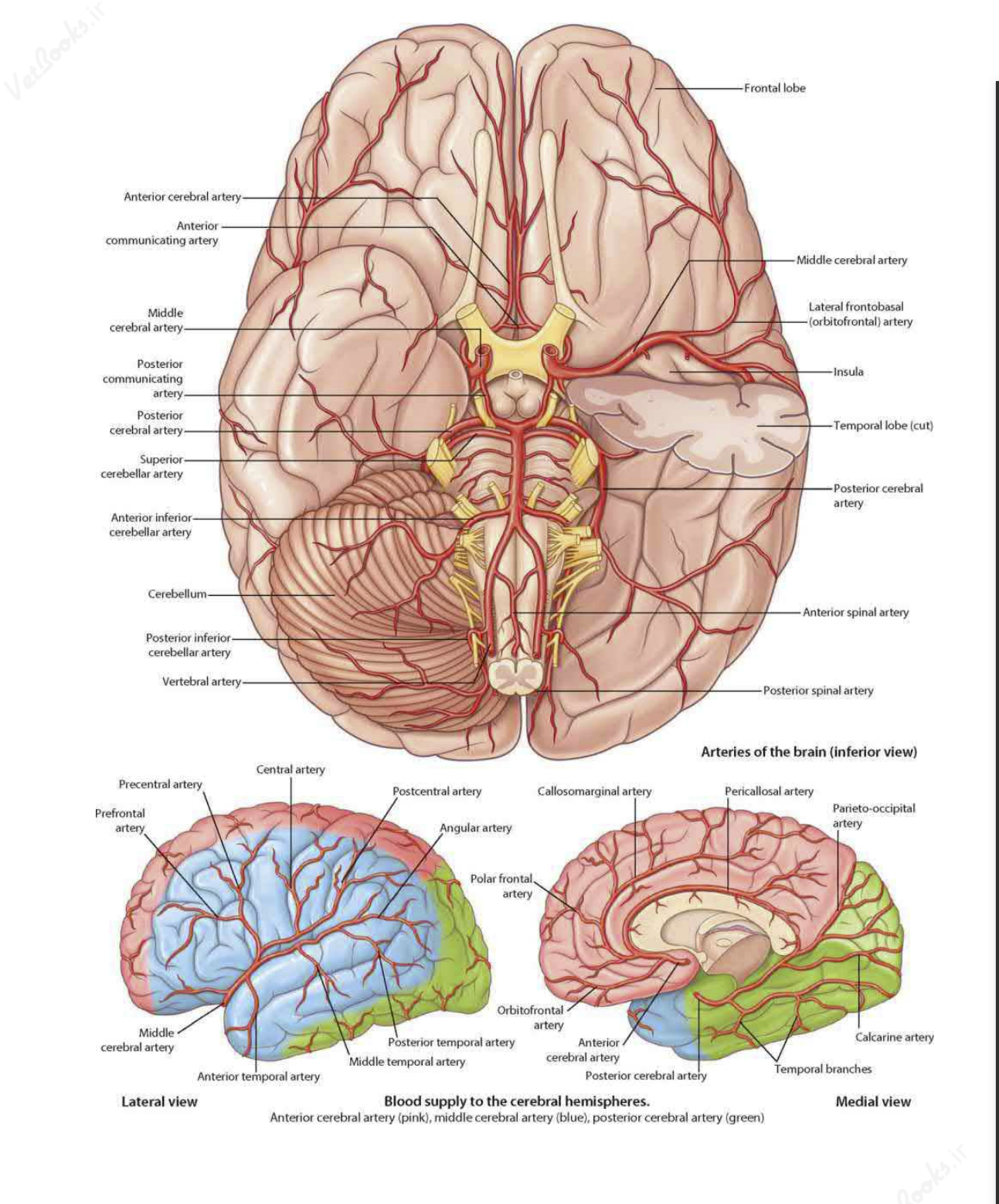}}
\end{figure}
\begin{figure}[H]%
\centering
\subfigure[\label{j}]{
\includegraphics[width=0.9\textwidth]{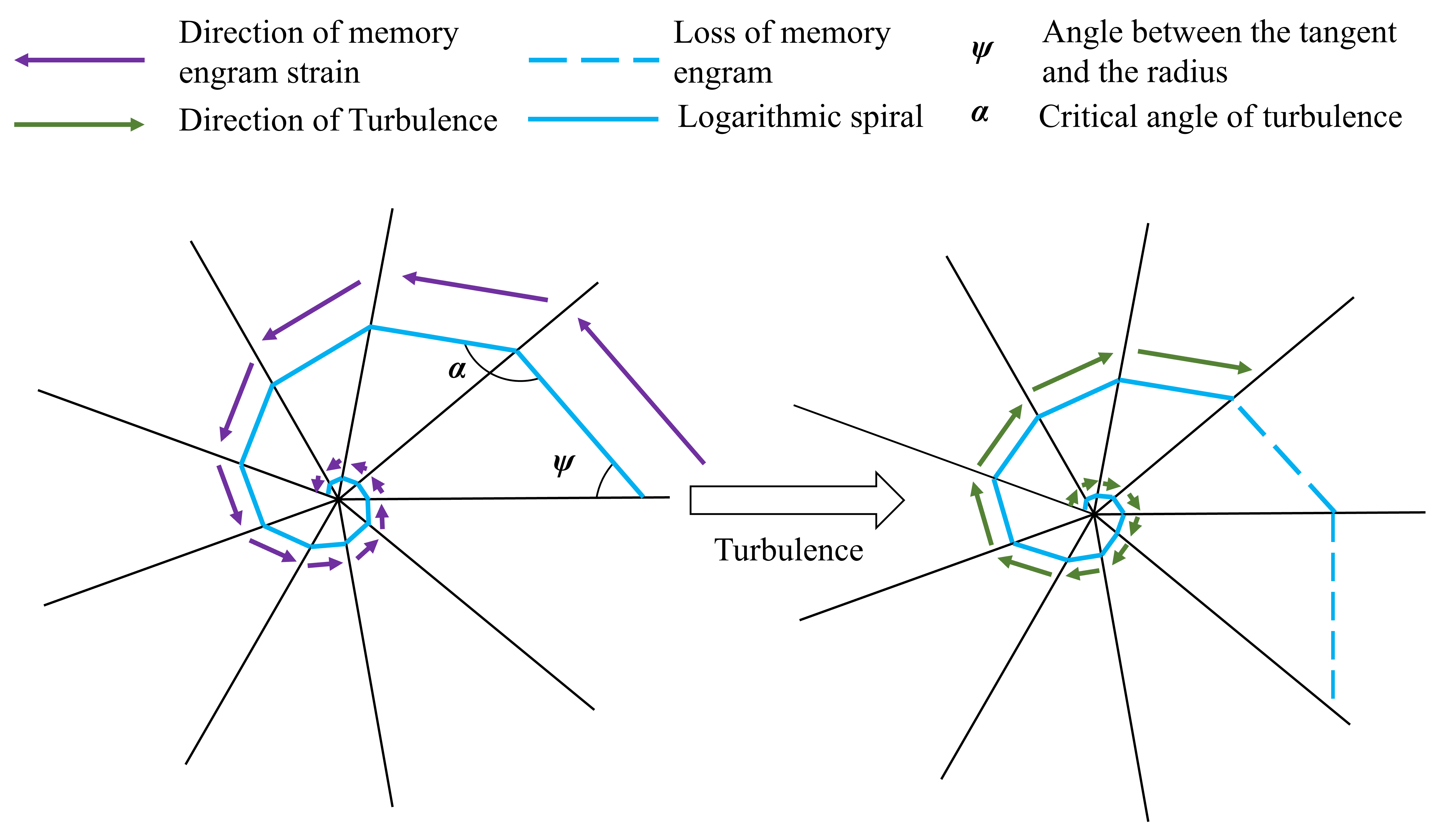}}
\caption{The brain's reverse turbulence is explained by figures: (a)Biological experiment results \cite{bib33} (b)Normal IQ (c)High IQ (d)Low IQ (e)$LEN$ points of relatively good and inferior memories (f)The critical angle between turbulence and laminar flow \cite{bib39} (g) Memory, cognition and affection Loss (h) Deep Learning model for upstream and downstream brain regions (i) Cerebral artery, from \textit{Gray's Atlas of Anatomy 3rd Edition} (j) Angles of logarithmic Spiral and loss of memory engram}
\label{fig20}
\end{figure}

\begin{figure}[H]%
\includegraphics[width=0.9\textwidth]{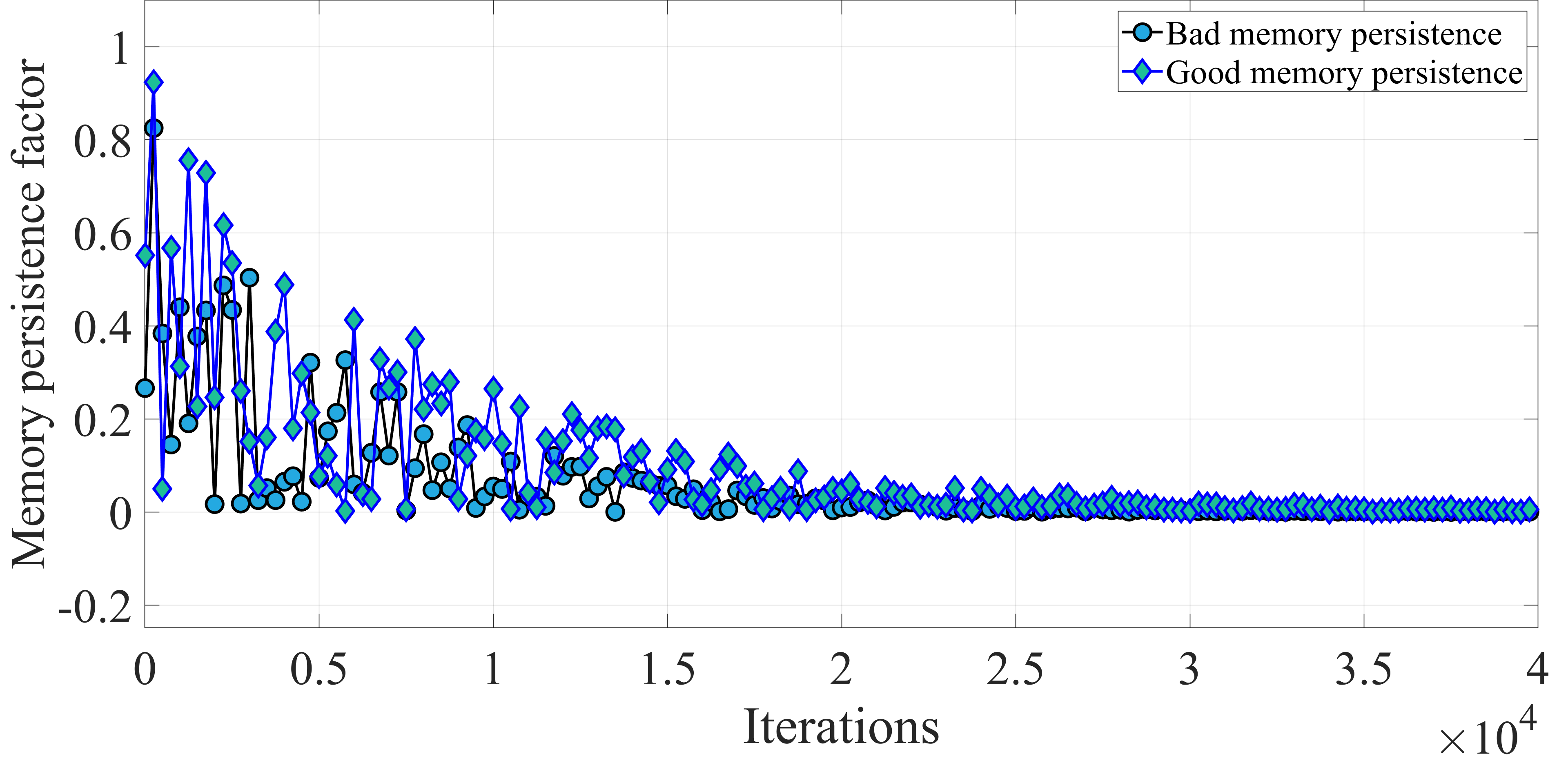}
\includegraphics[width=0.9\textwidth]{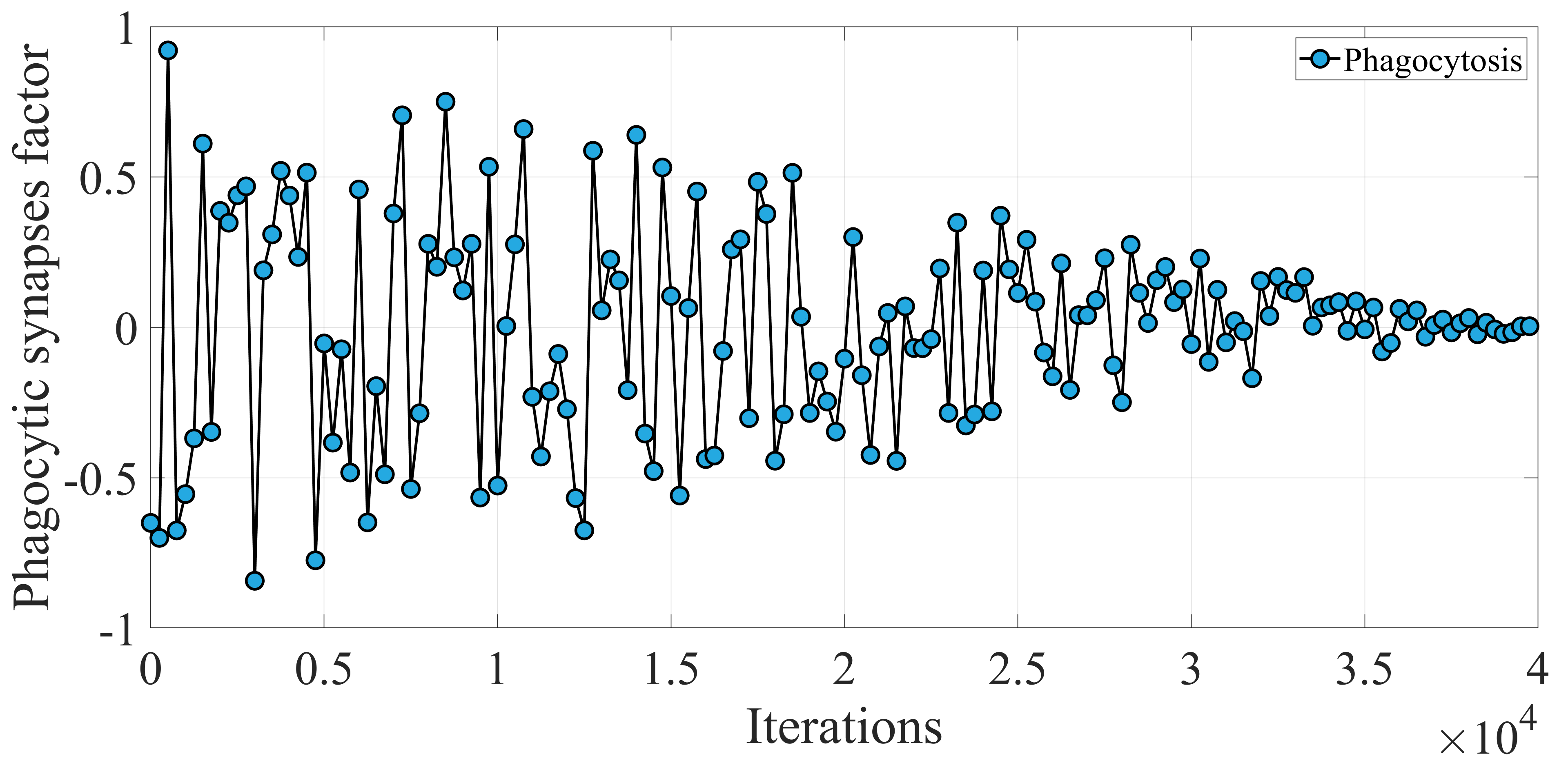}
\caption{Astrocytic synapse formation cortex memory persistence factor and astrocytes phagocytose synapses factor: (a) Astrocytic synapse formation cortex memory persistence factor (b) Astrocytes phagocytose synapses factor}
\label{fig1}
\end{figure}

Formula \eqref{eq4} updates  $r(m,k)$ by current and memory synaptic effective range weight, the best previous $r(m,k)_{best}$, and the relatively good $r(m,k)_{better}$. In the model featuring critical period, the hypothesis is relatively good brain plasticity affects current brain plasticity. $P={\rm sign}(rand-0.5)\times rand \times ({\rm j_{max}} - j)/{\rm j_{max}}$ implements plus or minus disturbance of brain plasticity astrocytes phagocytose synapses factor, and this also enables dynamic synaptic strength balance. The dynamic forgotten memory astrocytic synapse formation cortex memory persistence factor and disturbance astrocytes phagocytose synapses factor shown in Fig. \ref{fig1}. $({\rm j_{max}}- j)/{\rm j_{max}}$ illustrates we give decreasing weights of factor at critical period in $P$. ${\rm sign}(rand-0.5)$ gives Formula $P$ both positive and negative results. $P$ can be negative because of astrocytes phagocytose synapses, but considering the strength rebalance of synaptic effect, $P$ should be positive and negative. The $P$ decreases with the maturation response of astrocytes.

In our opinion, when you are considering one problem, whether positive or negative, the exacted working memory brain plasticity may reflect quantum, and exhibit exponential decay of wave function for a while because of barriers in hippocampus and cortices. It is important to maintain inner peace, for reducing these effects of exponential decay, and flatter curve.The actions of neurons may like particles, at the $LEN$ points of relatively good or inferior memories, mapping functions of these extracted memories brains plasticity may be quantum, the square of the amplitude is the distribution probability of the particle by quantum mechanics, and distribution probability of $y=R-r(m,k)_{i,better}$ or $y=R-r(m,k)_{i,worse}$ satisfies $\lvert \varPsi[f(y)] \rvert^2\propto \frac{exp[-2\left\lvert f(y) \right\rvert \times L(t)]}{L(t)} $ or $\lvert \varPsi(y) \rvert^2\propto \frac{exp[-2\left\lvert y \right\rvert \times L(t)]}{L(t)} $.The wave function describes synaptic effective range weights or brain plasticity, The length of time is $LEN$ form $(j-LEN+1)$th iteration to $j$th iteration, and by using Mont Carlo Method to calculate $r(m,k)_{temp, worse}={\rm sign} (rand-0.5)\times \beta \times \lvert r(m,k)_{best}+rand \times [r(m,k)_{best}-r(m,k)_{better}] - \sum_{i=j-LEN+1}^{j} \frac{r(m,k)_{i,worse}}{LEN} \rvert \times \ln(\frac{1}{rand}),L(t)=2 \times \beta \times \lvert r(m,k)_{best}+rand \times [r(m,k)_{best}-r(m,k)_{better}] -\sum_{i=j-LEN+1}^{j} \frac{r(m,k)_{i,worse}}{LEN}\rvert, r(m,k)_{temp,better}={\rm sign} (rand-0.5)\times \beta \times \lvert r(m,k)_{best}+rand \times [r(m,k)_{best}-r(m,k)_{better}] - \sum_{i=j-LEN+1}^{j} \frac{r(m,k)_{i,better}}{LEN} \rvert \times \ln(\frac{1}{rand}),L(t)=2 \times \beta \times \lvert r(m,k)_{best}+rand \times [r(m,k)_{best}-r(m,k)_{better}] -\sum_{i=j-LEN+1}^{j} \frac{r(m,k)_{i,better}}{LEN}\rvert, \beta=1$, and $\beta$ is shrinkage-expansion coefficient, a superposition of wave function of time. The length of time $LEN$, it may be short-term memory in quantum computing. $j$ is current iteration. $i$ is $i$th iteration. $MW_2 \times Q_{N}+MW_3 \times Q_{P}$ means a linear superposition of two wave functions. $r(m,k)_{i,better}$ is relatively good synaptic effective range weight in $i$th iteration, $r(m,k)_{i,worse}$ is relatively inferior synaptic effective range weight in $i$th iteration, $r(m,k)_{i,best}$ is best previous synaptic effective range weight in $i$th iteration. And $(rand>1/2)$ means it has $2^2=4$ scenarios of the relatively inferior and good two-qubit: \{1,1\}, \{1,0\}, \{0,1\} or \{0,0\}. The quantum uncertainty of relatively inferior and good range weights, because of scan resolution of the cortical folds. The effects of astrocytes phagocytose synapses $P$ and quantum computing $Q$ in calculations are very similar in minus or plus, the process of astrocytes phagocytose synapses may a normal computer by random or gradient, but also a quantum computer. It may be that the process of astrocytes phagocytose synapses is driven by positive and negative memory of brain plasticity, by excitatory/inhibitory interactions. The quantum computing to update the synaptic effective range weights by short-term memory, means $r(m,k)_{temp,better}$ or $r(m,k)_{temp,worse}$.

We can imagine our brain as the earth, and the production of geocentric lava occurs like short-term memory happen in the hippocampus, and the process is quantum. Earthquakes on the surface are released due to potential energy, just as strong short-term memory is selected and turns to long-term memory, and is stored in memory engram cells of different cortices, can be released. 

Sir Roger Penrose has put forward a bold conjecture that the potential features of quantum computation could explain enigmatic aspects of consciousness. The Artificial Brain model is actually the Heart-Brain model, and the heart serves as a medium, and the quantum entanglement between the heart and brain may be beyond absolute space-time. The heart produces positive $h(m,k)_{i,better}$ or negative $h(m,k)_{i,worse}$ pulse frequencies, and the pulse frequencies interaction potential signal strengthens or weakens synapses, which in turn changes the relatively good or inferior brain architecture $r(m,k)_{i,better}$ and $r(m,k)_{i,worse}$ This transcendental space-time quantum entanglement at heart frequencies accumulation $\sum_{i=j-LEN+1}^{j} \frac{h(m,k)_{i,better}}{LEN}$ and $\sum_{i=j-LEN+1}^{j} \frac{h(m,k)_{i,worse}}{LEN}$ feedback to the brain architectures accumulation $\sum_{i=j-LEN+1}^{j} \frac{r(m,k)_{i,better}}{LEN}$ and $\sum_{i=j-LEN+1}^{j} \frac{r(m,k)_{i,better}}{LEN}$ and satisfies exponential decay because of barriers within the brain \cite{bib36}.

Referring formula \eqref{eq2} and \eqref{eq4}, the memory consolidation formula \eqref{eq5} can be shown as follows, it is the relationship of short-term memory and long-term memory, and maximum of directional derivatives is gradient. From hippocampus to cortices, it is achieved from non-classical to classical in brain. Long-term memory is gradient of brain plasticity  $\frac{dr}{dn}$ which is too idealistic, there is an angle $\alpha_c=\tan^{-1} \frac{\frac{dr}{dt}}{\frac{dr}{dn}}$, $\alpha_c$ will increase because of human aging, the angle $\alpha_c$ is the critical value, greater than it is laminar flow and less than it is turbulence. And turbulence becomes laminar flow gradually throughout aging, the $\alpha_c=\tan^{-1} \frac{\frac{dr}{dt}}{\frac{dr}{dn}}$ was shown in Fig.\ref{fig20}(f) \cite{bib39}. If $\alpha_c \geq \pi$, symptoms of Alzheimer's disease will appear, and atrophy and hardening of the hippocampus because of reverse turbulence, was shown in Fig. \ref{fig20}(g), just like a star dying to form black hole. Our conjecture is Alzheimer's disease cognitive impairment is caused by search direction reversal, the negative gradient of Back Propagation is modified by positive gradient, and unable to converge, change of direction of transition was shown in Fig. \ref{fig20} (f)-(g) \cite{bib39}. The $\tan \alpha_c$  in formula \eqref{eq5} is equivalent to $MW_1$ or $MW_2$ or $MW_3$ in formula \eqref{eq2}. The simultaneous interaction of two different turbulence directions of $\alpha_c \geq \pi$ memory loss and $\alpha_c \leq \frac{\pi}{2}$  memory consolidation leads to $\beta$-amyloid plaques in brain. But the positive gradient of BP might be the more important reason for Alzheimer's disease than $\beta$-amyloid plaques in brain.

We studied Alzheimer's disease based on turbulence in brain information transmission and dysfunction of the neurotransmitter system. Disorders of neurotransmitters lead to abnormalities in turbulence between cortexes inside the brain - that is, the reverse turbulence. First we look at the arteries of the healthy brain Fig. \ref{fig20}(i), where the large and middle cerebral arteries are located inside the brain, and then the external branch arteries of the brain turn to small. The brain undergoes turbulent movements beyond a minimum critical value to spread information from the internal large and middle arteries to the small external branch arteries, that is, normal turbulence is from the downstream brain regions of the internal cortices to the upstream brain regions of the external cortices to achieve the Back Propagation of Deep Learning. Similarly, the large and middle ducts connecting the internal cortices are also larger in the downstream brain regions of the brain, while the ducts in the upstream brain regions of the external cortices are smaller.

For example, staying up late on the function of the kidneys and liver, and affects the excretion of intestinal toxins, so that the blood accumulates too many toxins and garbage and affects the metabolism of the brain. The blood and toxins are transmitted to the large and middle arteries of the brain through cardiac contraction, so that the large middle arteries appear cerebral atherosclerotic plaques and infarction, in order to unblock the plaques of these large and middle arteries, the heart must use a larger systolic blood pressure may lead to systolic hypertension. Similarly, because $\beta-$amyloid and tau proteins affect the cerebrospinal fluid, which in turn affects the large and middle ducts of the internal cortices. In this way, the branch arteries external our brain and the branch ducts between the cortices are relatively large because there is no garbage and toxins, while the large middle arteries inside the brain and the large and middle ducts in the cortices are relatively small. As a result, there is a reverse diffusion of turbulence between the cortices.

Heart failure is another condition in which reduced blood flow causes the large middle arteries and large middle ducts in the downstream brain regions relatively small.

Increased systolic blood pressure due to atherosclerosis, systolic hypertension, and heart failure leading to decreased blood flow and the large middle arteries in the brain turn to small, which may be responsible for Alzheimer's disease.

Why do patients with Alzheimer's disease experience circadian rhythm disorders, after the patient lies down, because only a smaller systolic blood pressure than the original can clear the large middle ducts inside the brain $\beta$-amyloid, tau protein, toxins and garbage of the large middle arteries, lying down also increases blood flow and expands the large middle artery and large middle duct, so that the direction of partial turbulence inside the brain returns to normal, makes thinking clear and memory restored and constant thinking, which in turn leads to circadian rhythm disorders.

Brain dynamics analogous to river dynamics, when the river that flows backwards, because the previous river flow slope is opposite to the current flow slope, it may lead to river siltation. The river siltation is like $\beta$-amyloid in brain. The river that flows backwards is like turbulence reverses in downstream brain regions.

\begin{equation}
\begin{aligned}
\frac{dr}{dn}\vert_{(x_0,y_0)}&=\frac{\partial r}{\partial x } \cos \alpha + \frac{\partial r}{\partial y } \cos \beta \vert_{(x_0,y_0)}\\
&=(\frac{\partial r}{\partial x}, \frac{\partial r}{\partial y})\vert_{(x_0,y_0)} \cdot (\cos \alpha, \cos \beta)\\
&={\rm grad} \, r(x_0,y_0) \cdot (\cos \alpha, \cos \beta)\\
&={\rm grad} \, r(x_0,y_0) \cdot \vec{e_n}\\
{\rm if} \, \alpha&=0, \vec{e_n}=(1,0) {\rm then} \, \max \frac{dr}{dn}\vert_{(x_0,y_0)} = \nabla r(x_0,y_0)\\
{\rm or} \, \frac{dr}{dn}\vert_{(x_0,y_0,z_0)} &= \frac{\partial r}{\partial x } \cos \alpha + \frac{\partial r}{\partial y } \cos \beta + \frac{\partial r}{\partial z }\vert_{(x_0,y_0,z_0)}\\
&=(\frac{\partial r}{\partial x}, \frac{\partial r}{\partial y}, \frac{\partial r}{\partial z})\vert_{(x_0,y_0,z_0)} \cdot (\cos \alpha, \cos \beta, \cos \gamma)\\
&={\rm grad} \, r(x_0,y_0,z_0) \cdot (\cos \alpha, \cos \beta, \cos \gamma)\\
&={\rm grad} \, r(x_0,y_0,z_0) \cdot \vec{e_n}\\
{\rm if} \, \alpha&=0, \vec{e_n}=(1,0,0) {\rm then} \, \max \frac{dr}{dn}\vert_{(x_0,y_0,z_0)} = \nabla r(x_0,y_0,z_0)\\
{\rm or} \, \alpha_c&= \tan ^{-1} \frac{\frac{dr}{dn}}{\frac{dr}{dt}}\\
\vec{e_n}&=(\cos \alpha_c, \sin \alpha_c, 0)\\
{\rm if} \, \alpha & \leqslant \alpha_c \, {\rm turbulence,} \, {\rm short-term} \, {\rm memory} \, {\rm turns} \, {\rm to} \, {\rm long-term} \, {\rm memory}\\
&{\rm else} \,\alpha > \alpha_c \, {\rm laminar} \, {\rm flow,} \, {\rm no} \, {\rm long-term} \, {\rm memory} \, {\rm can} \, {\rm be} \, {\rm transited} \\
&{\rm by} \, {\rm short-term} \, {\rm memory} \\
\frac{dr^2}{d^2n} \vert _{(x_1,y_1,z_1)} &=\frac{\partial r^2}{\partial^2 x} \cos \alpha \cos \alpha_2 \\
&+ \frac{\partial r^2}{\partial^2 y} \cos \beta \cos \beta_2 + \frac{\partial r^2}{\partial^2 z} \cos \gamma \cos \gamma_2 + \frac{\partial r}{\partial x} \frac{\partial r}{\partial y} \cos \alpha \cos \beta_2\\
&+ \frac{\partial r}{\partial x} \frac{\partial r}{\partial z} \cos \alpha \cos \gamma_2 + \frac{\partial r}{\partial y} \frac{\partial r}{\partial z} \cos \beta \cos \gamma_2 + \frac{\partial r}{\partial y} \frac{\partial r}{\partial x} \cos \beta \cos \alpha_2 \\
&+ \frac{\partial r}{\partial z} \frac{\partial r}{\partial x} \cos \gamma \cos \alpha_2 + \frac{\partial r}{\partial z} \frac{\partial r}{\partial y} \cos \gamma \cos \beta_2 \vert _{(x_1,y_1,z_1)}\\
\end{aligned}
\label{eq5}
\end{equation}

The pseudocode of PNN is as follows, which encompasses Forward and Back Propagation.

\renewcommand{\thefootnote}{\fnsymbol{footnote}}

\begin{figure}
        \centering
        \includegraphics[width=\textwidth]{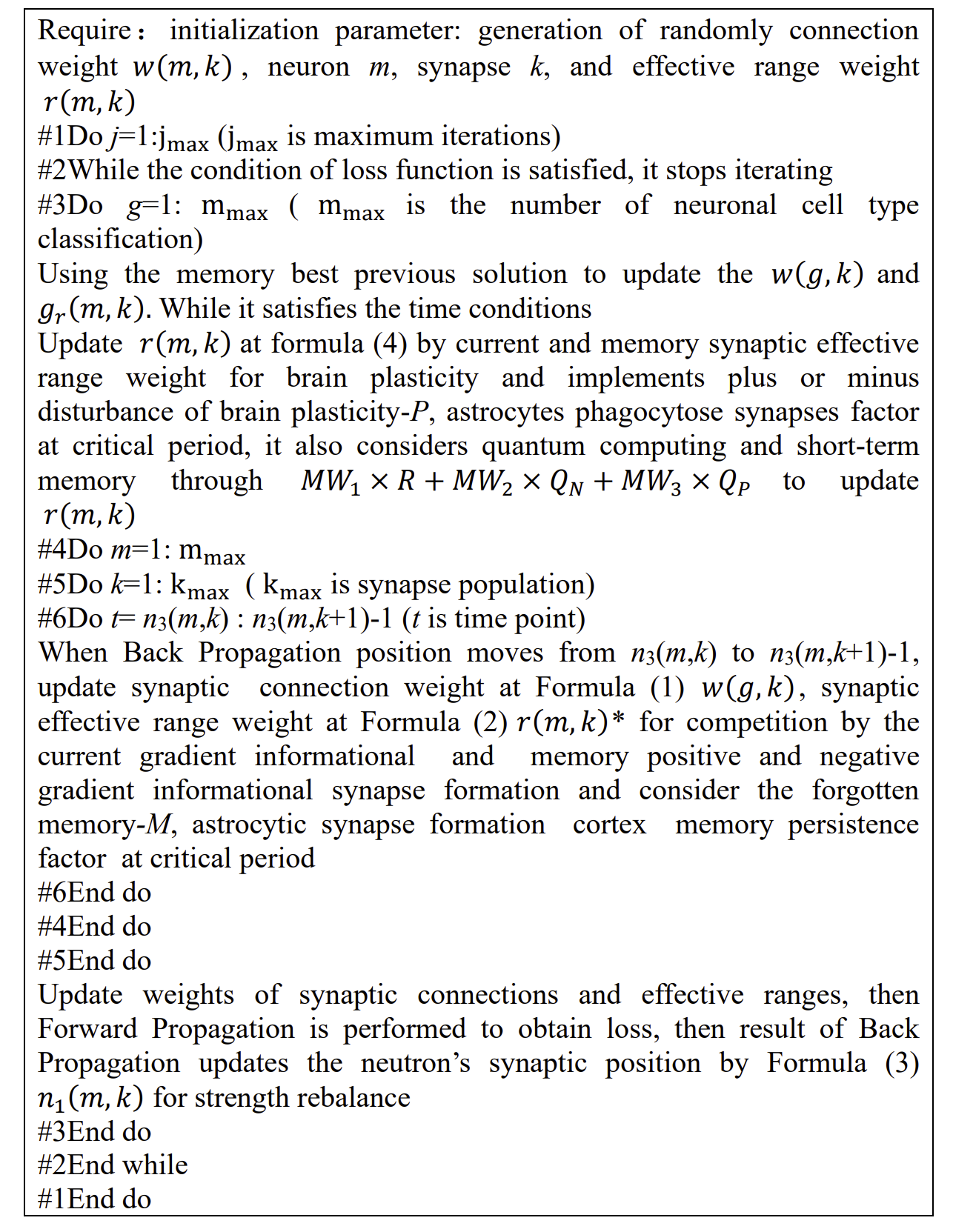}
        
\end{figure}

Because it is the reverse turbulence downstream of the brain, the positive gradient of renewal synaptic effective range is mainly distributed near the downstream brain regions of the internal cortices of the brain. The positive gradient of the connecting weights may be related to the reverse turbulence of immune cells leading to neuroinflammation. The $n$th to $n+m$th layers of the downstream brain regions use redundant objective function information of the $n+j$th to $n+m+j$th layers of the downstream brain regions.

\footnotetext{*As for updating $w(g,k)$ and $r(m,k)$, the global synaptic effective range weights $r(m,k)$ need to be changed by overall optimization, while the same type of connection weight $w(g,k)$ just needs to be changed by local optimization. i.e., research on disease dynamics, when infection rates (or cure rates) change, the time interval of the infection rate and cure rates both need to be updated at the same time.}

Then loops and iteration of Deep Learning, an early iteration of the Back Propagation is negative gradient which used to update synaptic connection weights and range weights. Then the later iteration of Back Propagation gradually becomes half negative gradient and half positive gradient, canceling out the excitement of the synapses, half of the $n+m$th layers of the downstream brain regions use redundant objective function information of the downstream brain regions, and half of the $n+m$th layers of the downstream brain regions use the objective function information of the normal upstream brain regions, so that the synapse loss and hallucinations of Alzheimer's disease occurs.

If the Back Propagation at the beginning of the iteration is negative gradient, the Back Propagation in the middle of the iteration gradually becomes half negative gradient and half positive gradient, half of the $n+m$th layers of the downstream brain regions use redundant objective function information of the downstream brain regions, and half of the $n+m$th layers of the downstream brain regions use the objective function information of the normal upstream brain regions, to reflect the synapse loss and hallucinations in early Alzheimer's disease, and the Back Propagation in the late iteration is positive gradient and the $n$th to $n+m$th layers all use the objective function information of redundant downstream brain regions, to reflect more severe Alzheimer's cognitive impairment. The model realized the early synapse loss and hallucinations in Alzheimer's disease to more severe cognitive impairment in the later stage.

The reverse turbulence affects memory engrams, the reverse turbulence has a certain effect on the memory engrams in the middle of iteration, and reverse turbulence has a greater effect on the memory engrams in the late iteration.

\section{Discussion}
\subsection{Comparison of PNNs in some scenarios}
Parameters regarding the abovementioned test is as follows: testing environment is WIN11 and MATLAB2020a, the parameters for updating the weights remain the same for the three tests, and the number of iterations is ${\rm j_{max}}=40000$. The tests employ cosine filter, in which input are filter from point $t_2+1$ to $t_2+4$, and prediction of output is filter at $t_2+5$ point. ${\rm l_{min}}=2$, ${\rm k_{max}}=9$, ${\rm l_{max}}=44$. Memory ensembles of long-term memory produced once each iteration such as $g_r(m,k)_{best}$ and $g_r(m,k)_{worse}$, but retrieval processes of short-term memory happened once by many iterations such as $r(m,k)_{best}$. Three methods were all considered the memory the best previous solution, so RRPNN and ORPNN both take into consideration the current and memory connection weight and synaptic effective range. But CRPNN only pays heed to the current and memory connection weight, as the synaptic effective range remains constant in synapse formation. Only ORPNN takes into account the current gradient informational and memory positive and negative gradient informational synapse formation, and it also considers current and memory brain plasticity. In the Fig.\ref{fig2}, the ORPNN model includes two scenarios, in addition to astrocytes phagocytose synapses factor, one only considering the best previous memory of synapse formation $g_r(m,k)_{best}$ ORPNN-MF(P)-PF, and the other considering both the best previous memory $g_r(m,k)_{best}$ and the relatively inferior memory $g_r(m,k)_{worse}$ of synapse formation ORPNN-MF(P\&N)-PF. MF(P) represents positive cortex memory persistence, and MF(P\&N) represents positive and negative cortex memories persistence.

Fig.\ref{fig21} shows respectively the flow charts of RNN, CRPNN, RRPNN and ORPNN

\begin{figure}%
\centering
\subfigure[RNN
$U$ is the weight from the input layer to the hidden layer, $V$ is the weight from the hidden layer to the output layer, $S$ is the vector of the hidden layer, $X$ is the vector of the input layer, and $O$ is the vector of the output layer. $W$ is the connection weight.]{
\includegraphics[width=0.9\textwidth]{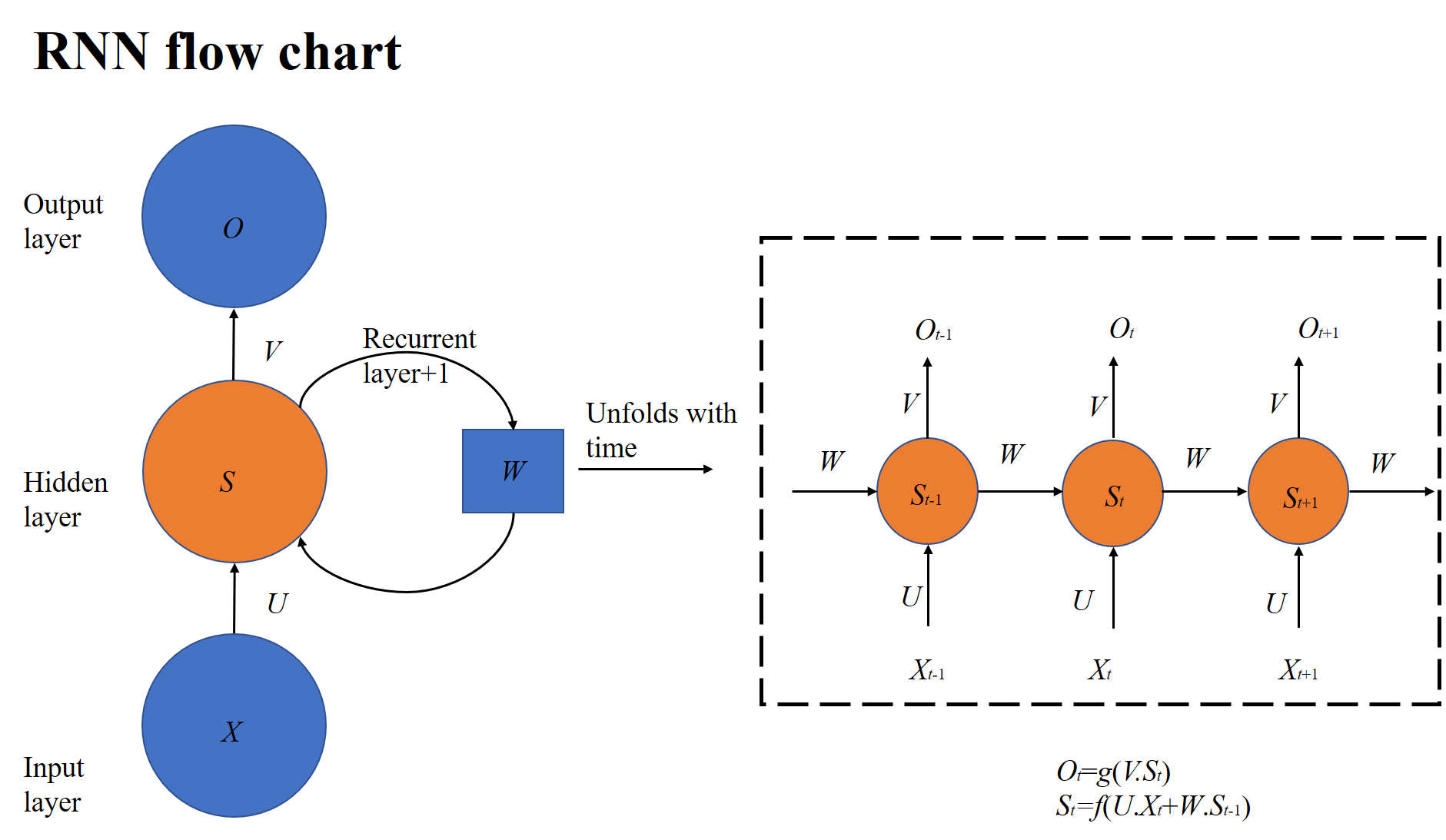}}
\subfigure[CRPNN and RRPNN 
CRPNN is very similar to RNN, except that the iteration time is changed from $+1$ to $+n_1$. The $n_1$ in RRPNN is a variable, and the length of $n_1$ is randomly generated within a certain range.]{
\includegraphics[width=0.9\textwidth]{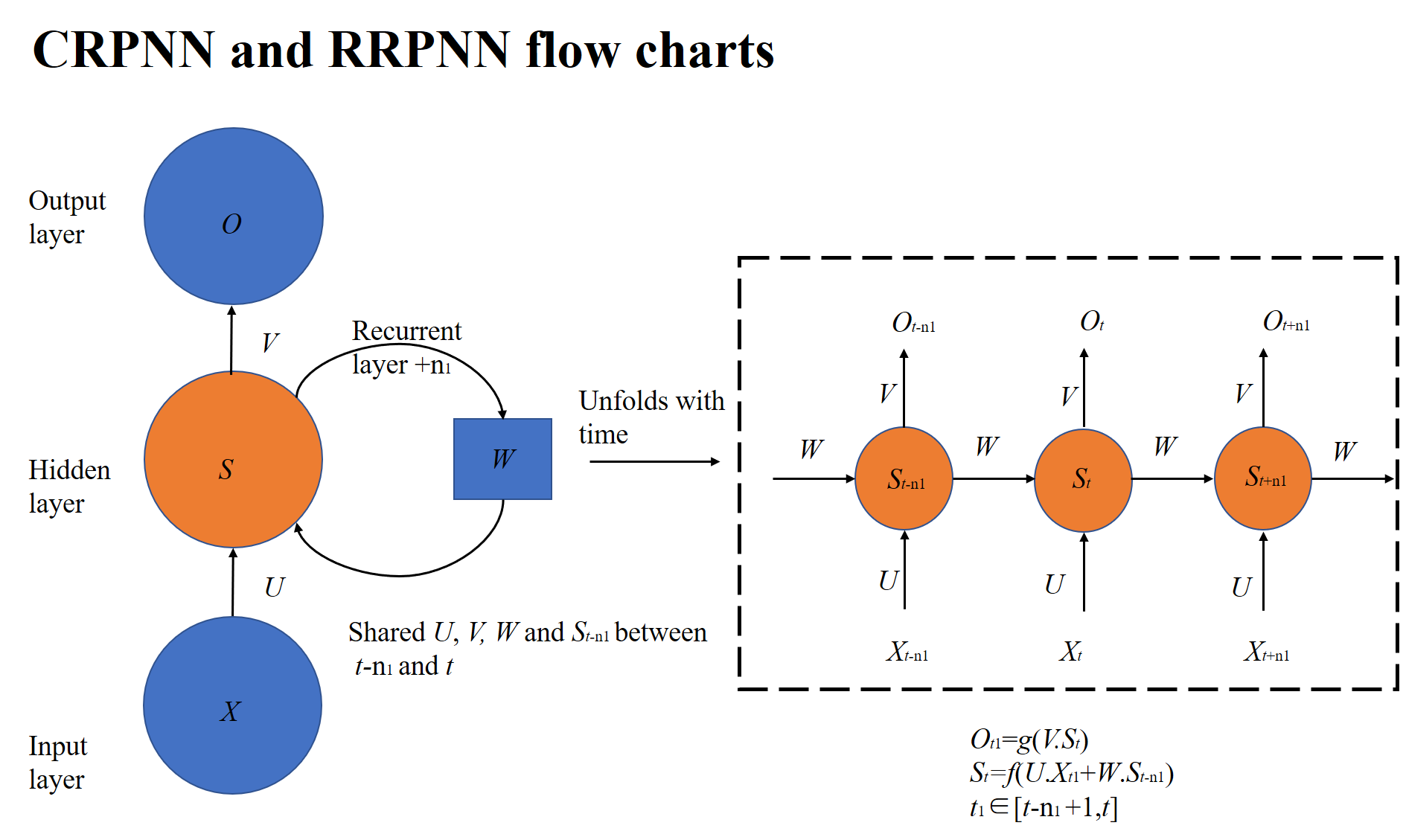}}
\end{figure}

\begin{figure}%
\centering
\subfigure[ORPNN  
Shared weights $U_k$, $V_k$, $W_k$, $R(W,k)$ and $St-n_1 (W,k)$  between $t-n_1 (W,k)$ and $t$, where $R(W,k)$ is the synaptic effective range weight. Astrocytes phagocytose synapses factor $P(W,k)$ realizes synaptic phagocytose, $k$ is taken as $1-{\rm k_{max}}$ to represent different synapses, and ${\rm k_{max}}$ is the size of synapse population. $n_1 (W,k)$ is the synaptic effective range of the variable $W$ and synapse $k$. Take positive or negative values to make $\sum_{k=1}^{k_{max}} P(W,k) =0$ achieve dynamic balance of synapses. The time length of  $\sum_{k=1}^{k_{max}} n_1(W,k) ={\rm l_{max}}$  achieves the static balance of synapses. ${\rm l_{max}}$  represents the sum of synaptic effective ranges. The cortex memory persistence factor $M(W,k)$  gets positive and negative memories in memory engram cells with the best previous gradient information and relatively inferior and good gradient information of synapse formation-$g_{r,P\&N}$, long-term memory. Another memory persistence factor $M(W,k)$ gets positive and negative memories with the relatively inferior and good brain plasticity-$r_P$ and $r_N$, short-term memory, which through quantum computing.]{
\includegraphics[width=0.9\textwidth]{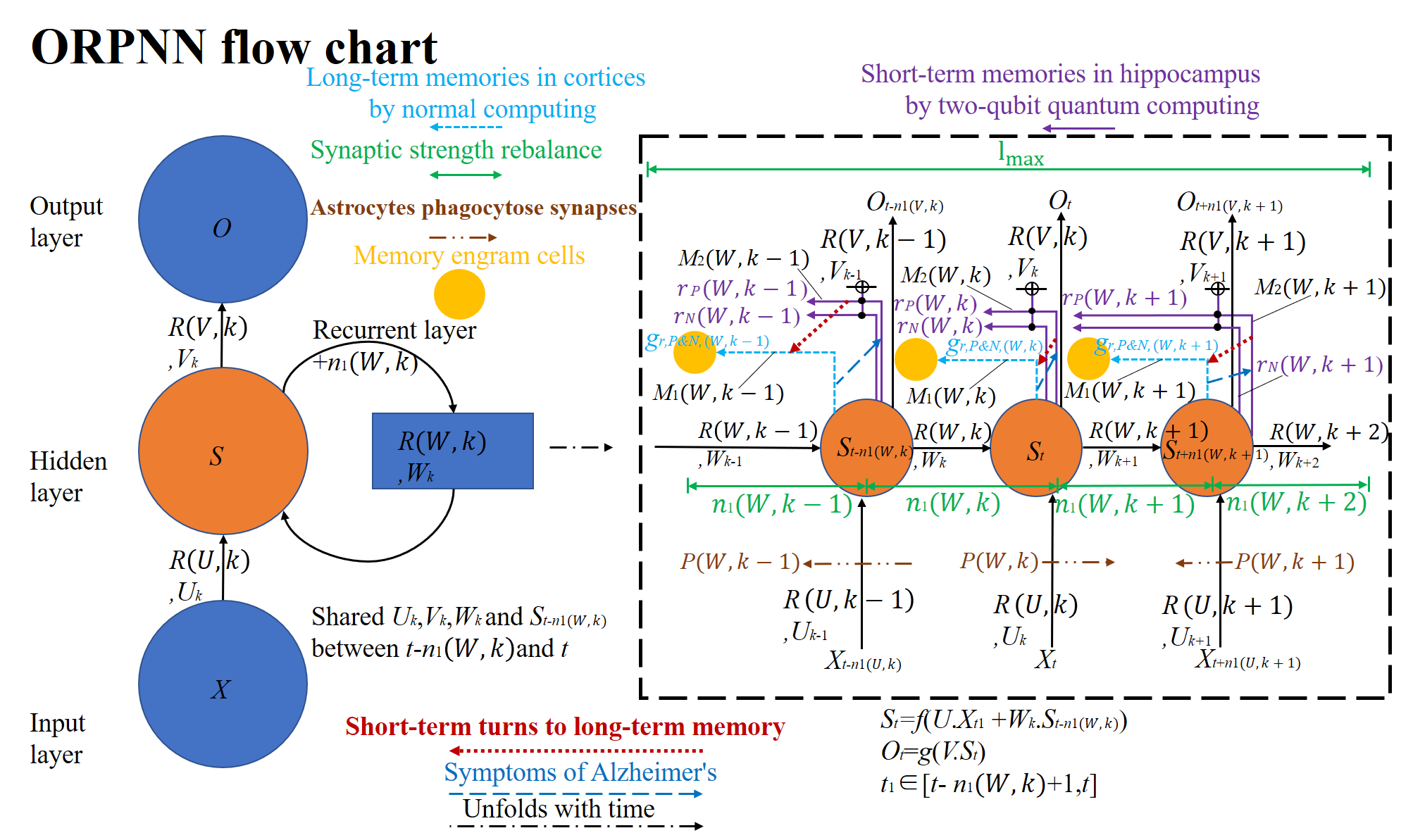}}
\caption{CRPNN, RRPNN and ORPNN flow charts. }
\label{fig21}
\end{figure}

A total of 3 tests were conducted, with the results shown in Fig.\ref{fig2}, in which the results of the variation of the logarithm of the loss of PNN with iterations are also provided.

Result of the first test: the correlation coefficient between CRPNN simulation data and actual data is 0.8057, the correlation coefficient between RRPNN simulation data and real data is 0.8706, and the correlation coefficient between M(P)ORPNN and M(P\&N) ORPNN simulation data and real data are 0.9709 and 0.9687. Result of the second test: the correlation coefficients are 0.8292, 0.8921, 0.9668 and 0.9646 respectively. Result of the third test: the correlation coefficients are 0.8031, 0.8728, 0.9622 and 0.9588 respectively.

The results of the RRPNN test turned out to be satisfactory, and though the connection weights are updated by equation \eqref{eq1} and the memory the best previous solution, more satisfactory optimization results could be found for random synaptic effective range locations.

The ORPNN showed strong fluctuation in the iterations, whereas the final convergence turned out better and faster than the RRPNN and the CRPNN.

\begin{figure}%
\centering
\includegraphics[width=0.9\textwidth]{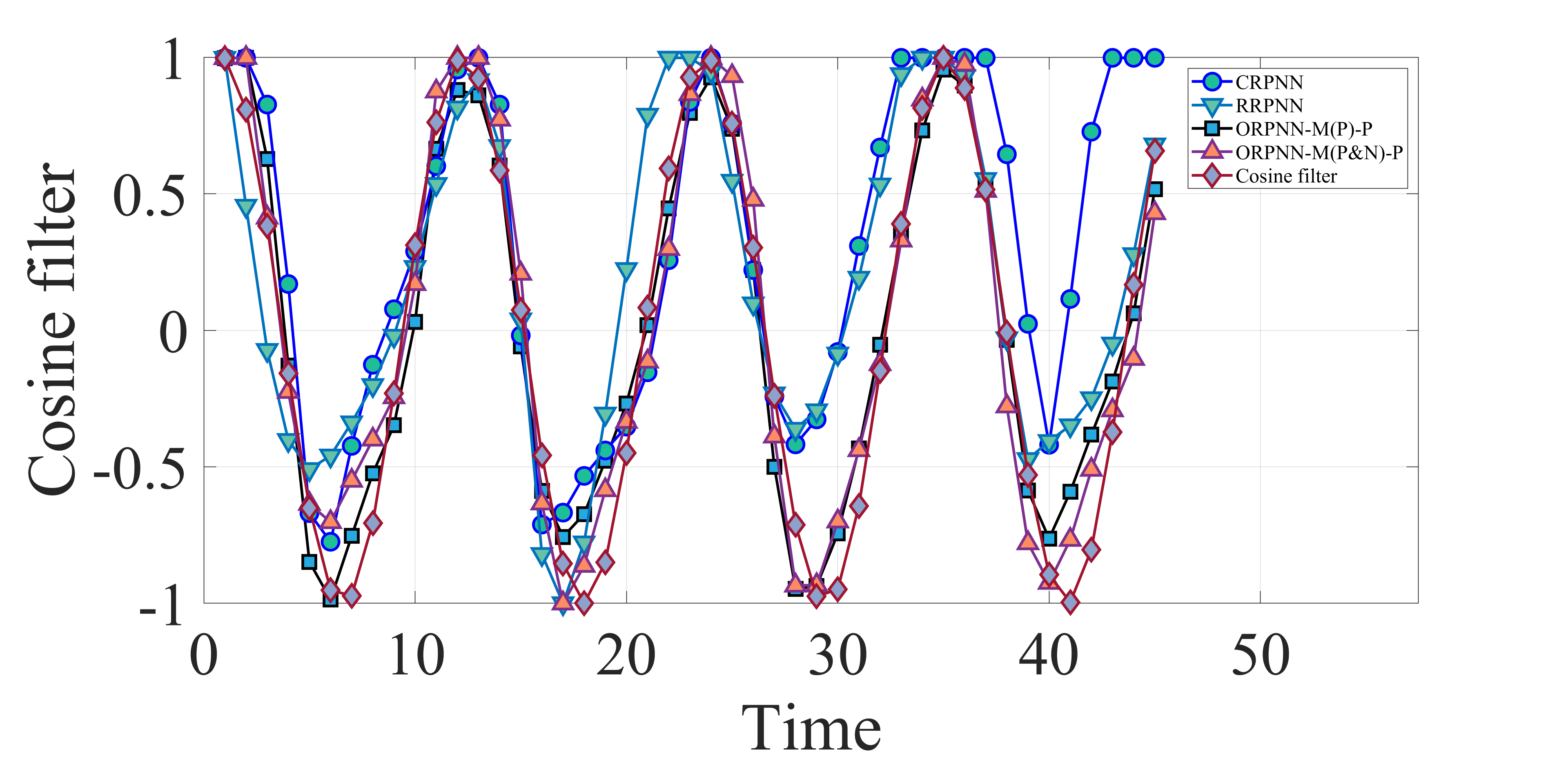}
\includegraphics[width=0.9\textwidth]{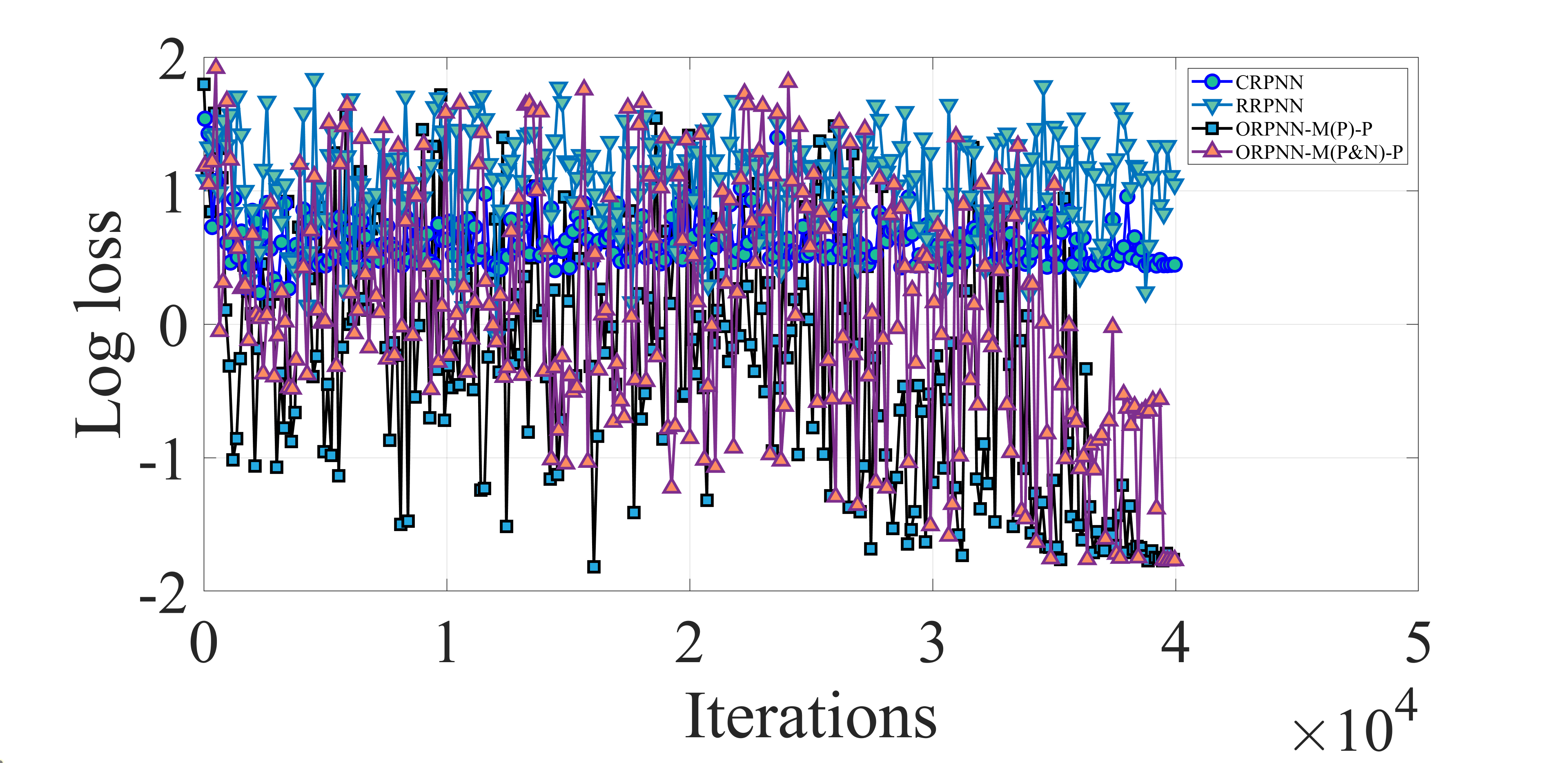}
\caption{first test results belong to CRPNN, RRPNN, and ORPNN respectively}
\label{fig2}
\end{figure}
\begin{figure}%
\centering
\includegraphics[width=0.9\textwidth]{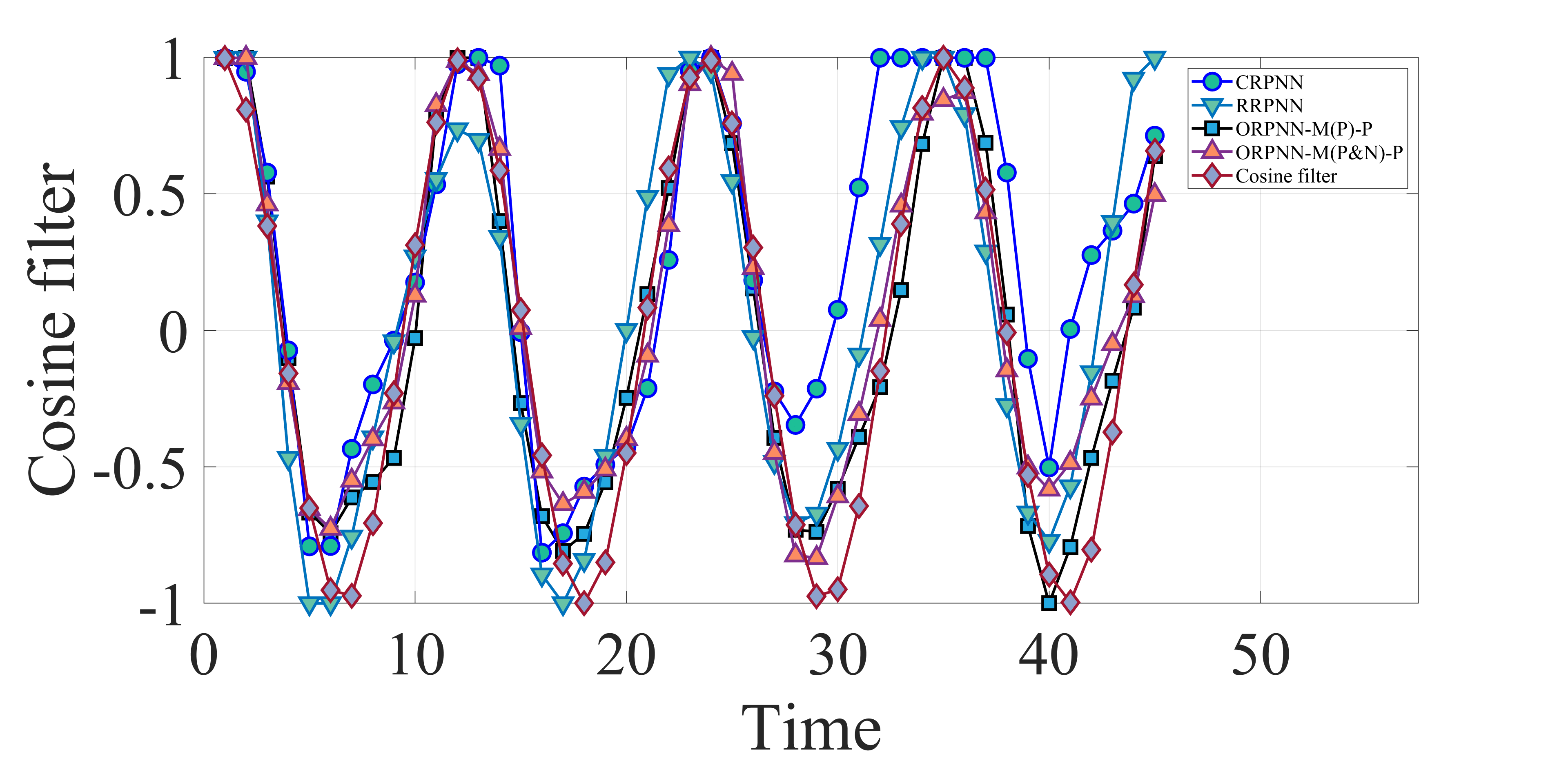}
\includegraphics[width=0.9\textwidth]{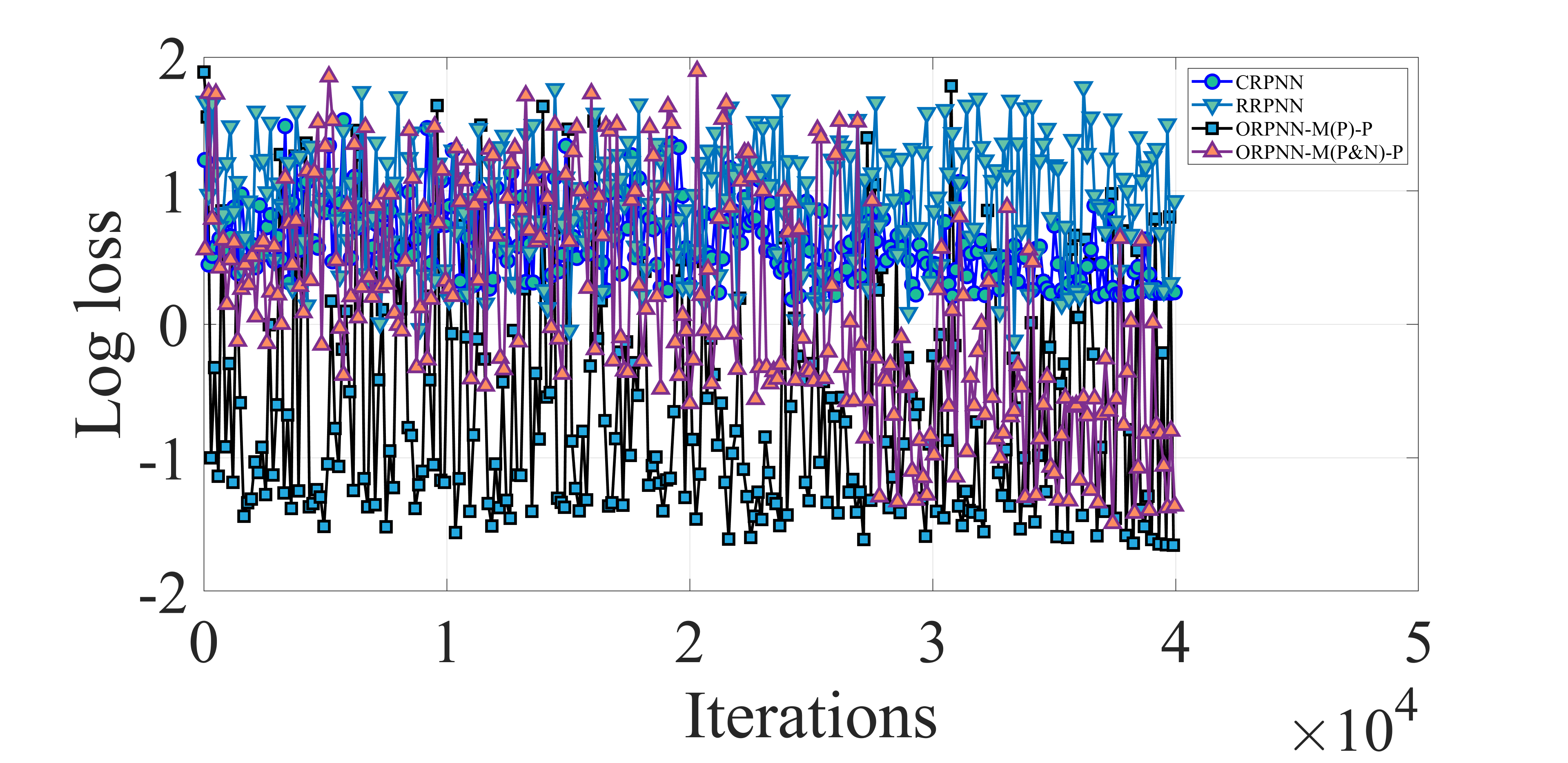}
\caption{second test results belong to CRPNN, RRPNN, and ORPNN respectively}
\label{fig3}
\end{figure}
\begin{figure}%
\centering
\includegraphics[width=0.9\textwidth]{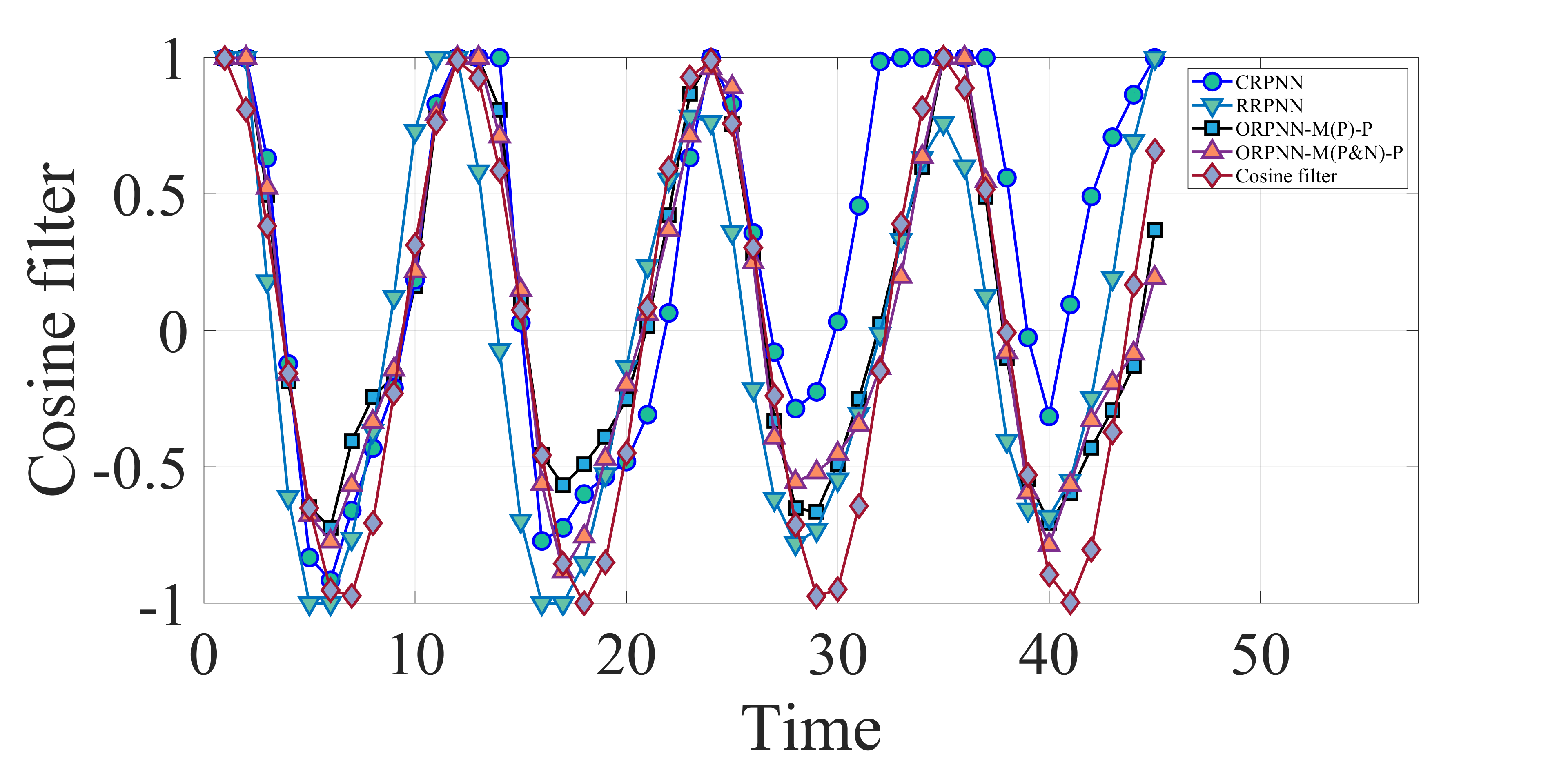}
\includegraphics[width=0.9\textwidth]{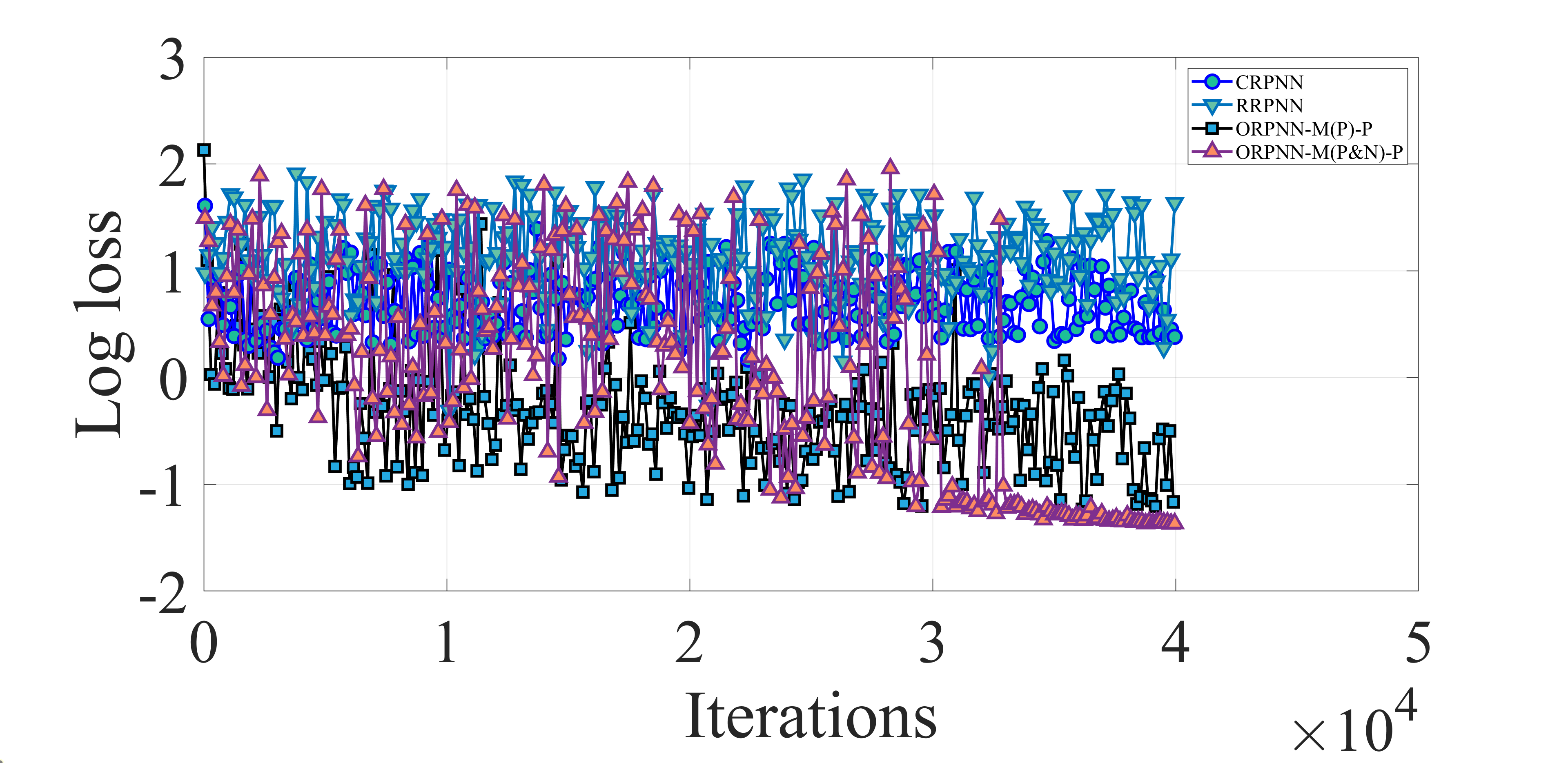}
\caption{third test results belong to CRPNN, RRPNN, and ORPNN respectively}
\label{fig4}
\end{figure}

Iteration of CRPNN is not convergent. RRPNN has slow convergence.

\begin{figure}[h]%
\centering
\includegraphics[width=0.9\textwidth]{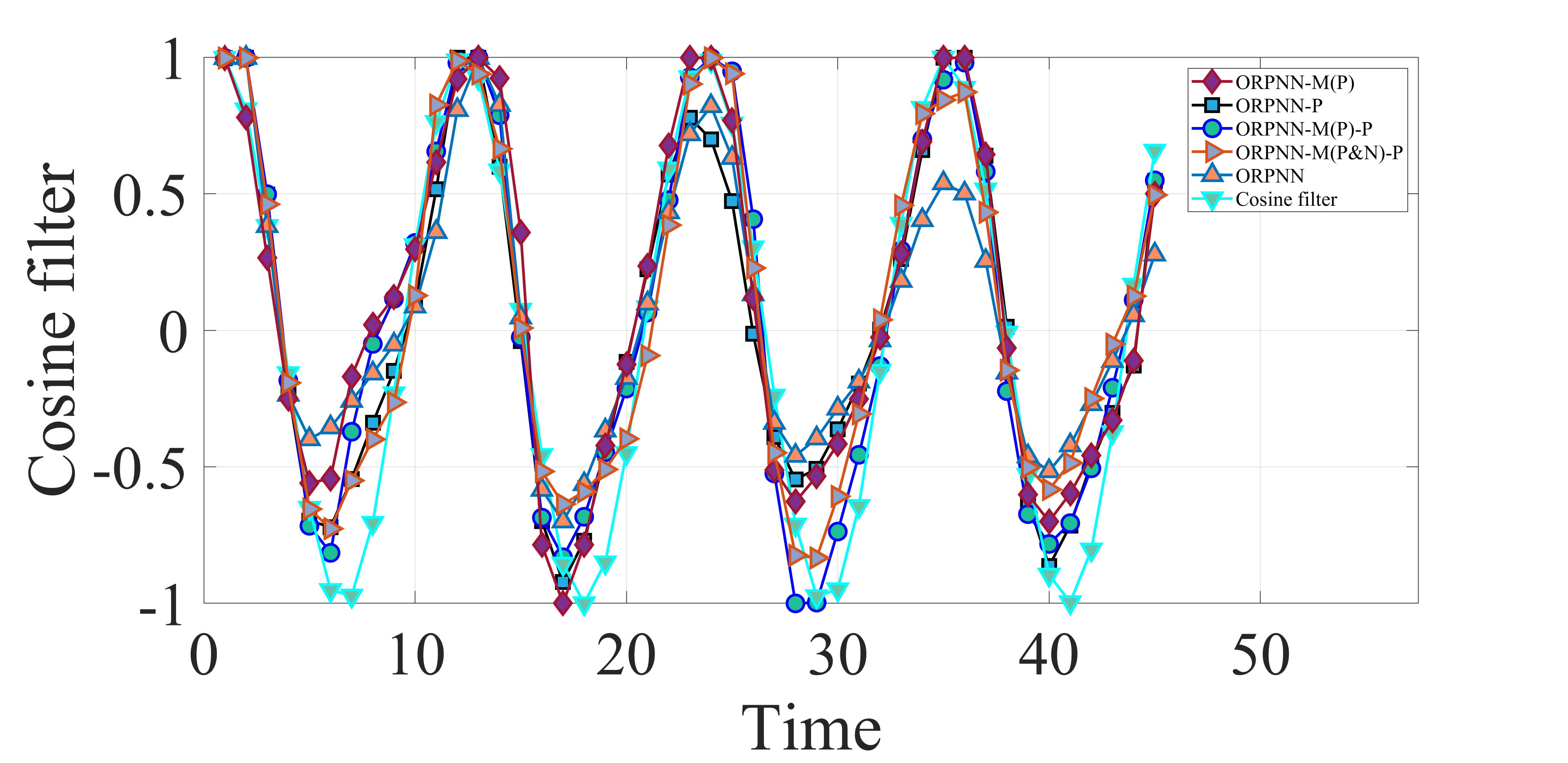}
\includegraphics[width=0.9\textwidth]{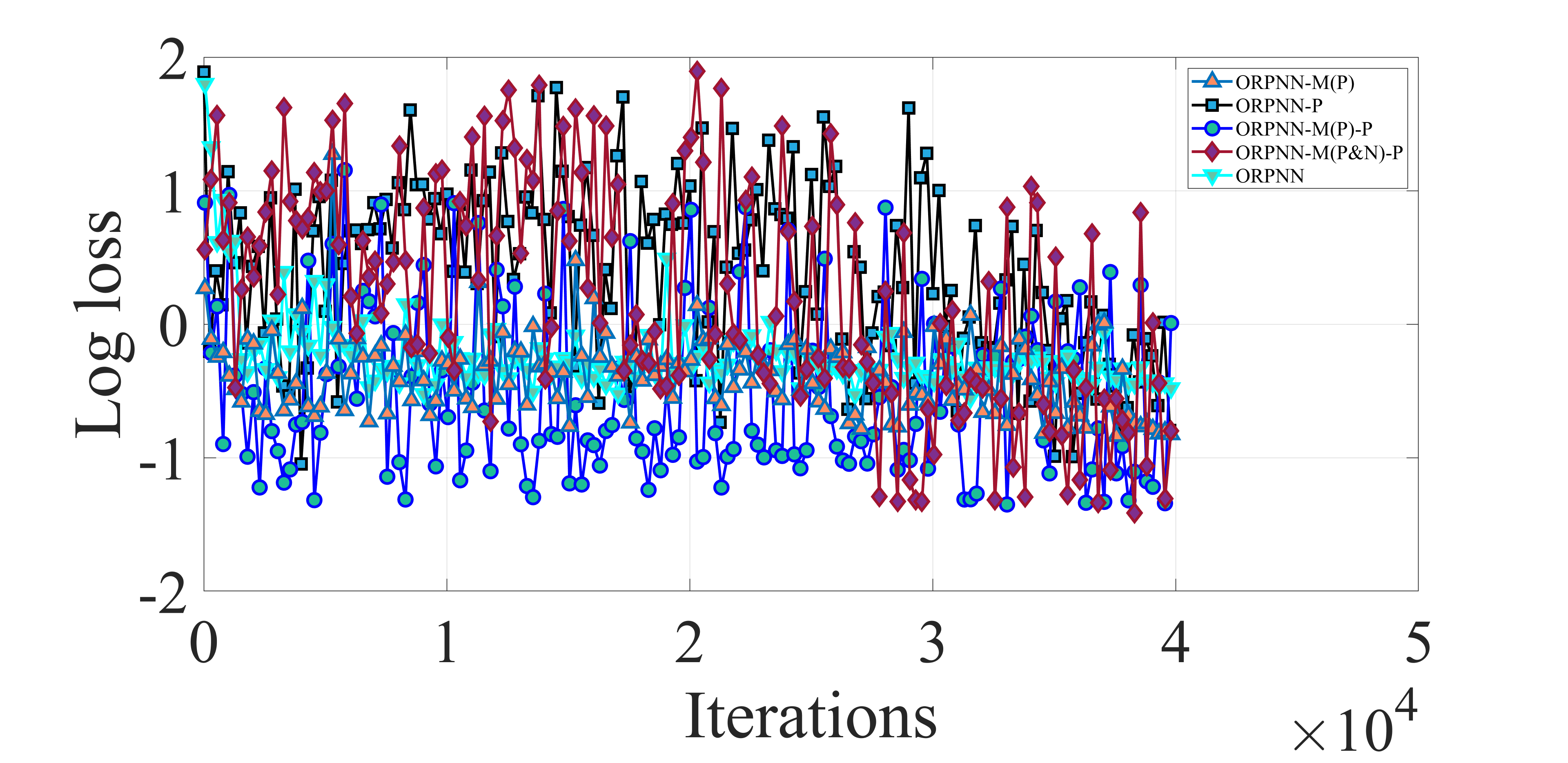}
\caption{Simulations at critical periods and the end of critical periods}
\label{fig5}
\end{figure}

\begin{figure}[h]%
\centering
\includegraphics[width=0.9\textwidth]{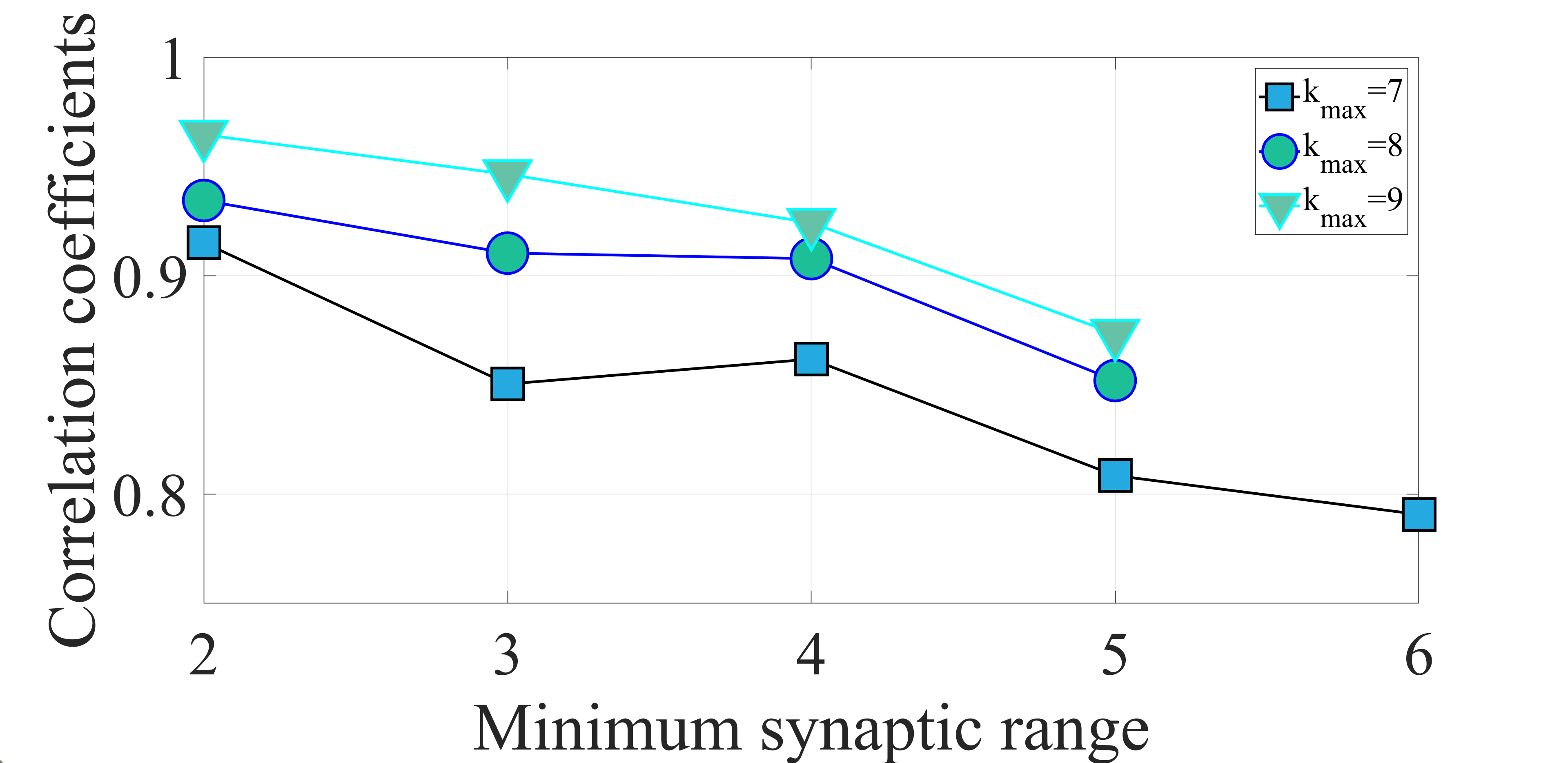}
\includegraphics[width=0.9\textwidth]{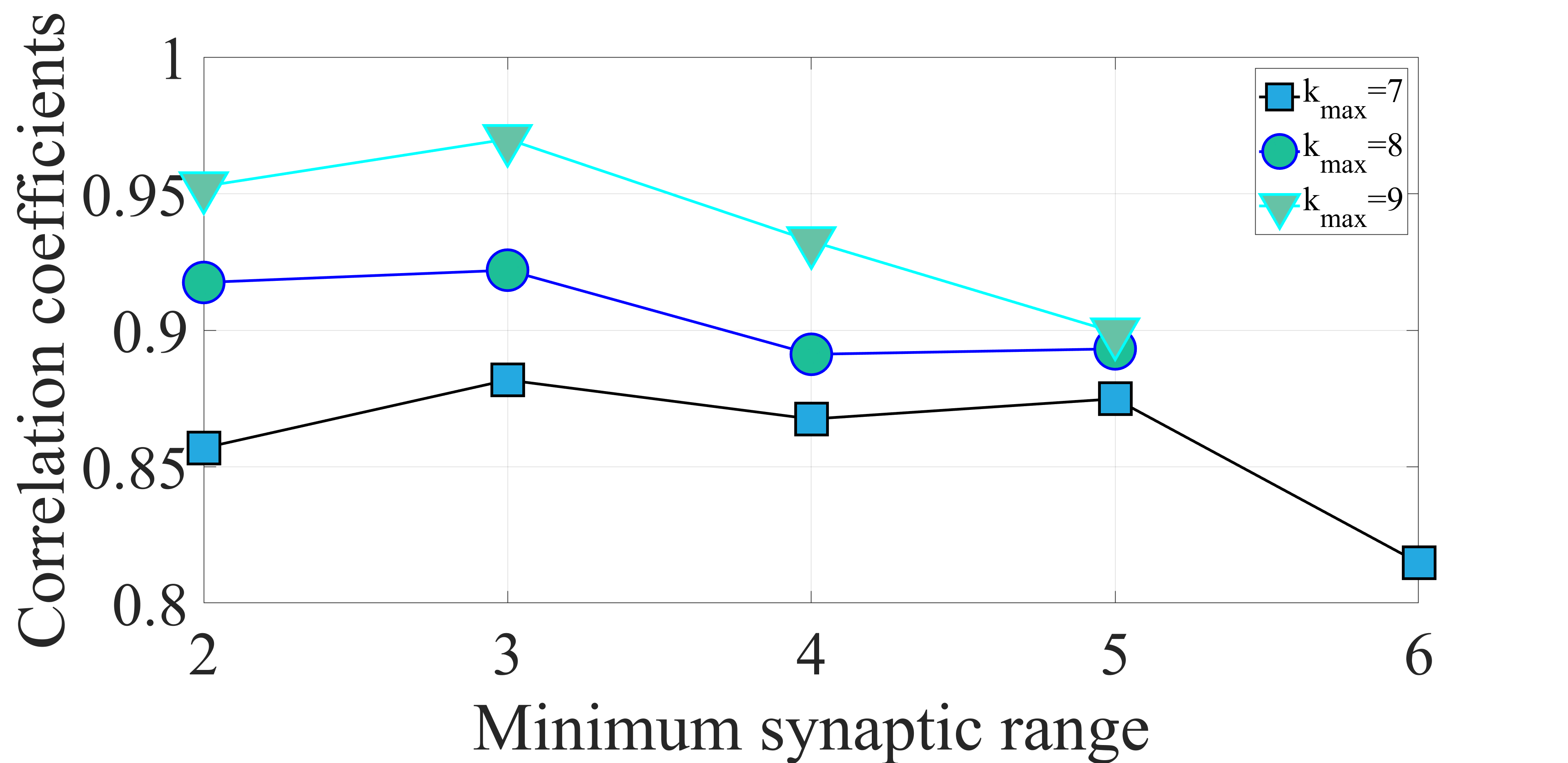}
\caption{Relationship between synapse formation (different ${\rm l_{min}}$ and  ${\rm k_{max}}$)and correlation coefficients:(a) ORPNN-M(P)-PF, (b) ORPNN-PF}
\label{fig6}
\end{figure}

\begin{figure}[h]%
\centering
\includegraphics[width=0.9\textwidth]{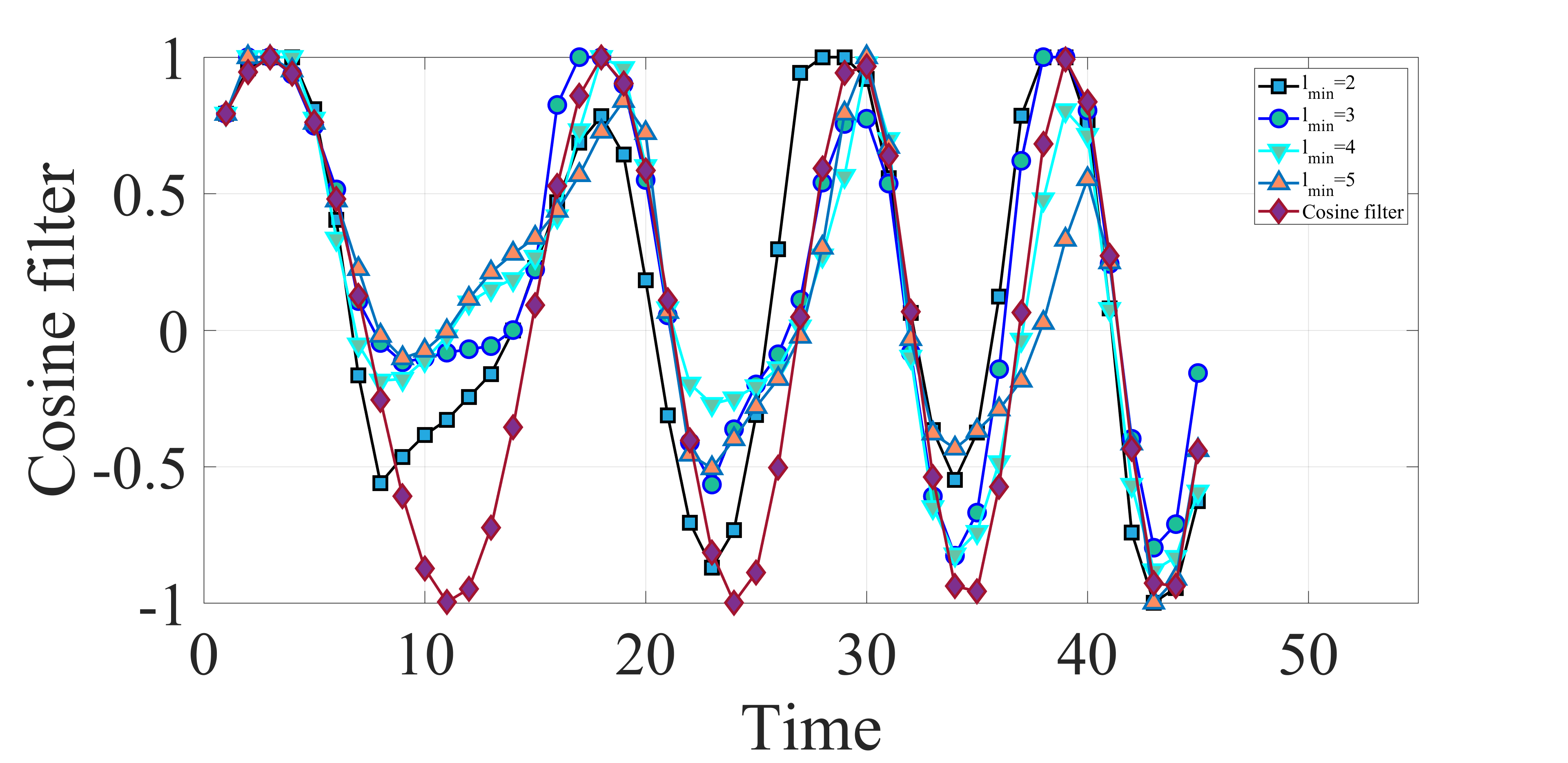}
\includegraphics[width=0.9\textwidth]{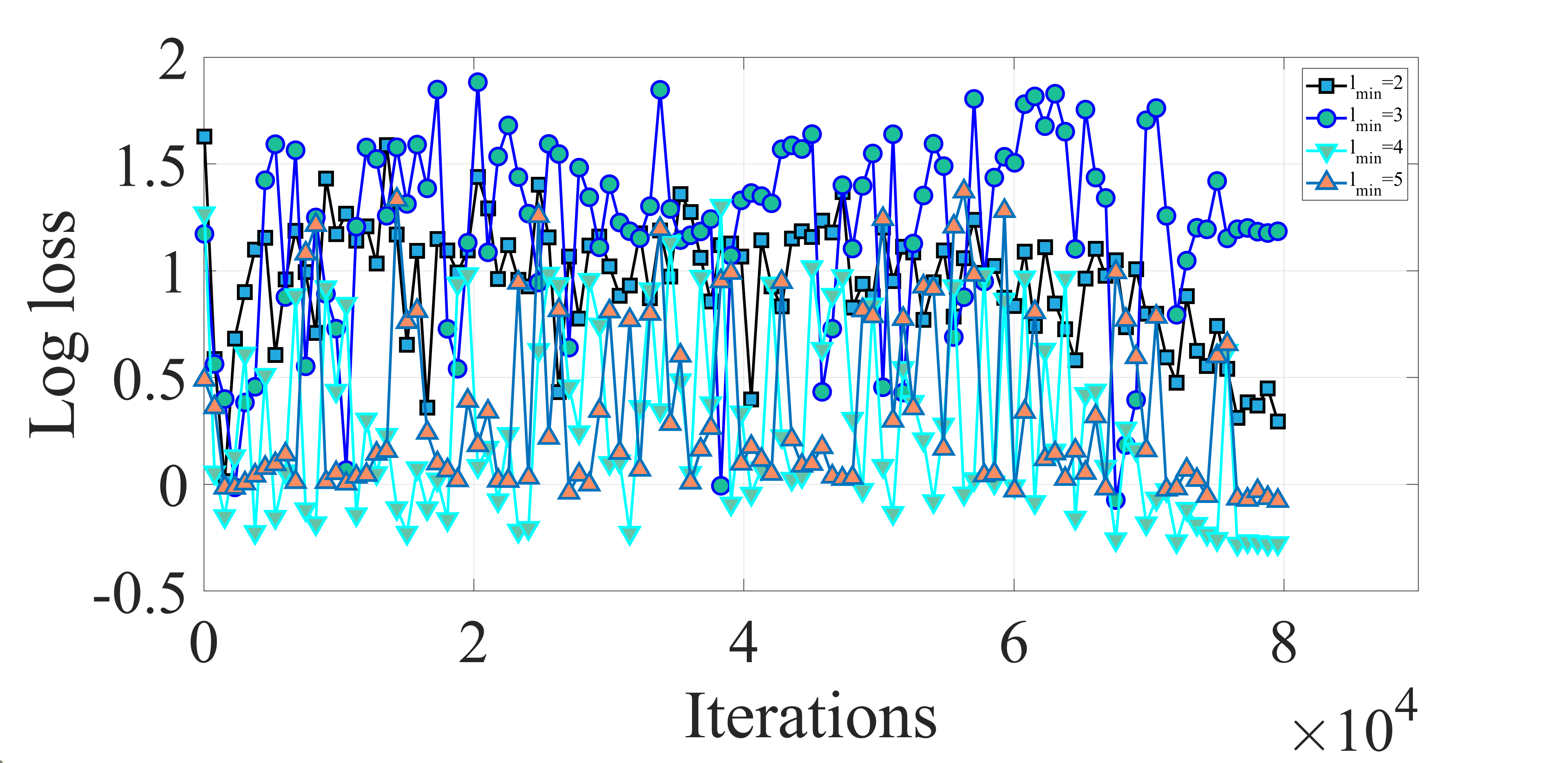}
\caption{ORPNN test results are ${\rm l_{min}}=2-5$ and ${\rm k_{max}}=9$ respectively (a)ORPNN}
\label{fig7}
\end{figure}

\begin{figure}[h]%
\centering
\includegraphics[width=0.9\textwidth]{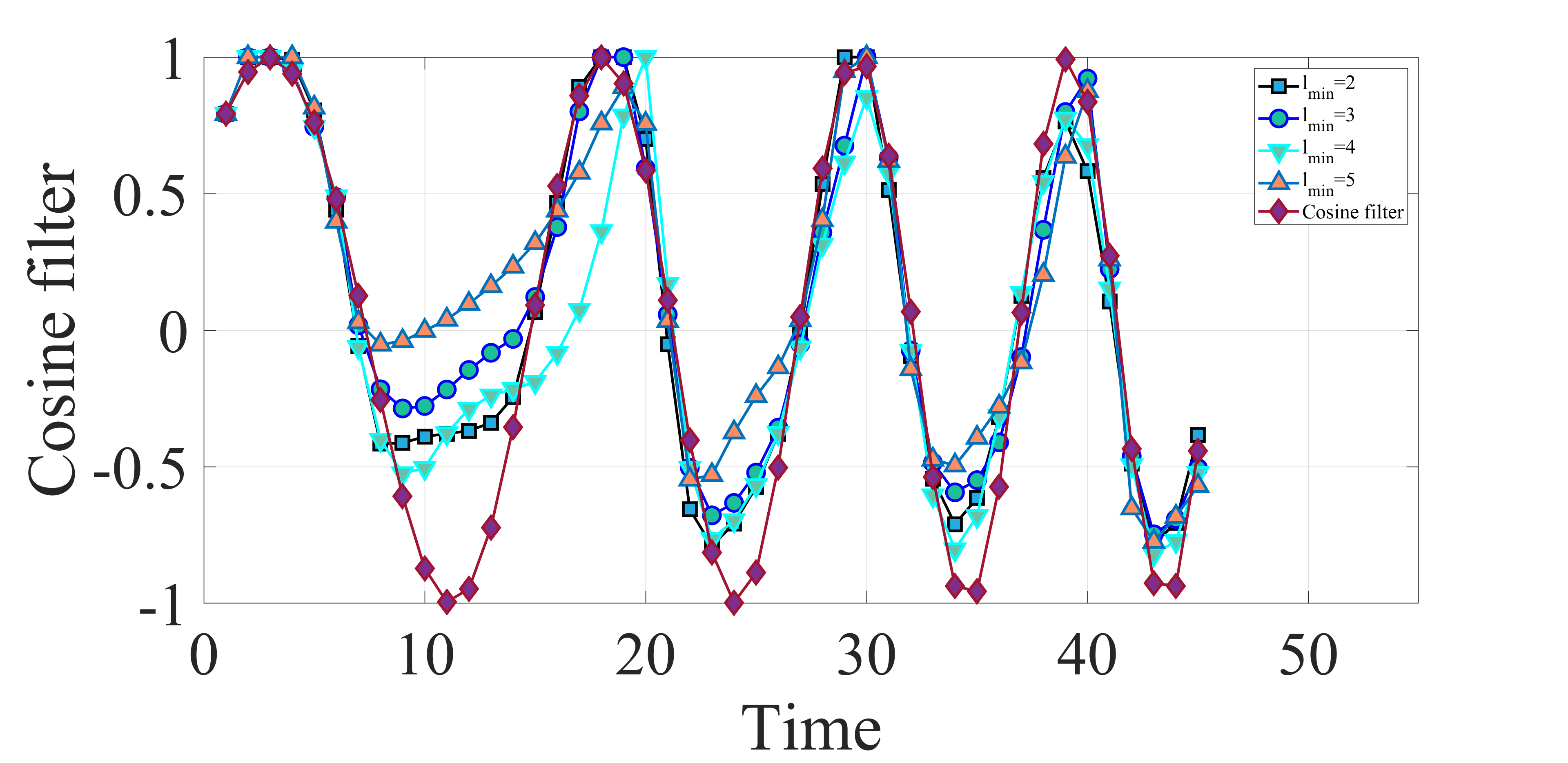}
\includegraphics[width=0.9\textwidth]{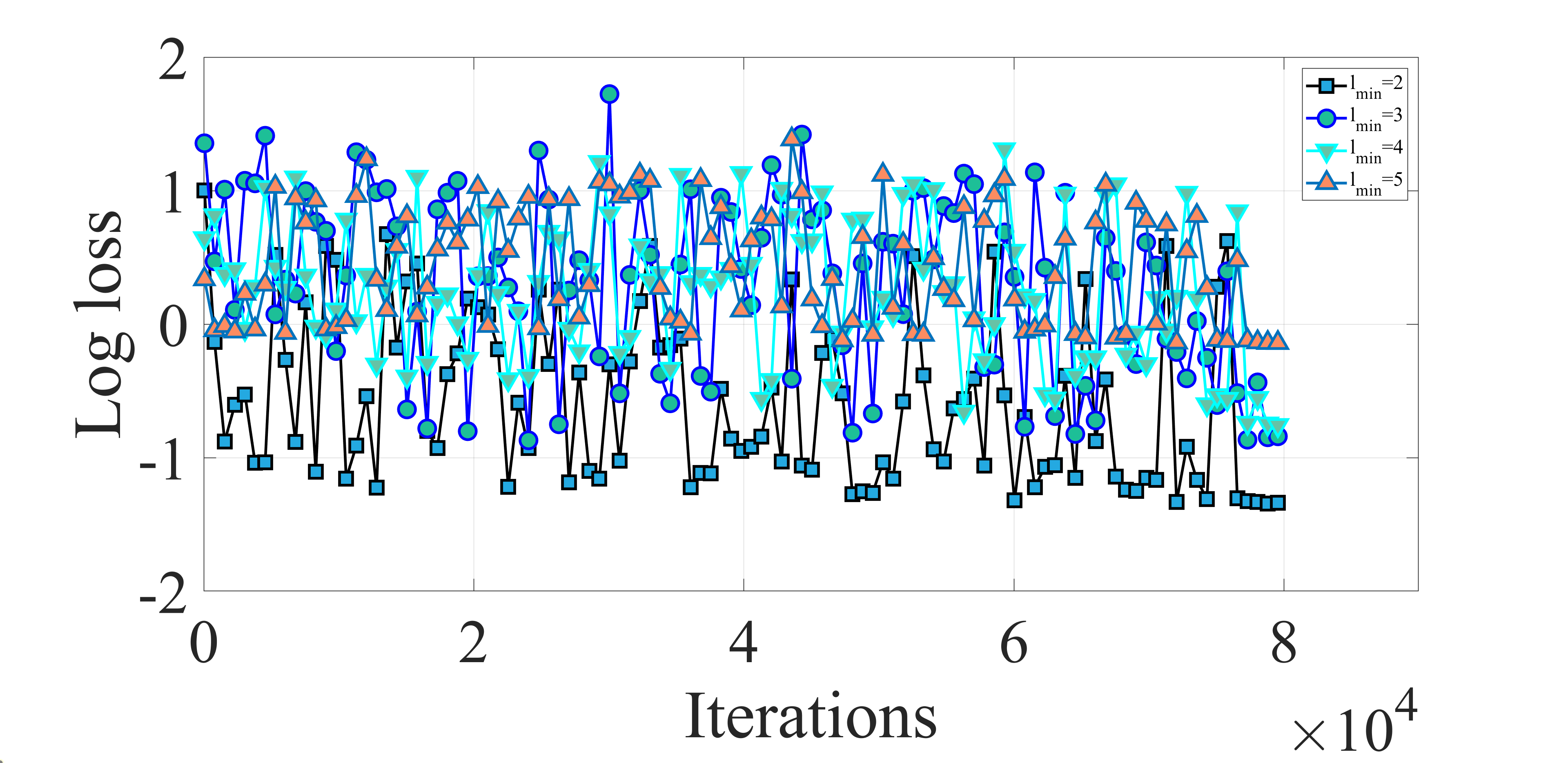}
\caption{ORPNN test results are ${\rm l_{min}}=2-5$ and ${\rm k_{max}}=9$ respectively (b)ORPNN-MF(P)-PF}
\label{fig8}
\end{figure}

\begin{figure}[h]%
\centering
\includegraphics[width=0.9\textwidth]{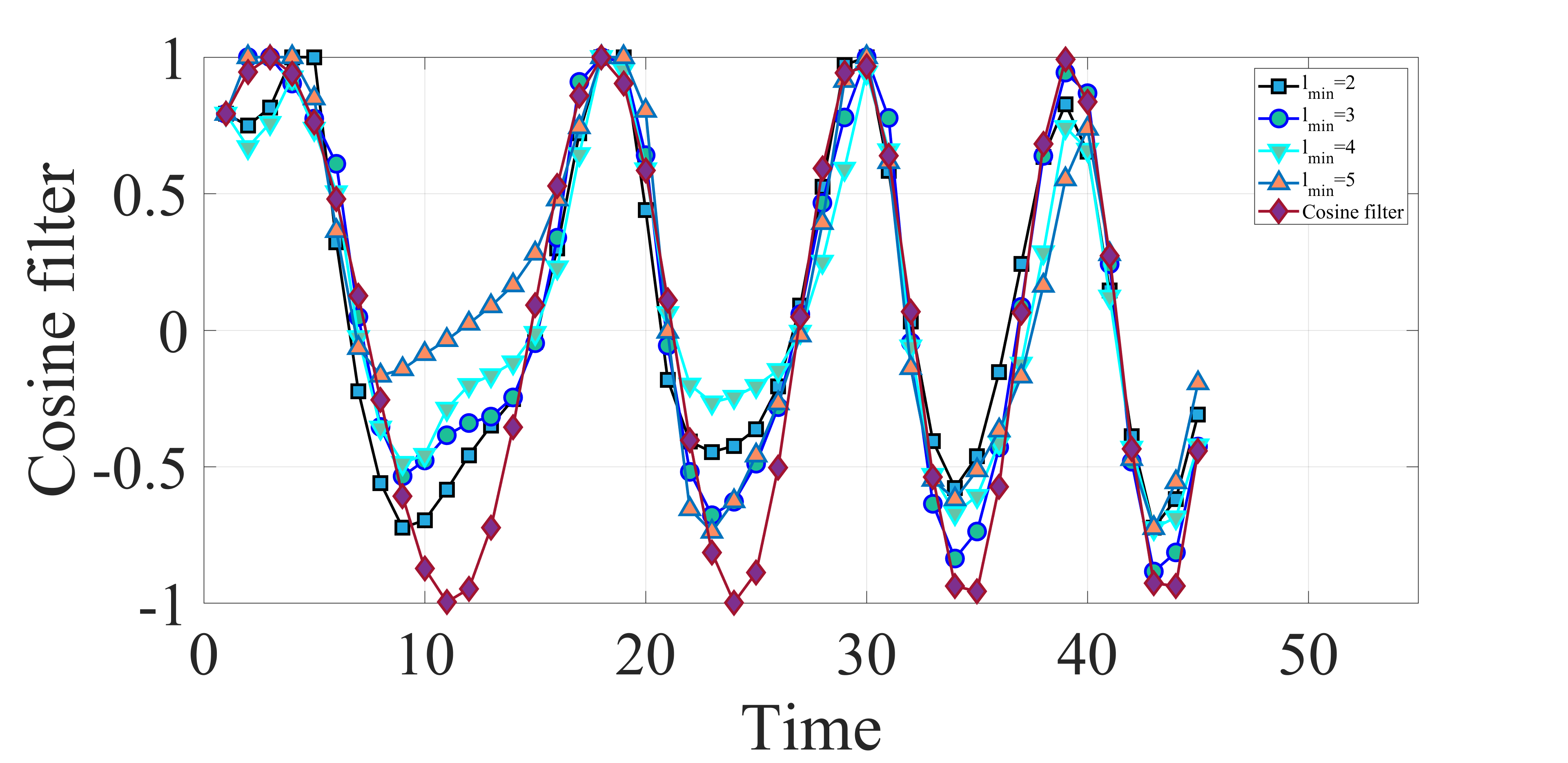}
\includegraphics[width=0.9\textwidth]{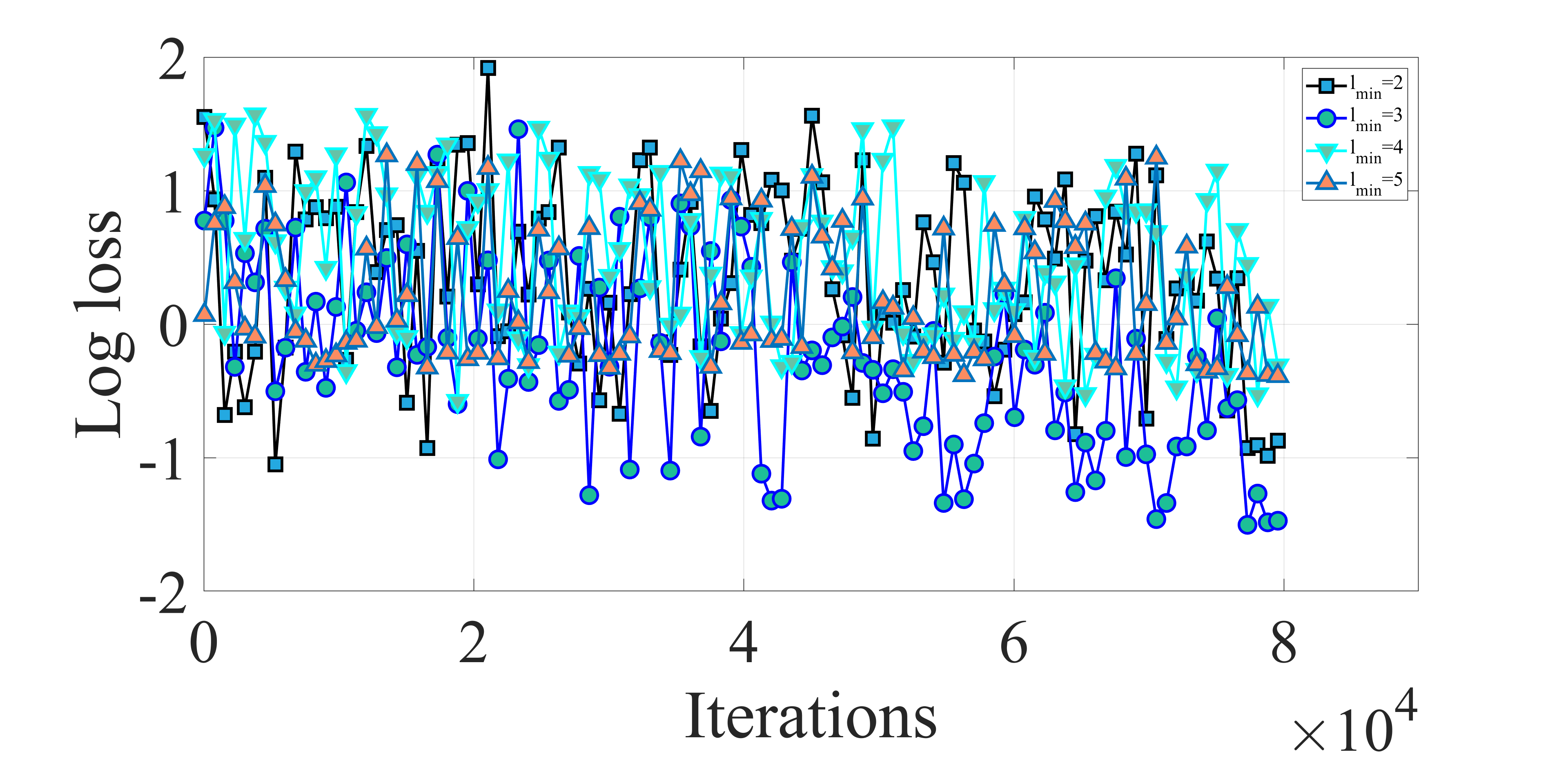}
\caption{ORPNN test results are ${\rm l_{min}}=2-5$ and ${\rm k_{max}}=9$ respectively (c)ORPNN-PF}
\label{fig9}
\end{figure}

\begin{figure}[h]%
\centering
\includegraphics[width=0.9\textwidth]{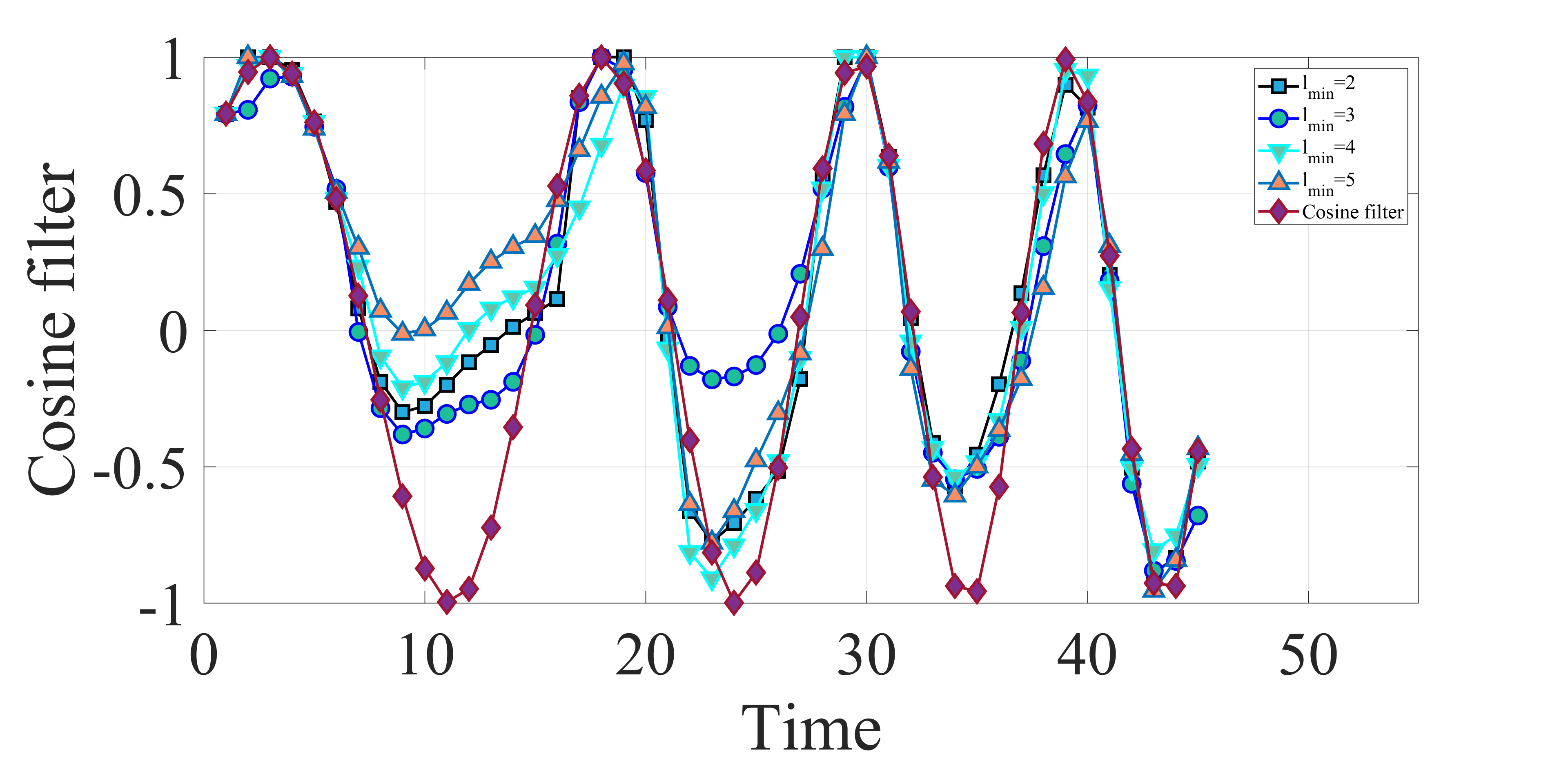}
\includegraphics[width=0.9\textwidth]{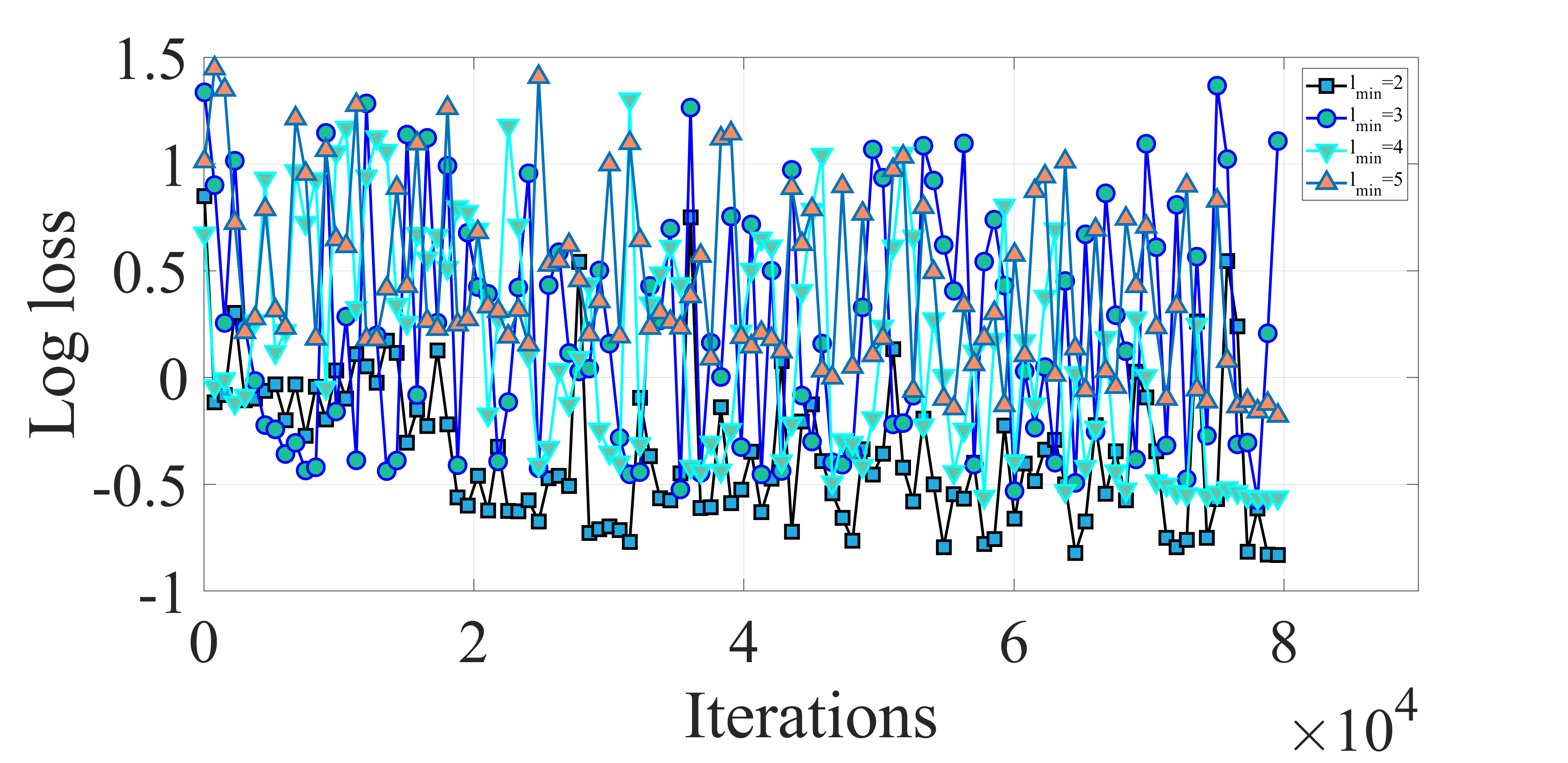}
\caption{ORPNN test results are ${\rm l_{min}}=2-5$ and ${\rm k_{max}}=9$ respectively. (d)ORPNN-MF(P)}
\label{fig10}
\end{figure}

\begin{figure}[h]%
\centering
\includegraphics[width=0.9\textwidth]{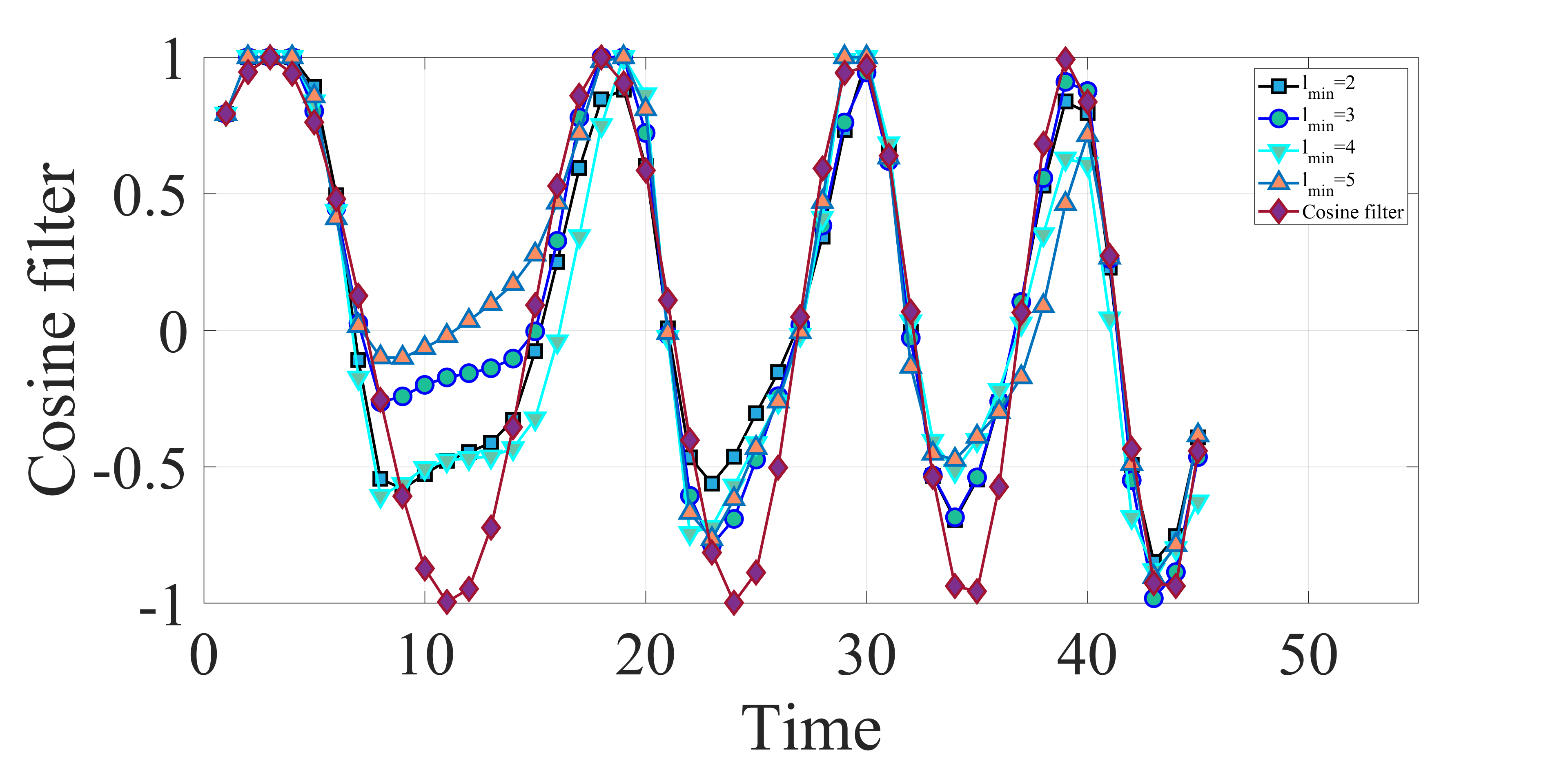}
\includegraphics[width=0.9\textwidth]{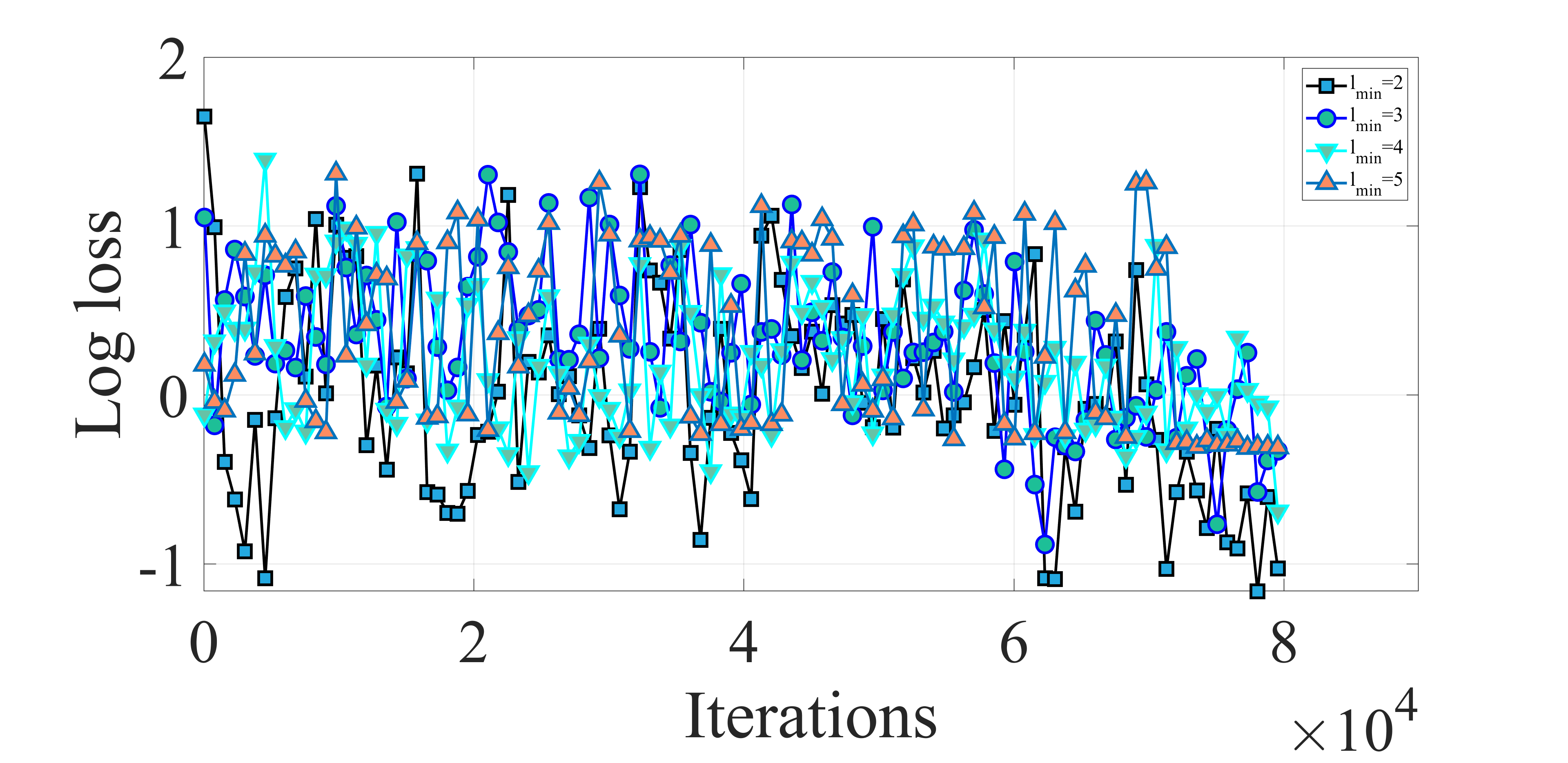}
\caption{ORPNN test results are ${\rm l_{min}}=2-5$ and ${\rm k_{max}}=9$ respectively. (e)ORPNN-MF(P\&N)}
\label{fig11}
\end{figure}

The number of iterations is ${\rm j_{max}}=40000$. In order to study the effect of the astrocytic synapse formation cortex memory persistence factor and astrocytes phagocytose synapses factor, Fig.\ref{fig5} is obtained. In the first scenario of ORPNN-PF, only the astrocytes phagocytose synapses factor is taken into account with the best previous $r(m,k)_{best}$ and a correlation coefficient of 0.9515. The second scenario ORPNN-MF(P)-PF takes into account both the astrocytic synapse formation positive memory persistence factor in formula \eqref{eq2} and astrocytes phagocytose synapses factor in formula \eqref{eq4}, cortex memory persistence factor considers positive memory. It may have better results and a correlation coefficient of 0.9573 at critical period. The third scenario ORPNN excludes the astrocytic synapse formation cortex memory persistence factor and astrocytes phagocytose synapses factor at the closure of critical period then astrocytes become mature. It may have worst results a correlation coefficient of 0.9208. The fourth scenario ORPNN-MF(P) only the astrocytic synapse formation positive cortex memory persistence factor is taken into account with the gradient of the best previous $g(m,k)_{best}$ and a correlation coefficient of 0.9346. The fifth scenario ORPNN-MF(P\&N)-PF takes into account both the astrocytic synapse formation cortex memory persistence factor in formula \eqref{eq2} and astrocytes phagocytose synapses factor in formula \eqref{eq4}, cortex memory persistence factor considers positive and negative memories. It may have best results and a correlation coefficient of 0.9646 at critical period. In Fig.\ref{fig5}, P means astrocytes phagocytose synapses factor, M(P\&N) means positive and negative cortex memories persistence factor, M(P) means positive cortex memory persistence factor.

\subsection{Synapse formation affects brain function - inhibiting dendrites}
We then modified the ORPNN model accordingly for analysis. By modifying ${\rm l_{min}}$,we managed to simulate the process of brain cognition from infancy to the senile phase. ${\rm l_{min}}$ exhibits elasticity at the beginning as a decimal value, then gradually decreases in plasticity. In this way the increase in minimum synaptic effective range from ${\rm l_{min}}=2$ to ${\rm l_{max}}/{\rm k_{max}} \approx 5$ made diversity of synaptic effective range progressively smaller. CRPNN corresponds to the senile phase period of brain cognition as it satisfies ${\rm l_{min}}= {\rm l_{max}}/{\rm k_{max}}=44/9\approx5$, and the funny thing is that the iteration is not convergent at CRPNN.

We obtained a possible explanation to the research findings of Dr. Luo's team through PNN. We drew the relationship of different ${\rm l_{min}}$ (minimum synaptic effective range), different ${\rm k_{max}}$ (synapse population) and correlation coefficients, and the results are shown in Fig. \ref{fig6} below. We also conducted another test, in which the cosine filter and cycle gradually diminish by time -variable cycle cosine filter, and the number of iterations is ${\rm j_{max}}=80000$. The functions of dendrites are to receive and process signals from other neurons. Our simulation results reveal that synapse formation will inhibit dendrites to a certain extent, furthering the findings of Dr. Luo's team. An explanation to this is that synaptic growth will lead to a reduction in the changes in synaptic effective range, and an overgrowth will inevitably disrupt the diversity and plasticity of human brain. ${\rm l_{min}}=2-5, {\rm k_{max}}=7-9$. Fig.\ref{fig6} concerns the relationship between synapse formation and correlation coefficients in ORPNN. The synapse formation could involve a decrease in ${\rm k_{max}}$ and an increase in ${\rm l_{min}}$. The PNN situation can be thought as the Finite Element Method, in which neurons of the brain can be regarded as nodes of FEM. As increased in the number of nodes and finer meshes can lead to more accurate computational results of FEM, more neurons and smaller minimum synaptic effective range can also help us obtain more desirable and accurate PNN results. When synapse population ${\rm k_{max}}$ decreases and sum of synaptic effective ranges ${\rm l_{max}}$ is a constant, the diversity of local synaptic effective range may be improved as the synaptic effective range enlarges from ${\rm l_{min}}$ to ${\rm l_{max}}/{\rm k_{max}}$, so ${\rm k_{max}}$ decreases may enhance correlation coefficients by accident, as you can observe from Fig. \ref{fig7} (a) ${\rm l_{min}}=3-4$, ${\rm k_{max}}=7-8$ and Fig. \ref{fig7} (b) ${\rm l_{min}}=2-3$, ${\rm k_{max}}=7-9$. When the minimum synapse length is too small, the probability of local synaptic accumulation increases, which also affects the optimization results, It can be seen from Table \ref{table2}.

\subsection{Effects of astrocytes are reflected in astrocytes phagocytose synapses factor and cortex memory persistence factor during critical and closure critical period}
As for the ORPNN-MF(P)-PF situation (ORPNN both containing astrocytic synapse formation positive cortex memory persistence factor and astrocytes phagocytose synapses factor at critical period), synaptic effective ranges with different ${\rm l_{min}}$ and different input variables are provided in Table \ref{table1}. When ${\rm l_{min}}=5$, the synapse grows to a certain extent, and synaptic effective range remains almost the same, thus leading to a worsened simulation result. When the synapse grows, standard deviation of synaptic effective range at the end of the iteration diminishes, and the correlation coefficients decreases to a certain extent when ${\rm l_{min}}=5$. From  ${\rm l_{min}}=2$ to ${\rm l_{min}}=5$, synaptic effective range gradually loses its diversity and plasticity. The number of iterations is ${\rm j_{max}}=80000$. And the results of synaptic effective range and correlation coefficient when iteration equals 20,000, 40,000, 60,000 and 80,000 were also observed, as 4444 in Table represents the synaptic effective ranges are all 4, and 0.9641, 0.9683, 0.9683 and 0.9645 are the correlation coefficients when iteration equals 20,000, 40,000, 60,000 and 80,000, respectively.

As for the ORPNN-PF situation (ORPNN only containing astrocytes phagocytose synapses factor), synaptic effective ranges with different ${\rm l_{min}}$ and different input variables are provided in Table \ref{table2}.  

Comparing with the other Tables, the effect of astrocyte made local synaptic effective range remain in an appropriate length at critical period, as one can observe in Tables \ref{table1}, \ref{table3}, \ref{table4} and \ref{table5}. Else some synapses have longer length and disorder, resulting in the alignment of other synaptic effective ranges, as one can observe in Table \ref{table2}. The simulations in Tables similar to that failure to the closure of critical period results in neurodevelopmental disorders, and then astrocytes become mature \cite{bib19}. The calculation of the correlation coefficient was better than that of Tables \ref{table2}, \ref{table4} and \ref{table5} but worse than Table \ref{table3} at critical period, as model of Deep Learning contains both astrocytic synapse formation cortex memory persistence factor and astrocytes phagocytose synapses factor in Table \ref{table1} and Fig.\ref{fig8}(b).

Table \ref{table2} shows that the local accumulation of synapse length is accompanied by Vanishing Gradients by Fig.\ref{fig7} and Tables. It can be seen from Table \ref{table2} that activity of synapse formation gradually diminishes with iterations and growth of the ${\rm l_{min}}$, the calculation of the correlation coefficient in Table \ref{table2} also turns out worse than in other Tables.

For Table \ref{table3} which considers only astrocytes phagocytose synapses factor, synapse formation showed good activity with iterations and ${\rm l_{min}}$ growth. We can find better results of correlation coefficients in Tables respectively. As the probability of Vanishing Gradients is small when considering astrocytes phagocytose synapses factor by Fig.\ref{fig9}(c) and Tables, the result of Table \ref{table3} converges even at ${\rm l_{min}}=5$, while that of other Tables converge poorly at ${\rm l_{min}}=5$.

Table \ref{table4} only deals with the cortex positive memory persistence factor, which maintains the slow change in synapse length, and the synapse formation shows a worse activity with iterations and minimum synaptic effective range growth than in Tables \ref{table1}-\ref{table3} and \ref{table5}. The Vanishing Gradients of the positive cortex memory persistence factor is worse than that of Tables \ref{table1}, \ref{table3} and \ref{table5}, and better than that of Table \ref{table2} by Fig.\ref{fig10}(d) and Tables.

Table \ref{table5} deals with the positive and negative cortex memories persistence factor, which maintains the active change of synapse length, and the synapse formation shows a better activity with iterations and minimum synaptic effective range growth than in Tables \ref{table1} and \ref{table3}. The Vanishing Gradients of the positive and negative cortex memories persistence factor is worse than that of Tables \ref{table1} and \ref{table3}, and better than that of Table \ref{table2} and \ref{table4} by Fig.\ref{fig11}(e) and Tables.

When ${\rm l_{min}}\approx{\rm l_{max}}/{\rm k_{max}}=44/9\approx5$, similar to the case of CRPNN, it can be seen from Tables \ref{table1}-\ref{table5} that the correlation coefficient effect is not so good at this time, and even Tables \ref{table1}, \ref{table2}, \ref{table5} is less ideal.

ORPNN contains both astrocytic synapse formation positive cortex memory persistence factor and astrocytes phagocytose synapses factor-ORPNN-MF(P)-PF. We propose the simulation of ${\rm l_{min}}=2-5$ and  ${\rm k_{max}}=9$respectively, and the correlation coefficients are 0.9645, 0.9464, 0.9242 and 0.8734 respectively. ORPNN contains neither astrocytic synapse formation cortex memory persistence factor nor astrocytose phagocytose synapses factor at the closure of critical period-ORPNN, and the correlation coefficients are 0.8730, 0.9282, 0.8846 and 0.8579 respectively. ORPNN only contains astrocytes phagocytose synapses factor-ORPNN-PF, and the correlation coefficients are 0.9527, 0.9700, 0.9326 and 0.8992 respectively. ORPNN only contains astrocytic synapse formation positive cortex memory persistence factor with positive memory - ORPNN-MF(P), and the correlation coefficients are 0.9382, 0.9265, 0.9115 and 0.8663 respectively. ORPNN only contains astrocytic synapse formation cortex memory persistence factor with negative and positive memories-ORPNN-MF(P\&N), and the correlation coefficients are 0.9592, 0.9423, 0.9228 and 0.8878 respectively. Different ${\rm l_{min}}$ affects simulations can be observed in Fig.\ref{fig7}-\ref{fig11}. The number of iterations is ${\rm j_{max}}=80000$.

\begin{table}[]
\caption{Synaptic effective range with different ${\rm l_{min}}$ and different variables-brain plasticity ORPNN-MF(P)-PF, test results are ${\rm l_{min}=2-5}$ and ${\rm k_{max}=9}$ respectively}
\label{table1}
\resizebox{\textwidth}{!}{
\begin{tabular}{|c|c|c|c|c|c|c|c|c|c|c|c|c|}
        \hline
        & \diagbox{V.L.}{S.L.} & S.1 & S.2 & S.3 & S.4 & S.5 & S.6 & S.7 & S.8 & S.9 & Std    & correlation coefficients\\
        \hline
                        \multirow{4}{*}{${\rm l_{min}}=2$} & V.1                                                                     & \textcolor{green}55\textcolor{green}55   & \textcolor{green}77\textcolor{green}77   & \textcolor{green}44\textcolor{green}44   & \textcolor{green}66\textcolor{green}66   & \textcolor{green}44\textcolor{green}55   & \textcolor{green}44\textcolor{green}44   & \textcolor{green}44\textcolor{green}44   & \textcolor{green}66\textcolor{green}66   & \textcolor{green}44\textcolor{green}33   & 1.2693 & 0.9641 \\
                        & V.2                                                                     & \textcolor{green}55\textcolor{green}45   & \textcolor{green}55\textcolor{green}55   & \textcolor{green}55\textcolor{green}55   & \textcolor{green}44\textcolor{green}44   & \textcolor{green}55\textcolor{green}55   & \textcolor{green}55\textcolor{green}55   & \textcolor{green}66\textcolor{green}66   & \textcolor{green}55\textcolor{green}55   & \textcolor{green}44\textcolor{green}54   & 0.6009 & 0.9683                        \\
                        & V.3                                                                     & \textcolor{green}45\textcolor{green}55   & \textcolor{green}66\textcolor{green}66   & \textcolor{green}44\textcolor{green}44   & \textcolor{green}34\textcolor{green}33   & \textcolor{green}44\textcolor{green}44   & \textcolor{green}55\textcolor{green}55   & \textcolor{green}44\textcolor{green}44   & \textcolor{green}66\textcolor{green}66   & \textcolor{green}86\textcolor{green}77   & 1.2693 & 0.9863                        \\
                        & V.4                                                                     & \textcolor{green}66\textcolor{green}66   & \textcolor{green}44\textcolor{green}44   & \textcolor{green}54\textcolor{green}44   & \textcolor{green}66\textcolor{green}66   & \textcolor{green}44\textcolor{green}44   & \textcolor{green}44\textcolor{green}44   & \textcolor{green}76\textcolor{green}66   & \textcolor{green}66\textcolor{green}56   &\textcolor{green}24\textcolor{green}54   & 1.0541 & 0.9645                        \\
\hline
                        \multirow{4}{*}{${\rm l_{min}}=3$} & V.1                                                                     & \textcolor{green}55\textcolor{green}55   & \textcolor{green}44\textcolor{green}44   & \textcolor{green}44\textcolor{green}44   & \textcolor{green}55\textcolor{green}55   & \textcolor{green}66\textcolor{green}66   & \textcolor{green}66\textcolor{green}66   & \textcolor{green}55\textcolor{green}55   & \textcolor{green}55\textcolor{green}55   & \textcolor{green}44\textcolor{green}44   & 0.7817 & 0.9409 \\
                        & V.2                                                                     & \textcolor{green}55\textcolor{green}55   & \textcolor{green}66\textcolor{green}66   & \textcolor{green}55\textcolor{green}55   & \textcolor{green}55\textcolor{green}55   & \textcolor{green}55\textcolor{green}55   & \textcolor{green}44\textcolor{green}44   & \textcolor{green}55\textcolor{green}55   & \textcolor{green}55\textcolor{green}55   & \textcolor{green}44\textcolor{green}44   & 0.6009 & 0.9456                        \\
                        & V.3                                                                     & \textcolor{green}65\textcolor{green}55   & \textcolor{green}66\textcolor{green}66   & \textcolor{green}44\textcolor{green}44   & \textcolor{green}66\textcolor{green}66   & \textcolor{green}55\textcolor{green}55   & \textcolor{green}55\textcolor{green}55   & \textcolor{green}44\textcolor{green}44   & \textcolor{green}55\textcolor{green}55   & \textcolor{green}34\textcolor{green}44   & 0.7817 & 0.9456                        \\
                        & V.4                                                                     & \textcolor{green}44\textcolor{green}44   & \textcolor{green}55\textcolor{green}55   & \textcolor{green}55\textcolor{green}55   & \textcolor{green}55\textcolor{green}55   & \textcolor{green}55\textcolor{green}55   & \textcolor{green}55\textcolor{green}55   & \textcolor{green}55\textcolor{green}55   & \textcolor{green}55\textcolor{green}55   & \textcolor{green}55\textcolor{green}55   & 0.3333 & 0.9464                        \\
\hline
                        \multirow{4}{*}{${\rm l_{min}}=4$} & V.1                                                                     & \textcolor{green}55\textcolor{green}55   & \textcolor{green}44\textcolor{green}44   & \textcolor{green}55\textcolor{green}55   & \textcolor{green}55\textcolor{green}55   & \textcolor{green}56\textcolor{green}66   & \textcolor{green}55\textcolor{green}55   & \textcolor{green}44\textcolor{green}44   & \textcolor{green}55\textcolor{green}66   & \textcolor{green}65\textcolor{green}44   & 0.7817 & 0.9033 \\
                        & V.2                                                                     & \textcolor{green}55\textcolor{green}55   & \textcolor{green}54\textcolor{green}44   & \textcolor{green}55\textcolor{green}55   & \textcolor{green}55\textcolor{green}54   & \textcolor{green}55\textcolor{green}55   & \textcolor{green}55\textcolor{green}55   & \textcolor{green}56\textcolor{green}56   & \textcolor{green}55\textcolor{green}55   & \textcolor{green}44\textcolor{green}55   & 0.6009 & 0.9267                        \\
                        & V.3                                                                     & \textcolor{green}56\textcolor{green}55   & \textcolor{green}55\textcolor{green}55   & \textcolor{green}54\textcolor{green}44   & \textcolor{green}44\textcolor{green}44   & \textcolor{green}54\textcolor{green}55   & \textcolor{green}66\textcolor{green}66   & \textcolor{green}55\textcolor{green}55   & \textcolor{green}55\textcolor{green}55   & \textcolor{green}45\textcolor{green}55   & 0.6009 & 0.9258                        \\
                        & V.4                                                                     & \textcolor{green}54\textcolor{green}44   & \textcolor{green}55\textcolor{green}55   & \textcolor{green}56\textcolor{green}66   & \textcolor{green}45\textcolor{green}55   & \textcolor{green}54\textcolor{green}44   & \textcolor{green}55\textcolor{green}55   & \textcolor{green}55\textcolor{green}55   & \textcolor{green}66\textcolor{green}66   & \textcolor{green}44\textcolor{green}44   & 0.7817 & 0.9242                        \\
\hline
                        \multirow{4}{*}{${\rm l_{min}}=5$} & V.1                                                                     & 5   & 5   & 5   & 5   & 5   & 5   & 5   & 5   & 4\footnotemark[1]  & 0.3333 & 0.8759 \\
                        & V.2                                                                     & 5   & 5   & 5   & 5   & 5   & 5   & 5   & 5   & 4   & 0.3333 & 0.8759                        \\
                        & V.3                                                                     & 5   & 5   & 5   & 5   & 5   & 5   & 5   & 5   & 4   & 0.3333 & 0.8759                        \\
                        & V.4                                                                     & 5   & 5   & 5   & 5   & 5   & 5   & 5   & 5   & 4   & 0.3333 & 0.8734              \\         
\hline
\end{tabular}
}
\footnotetext[1]{When ${\rm l_{min}}=5$ , $S.9=4<({\rm l_{min}}=5)$ as a result of condition of constraint: sum of synaptic effective ranges is constant=44 and reflects synaptic strength rebalance.}
\end{table}

\begin{table}[]
\caption{Synaptic effective range with different ${\rm l_{min}}$ and different variables-brain plasticity ORPNN, test results are ${\rm l_{min}=2-5}$ and ${\rm k_{max}=9}$ respectively}
\label{table2}
\resizebox{\textwidth}{!}{
\begin{tabular}{|c|c|c|c|c|c|c|c|c|c|c|c|c|}
        \hline
        & \diagbox{V.L.}{S.L.} & S.1 & S.2 & S.3 & S.4 & S.5 & S.6 & S.7 & S.8 & S.9 & Std    & correlation coefficients\\
        \hline
                        \multirow{4}{*}{${\rm l_{min}}=2$} & V.1                                                                     & \textcolor{green}23\textcolor{green}32   & \textcolor{green}22\textcolor{green}22   & \textcolor{green}44\textcolor{green}44   & \textcolor{green}33\textcolor{green}33   & \textcolor{green}45\textcolor{green}44   & \textcolor{green}33\textcolor{green}33   & \textcolor{green}{12}14\textcolor{green}{13}11   & \textcolor{green}85\textcolor{green}913   & \textcolor{green}65\textcolor{green}32   & 4.1366 & 0.8694 \\
                        & V.2                                                                     & \textcolor{green}23\textcolor{green}23   & \textcolor{green}22\textcolor{green}22   & \textcolor{green}55\textcolor{green}65   & \textcolor{green}45\textcolor{green}55   & \textcolor{green}66\textcolor{green}76   & \textcolor{green}22\textcolor{green}32   & \textcolor{green}88\textcolor{green}{10}8   & \textcolor{green}99\textcolor{green}45   & \textcolor{green}64\textcolor{green}48   & 2.2608 & 0.8694                        \\
                        & V.3                                                                     & \textcolor{green}22\textcolor{green}22   & \textcolor{green}33\textcolor{green}33   & \textcolor{green}55\textcolor{green}56   & \textcolor{green}43\textcolor{green}34   & \textcolor{green}33\textcolor{green}33   & \textcolor{green}33\textcolor{green}33   & \textcolor{green}98\textcolor{green}811   & \textcolor{green}79\textcolor{green}{10}9   & \textcolor{green}88\textcolor{green}73   & 3.1402 & 0.8694\footnotemark[3]                        \\
                        & V.4                                                                     & \textcolor{green}22\textcolor{green}22   & \textcolor{green}34\textcolor{green}44   & \textcolor{green}53\textcolor{green}33   & \textcolor{green}22\textcolor{green}22   & \textcolor{green}33\textcolor{green}33   & \textcolor{green}55\textcolor{green}55   & \textcolor{green}{11}13\textcolor{green}{12}12   & \textcolor{green}76\textcolor{green}310   &\textcolor{green}86\textcolor{green}{10}3   & 3.6209 & 0.8730                        \\
\hline
                        \multirow{4}{*}{${\rm l_{min}}=3$} & V.1                                                                     & \textcolor{green}44\textcolor{green}44   & \textcolor{green}44\textcolor{green}44   & \textcolor{green}54\textcolor{green}55   & \textcolor{green}43\textcolor{green}44   & \textcolor{green}33\textcolor{green}33   & \textcolor{green}97\textcolor{green}99   & \textcolor{green}44\textcolor{green}44   & \textcolor{green}58\textcolor{green}55   & \textcolor{green}67\textcolor{green}66   & 1.7638 & 0.9282 \\
                        & V.2                                                                     & \textcolor{green}44\textcolor{green}55   & \textcolor{green}44\textcolor{green}55   & \textcolor{green}33\textcolor{green}44   & \textcolor{green}33\textcolor{green}44   & \textcolor{green}44\textcolor{green}44   & \textcolor{green}33\textcolor{green}44   & \textcolor{green}86\textcolor{green}55   & \textcolor{green}79\textcolor{green}{10}10   & \textcolor{green}88\textcolor{green}33   & 2.0276 & 0.9282                        \\
                        & V.3                                                                     & \textcolor{green}33\textcolor{green}33   & \textcolor{green}54\textcolor{green}55   & \textcolor{green}43\textcolor{green}44   & \textcolor{green}44\textcolor{green}44   & \textcolor{green}33\textcolor{green}44   & \textcolor{green}44\textcolor{green}44   & \textcolor{green}611\textcolor{green}66\footnotemark[2]   & \textcolor{green}66\textcolor{green}66   & \textcolor{green}96\textcolor{green}88   & 1.5366 & 0.9282                        \\
                        & V.4                                                                     & \textcolor{green}33\textcolor{green}33   & \textcolor{green}33\textcolor{green}33   & \textcolor{green}44\textcolor{green}33   & \textcolor{green}43\textcolor{green}34   & \textcolor{green}34\textcolor{green}45   & \textcolor{green}44\textcolor{green}45   & \textcolor{green}44\textcolor{green}44   & \textcolor{green}44\textcolor{green}44   & \textcolor{green}{13}15\textcolor{green}{16}13   & 3.1402 & 0.9282\footnotemark[3]                        \\
\hline
                        \multirow{4}{*}{${\rm l_{min}}=4$} & V.1                                                                     & \textcolor{green}44\textcolor{green}44   & \textcolor{green}55\textcolor{green}55   & \textcolor{green}66\textcolor{green}66   & \textcolor{green}55\textcolor{green}55   & \textcolor{green}55\textcolor{green}55   & \textcolor{green}44\textcolor{green}44   & \textcolor{green}55\textcolor{green}55   & \textcolor{green}55\textcolor{green}55   & \textcolor{green}55\textcolor{green}55   & 0.6009 & 0.8846 \\
                        & V.2                                                                     & \textcolor{green}44\textcolor{green}44   & \textcolor{green}55\textcolor{green}55   & \textcolor{green}55\textcolor{green}55   & \textcolor{green}44\textcolor{green}44   & \textcolor{green}{10}10\textcolor{green}{10}10   & \textcolor{green}44\textcolor{green}44   & \textcolor{green}44\textcolor{green}44   & \textcolor{green}44\textcolor{green}44   & \textcolor{green}44\textcolor{green}44   & 1.9650 & 0.8846                        \\
                        & V.3                                                                     & \textcolor{green}55\textcolor{green}55   & \textcolor{green}55\textcolor{green}55   & \textcolor{green}55\textcolor{green}55   & \textcolor{green}55\textcolor{green}55   & \textcolor{green}55\textcolor{green}55   & \textcolor{green}55\textcolor{green}55   & \textcolor{green}55\textcolor{green}55   & \textcolor{green}55\textcolor{green}55   & \textcolor{green}44\textcolor{green}44   & 0.3333 & 0.8846                        \\
                        & V.4                                                                     & \textcolor{green}55\textcolor{green}55   & \textcolor{green}55\textcolor{green}55   & \textcolor{green}44\textcolor{green}44   & \textcolor{green}44\textcolor{green}44   & \textcolor{green}66\textcolor{green}66   & \textcolor{green}55\textcolor{green}55   & \textcolor{green}55\textcolor{green}55   & \textcolor{green}55\textcolor{green}55   & \textcolor{green}55\textcolor{green}55   & 0.6009 & 0.8846\footnotemark[3]        \\
\hline
                        \multirow{4}{*}{${\rm l_{min}}=5$} & V.1                                                                     & 5   & 5   & 5   & 5   & 5   & 5   & 5   & 5   & 4\footnotemark[1]  & 0.3333 & 0.8579 \\
                        & V.2                                                                     & 5   & 5   & 5   & 5   & 5   & 5   & 5   & 5   & 4   & 0.3333 & 0.8579                        \\
                        & V.3                                                                     & 5   & 5   & 5   & 5   & 5   & 5   & 5   & 5   & 4   & 0.3333 & 0.8579                        \\
                        & V.4                                                                     & 5   & 5   & 5   & 5   & 5   & 5   & 5   & 5   & 4   & 0.3333 & 0.8579\footnotemark[3]              \\         
\hline
\end{tabular}
}
\footnotetext[2]{We discovered the synaptic effective ranges are larger at local, resulting in the alignment of other synaptic effective ranges.}
\footnotetext[3]{Vanishing Gradients}
\end{table}

\begin{table}[]
\caption{Synaptic effective range with different ${\rm l_{min}}$ and different variables-brain plasticity ORPNN-PF, test results are ${\rm l_{min}=2-5}$ and ${\rm k_{max}=9}$ respectively}
\label{table3}
\resizebox{\textwidth}{!}{
\begin{tabular}{|c|c|c|c|c|c|c|c|c|c|c|c|c|}
        \hline
        & \diagbox{V.L.}{S.L.} & S.1 & S.2 & S.3 & S.4 & S.5 & S.6 & S.7 & S.8 & S.9 & Std    & correlation coefficients\\
        \hline
                        \multirow{4}{*}{${\rm l_{min}}=2$} & V.1                                                                     & \textcolor{green}34\textcolor{green}33   & \textcolor{green}44\textcolor{green}44   & \textcolor{green}54\textcolor{green}44   & \textcolor{green}66\textcolor{green}66   & \textcolor{green}77\textcolor{green}77   & \textcolor{green}66\textcolor{green}66   & \textcolor{green}33\textcolor{green}33   & \textcolor{green}66\textcolor{green}66   & \textcolor{green}44\textcolor{green}55   & 1.4530 & 0.9472 \\
                        & V.2                                                                     & \textcolor{green}76\textcolor{green}77   & \textcolor{green}44\textcolor{green}44   & \textcolor{green}77\textcolor{green}77   & \textcolor{green}44\textcolor{green}44   & \textcolor{green}65\textcolor{green}56   & \textcolor{green}54\textcolor{green}44   & \textcolor{green}44\textcolor{green}44   & \textcolor{green}45\textcolor{green}44   & \textcolor{green}35\textcolor{green}54   & 1.3642 & 0.9472                        \\
                        & V.3                                                                     & \textcolor{green}32\textcolor{green}22   & \textcolor{green}66\textcolor{green}66   & \textcolor{green}55\textcolor{green}55   & \textcolor{green}55\textcolor{green}55   & \textcolor{green}77\textcolor{green}77   & \textcolor{green}66\textcolor{green}66   & \textcolor{green}45\textcolor{green}55   & \textcolor{green}44\textcolor{green}44   & \textcolor{green}44\textcolor{green}44   & 1.4530 & 0.9497                        \\
                        & V.4                                                                     & \textcolor{green}33\textcolor{green}33   & \textcolor{green}44\textcolor{green}44   & \textcolor{green}67\textcolor{green}77   & \textcolor{green}55\textcolor{green}55   & \textcolor{green}54\textcolor{green}44   & \textcolor{green}55\textcolor{green}55   & \textcolor{green}77\textcolor{green}77   & \textcolor{green}55\textcolor{green}66   &\textcolor{green}44\textcolor{green}33   & 1.5366 & 0.9527                        \\
\hline
                        \multirow{4}{*}{${\rm l_{min}}=3$} & V.1                                                                     & \textcolor{green}44\textcolor{green}44   & \textcolor{green}44\textcolor{green}44   & \textcolor{green}66\textcolor{green}66   & \textcolor{green}66\textcolor{green}77   & \textcolor{green}55\textcolor{green}55   & \textcolor{green}66\textcolor{green}66   & \textcolor{green}44\textcolor{green}44   & \textcolor{green}55\textcolor{green}55   & \textcolor{green}44\textcolor{green}33   & 1.2693 & 0.9672 \\
                        & V.2                                                                     & \textcolor{green}44\textcolor{green}44   & \textcolor{green}45\textcolor{green}44   & \textcolor{green}55\textcolor{green}55   & \textcolor{green}65\textcolor{green}55   & \textcolor{green}55\textcolor{green}55   & \textcolor{green}55\textcolor{green}55   & \textcolor{green}55\textcolor{green}55   & \textcolor{green}66\textcolor{green}66   & \textcolor{green}44\textcolor{green}55   & 0.6009 & 0.9686                        \\
                        & V.3                                                                     & \textcolor{green}66\textcolor{green}66   & \textcolor{green}44\textcolor{green}44   & \textcolor{green}44\textcolor{green}44   & \textcolor{green}55\textcolor{green}55   & \textcolor{green}55\textcolor{green}55   & \textcolor{green}66\textcolor{green}66   & \textcolor{green}55\textcolor{green}55   & \textcolor{green}55\textcolor{green}55   & \textcolor{green}44\textcolor{green}44   & 0.7817 & 0.9681                        \\
                        & V.4                                                                     & \textcolor{green}44\textcolor{green}44   & \textcolor{green}45\textcolor{green}44   & \textcolor{green}44\textcolor{green}44   & \textcolor{green}63\textcolor{green}66   & \textcolor{green}66\textcolor{green}66   & \textcolor{green}55\textcolor{green}55   & \textcolor{green}56\textcolor{green}55   & \textcolor{green}55\textcolor{green}55   & \textcolor{green}56\textcolor{green}55   & 0.7817 & 0.9700                        \\
\hline
                        \multirow{4}{*}{${\rm l_{min}}=4$} & V.1                                                                     & \textcolor{green}55\textcolor{green}55   & \textcolor{green}44\textcolor{green}44   & \textcolor{green}55\textcolor{green}55   & \textcolor{green}66\textcolor{green}66   & \textcolor{green}55\textcolor{green}55   & \textcolor{green}44\textcolor{green}44   & \textcolor{green}45\textcolor{green}55   & \textcolor{green}65\textcolor{green}66   & \textcolor{green}55\textcolor{green}44   & 0.7817 & 0.9218 \\
                        & V.2                                                                     & \textcolor{green}55\textcolor{green}55   & \textcolor{green}55\textcolor{green}55   & \textcolor{green}55\textcolor{green}55   & \textcolor{green}55\textcolor{green}55   & \textcolor{green}45\textcolor{green}55   & \textcolor{green}55\textcolor{green}55   & \textcolor{green}55\textcolor{green}55   & \textcolor{green}55\textcolor{green}55   & \textcolor{green}54\textcolor{green}44   & 0.3333 & 0.9218                        \\
                        & V.3                                                                     & \textcolor{green}55\textcolor{green}55   & \textcolor{green}55\textcolor{green}55   & \textcolor{green}66\textcolor{green}66   & \textcolor{green}55\textcolor{green}55   & \textcolor{green}55\textcolor{green}55   & \textcolor{green}44\textcolor{green}44   & \textcolor{green}44\textcolor{green}44   & \textcolor{green}55\textcolor{green}55   & \textcolor{green}55\textcolor{green}55   & 0.6009 & 0.9218                        \\
                        & V.4                                                                     & \textcolor{green}44\textcolor{green}44   & \textcolor{green}55\textcolor{green}55   & \textcolor{green}44\textcolor{green}44   & \textcolor{green}55\textcolor{green}55   & \textcolor{green}66\textcolor{green}66   & \textcolor{green}55\textcolor{green}55   & \textcolor{green}55\textcolor{green}55   & \textcolor{green}44\textcolor{green}54   & \textcolor{green}66\textcolor{green}56   & 0.7817 & 0.9326                        \\
\hline
                        \multirow{4}{*}{${\rm l_{min}}=5$} & V.1                                                                     & 5   & 5   & 5   & 5   & 5   & 5   & 5   & 5   & 4\footnotemark[1]  & 0.3333 & 0.8991 \\
                        & V.2                                                                     & 5   & 5   & 5   & 5   & 5   & 5   & 5   & 5   & 4   & 0.3333 & 0.8991                        \\
                        & V.3                                                                     & 5   & 5   & 5   & 5   & 5   & 5   & 5   & 5   & 4   & 0.3333 & 0.8991                        \\
                        & V.4                                                                     & 5   & 5   & 5   & 5   & 5   & 5   & 5   & 5   & 4   & 0.3333 & 0.8992              \\         
\hline
\end{tabular}
}
\end{table}

\begin{table}[]
\caption{Synaptic effective range with different ${\rm l_{min}}$ and different variables-brain plasticity ORPNN-MF(P), test results are ${\rm l_{min}=2-5}$ and ${\rm k_{max}=9}$ respectively}
\label{table4}
\resizebox{\textwidth}{!}{
\begin{tabular}{|c|c|c|c|c|c|c|c|c|c|c|c|c|}
        \hline
        & \diagbox{V.L.}{S.L.} & S.1 & S.2 & S.3 & S.4 & S.5 & S.6 & S.7 & S.8 & S.9 & Std    & correlation coefficients\\
        \hline
                        \multirow{4}{*}{${\rm l_{min}}=2$} & V.1                                                                     & \textcolor{green}77\textcolor{green}77   & \textcolor{green}44\textcolor{green}44   & \textcolor{green}44\textcolor{green}44   & \textcolor{green}44\textcolor{green}44   & \textcolor{green}55\textcolor{green}55   & \textcolor{green}44\textcolor{green}44   & \textcolor{green}66\textcolor{green}66   & \textcolor{green}55\textcolor{green}55   & \textcolor{green}55\textcolor{green}55   & 1.0541 & 0.9194 \\
                        & V.2                                                                     & \textcolor{green}55\textcolor{green}55   & \textcolor{green}44\textcolor{green}44   & \textcolor{green}55\textcolor{green}55   & \textcolor{green}66\textcolor{green}66   & \textcolor{green}44\textcolor{green}44   & \textcolor{green}44\textcolor{green}44   & \textcolor{green}77\textcolor{green}77   & \textcolor{green}55\textcolor{green}55   & \textcolor{green}44\textcolor{green}44   & 1.0541 & 0.9350                        \\
                        & V.3                                                                     & \textcolor{green}55\textcolor{green}55   & \textcolor{green}44\textcolor{green}44   & \textcolor{green}55\textcolor{green}55   & \textcolor{green}77\textcolor{green}77   & \textcolor{green}66\textcolor{green}66   & \textcolor{green}44\textcolor{green}44   & \textcolor{green}44\textcolor{green}44   & \textcolor{green}55\textcolor{green}55   & \textcolor{green}44\textcolor{green}44   & 1.0541 & 0.9302                        \\
                        & V.4                                                                     & \textcolor{green}66\textcolor{green}66   & \textcolor{green}66\textcolor{green}66   & \textcolor{green}44\textcolor{green}44   & \textcolor{green}44\textcolor{green}44   & \textcolor{green}44\textcolor{green}44   & \textcolor{green}44\textcolor{green}44   & \textcolor{green}77\textcolor{green}77   & \textcolor{green}44\textcolor{green}44   & \textcolor{green}55\textcolor{green}55   & 1.1667 & 0.9382                        \\
\hline
                        \multirow{4}{*}{${\rm l_{min}}=3$} & V.1                                                                     & \textcolor{green}55\textcolor{green}55   & \textcolor{green}44\textcolor{green}44   & \textcolor{green}44\textcolor{green}44   & \textcolor{green}66\textcolor{green}66   & \textcolor{green}66\textcolor{green}66   & \textcolor{green}55\textcolor{green}55   & \textcolor{green}55\textcolor{green}55   & \textcolor{green}55\textcolor{green}55   & \textcolor{green}44\textcolor{green}44   & 0.7817 & 0.9188 \\
                        & V.2                                                                     & \textcolor{green}55\textcolor{green}55   & \textcolor{green}55\textcolor{green}55   & \textcolor{green}55\textcolor{green}55   & \textcolor{green}55\textcolor{green}55   & \textcolor{green}44\textcolor{green}44   & \textcolor{green}55\textcolor{green}55   & \textcolor{green}55\textcolor{green}55   & \textcolor{green}55\textcolor{green}55   & \textcolor{green}33\textcolor{green}33   & 0.3333 & 0.9352                        \\
                        & V.3                                                                     & \textcolor{green}55\textcolor{green}55   & \textcolor{green}55\textcolor{green}55   & \textcolor{green}44\textcolor{green}44   & \textcolor{green}66\textcolor{green}66   & \textcolor{green}55\textcolor{green}55   & \textcolor{green}44\textcolor{green}44   & \textcolor{green}44\textcolor{green}44   & \textcolor{green}66\textcolor{green}66   & \textcolor{green}55\textcolor{green}55   & 0.7817 & 0.9335                        \\
                        & V.4                                                                     & \textcolor{green}44\textcolor{green}44   & \textcolor{green}44\textcolor{green}44   & \textcolor{green}55\textcolor{green}55   & \textcolor{green}55\textcolor{green}55   & \textcolor{green}55\textcolor{green}55   & \textcolor{green}66\textcolor{green}66   & \textcolor{green}55\textcolor{green}55   & \textcolor{green}55\textcolor{green}55   & \textcolor{green}55\textcolor{green}55   & 0.6009 & 0.9265                        \\
\hline
                        \multirow{4}{*}{${\rm l_{min}}=4$} & V.1                                                                     & \textcolor{green}55\textcolor{green}55   & \textcolor{green}55\textcolor{green}55   & \textcolor{green}55\textcolor{green}55   & \textcolor{green}55\textcolor{green}55   & \textcolor{green}55\textcolor{green}55   & \textcolor{green}55\textcolor{green}55  & \textcolor{green}55\textcolor{green}55   & \textcolor{green}55\textcolor{green}55   & \textcolor{green}44\textcolor{green}44   & 0.3333 & 0.8991 \\
                        & V.2                                                                     & \textcolor{green}55\textcolor{green}55   & \textcolor{green}44\textcolor{green}44   & \textcolor{green}55\textcolor{green}55   & \textcolor{green}55\textcolor{green}55   & \textcolor{green}55\textcolor{green}55   & \textcolor{green}55\textcolor{green}55  & \textcolor{green}55\textcolor{green}55   & \textcolor{green}55\textcolor{green}55   & \textcolor{green}44\textcolor{green}44   & 0.3333 & 0.9073                        \\
                        & V.3                                                                     & \textcolor{green}55\textcolor{green}55   & \textcolor{green}55\textcolor{green}55   & \textcolor{green}55\textcolor{green}55   & \textcolor{green}55\textcolor{green}55   & \textcolor{green}55\textcolor{green}55   & \textcolor{green}44\textcolor{green}44  & \textcolor{green}55\textcolor{green}55   & \textcolor{green}55\textcolor{green}55   & \textcolor{green}55\textcolor{green}55   & 0.3333 & 0.9117                        \\
                        & V.4                                                                     & \textcolor{green}55\textcolor{green}55   & \textcolor{green}55\textcolor{green}55   & \textcolor{green}55\textcolor{green}55   & \textcolor{green}55\textcolor{green}55   & \textcolor{green}44\textcolor{green}44   & \textcolor{green}55\textcolor{green}55  & \textcolor{green}55\textcolor{green}55   & \textcolor{green}55\textcolor{green}55   & \textcolor{green}55\textcolor{green}55   & 0.3333 & 0.9115                        \\
\hline
                        \multirow{4}{*}{${\rm l_{min}}=5$} & V.1                                                                     & 5   & 5   & 5   & 5   & 5   & 5   & 5   & 5   & 4\footnotemark[1]  & 0.3333 & 0.8206 \\
                        & V.2                                                                     & 5   & 5   & 5   & 5   & 5   & 5   & 5   & 5   & 4   & 0.3333 & 0.8278                        \\
                        & V.3                                                                     & 5   & 5   & 5   & 5   & 5   & 5   & 5   & 5   & 4   & 0.3333 & 0.8738                        \\
                        & V.4                                                                     & 5   & 5   & 5   & 5   & 5   & 5   & 5   & 5   & 4   & 0.3333 & 0.8663              \\         
\hline
\end{tabular}
}
\end{table}

\begin{table}[]
\caption{Synaptic effective range with different ${\rm l_{min}}$ and different variables-brain plasticity ORPNN-MF(P\&N), test results are ${\rm l_{min}=2-5}$ and ${\rm k_{max}=9}$ respectively}
\label{table5}
\resizebox{\textwidth}{!}{
\begin{tabular}{|c|c|c|c|c|c|c|c|c|c|c|c|c|}
        \hline
        & \diagbox{V.L.}{S.L.} & S.1 & S.2 & S.3 & S.4 & S.5 & S.6 & S.7 & S.8 & S.9 & Std    & correlation coefficients\\
        \hline
                        \multirow{4}{*}{${\rm l_{min}}=2$} & V.1                                                                     & \textcolor{green}55\textcolor{green}55   & \textcolor{green}66\textcolor{green}64   & \textcolor{green}34\textcolor{green}33   & \textcolor{green}66\textcolor{green}66   & \textcolor{green}44\textcolor{green}44   & \textcolor{green}54\textcolor{green}44   & \textcolor{green}63\textcolor{green}55   & \textcolor{green}66\textcolor{green}66   & \textcolor{green}36\textcolor{green}57   & 1.2693 & 0.9417 \\
                        & V.2                                                                     & \textcolor{green}34\textcolor{green}33   & \textcolor{green}76\textcolor{green}66   & \textcolor{green}35\textcolor{green}55   & \textcolor{green}65\textcolor{green}65   & \textcolor{green}65\textcolor{green}55   & \textcolor{green}65\textcolor{green}66   & \textcolor{green}55\textcolor{green}55   & \textcolor{green}35\textcolor{green}33   & \textcolor{green}54\textcolor{green}56   & 1.1667 & 0.9650                        \\
                        & V.3                                                                     & \textcolor{green}55\textcolor{green}66   & \textcolor{green}65\textcolor{green}45   & \textcolor{green}56\textcolor{green}56   & \textcolor{green}66\textcolor{green}53   & \textcolor{green}55\textcolor{green}56   & \textcolor{green}55\textcolor{green}56   & \textcolor{green}66\textcolor{green}66   & \textcolor{green}33\textcolor{green}33   & \textcolor{green}33\textcolor{green}53   & 1.4530 & 0.9650                        \\
                        & V.4                                                                     & \textcolor{green}45\textcolor{green}66   & \textcolor{green}55\textcolor{green}55   & \textcolor{green}33\textcolor{green}33   & \textcolor{green}66\textcolor{green}66   & \textcolor{green}56\textcolor{green}55   & \textcolor{green}56\textcolor{green}55   & \textcolor{green}55\textcolor{green}55   & \textcolor{green}54\textcolor{green}55   & \textcolor{green}64\textcolor{green}44   & 0.9280 & 0.9592                        \\
\hline
                        \multirow{4}{*}{${\rm l_{min}}=3$} & V.1                                                                     & \textcolor{green}55\textcolor{green}55   & \textcolor{green}56\textcolor{green}55   & \textcolor{green}66\textcolor{green}65   & \textcolor{green}65\textcolor{green}55   & \textcolor{green}45\textcolor{green}65   & \textcolor{green}55\textcolor{green}55   & \textcolor{green}44\textcolor{green}44   & \textcolor{green}54\textcolor{green}45   & \textcolor{green}44\textcolor{green}45   & 0.3333 & 0.9359 \\
                        & V.2                                                                     & \textcolor{green}56\textcolor{green}66   & \textcolor{green}56\textcolor{green}66   & \textcolor{green}54\textcolor{green}44   & \textcolor{green}55\textcolor{green}55   & \textcolor{green}55\textcolor{green}45   & \textcolor{green}66\textcolor{green}66   & \textcolor{green}44\textcolor{green}44   & \textcolor{green}45\textcolor{green}55   & \textcolor{green}53\textcolor{green}43   & 1.0541 & 0.9415                        \\
                        & V.3                                                                     & \textcolor{green}53\textcolor{green}34   & \textcolor{green}55\textcolor{green}55   & \textcolor{green}56\textcolor{green}66   & \textcolor{green}55\textcolor{green}55   & \textcolor{green}55\textcolor{green}55   & \textcolor{green}46\textcolor{green}65   & \textcolor{green}55\textcolor{green}55   & \textcolor{green}55\textcolor{green}55   & \textcolor{green}54\textcolor{green}44   & 0.6009 & 0.9415                        \\
                        & V.4                                                                     & \textcolor{green}53\textcolor{green}55   & \textcolor{green}56\textcolor{green}55   & \textcolor{green}45\textcolor{green}44   & \textcolor{green}56\textcolor{green}55   & \textcolor{green}44\textcolor{green}45   & \textcolor{green}55\textcolor{green}55   & \textcolor{green}55\textcolor{green}55   & \textcolor{green}55\textcolor{green}55   & \textcolor{green}65\textcolor{green}65   & 0.3333 & 0.9423                        \\
\hline
                        \multirow{4}{*}{${\rm l_{min}}=4$} & V.1                                                                     & \textcolor{green}55\textcolor{green}55   & \textcolor{green}55\textcolor{green}55   & \textcolor{green}55\textcolor{green}55   & \textcolor{green}54\textcolor{green}45   & \textcolor{green}55\textcolor{green}55   & \textcolor{green}55\textcolor{green}55  & \textcolor{green}55\textcolor{green}55   & \textcolor{green}55\textcolor{green}55   & \textcolor{green}45\textcolor{green}54   & 0.3333 & 0.9068 \\
                        & V.2                                                                     & \textcolor{green}55\textcolor{green}55   & \textcolor{green}55\textcolor{green}55   & \textcolor{green}55\textcolor{green}55   & \textcolor{green}55\textcolor{green}55   & \textcolor{green}55\textcolor{green}55   & \textcolor{green}55\textcolor{green}55  & \textcolor{green}55\textcolor{green}55   & \textcolor{green}55\textcolor{green}55   & \textcolor{green}44\textcolor{green}44   & 0.3333 & 0.9068                        \\
                        & V.3                                                                     & \textcolor{green}55\textcolor{green}55   & \textcolor{green}55\textcolor{green}55   & \textcolor{green}55\textcolor{green}55   & \textcolor{green}55\textcolor{green}55   & \textcolor{green}55\textcolor{green}55   & \textcolor{green}55\textcolor{green}55  & \textcolor{green}55\textcolor{green}55   & \textcolor{green}55\textcolor{green}55   & \textcolor{green}44\textcolor{green}44   & 0.3333 & 0.9068                        \\
                        & V.4                                                                     & \textcolor{green}56\textcolor{green}65   & \textcolor{green}55\textcolor{green}55   & \textcolor{green}55\textcolor{green}44   & \textcolor{green}55\textcolor{green}55   & \textcolor{green}55\textcolor{green}55   & \textcolor{green}55\textcolor{green}55  & \textcolor{green}55\textcolor{green}55   & \textcolor{green}55\textcolor{green}55   & \textcolor{green}43\textcolor{green}45   & 0.3333 & 0.9228                        \\
\hline
                        \multirow{4}{*}{${\rm l_{min}}=5$} & V.1                                                                     & 5   & 5   & 5   & 5   & 5   & 5   & 5   & 5   & 4\footnotemark[1]  & 0.3333 & 0.8894 \\
                        & V.2                                                                     & 5   & 5   & 5   & 5   & 5   & 5   & 5   & 5   & 4   & 0.3333 & 0.8894                        \\
                        & V.3                                                                     & 5   & 5   & 5   & 5   & 5   & 5   & 5   & 5   & 4   & 0.3333 & 0.8894                        \\
                        & V.4                                                                     & 5   & 5   & 5   & 5   & 5   & 5   & 5   & 5   & 4   & 0.3333 & 0.8878              \\         
\hline
\end{tabular}
}
\end{table}

\section{A simple PNN}

A simple PNN, only correlation coefficient and astrocytes phagocytose synapses are considered rather than gradients update. Here, the PNN is simplified to only consider the astrocytes phagocytose synapses formula \eqref{eq4} with updated synaptic effective range weight and formula \eqref{eq1} with updated shared connection weight. 

Two approaches are proposed: one approach is to evaluate the $f(t)$ and $h(t)$ correlation coefficient of $r(m,k)$ corresponding to the time range, each variable and synaptic position corresponding to one time range and $t\in[n_3(m,k),n_3(m,k+1)-1]$, and then considering whether or not to cancel the update of the previous-generation $r(m,k)$ depending on whether current correlation coefficient is improved compared to the previous generation; the other approach favors not processing $r(m,k)$ at all.

\begin{figure}%
\centering
\includegraphics[width=0.9\textwidth]{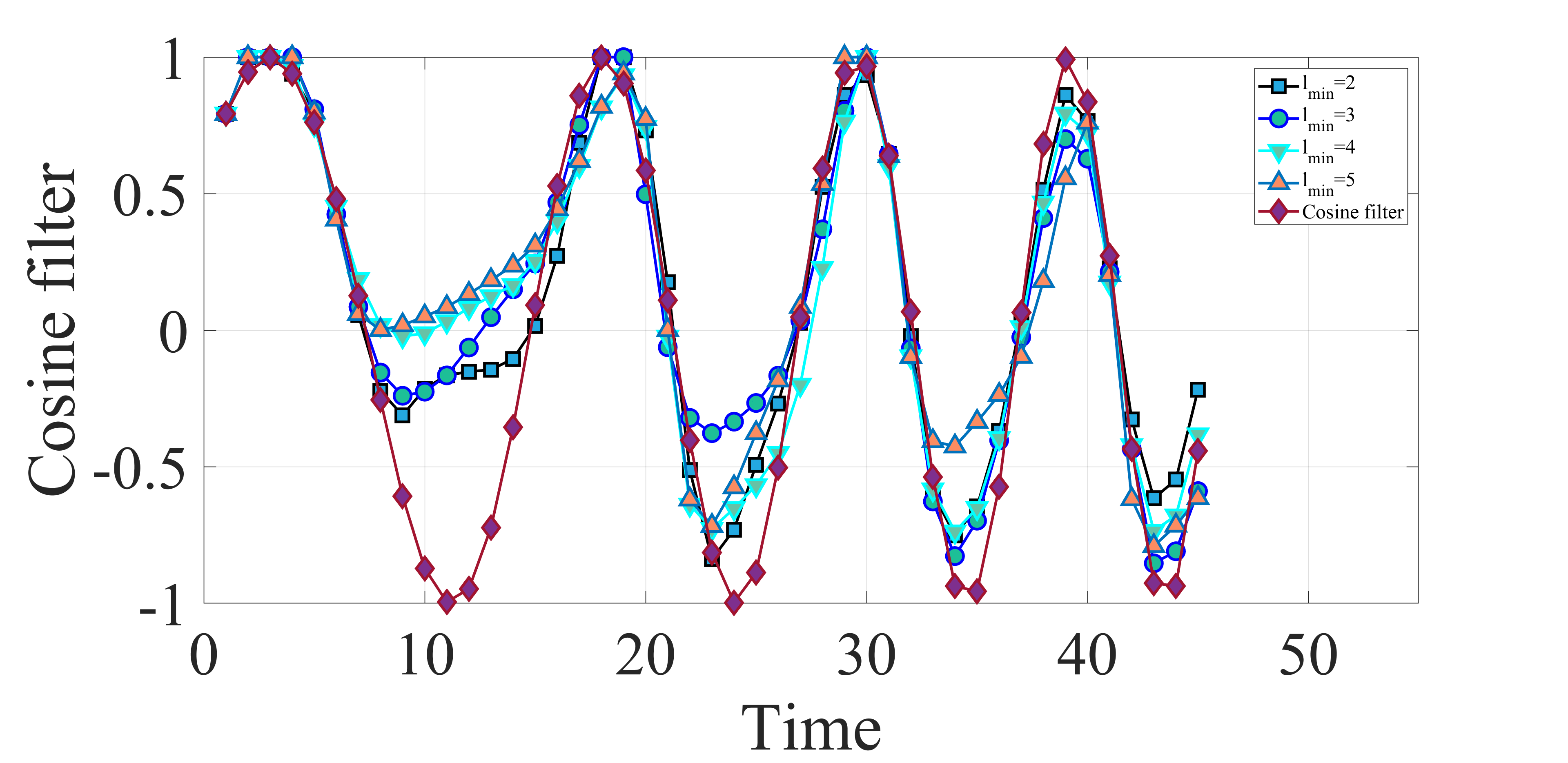}
\includegraphics[width=0.9\textwidth]{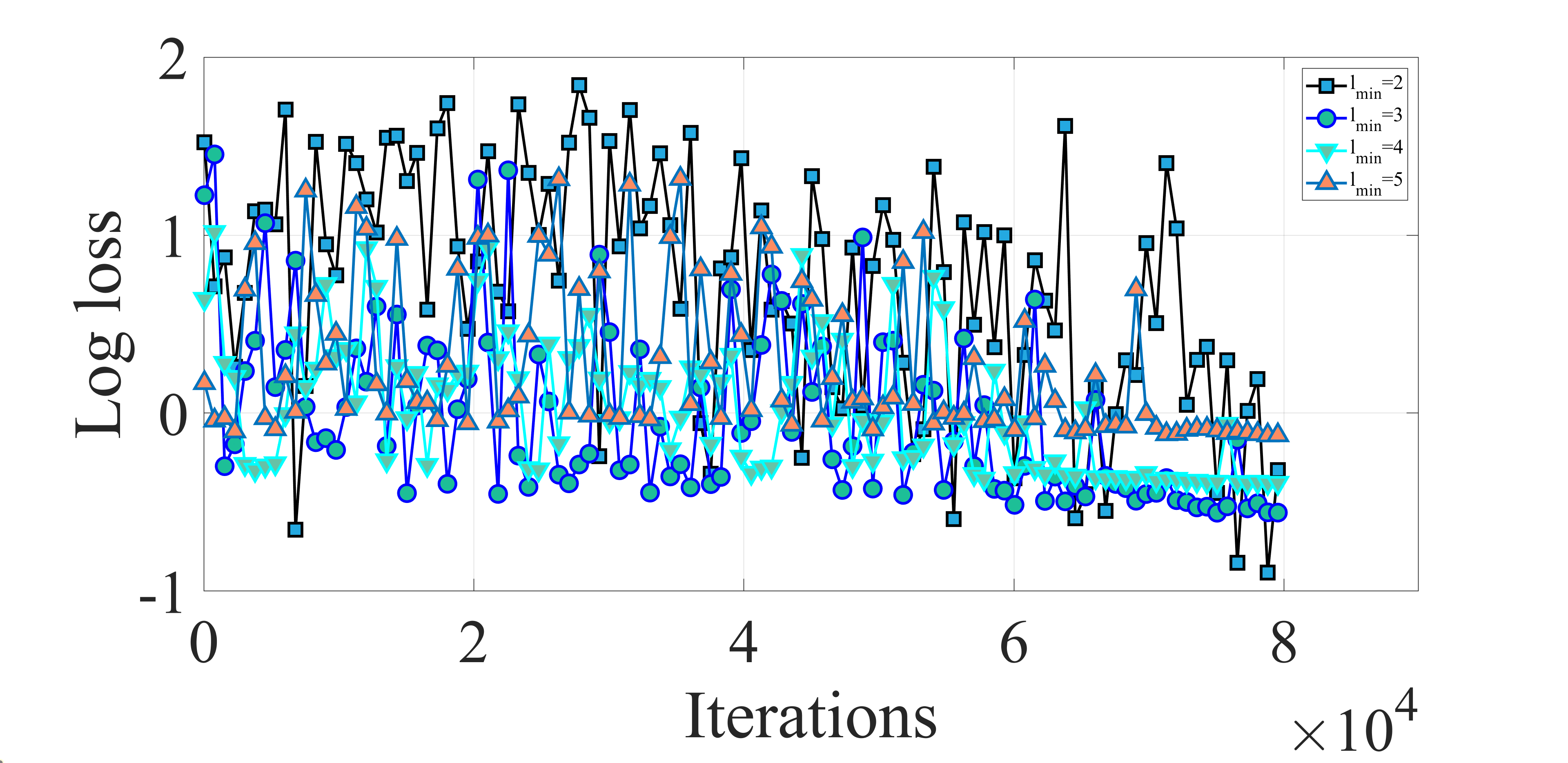}
\caption{Simple ORPNN-PF without correlation coefficient correction, test results are ${\rm l_{min}}=2-5$ and ${\rm k_{max}}=9$ respectively.}
\label{fig12}
\end{figure}

\begin{figure}%
\centering
\includegraphics[width=0.9\textwidth]{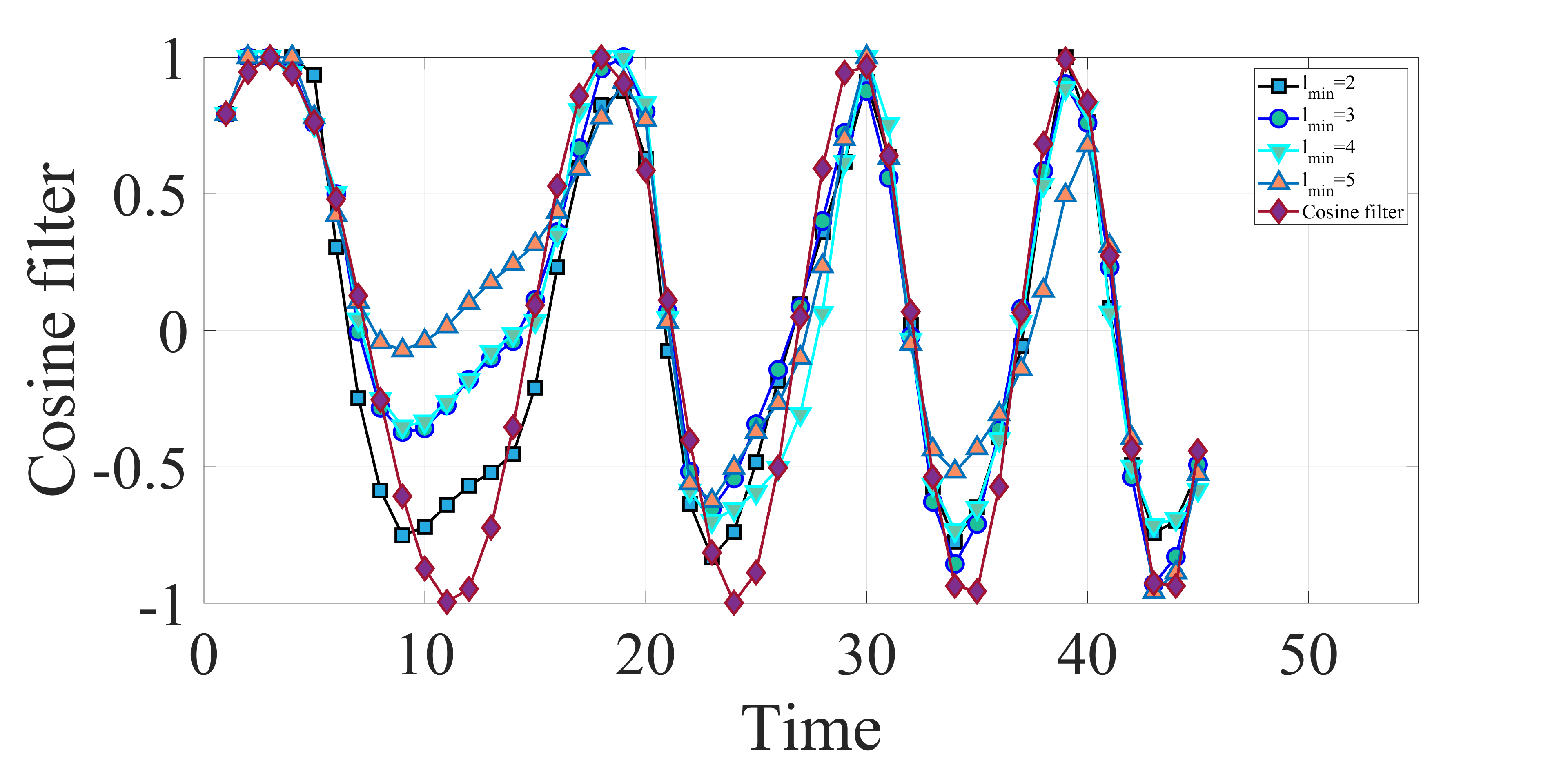}
\includegraphics[width=0.9\textwidth]{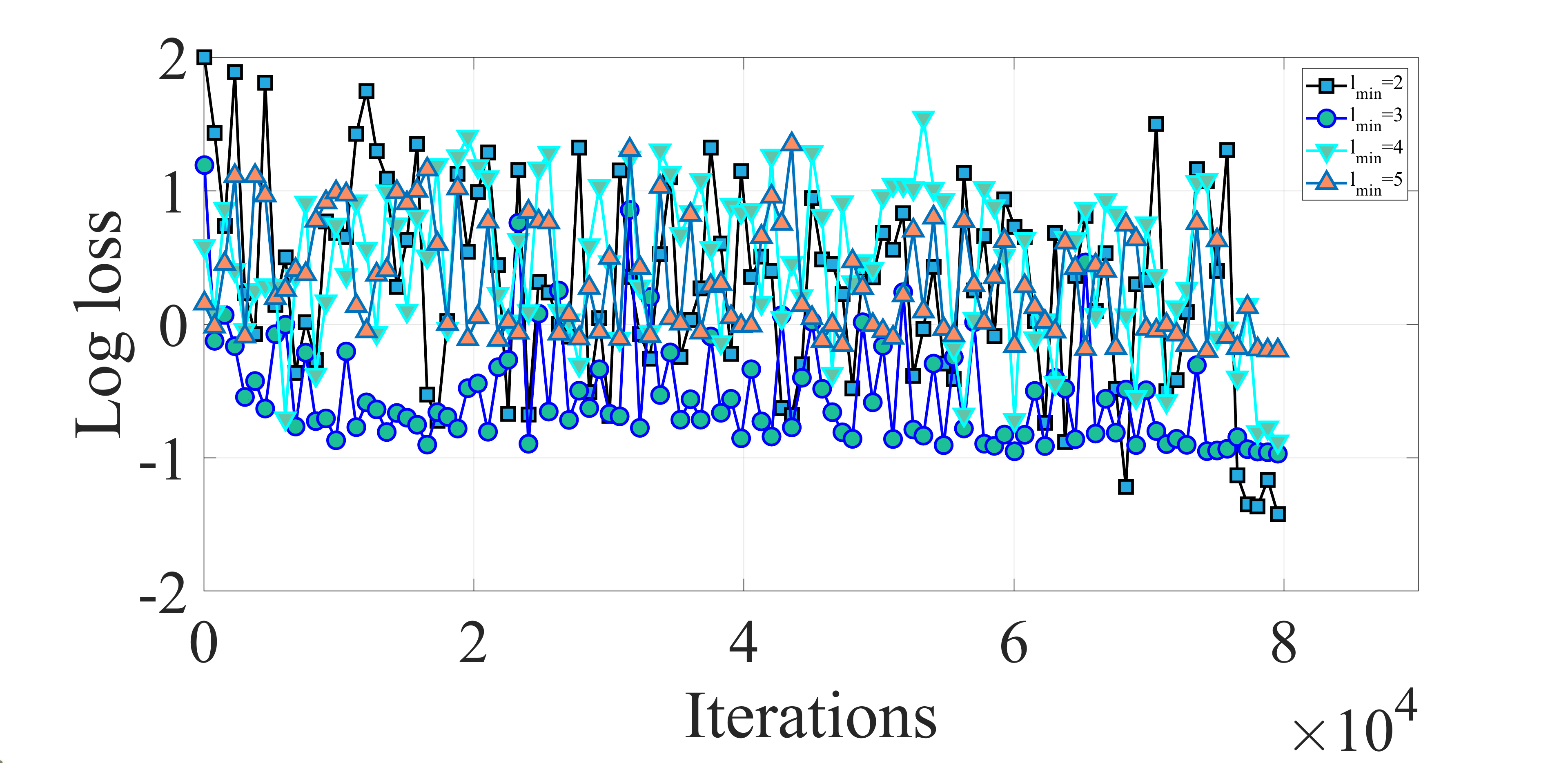}
\caption{Simple ORPNN-PF with correlation coefficient correction, test results are ${\rm l_{min}}=2-5$ and ${\rm k_{max}}=9$ respectively.}
\label{fig13}
\end{figure}

By comparing Table.\ref{table6}-\ref{table7} and Fig.\ref{fig12}-\ref{fig13}, the first approach garners better simulation results, and the processing helps activate the synaptic effective range that varies with iteration.

\begin{table}
\caption{Synaptic effective range with different ${\rm l_{min}}$ and different variables-brain plasticity simple ORPNN-PF without correlation coefficient correction, test results are ${\rm l_{min}=2-5}$ and ${\rm k_{max}=9}$ respectively}
\label{table6}
\resizebox{\textwidth}{!}{
\begin{tabular}{|c|c|c|c|c|c|c|c|c|c|c|c|c|}
        \hline
        & \diagbox{V.L.}{S.L.} & S.1 & S.2 & S.3 & S.4 & S.5 & S.6 & S.7 & S.8 & S.9 & Std    & correlation coefficients\\
        \hline
                        \multirow{4}{*}{${\rm l_{min}}=2$} & V.1                                                                     & \textcolor{green}55\textcolor{green}55   & \textcolor{green}44\textcolor{green}44   & \textcolor{green}43\textcolor{green}33   & \textcolor{green}54\textcolor{green}44   & \textcolor{green}77\textcolor{green}77   & \textcolor{green}44\textcolor{green}44   & \textcolor{green}66\textcolor{green}66   & \textcolor{green}44\textcolor{green}45   & \textcolor{green}57\textcolor{green}76   & 1.2693 & 0.9404 \\
                        & V.2                                                                     & \textcolor{green}33\textcolor{green}33   & \textcolor{green}33\textcolor{green}33   & \textcolor{green}44\textcolor{green}44   & \textcolor{green}55\textcolor{green}55   & \textcolor{green}66\textcolor{green}66   & \textcolor{green}99\textcolor{green}99   & \textcolor{green}44\textcolor{green}44   & \textcolor{green}66\textcolor{green}66   & \textcolor{green}44\textcolor{green}44   & 1.9003 & 0.9473                        \\
                        & V.3                                                                     & \textcolor{green}54\textcolor{green}44   & \textcolor{green}33\textcolor{green}33   & \textcolor{green}33\textcolor{green}33   & \textcolor{green}44\textcolor{green}44   & \textcolor{green}78\textcolor{green}87   & \textcolor{green}65\textcolor{green}55   & \textcolor{green}44\textcolor{green}34   & \textcolor{green}86\textcolor{green}66   & \textcolor{green}47\textcolor{green}88   & 1.7638 & 0.9542                        \\
                        & V.4                                                                     & \textcolor{green}33\textcolor{green}33   & \textcolor{green}54\textcolor{green}44   & \textcolor{green}54\textcolor{green}44   & \textcolor{green}55\textcolor{green}55   & \textcolor{green}65\textcolor{green}55   & \textcolor{green}55\textcolor{green}55   & \textcolor{green}55\textcolor{green}55   & \textcolor{green}78\textcolor{green}88   & \textcolor{green}35\textcolor{green}55   & 1.3642 & 0.9525                        \\
\hline
                        \multirow{4}{*}{${\rm l_{min}}=3$} & V.1                                                                     & \textcolor{green}66\textcolor{green}66   & \textcolor{green}44\textcolor{green}44   & \textcolor{green}66\textcolor{green}66   & \textcolor{green}55\textcolor{green}55   & \textcolor{green}55\textcolor{green}55   & \textcolor{green}34\textcolor{green}44   & \textcolor{green}55\textcolor{green}55   & \textcolor{green}66\textcolor{green}66   & \textcolor{green}43\textcolor{green}33   & 1.0541 & 0.9235 \\
                        & V.2                                                                     & \textcolor{green}76\textcolor{green}66   & \textcolor{green}44\textcolor{green}44   & \textcolor{green}44\textcolor{green}44   & \textcolor{green}55\textcolor{green}55   & \textcolor{green}44\textcolor{green}44   & \textcolor{green}44\textcolor{green}44   & \textcolor{green}66\textcolor{green}66   & \textcolor{green}66\textcolor{green}55   & \textcolor{green}45\textcolor{green}66   & 0.9280 & 0.9235                        \\
                        & V.3                                                                     & \textcolor{green}66\textcolor{green}66   & \textcolor{green}66\textcolor{green}66   & \textcolor{green}44\textcolor{green}44   & \textcolor{green}66\textcolor{green}66   & \textcolor{green}55\textcolor{green}55   & \textcolor{green}55\textcolor{green}55   & \textcolor{green}44\textcolor{green}44   & \textcolor{green}44\textcolor{green}44   & \textcolor{green}44\textcolor{green}44   & 0.9280 & 0.9235                        \\
                        & V.4                                                                     & \textcolor{green}55\textcolor{green}55   & \textcolor{green}55\textcolor{green}55   & \textcolor{green}55\textcolor{green}55   & \textcolor{green}44\textcolor{green}44   & \textcolor{green}55\textcolor{green}55   & \textcolor{green}55\textcolor{green}55   & \textcolor{green}66\textcolor{green}66   & \textcolor{green}55\textcolor{green}55   & \textcolor{green}44\textcolor{green}44   & 0.6009 & 0.9217                        \\
\hline
                        \multirow{4}{*}{${\rm l_{min}}=4$} & V.1                                                                     & \textcolor{green}44\textcolor{green}44   & \textcolor{green}55\textcolor{green}55   & \textcolor{green}55\textcolor{green}55   & \textcolor{green}55\textcolor{green}55   & \textcolor{green}64\textcolor{green}66   & \textcolor{green}55\textcolor{green}55  & \textcolor{green}44\textcolor{green}44   & \textcolor{green}66\textcolor{green}66   & \textcolor{green}46\textcolor{green}44   & 0.7817 & 0.8947 \\
                        & V.2                                                                     & \textcolor{green}55\textcolor{green}55   & \textcolor{green}55\textcolor{green}55   & \textcolor{green}55\textcolor{green}55   & \textcolor{green}55\textcolor{green}55   & \textcolor{green}44\textcolor{green}44   & \textcolor{green}55\textcolor{green}55  & \textcolor{green}55\textcolor{green}55   & \textcolor{green}44\textcolor{green}44   & \textcolor{green}66\textcolor{green}66   & 0.6009 & 0.9084                        \\
                        & V.3                                                                     & \textcolor{green}66\textcolor{green}66   & \textcolor{green}55\textcolor{green}55   & \textcolor{green}55\textcolor{green}55   & \textcolor{green}55\textcolor{green}55   & \textcolor{green}44\textcolor{green}44   & \textcolor{green}45\textcolor{green}55  & \textcolor{green}66\textcolor{green}55   & \textcolor{green}55\textcolor{green}55   & \textcolor{green}43\textcolor{green}44   & 0.6009 & 0.9005                        \\
                        & V.4                                                                     & \textcolor{green}55\textcolor{green}55   & \textcolor{green}44\textcolor{green}44   & \textcolor{green}55\textcolor{green}55   & \textcolor{green}65\textcolor{green}66   & \textcolor{green}45\textcolor{green}44   & \textcolor{green}55\textcolor{green}55  & \textcolor{green}55\textcolor{green}55   & \textcolor{green}55\textcolor{green}55   & \textcolor{green}55\textcolor{green}55   & 0.6009 & 0.8975                        \\
\hline
                        \multirow{4}{*}{${\rm l_{min}}=5$} & V.1                                                                     & 5   & 5   & 5   & 5   & 5   & 5   & 5   & 5   & 4\footnotemark[1]  & 0.3333 & 0.8671 \\
                        & V.2                                                                     & 5   & 5   & 5   & 5   & 5   & 5   & 5   & 5   & 4   & 0.3333 & 0.8671                        \\
                        & V.3                                                                     & 5   & 5   & 5   & 5   & 5   & 5   & 5   & 5   & 4   & 0.3333 & 0.8671                        \\
                        & V.4                                                                     & 5   & 5   & 5   & 5   & 5   & 5   & 5   & 5   & 4   & 0.3333 & 0.8671              \\         
\hline
\end{tabular}
}
\end{table}

\begin{table}[]
\caption{Synaptic effective range with different ${\rm l_{min}}$ and different variables-brain plasticity simple ORPNN-PF with correlation coefficient correction, test results are ${\rm l_{min}=2-5}$ and ${\rm k_{max}=9}$ respectively}
\label{table7}
\resizebox{\textwidth}{!}{
\begin{tabular}{|c|c|c|c|c|c|c|c|c|c|c|c|c|}
        \hline
        & \diagbox{V.L.}{S.L.} & S.1 & S.2 & S.3 & S.4 & S.5 & S.6 & S.7 & S.8 & S.9 & Std    & correlation coefficients\\
\hline
                        \multirow{4}{*}{${\rm l_{min}}=2$} & V.1                                                                     & \textcolor{green}23\textcolor{green}33   & \textcolor{green}77\textcolor{green}77   & \textcolor{green}78\textcolor{green}88   & \textcolor{green}65\textcolor{green}44   & \textcolor{green}56\textcolor{green}66   & \textcolor{green}44\textcolor{green}54   & \textcolor{green}33\textcolor{green}33   & \textcolor{green}44\textcolor{green}44   & \textcolor{green}64\textcolor{green}45   & 1.7638 & 0.9427 \\
                        & V.2                                                                     & \textcolor{green}66\textcolor{green}66   & \textcolor{green}33\textcolor{green}33   & \textcolor{green}65\textcolor{green}66   & \textcolor{green}66\textcolor{green}67   & \textcolor{green}33\textcolor{green}33   & \textcolor{green}46\textcolor{green}66   & \textcolor{green}54\textcolor{green}54   & \textcolor{green}73\textcolor{green}33   & \textcolor{green}48\textcolor{green}66   & 1.6159 & 0.9545                        \\
                        & V.3                                                                     & \textcolor{green}54\textcolor{green}44   & \textcolor{green}44\textcolor{green}55   & \textcolor{green}43\textcolor{green}33   & \textcolor{green}68\textcolor{green}88   & \textcolor{green}44\textcolor{green}43   & \textcolor{green}54\textcolor{green}44   & \textcolor{green}57\textcolor{green}46   & \textcolor{green}45\textcolor{green}44   & \textcolor{green}75\textcolor{green}87   & 1.7638 & 0.9545                        \\
                        & V.4                                                                     & \textcolor{green}43\textcolor{green}33   & \textcolor{green}44\textcolor{green}44   & \textcolor{green}67\textcolor{green}77   & \textcolor{green}46\textcolor{green}66   & \textcolor{green}97\textcolor{green}77   & \textcolor{green}33\textcolor{green}33   & \textcolor{green}54\textcolor{green}55   & \textcolor{green}44\textcolor{green}33   & \textcolor{green}56\textcolor{green}66   & 1.6915 & 0.9622                        \\
\hline
                        \multirow{4}{*}{${\rm l_{min}}=3$} & V.1                                                                     & \textcolor{green}55\textcolor{green}55   & \textcolor{green}44\textcolor{green}44   & \textcolor{green}66\textcolor{green}66   & \textcolor{green}66\textcolor{green}66   & \textcolor{green}55\textcolor{green}55   & \textcolor{green}55\textcolor{green}55   & \textcolor{green}55\textcolor{green}45   & \textcolor{green}44\textcolor{green}44   & \textcolor{green}44\textcolor{green}54   & 0.7817 & 0.9451 \\
                        & V.2                                                                     & \textcolor{green}55\textcolor{green}55   & \textcolor{green}66\textcolor{green}66   & \textcolor{green}45\textcolor{green}54   & \textcolor{green}55\textcolor{green}55   & \textcolor{green}43\textcolor{green}33   & \textcolor{green}56\textcolor{green}66   & \textcolor{green}55\textcolor{green}55   & \textcolor{green}44\textcolor{green}44   & \textcolor{green}65\textcolor{green}56   & 1.0541 & 0.9451                        \\
                        & V.3                                                                     & \textcolor{green}66\textcolor{green}66   & \textcolor{green}44\textcolor{green}44   & \textcolor{green}55\textcolor{green}55   & \textcolor{green}66\textcolor{green}66   & \textcolor{green}44\textcolor{green}44   & \textcolor{green}33\textcolor{green}33   & \textcolor{green}66\textcolor{green}66   & \textcolor{green}44\textcolor{green}44   & \textcolor{green}66\textcolor{green}66   & 1.1667 & 0.9451                        \\
                        & V.4                                                                     & \textcolor{green}44\textcolor{green}44   & \textcolor{green}55\textcolor{green}55   & \textcolor{green}55\textcolor{green}55   & \textcolor{green}66\textcolor{green}66   & \textcolor{green}55\textcolor{green}55   & \textcolor{green}55\textcolor{green}55   & \textcolor{green}55\textcolor{green}55   & \textcolor{green}55\textcolor{green}55   & \textcolor{green}44\textcolor{green}44   & 0.6009 & 0.9488                        \\
\hline
                        \multirow{4}{*}{${\rm l_{min}}=4$} & V.1                                                                     & \textcolor{green}56\textcolor{green}55   & \textcolor{green}54\textcolor{green}44   & \textcolor{green}55\textcolor{green}55   & \textcolor{green}54\textcolor{green}44   & \textcolor{green}44\textcolor{green}44   & \textcolor{green}45\textcolor{green}55  & \textcolor{green}55\textcolor{green}55   & \textcolor{green}66\textcolor{green}66   & \textcolor{green}35\textcolor{green}66   & 0.7817 & 0.9288 \\
                        & V.2                                                                     & \textcolor{green}55\textcolor{green}55   & \textcolor{green}55\textcolor{green}55   & \textcolor{green}78\textcolor{green}77   & \textcolor{green}54\textcolor{green}44   & \textcolor{green}55\textcolor{green}55   & \textcolor{green}44\textcolor{green}55  & \textcolor{green}54\textcolor{green}44   & \textcolor{green}44\textcolor{green}44   & \textcolor{green}45\textcolor{green}55   & 0.9280 & 0.9368                        \\
                        & V.3                                                                     & \textcolor{green}66\textcolor{green}66   & \textcolor{green}44\textcolor{green}44   & \textcolor{green}66\textcolor{green}66   & \textcolor{green}44\textcolor{green}44   & \textcolor{green}55\textcolor{green}55   & \textcolor{green}44\textcolor{green}44  & \textcolor{green}55\textcolor{green}55   & \textcolor{green}55\textcolor{green}55   & \textcolor{green}55\textcolor{green}55   & 0.7817 & 0.9261                        \\
                        & V.4                                                                     & \textcolor{green}44\textcolor{green}44   & \textcolor{green}77\textcolor{green}77   & \textcolor{green}54\textcolor{green}44   & \textcolor{green}55\textcolor{green}55   & \textcolor{green}44\textcolor{green}44   & \textcolor{green}55\textcolor{green}55  & \textcolor{green}55\textcolor{green}55   & \textcolor{green}55\textcolor{green}55   & \textcolor{green}55\textcolor{green}55   & 0.9280 & 0.9364                        \\
\hline
                        \multirow{4}{*}{${\rm l_{min}}=5$} & V.1                                                                     & 5   & 5   & 5   & 5   & 5   & 5   & 5   & 5   & 4\footnotemark[1]  & 0.3333 & 0.8632 \\
                        & V.2                                                                     & 5   & 5   & 5   & 5   & 5   & 5   & 5   & 5   & 4   & 0.3333 & 0.8632                        \\
                        & V.3                                                                     & 5   & 5   & 5   & 5   & 5   & 5   & 5   & 5   & 4   & 0.3333 & 0.8851                        \\
                        & V.4                                                                     & 5   & 5   & 5   & 5   & 5   & 5   & 5   & 5   & 4   & 0.3333 & 0.8728              \\         
\hline
\end{tabular}
}
\end{table}

\section{The simplified higher-lower-order Deep Learning model}

The relatively inferior and good solution is also introduced as the emotional memory. If the memory of the downstream first cortex is the brain architecture, then the memory of the upstream $n$th cortex may be approximately the $n-1$st derivative of the brain architecture, and the emotional memory will gradually decline from the downstream to the upstream brain regions or from output to input. For example, the short-term emotional memory of the downstream brain regions may be the brain architecture of the relatively good s or relatively inferior, while the long-term emotional memory of the upstream brain regions may be the gradient of the brain architecture of the relatively good or relatively inferior. For the higher-order brain functions, we use the second-order Taylor approximation to derive the Newton Method, and for the lower-order brain, we use the first-order Taylor approximation to obtain the Gradient Descent Method. 

Considering the high-order cognition. The simple signals in the prefrontal cortex with the second-order Taylor approximation, the Deep Learning model is slightly more complex\cite{bib41}, and of course there are ways to avoid directly calculating the Hessian Matrix, such as using the Quasi-Newtonian Method.

The formula \eqref{eq6}, \eqref{eq7} and \eqref{eq8} for the Deep Learning model for upstream and downstream brain regions. If the upstream brain regions consider second-order Taylor approximation, $g(m,k)$ needs to be replaced with $[h(m,k)]^{-1} \times g(m,k)$. 

We consider a simpler method to implement higher-order rational brain and lower-order emotional brain to obtain a general and simple Deep Learning model, only consider the higher-order memory and affection\cite{bib41}. The Back Propagation only considers the Gradient Descent Method. See formula \eqref{eq11}. The synaptic connection weights $w(m,k)$ and range weights $r(m,k)$, $k_1$ from the upstream brain regions to the downstream brain regions $k_{\rm max}$, and from the signals of input layer $l_1$ to the output layer $l_{\rm max}$ which belong to a different brain region $k$. The mnemonic parameters of upstream brain regions, which are related to low-frequency waves, $ p_{g_r(m,k)_{better}} > p_{r(m,k)_{better}}b > p_{g_r(m,k)_{worse}} > p_{r(m,k)_{worse}}$, and the mnemonic parameters of downstream brain regions, which are related to high-frequency waves, $ p_{g_r(m,k)_{better}} < p_{r(m,k)_{better}}b < p_{g_r(m,k)_{worse}} < p_{r(m,k)_{worse}}$. The energy parameters of short-term memory and long-term memory both will decrease when the number of regions increases the upstream brain regions of long-term memory have more low-frequency waves, and the energy parameters of the upstream brain regions are greater than downstream brain regions due to better cognitive abilities. The simplified higher lower order Deep Learning model is shown in Fig. \ref{fig20}.

In addition to the decreasing calculations by formula \eqref{eq11}, the energy parameters $p_5$ or $p_6$ can both be written in the activation function by considering the movement of the logarithmic spiral, that is $f(t+1)=f(t)+\tanh[{\rm \Delta} f(t)]$ is replaced by $f(t+1)=f(t)+\tanh[{\rm \Delta}f(p_5 t)]$ , $p_5=\frac {d^{k_{max}-k} ae^{b\theta}}{d\theta^{k_{max}-k}}=ab^{k_{max}-k} e^{b\theta}$, so the derivation gets $f'(t+1)=ab^{k_{max}-k+1} e^{b\theta} [1-f(p_5 t)^2 ]$, higher-order energy $b>1$. $p_6=\frac{d^k ae^{b\theta}}{d\theta^k}=ab^k e^{b\theta}$, and $f'(t+1)=ab^{k+1} e^{b\theta} [1-f(p_6 t)^2]$, lower-order energy $b<1$. The Synaptic activities in different brain regions are shown in Fig. \ref{fig18}, Synapses are more active in upstream brain regions. We can also calculate mnemonic parameter $p_1=\frac{d^{k_{max}-k} ae^{b\theta}}{d\theta^{k_{max}-k}}=ab^{k_{max}-k} e^{b\theta}$,$b>1$ and $p_4=\frac{d^{k_{max}-k} ae^{b\theta}} {d\theta^{k_{max}-k}}=ab^{k_{max}-k} e^{b\theta}$,$b<1$. This allows us to calculate $p_1$ to $p_6$ by the movement of logarithmic spiral in brain regions, with each $a$ and $b$ of the 6 parameters.

The $p_1=\sin(\Phi_1 + \frac{\pi \times k}{k_{\rm max}})/4+1.25$ in formula \eqref{eq11} as the parameter of the relatively good long-term memory from the upstream brain regions to the downstream brain regions, which is weakening from 1.5 to 1, so that the upstream brain regions are more rational and higher-order relate to low-frequency waves. The $p_4=\sin(\Phi_3 + \frac{\pi \times k}{k_{\rm max}})/5+1.1$ in formula \eqref{eq11} as the parameter of the relatively inferior short-term memory from the upstream brain regions to the downstream brain regions, which is increasing from 0.9 to 1.3, so that the downstream brain regions are more lower-order and emotional relate to high-frequency waves, $\Phi_1 = \frac{\pi}{2}, \Phi_2 = \frac{3\pi}{2}$, because of more higher-order relatively good memory $g_r (m,k)_{better}$ in upstream brain regions, and more lower-order relatively inferior memory $r(m,k)_{worse}$ in downstream brain regions. The higher lower order affection and memory parameters in a circuit are shown in Fig. \ref{fig15}. The lower order of relatively good $r(m,k)_{worse}$ is calculated by the $p_2$, and the higher order of relatively inferior $g_r (m,k)_{worse}$ is calculated by $p_3$. The $p_2$ and $p_3$ are the two equal points of $p_1$ and $p_4$.

The $(rand>0.35+\frac{0.45 \times k}{k_{\rm max}} )$ in formula \eqref{eq11} indicated that neural circuits had smaller frequencies of retrieving short-term memory $r(m,k)_{better}$ and $r(m,k)_{worse}$, it becomes smaller as the number of regions increases due to the low energy in the downstream brain regions. But larger frequencies of retrieving long-term memory $g_r (m,k)_{better}$ and $g_r (m,k)_{worse}$.

The Quantum Entanglement may occur between the positive and negative emotions by heartbeat and the synaptic strength are affected by changes in the brain potentials, and superposition of the wave functions, then the wave function propagates and collapses due to the barriers and reflection of exponential decay in the brain.

\begin{equation}
\begin{aligned}
r(m,k) = r(m,k) + [r(m,k)_{better}-r(m,k)] \times rand \times(rand>0.35) + P
\end{aligned}
\label{eq6}
\end{equation}

\begin{equation}
\begin{aligned}
w(m,k)&=w(m,k)-g_w(m,k)\\
&+[MW_1 \times g_w{(m,k)}_{better}+MW_2 \times g_w{(m,k)}_{worse}-g_w(m,k)]\\
&\times M, MW_1=0.5,MW_2=0.5\\
\end{aligned}
\label{eq7}
\end{equation}

\begin{equation}
\begin{aligned}
r(m,k)&=r(m,k)-g_r(m,k)\\
&+[MW_1 \times g_r{(m,k)}_{better}+MW_2 \times g_r{(m,k)}_{worse}-g_r(m,k)]\\
&\times M, MW_1=0.5,MW_2=0.5\\
\end{aligned}
\label{eq8}
\end{equation}



\begin{equation}
\begin{aligned}
r(m,k)&=r(m,k)-g_r(m,k)\\
&+[MW_1 \times g_r{(m,k)}_{better} \times p_1 + MW_2 \times g_r{(m,k)}_{worse}\times p_3\\
&-(p_1+p_3) \times g_r(m,k)] \times M \times p_5\\
&+[MW_3 \times r{(m,k)}_{better} \times p_2 + MW_4 \times r{(m,k)}_{worse}\times p_4\\
&-(p_2+p_4) \times r(m,k)] \times M \times p_6 \times (rand>0.35 + \frac{0.45 \times k}{k_{\rm max}}), MW_1\\
&=0.5, MW_2=0.5,MW_3=0.5,MW_4=0.5\\
\end{aligned}
\label{eq11}
\end{equation}

\begin{equation}
        p_{1}=[\sin(\Phi_1 + \frac{\pi \times k}{k_{\rm max}})/4+1.25]
\end{equation}

\begin{equation}
        p_{2}=p_1+\frac{(p_4-p_1)}{3}
\end{equation}

\begin{equation}
        p_{3}=p_1+\frac{2 \times (p_4-p_1)}{3}
\end{equation}

\begin{equation}
        p_{4}=[\sin(\Phi_3 - \frac{\pi \times k}{k_{\rm max}})/5+1.1]
\end{equation}

\begin{equation}
        p_{5}=[\sin(\Phi_1 + \frac{\pi \times k}{k_{\rm max}})/2+0.7]
\end{equation}

\begin{equation}
        p_{6}=[\sin(\Phi_1 + \frac{\pi \times k}{k_{\rm max}})/2+0.7]
\end{equation}

\begin{figure}%
\centering
\includegraphics[width=0.9\textwidth]{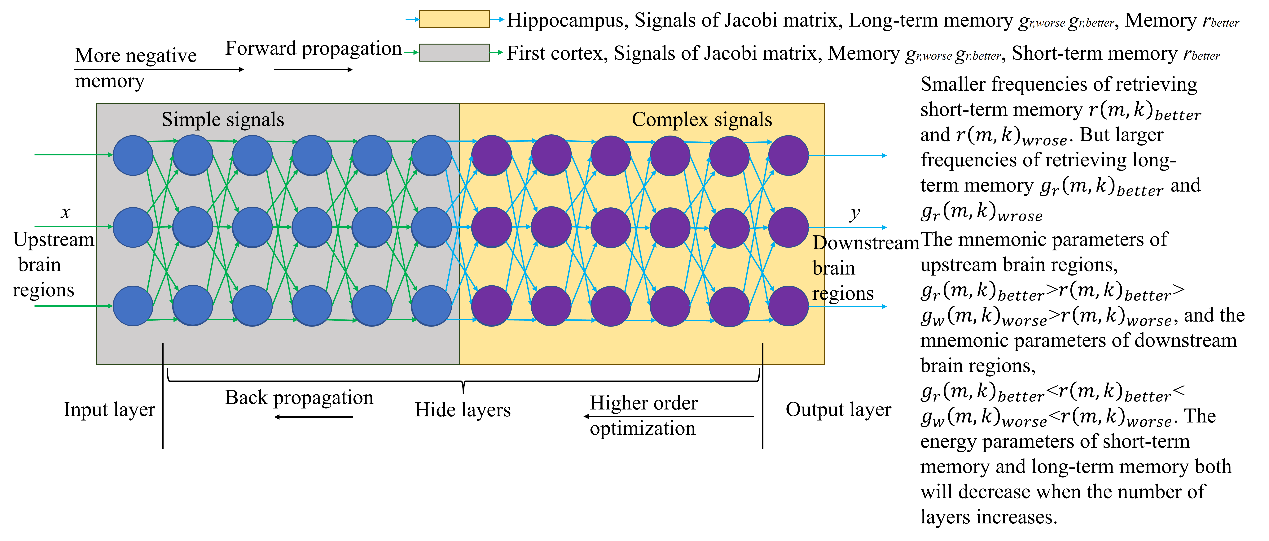}
\includegraphics[width=0.9\textwidth]{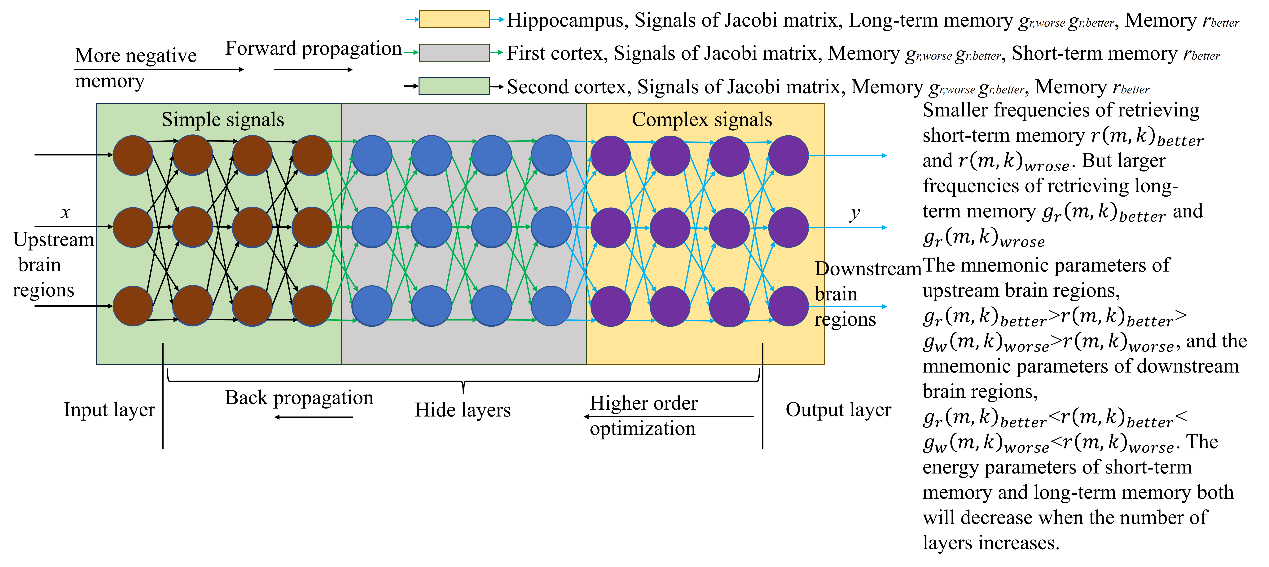}
\caption{The simplified higher lower order Deep Learning model}
\label{fig14}
\end{figure}

\begin{figure}%
        \centering
        \includegraphics[width=0.9\textwidth]{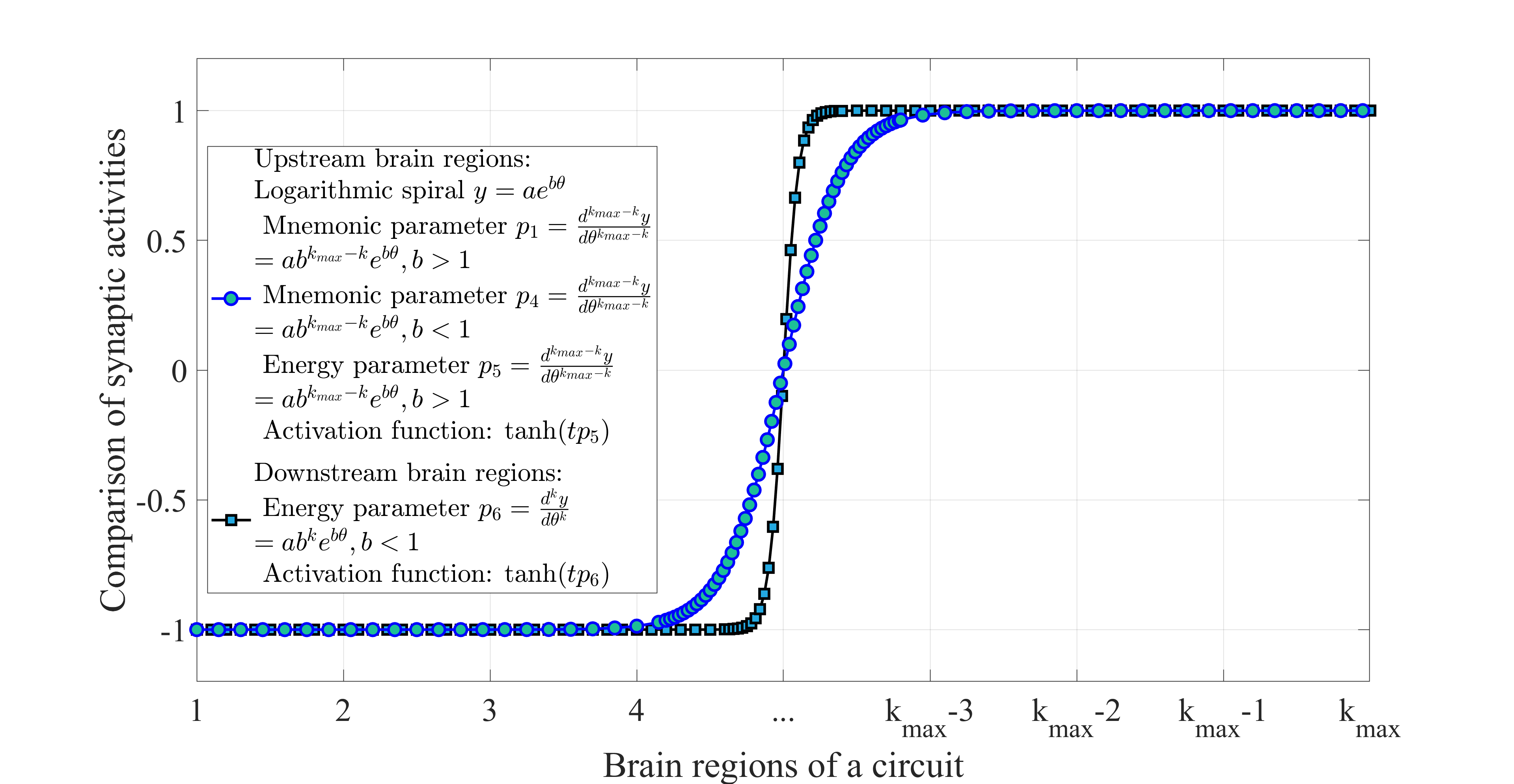}
        \caption{The Synaptic activities in different brain regions}
        \label{fig18}
\end{figure}

\begin{figure}%
\centering
\includegraphics[width=0.9\textwidth]{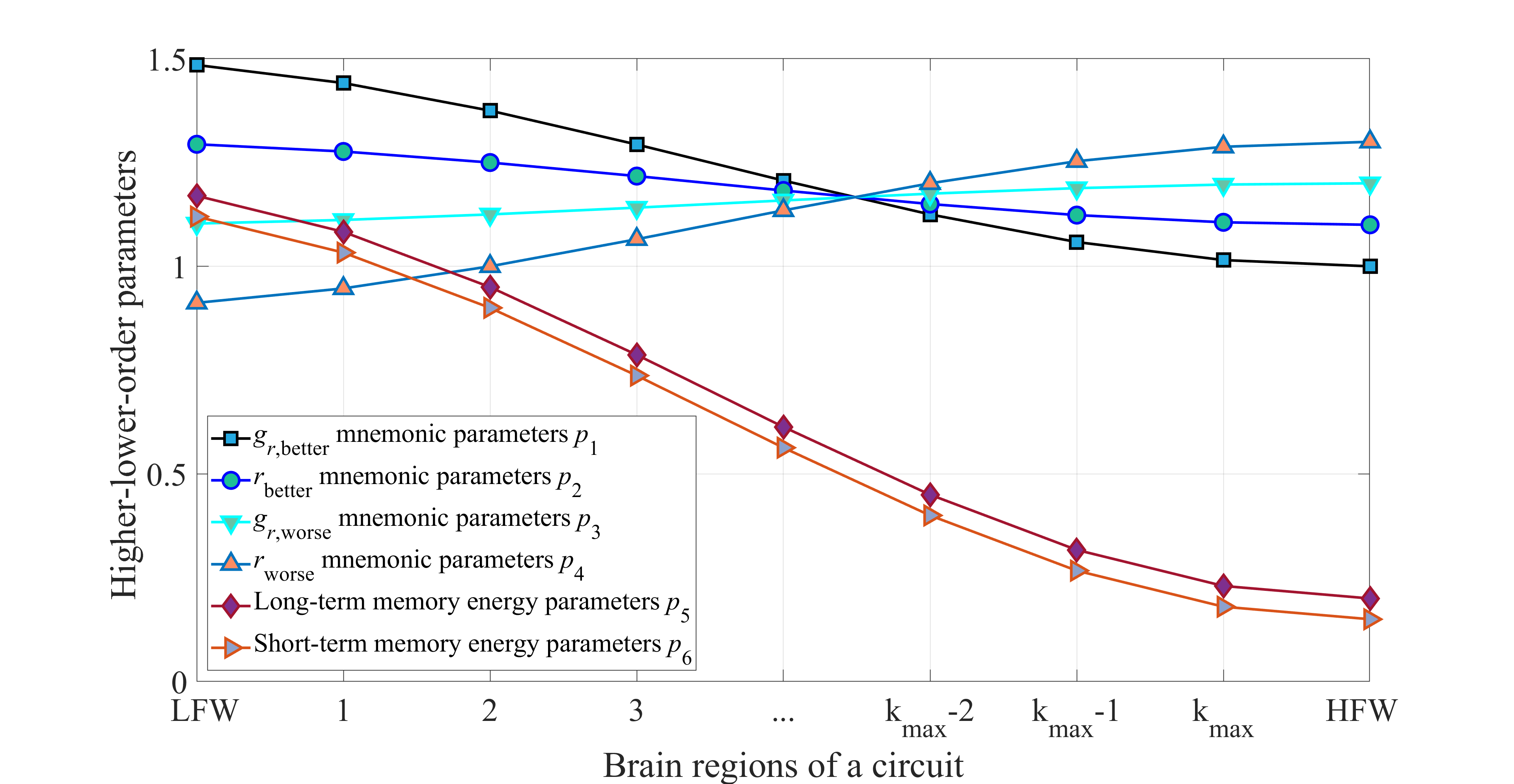}
\caption{The higher lower order affection and memory parameters in a circuit}
\label{fig15}
\end{figure}

\begin{figure}
\centering
\begin{minipage}{0.8\textwidth}
        \centering
        \includegraphics[width=0.8\textwidth]{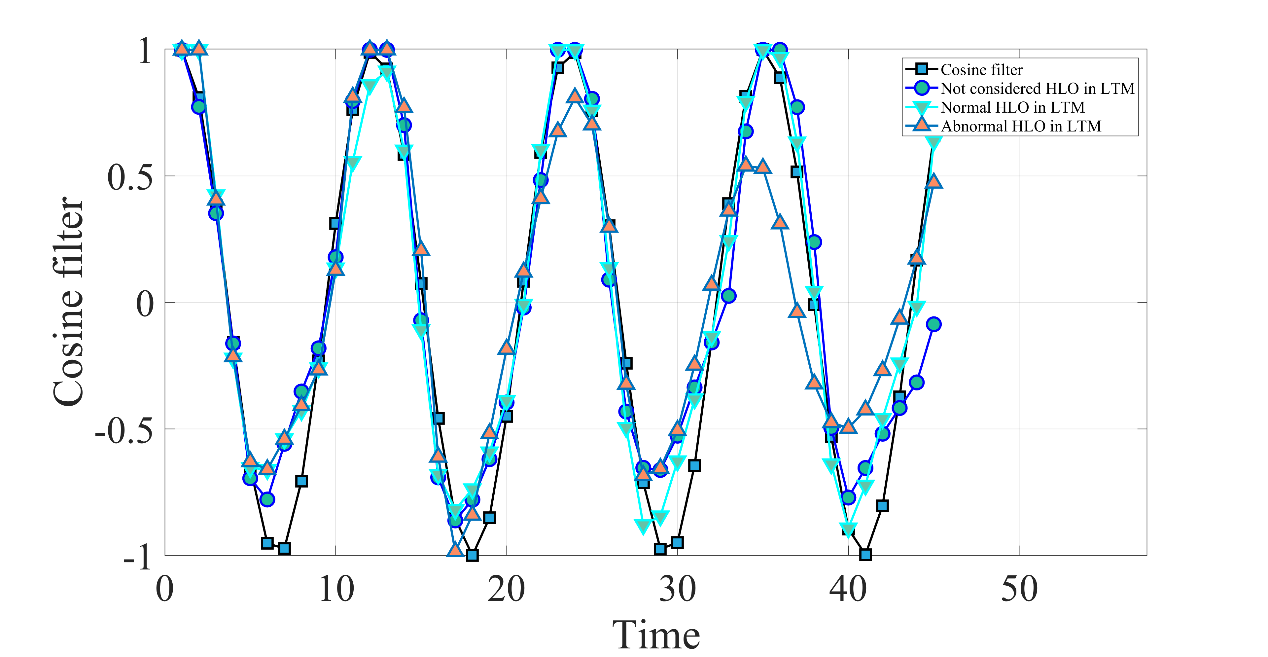}
\end{minipage}
\begin{minipage}{0.8\textwidth}
        \centering
        \includegraphics[width=0.8\textwidth]{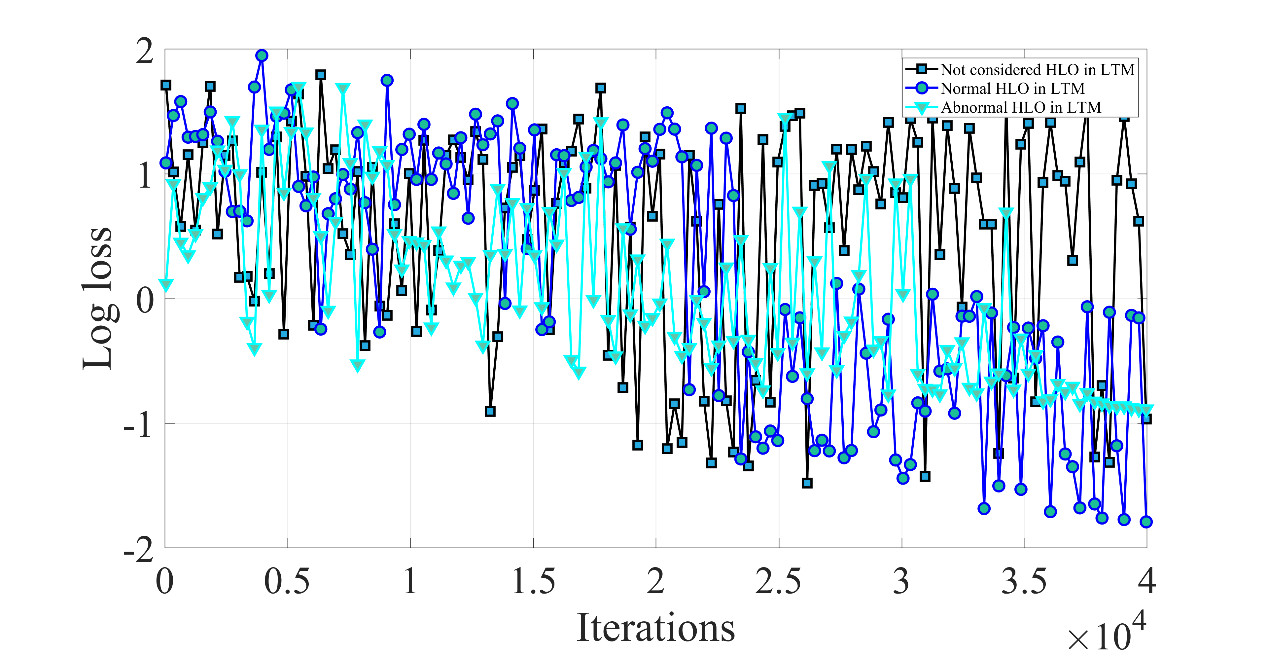}
        \caption{Only long-term memory is considered}
\end{minipage}
\begin{minipage}{0.8\textwidth}
        \centering
        \includegraphics[width=0.8\textwidth]{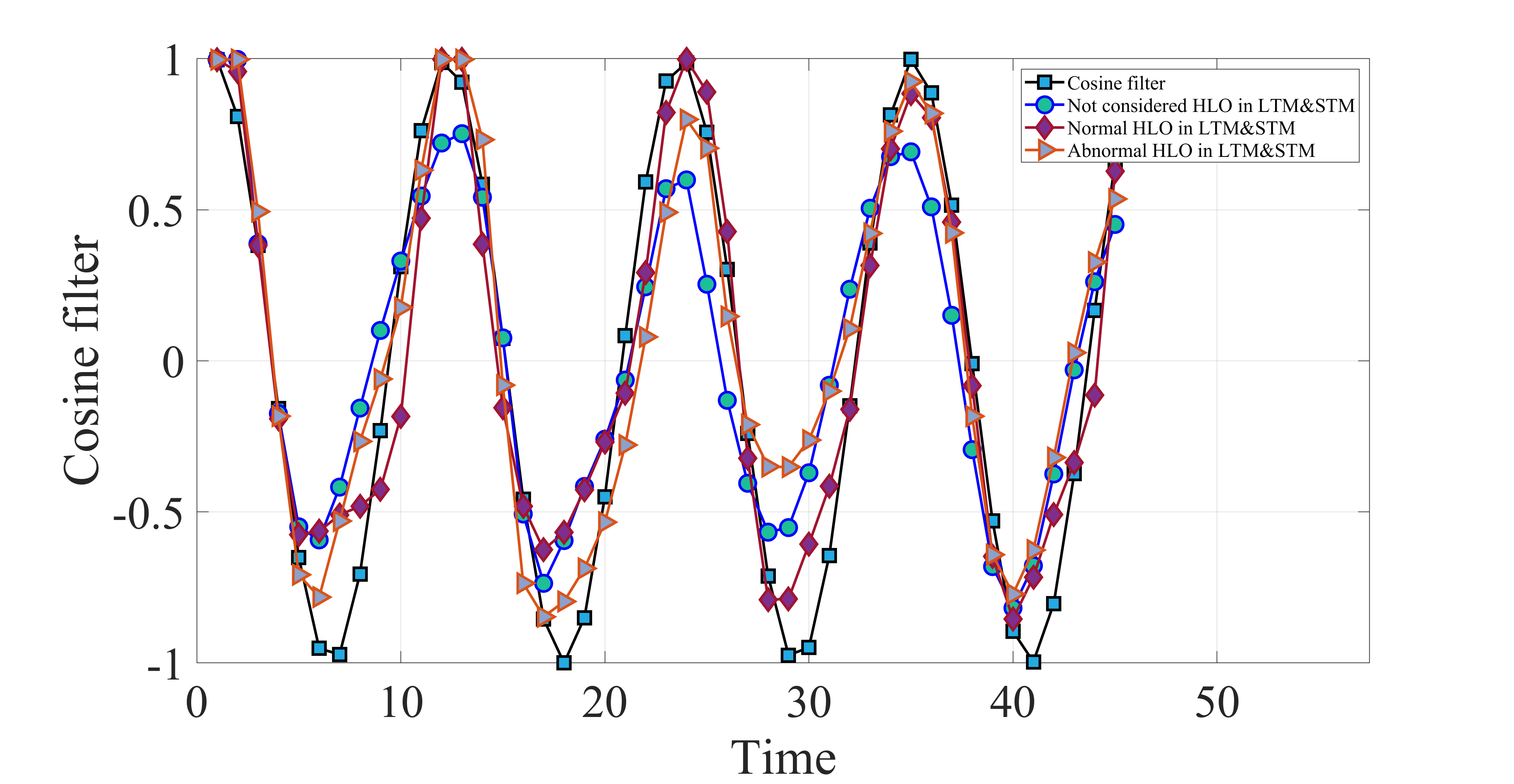}
\end{minipage}
\begin{minipage}{0.8\textwidth}
        \centering
        \includegraphics[width=0.8\textwidth]{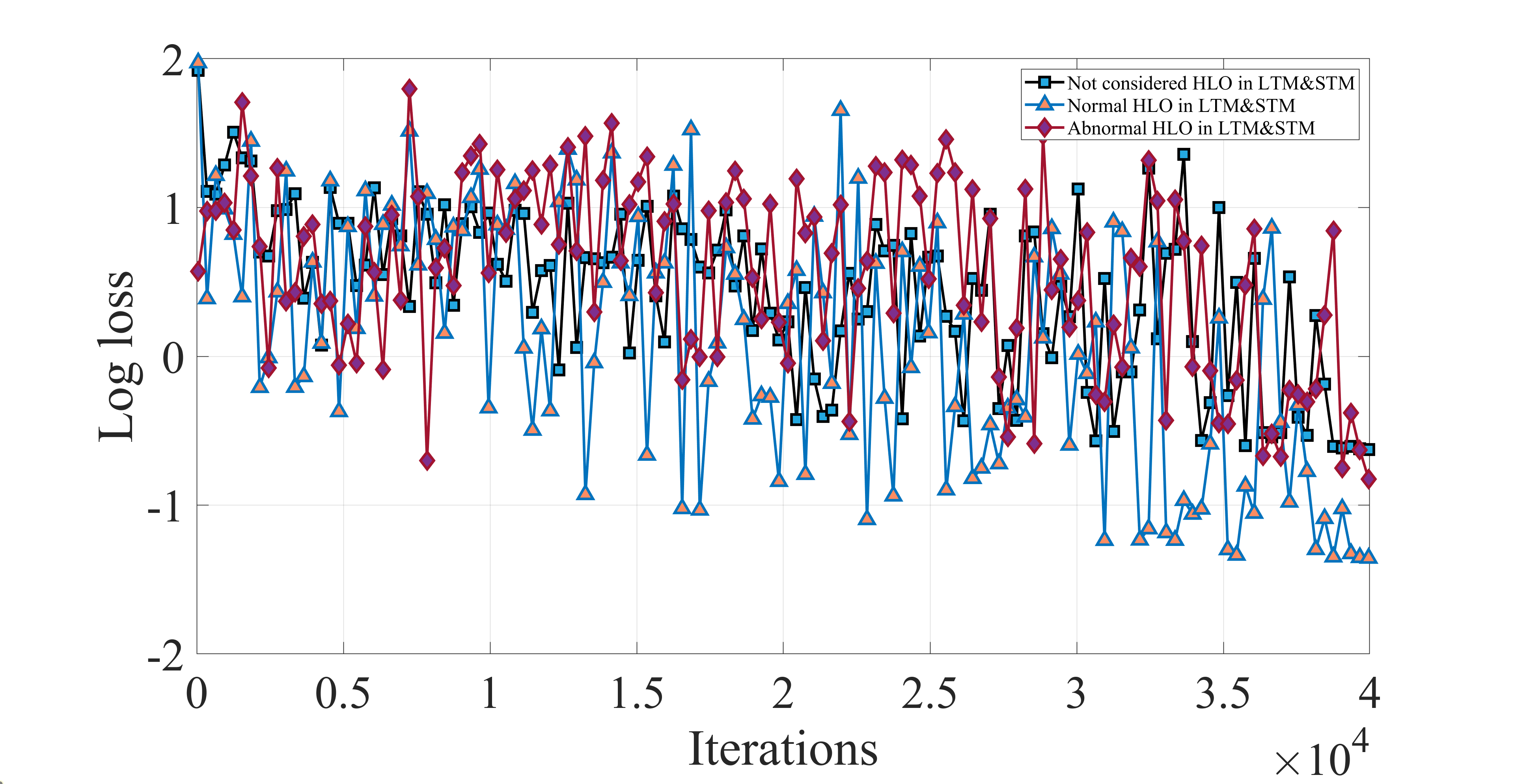}
        \caption{Both short-term and long-term memory are considered}
\end{minipage}
\caption{The comparisons of higher lower order brain by the probability of $rand>0.35$ astrocytes phagocytose synapses}
\label{fig16}
\end{figure}

\begin{figure}%
        \centering
        \begin{minipage}{0.8\textwidth}
                \centering
                \includegraphics[width=0.8\textwidth]{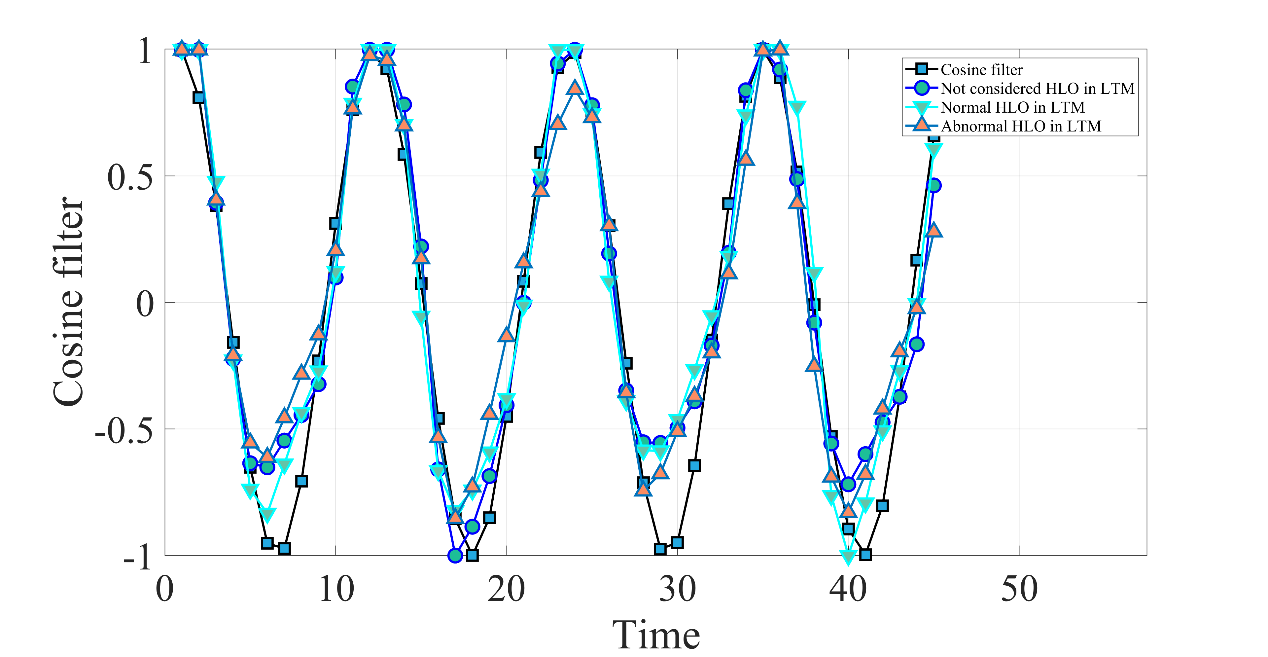}
        \end{minipage}
        \begin{minipage}{0.8\textwidth}
                \centering
                \includegraphics[width=0.8\textwidth]{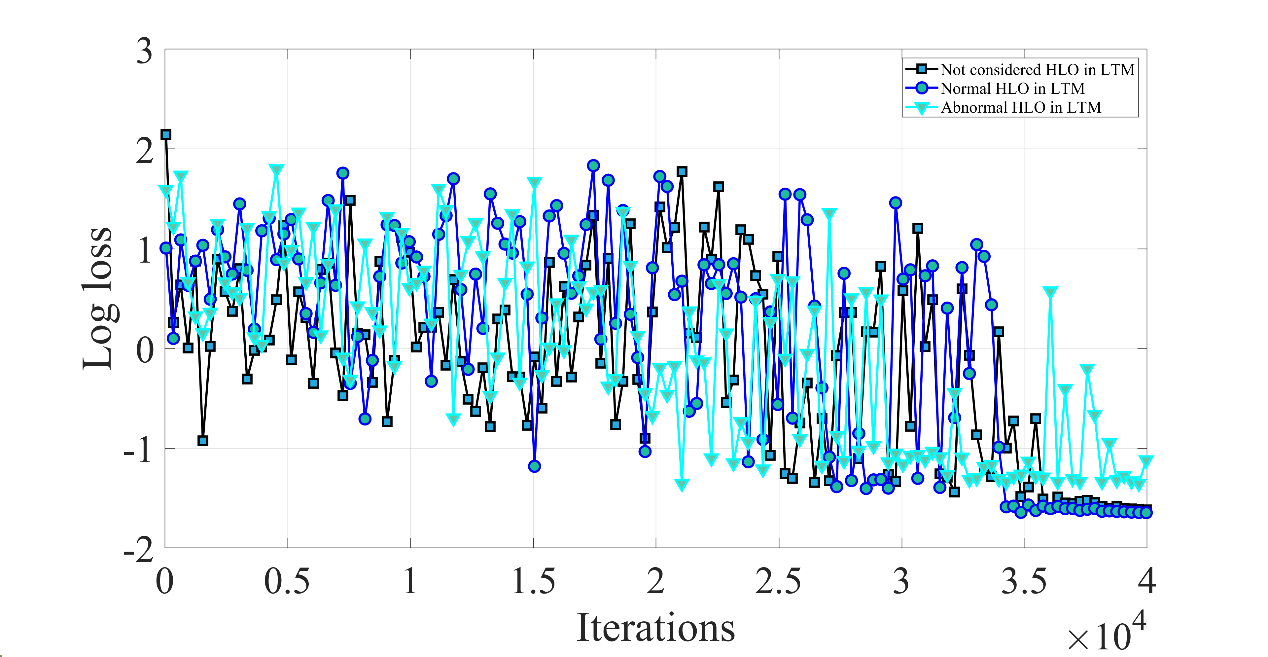}
                \caption{Only long-term memory is considered}
        \end{minipage}
        \begin{minipage}{0.8\textwidth}
                \centering
                \includegraphics[width=0.8\textwidth]{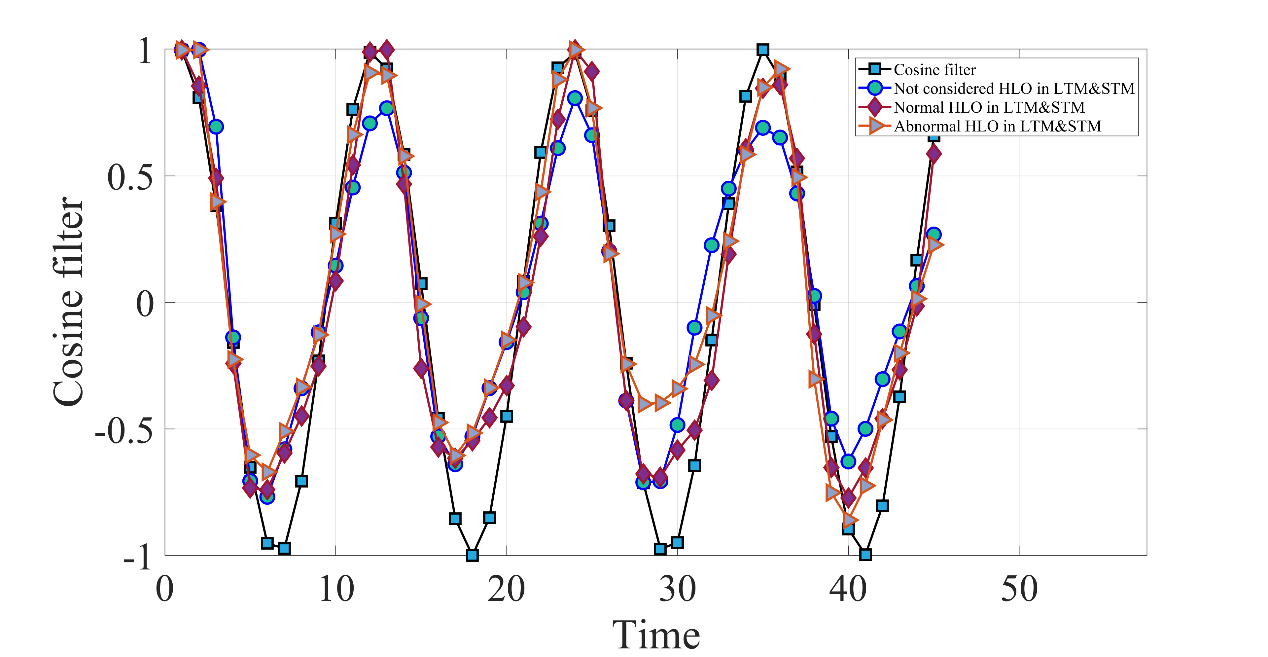}
        \end{minipage}
        \begin{minipage}{0.8\textwidth}
                \centering
                \includegraphics[width=0.8\textwidth]{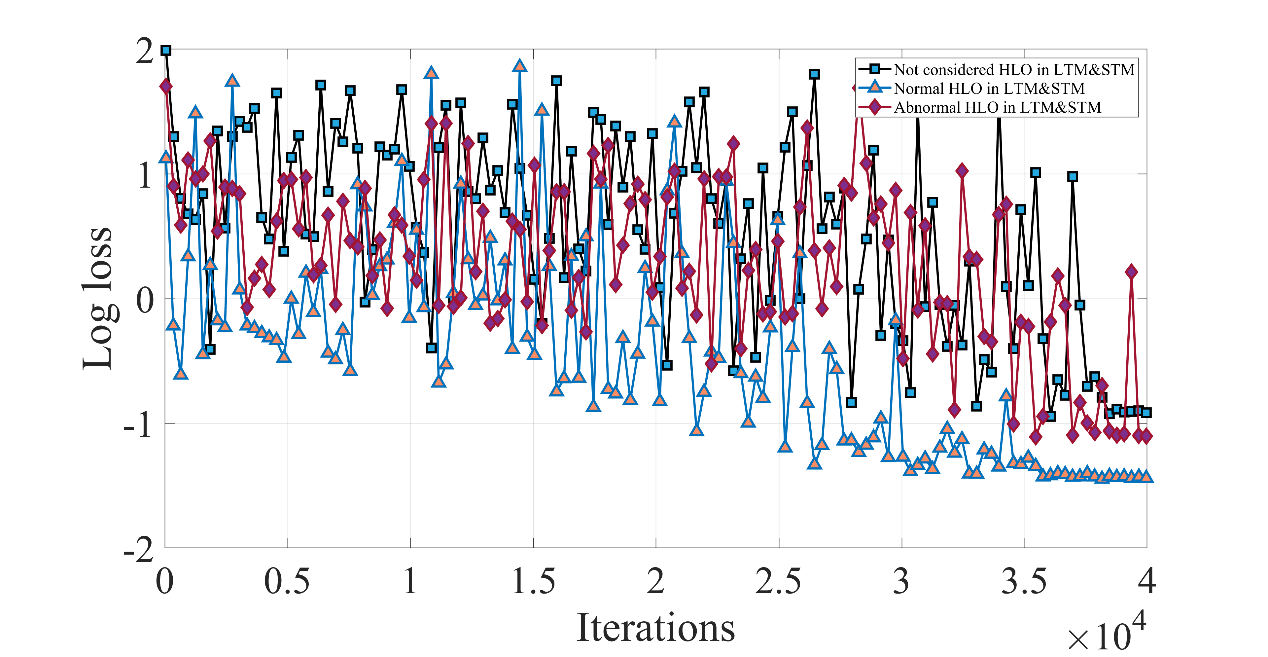}
                \caption{Both short-term and long-term memory are considered}
        \end{minipage}
\caption{The comparisons of higher lower order brain by the probability of $rand>0.1$ astrocytes phagocytose synapses}
\label{fig17}
\end{figure}

The simulation results of formula \eqref{eq6},\eqref{eq7},\eqref{eq8} and formula \eqref{eq11} are shown in Fig. \ref{fig16},\ref{fig17}, including the higher-lower-order brain situations of both short-term and long-term memory are considered, and only long-term memory is considered. If the probability of $(rand>0.35)$ in formula \eqref{eq11} is changed to $(rand>0.1)$, when more times of astrocytes phagocytose synapses, the comparisons of Fig. \ref{fig17} and Table \ref{table10}, \ref{table11} are not obvious.

Table \ref{table8}, \ref{table9}, \ref{table10}, \ref{table11} give synaptic strength and correlation coefficients of each algorithm after termination of iterations, respectively.

\begin{table}[]
\caption{The synaptic strength of methods by the probability of $rand>0.35$ astrocytes phagocytose synapses}
\label{table8}
\centering
\renewcommand\arraystretch{0.6}
\resizebox{\textwidth}{!}{
\begin{tabular}{|c|c|ccccccccc|}
\hline
{\begin{tabular}{@{}c@{}}Method\end{tabular}} & {\begin{tabular}{@{}c@{}}Neuron types\end{tabular}} & \multicolumn{9}{|c|}{Synaptic strength} \\ \cline{3-11} 
 &  & \multicolumn{1}{c|}{1} & \multicolumn{1}{c|}{2} & \multicolumn{1}{c|}{3} & \multicolumn{1}{c|}{4} & \multicolumn{1}{c|}{5} & \multicolumn{1}{c|}{6} & \multicolumn{1}{c|}{7} & \multicolumn{1}{c|}{8} & 9 \\ \hline
\multirow{4}{*}{NCHLOILTM} & 1 & \multicolumn{1}{c|}{5} & \multicolumn{1}{c|}{5} & \multicolumn{1}{c|}{5} & \multicolumn{1}{c|}{4} & \multicolumn{1}{c|}{5} & \multicolumn{1}{c|}{5} & \multicolumn{1}{c|}{5} & \multicolumn{1}{c|}{5} & 5 \\ \cline{2-11} 
 & 2 & \multicolumn{1}{c|}{4} & \multicolumn{1}{c|}{3} & \multicolumn{1}{c|}{5} & \multicolumn{1}{c|}{6} & \multicolumn{1}{c|}{5} & \multicolumn{1}{c|}{5} & \multicolumn{1}{c|}{5} & \multicolumn{1}{c|}{5} & 6 \\ \cline{2-11} 
 & 3 & \multicolumn{1}{c|}{3} & \multicolumn{1}{c|}{5} & \multicolumn{1}{c|}{6} & \multicolumn{1}{c|}{6} & \multicolumn{1}{c|}{5} & \multicolumn{1}{c|}{4} & \multicolumn{1}{c|}{6} & \multicolumn{1}{c|}{3} & 6 \\ \cline{2-11} 
 & 4 & \multicolumn{1}{c|}{5} & \multicolumn{1}{c|}{5} & \multicolumn{1}{c|}{5} & \multicolumn{1}{c|}{4} & \multicolumn{1}{c|}{5} & \multicolumn{1}{c|}{5} & \multicolumn{1}{c|}{5} & \multicolumn{1}{c|}{5} & 5 \\ \hline
\multirow{4}{*}{NHLOILTM} & 1 & \multicolumn{1}{c|}{4} & \multicolumn{1}{c|}{5} & \multicolumn{1}{c|}{6} & \multicolumn{1}{c|}{5} & \multicolumn{1}{c|}{5} & \multicolumn{1}{c|}{3} & \multicolumn{1}{c|}{6} & \multicolumn{1}{c|}{6} & 4 \\ \cline{2-11} 
 & 2 & \multicolumn{1}{c|}{4} & \multicolumn{1}{c|}{5} & \multicolumn{1}{c|}{4} & \multicolumn{1}{c|}{5} & \multicolumn{1}{c|}{4} & \multicolumn{1}{c|}{6} & \multicolumn{1}{c|}{5} & \multicolumn{1}{c|}{5} & 6 \\ \cline{2-11} 
 & 3 & \multicolumn{1}{c|}{5} & \multicolumn{1}{c|}{5} & \multicolumn{1}{c|}{6} & \multicolumn{1}{c|}{5} & \multicolumn{1}{c|}{6} & \multicolumn{1}{c|}{3} & \multicolumn{1}{c|}{4} & \multicolumn{1}{c|}{5} & 5 \\ \cline{2-11} 
 & 4 & \multicolumn{1}{c|}{4} & \multicolumn{1}{c|}{5} & \multicolumn{1}{c|}{6} & \multicolumn{1}{c|}{4} & \multicolumn{1}{c|}{4} & \multicolumn{1}{c|}{4} & \multicolumn{1}{c|}{5} & \multicolumn{1}{c|}{6} & 6 \\ \hline
\multirow{4}{*}{AHLOILTM} & 1 & \multicolumn{1}{c|}{5} & \multicolumn{1}{c|}{5} & \multicolumn{1}{c|}{5} & \multicolumn{1}{c|}{3} & \multicolumn{1}{c|}{5} & \multicolumn{1}{c|}{5} & \multicolumn{1}{c|}{5} & \multicolumn{1}{c|}{5} & 6 \\ \cline{2-11} 
 & 2 & \multicolumn{1}{c|}{4} & \multicolumn{1}{c|}{6} & \multicolumn{1}{c|}{5} & \multicolumn{1}{c|}{4} & \multicolumn{1}{c|}{4} & \multicolumn{1}{c|}{5} & \multicolumn{1}{c|}{6} & \multicolumn{1}{c|}{5} & 5 \\ \cline{2-11} 
 & 3 & \multicolumn{1}{c|}{4} & \multicolumn{1}{c|}{5} & \multicolumn{1}{c|}{4} & \multicolumn{1}{c|}{6} & \multicolumn{1}{c|}{6} & \multicolumn{1}{c|}{5} & \multicolumn{1}{c|}{4} & \multicolumn{1}{c|}{5} & 5 \\ \cline{2-11} 
 & 4 & \multicolumn{1}{c|}{3} & \multicolumn{1}{c|}{6} & \multicolumn{1}{c|}{6} & \multicolumn{1}{c|}{5} & \multicolumn{1}{c|}{6} & \multicolumn{1}{c|}{3} & \multicolumn{1}{c|}{6} & \multicolumn{1}{c|}{4} & 5 \\ \hline
\multirow{4}{*}{NCHLOILTM \& STM} & 1 & \multicolumn{1}{c|}{6} & \multicolumn{1}{c|}{3} & \multicolumn{1}{c|}{5} & \multicolumn{1}{c|}{5} & \multicolumn{1}{c|}{6} & \multicolumn{1}{c|}{4} & \multicolumn{1}{c|}{3} & \multicolumn{1}{c|}{6} & 6 \\ \cline{2-11} 
 & 2 & \multicolumn{1}{c|}{4} & \multicolumn{1}{c|}{3} & \multicolumn{1}{c|}{6} & \multicolumn{1}{c|}{5} & \multicolumn{1}{c|}{3} & \multicolumn{1}{c|}{5} & \multicolumn{1}{c|}{7} & \multicolumn{1}{c|}{5} & 6 \\ \cline{2-11} 
 & 3 & \multicolumn{1}{c|}{5} & \multicolumn{1}{c|}{5} & \multicolumn{1}{c|}{4} & \multicolumn{1}{c|}{6} & \multicolumn{1}{c|}{6} & \multicolumn{1}{c|}{5} & \multicolumn{1}{c|}{5} & \multicolumn{1}{c|}{5} & 3 \\ \cline{2-11} 
 & 4 & \multicolumn{1}{c|}{2} & \multicolumn{1}{c|}{6} & \multicolumn{1}{c|}{7} & \multicolumn{1}{c|}{4} & \multicolumn{1}{c|}{4} & \multicolumn{1}{c|}{3} & \multicolumn{1}{c|}{6} & \multicolumn{1}{c|}{6} & 6 \\ \hline
\multirow{4}{*}{NHLOILTM \& STM} & 1 & \multicolumn{1}{c|}{3} & \multicolumn{1}{c|}{5} & \multicolumn{1}{c|}{5} & \multicolumn{1}{c|}{5} & \multicolumn{1}{c|}{4} & \multicolumn{1}{c|}{6} & \multicolumn{1}{c|}{6} & \multicolumn{1}{c|}{5} & 5 \\ \cline{2-11} 
 & 2 & \multicolumn{1}{c|}{4} & \multicolumn{1}{c|}{4} & \multicolumn{1}{c|}{4} & \multicolumn{1}{c|}{5} & \multicolumn{1}{c|}{5} & \multicolumn{1}{c|}{6} & \multicolumn{1}{c|}{5} & \multicolumn{1}{c|}{4} & 7 \\ \cline{2-11} 
 & 3 & \multicolumn{1}{c|}{4} & \multicolumn{1}{c|}{4} & \multicolumn{1}{c|}{6} & \multicolumn{1}{c|}{4} & \multicolumn{1}{c|}{5} & \multicolumn{1}{c|}{4} & \multicolumn{1}{c|}{4} & \multicolumn{1}{c|}{6} & 7 \\ \cline{2-11} 
 & 4 & \multicolumn{1}{c|}{3} & \multicolumn{1}{c|}{4} & \multicolumn{1}{c|}{4} & \multicolumn{1}{c|}{4} & \multicolumn{1}{c|}{6} & \multicolumn{1}{c|}{5} & \multicolumn{1}{c|}{5} & \multicolumn{1}{c|}{7} & 6 \\ \hline
\multirow{4}{*}{AHLOILTM \& STM} & 1 & \multicolumn{1}{c|}{4} & \multicolumn{1}{c|}{5} & \multicolumn{1}{c|}{4} & \multicolumn{1}{c|}{3} & \multicolumn{1}{c|}{7} & \multicolumn{1}{c|}{5} & \multicolumn{1}{c|}{5} & \multicolumn{1}{c|}{6} & 5 \\ \cline{2-11} 
 & 2 & \multicolumn{1}{c|}{5} & \multicolumn{1}{c|}{5} & \multicolumn{1}{c|}{4} & \multicolumn{1}{c|}{6} & \multicolumn{1}{c|}{5} & \multicolumn{1}{c|}{4} & \multicolumn{1}{c|}{5} & \multicolumn{1}{c|}{5} & 5 \\ \cline{2-11} 
 & 3 & \multicolumn{1}{c|}{5} & \multicolumn{1}{c|}{6} & \multicolumn{1}{c|}{5} & \multicolumn{1}{c|}{5} & \multicolumn{1}{c|}{6} & \multicolumn{1}{c|}{4} & \multicolumn{1}{c|}{5} & \multicolumn{1}{c|}{5} & 3 \\ \cline{2-11} 
 & 4 & \multicolumn{1}{c|}{4} & \multicolumn{1}{c|}{6} & \multicolumn{1}{c|}{4} & \multicolumn{1}{c|}{4} & \multicolumn{1}{c|}{5} & \multicolumn{1}{c|}{5} & \multicolumn{1}{c|}{4} & \multicolumn{1}{c|}{5} & 7 \\ \hline
\end{tabular}%
}
\end{table}

\begin{table}[]
        \caption{The correlation coefficients of methods by the probability of $rand>0.35$ astrocytes phagocytose synapses}
        \resizebox{\textwidth}{!}{
        \label{table9}
        \begin{tabular}{|c|c|c|c|c|c|c|}
        \hline
        Method                  & \begin{tabular}[c]{@{}c@{}}NCHLOI\\    \\ LTM\end{tabular} & \begin{tabular}[c]{@{}c@{}}NHLOI\\    \\ LTM\end{tabular} & \begin{tabular}[c]{@{}c@{}}AHLOI\\    \\ LTM\end{tabular} & \begin{tabular}[c]{@{}c@{}}NCHLOI\\    \\ LTM\&STM\end{tabular} & \begin{tabular}[c]{@{}c@{}}NHLOI\\    \\ LTM\& STM\end{tabular} & \begin{tabular}[c]{@{}c@{}}AHLOI\\    \\ LTM\& STM\end{tabular} \\ \hline
        Correlation coefficient & 0.9482                                                     & 0.9723                                                    & 0.9343                                                    & 0.9219                                                          & 0.9587                                                          & 0.9303                                                          \\ \hline
        \end{tabular}
        }
\end{table}

\begin{table}[]
        \caption{The synaptic strength of methods by the probability of $rand>0.1$ astrocytes phagocytose synapses}
        \resizebox{\textwidth}{!}{
        \label{table10}
        \begin{tabular}{|c|c|ccccccccc|}
        \hline
        \multirow{2}{*}{Method}                                                          & \multirow{2}{*}{Neuron types} & \multicolumn{9}{c|}{Synaptic strength}                                                                                                                                                                    \\ \cline{3-11} 
                                                                                         &                               & \multicolumn{1}{c|}{1} & \multicolumn{1}{c|}{2} & \multicolumn{1}{c|}{3} & \multicolumn{1}{c|}{4} & \multicolumn{1}{c|}{5} & \multicolumn{1}{c|}{6} & \multicolumn{1}{c|}{7} & \multicolumn{1}{c|}{8} & 9 \\ \hline
        \multirow{4}{*}{NCHLOILTM}                                                       & 1                             & \multicolumn{1}{c|}{5} & \multicolumn{1}{c|}{5} & \multicolumn{1}{c|}{5} & \multicolumn{1}{c|}{5} & \multicolumn{1}{c|}{4} & \multicolumn{1}{c|}{6} & \multicolumn{1}{c|}{4} & \multicolumn{1}{c|}{5} & 5 \\ \cline{2-11} 
                                                                                         & 2                             & \multicolumn{1}{c|}{5} & \multicolumn{1}{c|}{5} & \multicolumn{1}{c|}{6} & \multicolumn{1}{c|}{4} & \multicolumn{1}{c|}{5} & \multicolumn{1}{c|}{3} & \multicolumn{1}{c|}{5} & \multicolumn{1}{c|}{6} & 5 \\ \cline{2-11} 
                                                                                         & 3                             & \multicolumn{1}{c|}{4} & \multicolumn{1}{c|}{5} & \multicolumn{1}{c|}{5} & \multicolumn{1}{c|}{5} & \multicolumn{1}{c|}{5} & \multicolumn{1}{c|}{5} & \multicolumn{1}{c|}{5} & \multicolumn{1}{c|}{5} & 5 \\ \cline{2-11} 
                                                                                         & 4                             & \multicolumn{1}{c|}{3} & \multicolumn{1}{c|}{6} & \multicolumn{1}{c|}{6} & \multicolumn{1}{c|}{3} & \multicolumn{1}{c|}{6} & \multicolumn{1}{c|}{5} & \multicolumn{1}{c|}{4} & \multicolumn{1}{c|}{5} & 6 \\ \hline
        \multirow{4}{*}{NHLOILTM}                                                        & 1                             & \multicolumn{1}{c|}{4} & \multicolumn{1}{c|}{6} & \multicolumn{1}{c|}{5} & \multicolumn{1}{c|}{4} & \multicolumn{1}{c|}{5} & \multicolumn{1}{c|}{5} & \multicolumn{1}{c|}{5} & \multicolumn{1}{c|}{5} & 5 \\ \cline{2-11} 
                                                                                         & 2                             & \multicolumn{1}{c|}{4} & \multicolumn{1}{c|}{5} & \multicolumn{1}{c|}{6} & \multicolumn{1}{c|}{4} & \multicolumn{1}{c|}{5} & \multicolumn{1}{c|}{5} & \multicolumn{1}{c|}{4} & \multicolumn{1}{c|}{5} & 6 \\ \cline{2-11} 
                                                                                         & 3                             & \multicolumn{1}{c|}{3} & \multicolumn{1}{c|}{5} & \multicolumn{1}{c|}{5} & \multicolumn{1}{c|}{5} & \multicolumn{1}{c|}{6} & \multicolumn{1}{c|}{4} & \multicolumn{1}{c|}{6} & \multicolumn{1}{c|}{5} & 5 \\ \cline{2-11} 
                                                                                         & 4                             & \multicolumn{1}{c|}{4} & \multicolumn{1}{c|}{5} & \multicolumn{1}{c|}{5} & \multicolumn{1}{c|}{4} & \multicolumn{1}{c|}{5} & \multicolumn{1}{c|}{5} & \multicolumn{1}{c|}{5} & \multicolumn{1}{c|}{5} & 6 \\ \hline
        \multirow{4}{*}{AHLOILTM}                                                        & 1                             & \multicolumn{1}{c|}{4} & \multicolumn{1}{c|}{6} & \multicolumn{1}{c|}{6} & \multicolumn{1}{c|}{4} & \multicolumn{1}{c|}{5} & \multicolumn{1}{c|}{5} & \multicolumn{1}{c|}{6} & \multicolumn{1}{c|}{4} & 4 \\ \cline{2-11} 
                                                                                         & 2                             & \multicolumn{1}{c|}{6} & \multicolumn{1}{c|}{5} & \multicolumn{1}{c|}{5} & \multicolumn{1}{c|}{5} & \multicolumn{1}{c|}{4} & \multicolumn{1}{c|}{6} & \multicolumn{1}{c|}{6} & \multicolumn{1}{c|}{3} & 4 \\ \cline{2-11} 
                                                                                         & 3                             & \multicolumn{1}{c|}{4} & \multicolumn{1}{c|}{6} & \multicolumn{1}{c|}{6} & \multicolumn{1}{c|}{4} & \multicolumn{1}{c|}{6} & \multicolumn{1}{c|}{6} & \multicolumn{1}{c|}{3} & \multicolumn{1}{c|}{4} & 5 \\ \cline{2-11} 
                                                                                         & 4                             & \multicolumn{1}{c|}{4} & \multicolumn{1}{c|}{5} & \multicolumn{1}{c|}{6} & \multicolumn{1}{c|}{5} & \multicolumn{1}{c|}{6} & \multicolumn{1}{c|}{4} & \multicolumn{1}{c|}{4} & \multicolumn{1}{c|}{5} & 5 \\ \hline
        \multirow{4}{*}{\begin{tabular}[c]{@{}c@{}}NCHLOILTM\\    \\ \&STM\end{tabular}} & 1                             & \multicolumn{1}{c|}{4} & \multicolumn{1}{c|}{5} & \multicolumn{1}{c|}{5} & \multicolumn{1}{c|}{4} & \multicolumn{1}{c|}{5} & \multicolumn{1}{c|}{6} & \multicolumn{1}{c|}{7} & \multicolumn{1}{c|}{4} & 4 \\ \cline{2-11} 
                                                                                         & 2                             & \multicolumn{1}{c|}{2} & \multicolumn{1}{c|}{6} & \multicolumn{1}{c|}{5} & \multicolumn{1}{c|}{6} & \multicolumn{1}{c|}{4} & \multicolumn{1}{c|}{6} & \multicolumn{1}{c|}{4} & \multicolumn{1}{c|}{6} & 5 \\ \cline{2-11} 
                                                                                         & 3                             & \multicolumn{1}{c|}{5} & \multicolumn{1}{c|}{5} & \multicolumn{1}{c|}{4} & \multicolumn{1}{c|}{6} & \multicolumn{1}{c|}{4} & \multicolumn{1}{c|}{4} & \multicolumn{1}{c|}{5} & \multicolumn{1}{c|}{7} & 4 \\ \cline{2-11} 
                                                                                         & 4                             & \multicolumn{1}{c|}{3} & \multicolumn{1}{c|}{5} & \multicolumn{1}{c|}{6} & \multicolumn{1}{c|}{5} & \multicolumn{1}{c|}{5} & \multicolumn{1}{c|}{5} & \multicolumn{1}{c|}{4} & \multicolumn{1}{c|}{5} & 6 \\ \hline
        \multirow{4}{*}{\begin{tabular}[c]{@{}c@{}}NHLOILTM\\    \\ \& STM\end{tabular}} & 1                             & \multicolumn{1}{c|}{3} & \multicolumn{1}{c|}{4} & \multicolumn{1}{c|}{5} & \multicolumn{1}{c|}{5} & \multicolumn{1}{c|}{5} & \multicolumn{1}{c|}{6} & \multicolumn{1}{c|}{5} & \multicolumn{1}{c|}{6} & 5 \\ \cline{2-11} 
                                                                                         & 2                             & \multicolumn{1}{c|}{3} & \multicolumn{1}{c|}{4} & \multicolumn{1}{c|}{5} & \multicolumn{1}{c|}{5} & \multicolumn{1}{c|}{5} & \multicolumn{1}{c|}{5} & \multicolumn{1}{c|}{5} & \multicolumn{1}{c|}{6} & 6 \\ \cline{2-11} 
                                                                                         & 3                             & \multicolumn{1}{c|}{3} & \multicolumn{1}{c|}{5} & \multicolumn{1}{c|}{5} & \multicolumn{1}{c|}{5} & \multicolumn{1}{c|}{6} & \multicolumn{1}{c|}{4} & \multicolumn{1}{c|}{5} & \multicolumn{1}{c|}{5} & 6 \\ \cline{2-11} 
                                                                                         & 4                             & \multicolumn{1}{c|}{3} & \multicolumn{1}{c|}{4} & \multicolumn{1}{c|}{6} & \multicolumn{1}{c|}{5} & \multicolumn{1}{c|}{5} & \multicolumn{1}{c|}{4} & \multicolumn{1}{c|}{4} & \multicolumn{1}{c|}{6} & 7 \\ \hline
        \multirow{4}{*}{\begin{tabular}[c]{@{}c@{}}AHLOILTM\\    \\ \& STM\end{tabular}} & 1                             & \multicolumn{1}{c|}{4} & \multicolumn{1}{c|}{5} & \multicolumn{1}{c|}{5} & \multicolumn{1}{c|}{4} & \multicolumn{1}{c|}{6} & \multicolumn{1}{c|}{5} & \multicolumn{1}{c|}{5} & \multicolumn{1}{c|}{5} & 5 \\ \cline{2-11} 
                                                                                         & 2                             & \multicolumn{1}{c|}{5} & \multicolumn{1}{c|}{4} & \multicolumn{1}{c|}{7} & \multicolumn{1}{c|}{5} & \multicolumn{1}{c|}{5} & \multicolumn{1}{c|}{5} & \multicolumn{1}{c|}{5} & \multicolumn{1}{c|}{5} & 3 \\ \cline{2-11} 
                                                                                         & 3                             & \multicolumn{1}{c|}{4} & \multicolumn{1}{c|}{5} & \multicolumn{1}{c|}{5} & \multicolumn{1}{c|}{6} & \multicolumn{1}{c|}{4} & \multicolumn{1}{c|}{7} & \multicolumn{1}{c|}{4} & \multicolumn{1}{c|}{5} & 4 \\ \cline{2-11} 
                                                                                         & 4                             & \multicolumn{1}{c|}{4} & \multicolumn{1}{c|}{4} & \multicolumn{1}{c|}{6} & \multicolumn{1}{c|}{5} & \multicolumn{1}{c|}{5} & \multicolumn{1}{c|}{5} & \multicolumn{1}{c|}{6} & \multicolumn{1}{c|}{4} & 5 \\ \hline
        \end{tabular}
        }
\end{table}

\begin{table}[]
        \caption{The correlation coefficients of methods by the probability of $rand>0.1$ astrocytes phagocytose synapses}
        \resizebox{\textwidth}{!}{
        \label{table11}
        \begin{tabular}{|c|c|c|c|c|c|c|}
        \hline
        Method                  & \begin{tabular}[c]{@{}c@{}}NCHLOI\\    \\ LTM\end{tabular} & \begin{tabular}[c]{@{}c@{}}NHLOI\\    \\ LTM\end{tabular} & \begin{tabular}[c]{@{}c@{}}AHLOI\\    \\ LTM\end{tabular} & \begin{tabular}[c]{@{}c@{}}NCHLOI\\    \\ LTM\&STM\end{tabular} & \begin{tabular}[c]{@{}c@{}}NHLOI\\    \\ LTM\& STM\end{tabular} & \begin{tabular}[c]{@{}c@{}}AHLOI\\    \\ LTM\& STM\end{tabular} \\ \hline
        Correlation coefficient & 0.9676                                                     & 0.9682                                                    & 0.9593                                                    & 0.9476                                                          & 0.9633                                                          & 0.9529                                                          \\ \hline
        \end{tabular}
        }
\end{table}

\begin{figure}%
        \centering
        \begin{minipage}{0.8\textwidth}
                \centering
                \includegraphics[width=0.8\textwidth]{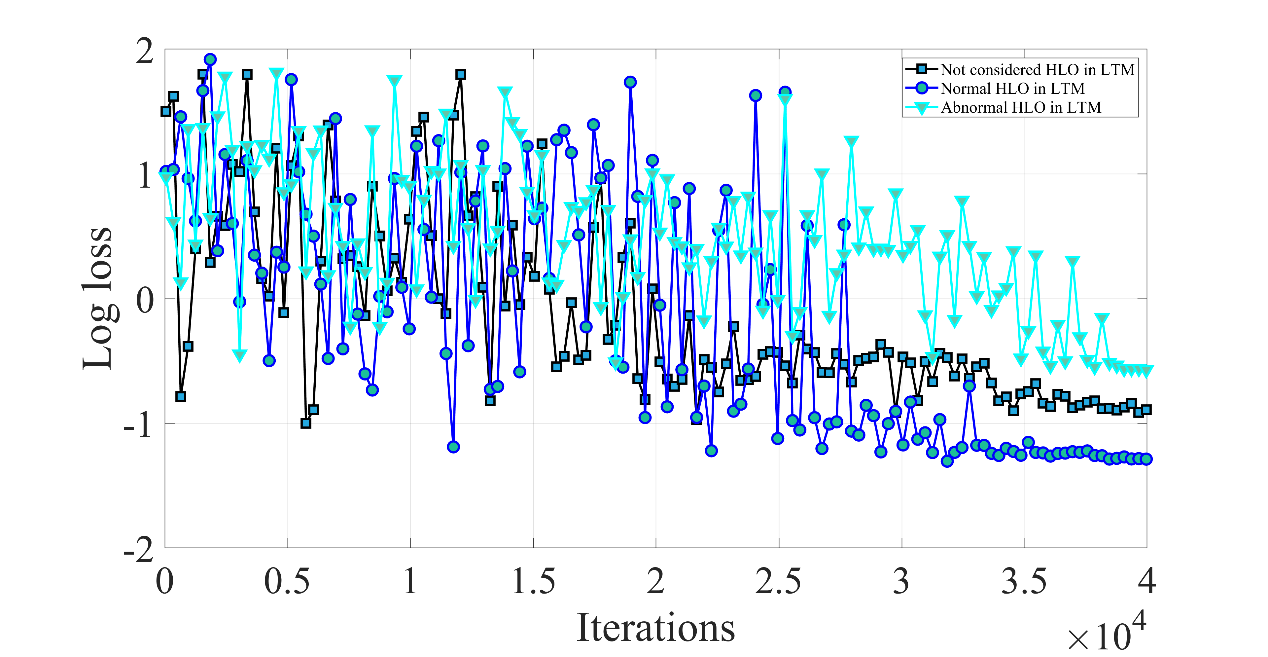}
        \end{minipage}
        \begin{minipage}{0.8\textwidth}
                \centering
                \includegraphics[width=0.8\textwidth]{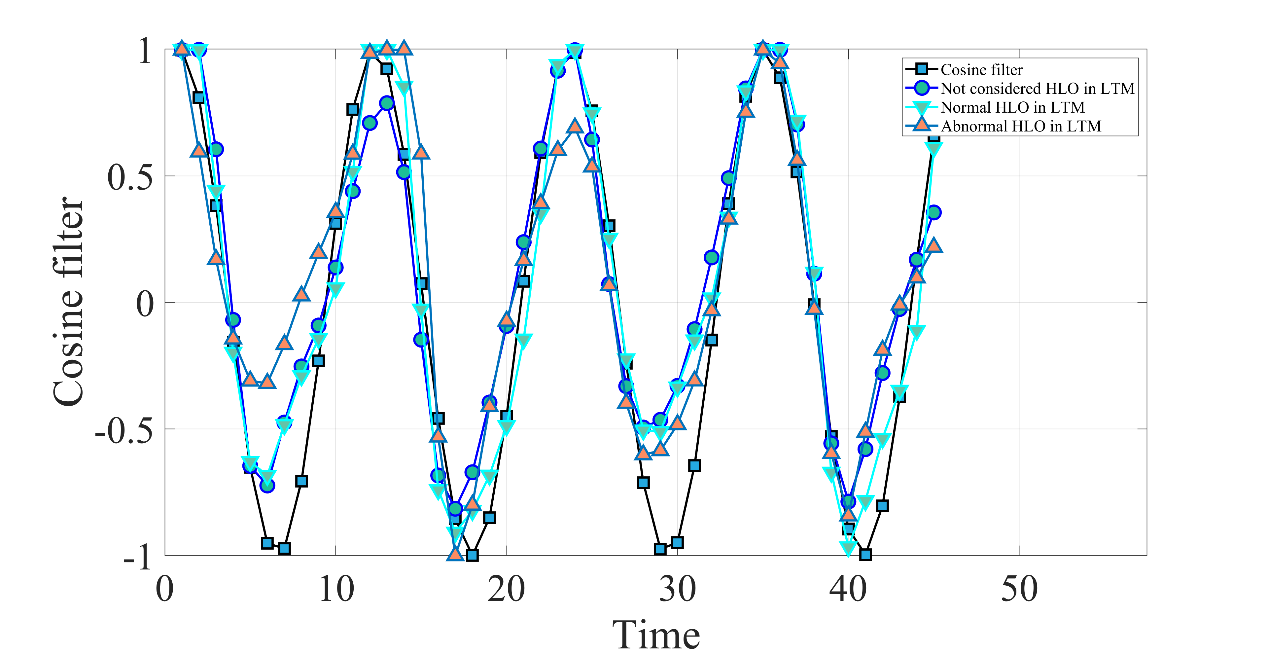}
                \caption{Only long-term memory is considered}
        \end{minipage}
        \begin{minipage}{0.8\textwidth}
                \centering
                \includegraphics[width=0.8\textwidth]{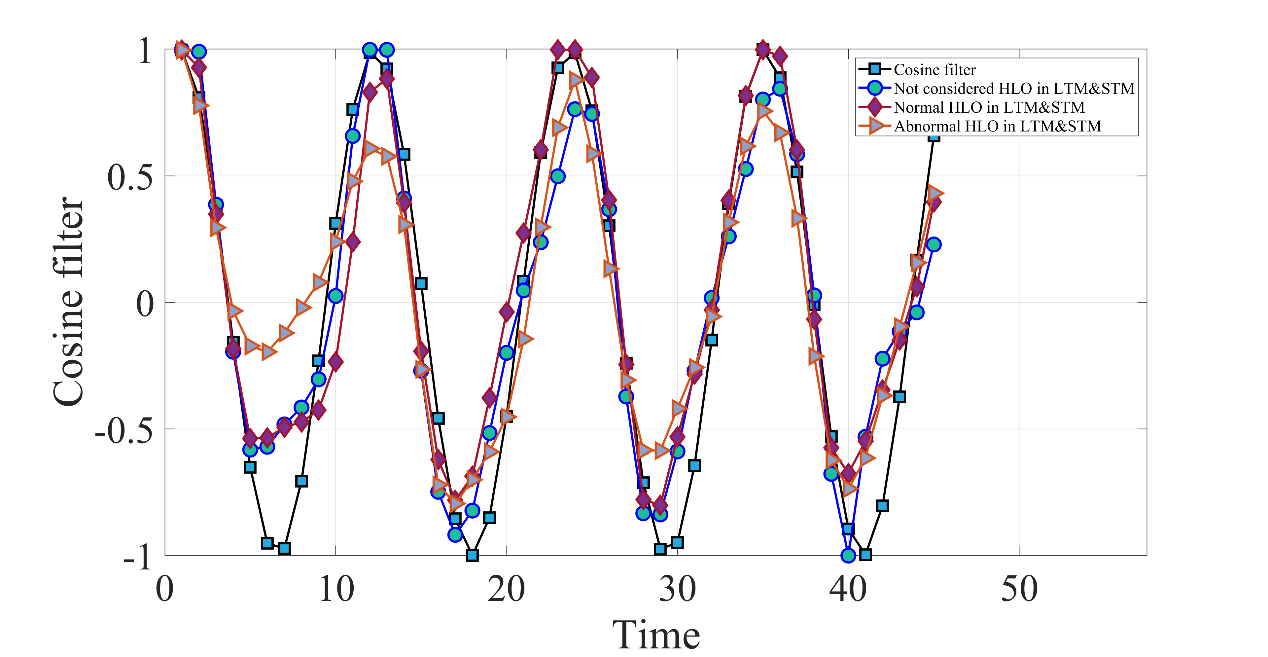}
        \end{minipage}
        \begin{minipage}{0.8\textwidth}
                \centering
                \includegraphics[width=0.8\textwidth]{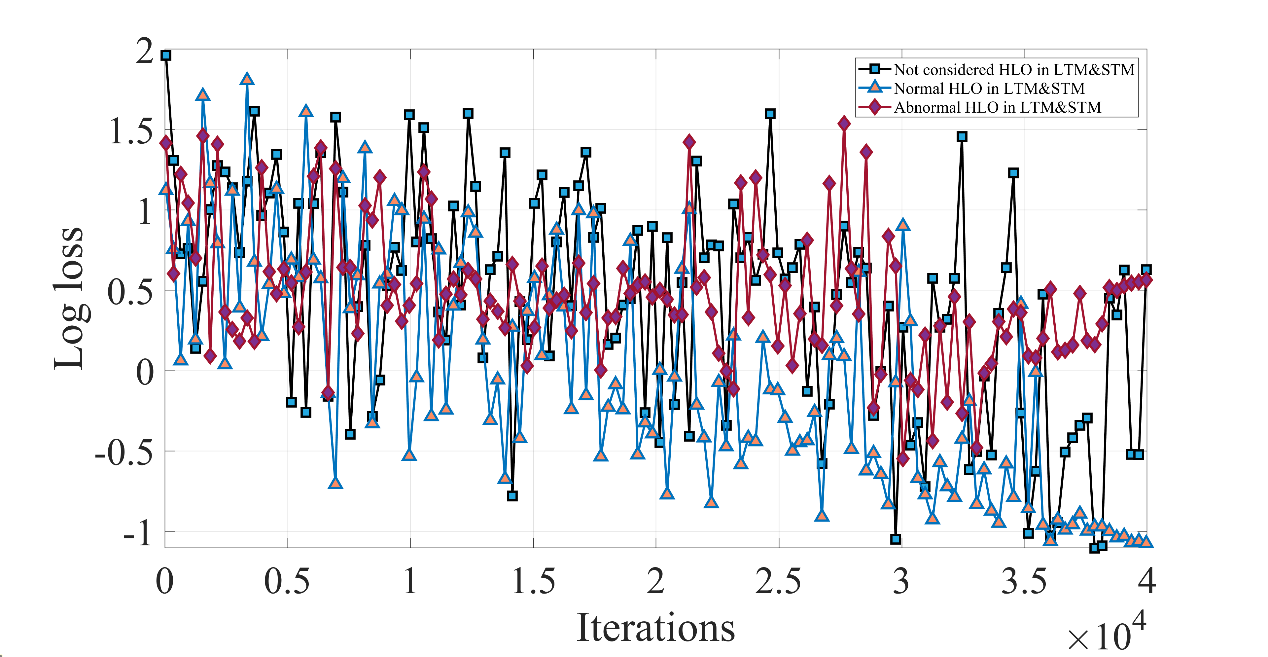}
                \caption{Both short-term and long-term memory are considered}
        \end{minipage}
\caption{The comparisons of higher lower order brain by the probability of $rand>1$ astrocytes phagocytose synapses, the cosine filter simulation includes the same frequency of the upstream and downstream brain regions}
\label{fig19}
\end{figure}

Higher-lower-order model study the distribution of brain regions of high-frequency and low-frequency waves\cite{bib42}. We consider the case where glial cells do not phagocytose synapses, take $rand>1$, and the probability of glial cells phagocytose synapses is 0. The cosine filter simulation includes the same frequency of upstream and downstream brain regions, see Table \ref{table12} for synaptic strength, and Table \ref{table13} for the correlation coefficients between the actual and expected output. Fig. \ref{fig19} compares the fitting effect and the iterations of the loss function by different methods, which include the higher-lower order cases in the upstream and downstream brain regions are both not considered, the normal brain and the brain of Alzheimer's patients. And the cosine filtering of the low frequency of the upstream brain regions and the high frequency of the downstream brain regions, as shown in Table \ref{table14}-\ref{table15}, Fig. \ref{fig24}.

\begin{table}[]
        \caption{The synaptic strength of methods by the probability of $rand>1$ astrocytes phagocytose synapses, the cosine filter simulation includes the same frequency of the upstream and downstream brain regions}
        \resizebox{\textwidth}{!}{
        \label{table12}
        \begin{tabular}{|c|c|ccccccccc|}
                \hline
                \multirow{2}{*}{Method} & \multirow{2}{*}{Neuron types} & \multicolumn{9}{c|}{Synaptic strength} \\ \cline{3-11} 
                 &  & \multicolumn{1}{c|}{1} & \multicolumn{1}{c|}{2} & \multicolumn{1}{c|}{3} & \multicolumn{1}{c|}{4} & \multicolumn{1}{c|}{5} & \multicolumn{1}{c|}{6} & \multicolumn{1}{c|}{7} & \multicolumn{1}{c|}{8} & 9 \\ \hline
                \multirow{4}{*}{NCHLOILTM} & 1 & \multicolumn{1}{c|}{3} & \multicolumn{1}{c|}{5} & \multicolumn{1}{c|}{5} & \multicolumn{1}{c|}{6} & \multicolumn{1}{c|}{6} & \multicolumn{1}{c|}{5} & \multicolumn{1}{c|}{5} & \multicolumn{1}{c|}{3} & 6 \\ \cline{2-11} 
                 & 2 & \multicolumn{1}{c|}{4} & \multicolumn{1}{c|}{6} & \multicolumn{1}{c|}{4} & \multicolumn{1}{c|}{4} & \multicolumn{1}{c|}{5} & \multicolumn{1}{c|}{5} & \multicolumn{1}{c|}{6} & \multicolumn{1}{c|}{5} & 5 \\ \cline{2-11} 
                 & 3 & \multicolumn{1}{c|}{3} & \multicolumn{1}{c|}{5} & \multicolumn{1}{c|}{6} & \multicolumn{1}{c|}{5} & \multicolumn{1}{c|}{5} & \multicolumn{1}{c|}{5} & \multicolumn{1}{c|}{5} & \multicolumn{1}{c|}{4} & 6 \\ \cline{2-11} 
                 & 4 & \multicolumn{1}{c|}{4} & \multicolumn{1}{c|}{4} & \multicolumn{1}{c|}{6} & \multicolumn{1}{c|}{5} & \multicolumn{1}{c|}{5} & \multicolumn{1}{c|}{5} & \multicolumn{1}{c|}{4} & \multicolumn{1}{c|}{5} & 6 \\ \hline
                \multirow{4}{*}{NHLOILTM} & 1 & \multicolumn{1}{c|}{3} & \multicolumn{1}{c|}{4} & \multicolumn{1}{c|}{6} & \multicolumn{1}{c|}{5} & \multicolumn{1}{c|}{6} & \multicolumn{1}{c|}{5} & \multicolumn{1}{c|}{3} & \multicolumn{1}{c|}{6} & 6 \\ \cline{2-11} 
                 & 2 & \multicolumn{1}{c|}{4} & \multicolumn{1}{c|}{5} & \multicolumn{1}{c|}{3} & \multicolumn{1}{c|}{6} & \multicolumn{1}{c|}{6} & \multicolumn{1}{c|}{3} & \multicolumn{1}{c|}{6} & \multicolumn{1}{c|}{5} & 6 \\ \cline{2-11} 
                 & 3 & \multicolumn{1}{c|}{4} & \multicolumn{1}{c|}{6} & \multicolumn{1}{c|}{5} & \multicolumn{1}{c|}{5} & \multicolumn{1}{c|}{3} & \multicolumn{1}{c|}{5} & \multicolumn{1}{c|}{6} & \multicolumn{1}{c|}{6} & 4 \\ \cline{2-11} 
                 & 4 & \multicolumn{1}{c|}{4} & \multicolumn{1}{c|}{6} & \multicolumn{1}{c|}{4} & \multicolumn{1}{c|}{4} & \multicolumn{1}{c|}{4} & \multicolumn{1}{c|}{5} & \multicolumn{1}{c|}{5} & \multicolumn{1}{c|}{6} & 6 \\ \hline
                \multirow{4}{*}{AHLOILTM} & 1 & \multicolumn{1}{c|}{3} & \multicolumn{1}{c|}{4} & \multicolumn{1}{c|}{6} & \multicolumn{1}{c|}{6} & \multicolumn{1}{c|}{6} & \multicolumn{1}{c|}{5} & \multicolumn{1}{c|}{5} & \multicolumn{1}{c|}{4} & 5 \\ \cline{2-11} 
                 & 2 & \multicolumn{1}{c|}{4} & \multicolumn{1}{c|}{5} & \multicolumn{1}{c|}{5} & \multicolumn{1}{c|}{4} & \multicolumn{1}{c|}{4} & \multicolumn{1}{c|}{6} & \multicolumn{1}{c|}{5} & \multicolumn{1}{c|}{6} & 5 \\ \cline{2-11} 
                 & 3 & \multicolumn{1}{c|}{5} & \multicolumn{1}{c|}{6} & \multicolumn{1}{c|}{4} & \multicolumn{1}{c|}{5} & \multicolumn{1}{c|}{5} & \multicolumn{1}{c|}{6} & \multicolumn{1}{c|}{4} & \multicolumn{1}{c|}{4} & 5 \\ \cline{2-11} 
                 & 4 & \multicolumn{1}{c|}{5} & \multicolumn{1}{c|}{6} & \multicolumn{1}{c|}{4} & \multicolumn{1}{c|}{5} & \multicolumn{1}{c|}{5} & \multicolumn{1}{c|}{4} & \multicolumn{1}{c|}{4} & \multicolumn{1}{c|}{6} & 5 \\ \hline
                \multirow{4}{*}{NCHLOILTM \& STM} & 1 & \multicolumn{1}{c|}{3} & \multicolumn{1}{c|}{5} & \multicolumn{1}{c|}{4} & \multicolumn{1}{c|}{6} & \multicolumn{1}{c|}{4} & \multicolumn{1}{c|}{5} & \multicolumn{1}{c|}{6} & \multicolumn{1}{c|}{6} & 5 \\ \cline{2-11} 
                 & 2 & \multicolumn{1}{c|}{4} & \multicolumn{1}{c|}{6} & \multicolumn{1}{c|}{5} & \multicolumn{1}{c|}{3} & \multicolumn{1}{c|}{5} & \multicolumn{1}{c|}{6} & \multicolumn{1}{c|}{5} & \multicolumn{1}{c|}{4} & 6 \\ \cline{2-11} 
                 & 3 & \multicolumn{1}{c|}{4} & \multicolumn{1}{c|}{5} & \multicolumn{1}{c|}{5} & \multicolumn{1}{c|}{5} & \multicolumn{1}{c|}{6} & \multicolumn{1}{c|}{5} & \multicolumn{1}{c|}{4} & \multicolumn{1}{c|}{5} & 5 \\ \cline{2-11} 
                 & 4 & \multicolumn{1}{c|}{5} & \multicolumn{1}{c|}{4} & \multicolumn{1}{c|}{4} & \multicolumn{1}{c|}{6} & \multicolumn{1}{c|}{5} & \multicolumn{1}{c|}{3} & \multicolumn{1}{c|}{6} & \multicolumn{1}{c|}{6} & 5 \\ \hline
                \multirow{4}{*}{NHLOILTM \& STM} & 1 & \multicolumn{1}{c|}{5} & \multicolumn{1}{c|}{5} & \multicolumn{1}{c|}{5} & \multicolumn{1}{c|}{5} & \multicolumn{1}{c|}{4} & \multicolumn{1}{c|}{6} & \multicolumn{1}{c|}{4} & \multicolumn{1}{c|}{5} & 5 \\ \cline{2-11} 
                 & 2 & \multicolumn{1}{c|}{4} & \multicolumn{1}{c|}{3} & \multicolumn{1}{c|}{5} & \multicolumn{1}{c|}{4} & \multicolumn{1}{c|}{5} & \multicolumn{1}{c|}{5} & \multicolumn{1}{c|}{5} & \multicolumn{1}{c|}{7} & 6 \\ \cline{2-11} 
                 & 3 & \multicolumn{1}{c|}{3} & \multicolumn{1}{c|}{4} & \multicolumn{1}{c|}{3} & \multicolumn{1}{c|}{6} & \multicolumn{1}{c|}{5} & \multicolumn{1}{c|}{4} & \multicolumn{1}{c|}{7} & \multicolumn{1}{c|}{6} & 6 \\ \cline{2-11} 
                 & 4 & \multicolumn{1}{c|}{3} & \multicolumn{1}{c|}{4} & \multicolumn{1}{c|}{4} & \multicolumn{1}{c|}{6} & \multicolumn{1}{c|}{5} & \multicolumn{1}{c|}{5} & \multicolumn{1}{c|}{4} & \multicolumn{1}{c|}{6} & 7 \\ \hline
                \multirow{4}{*}{AHLOILTM \& STM} & 1 & \multicolumn{1}{c|}{6} & \multicolumn{1}{c|}{6} & \multicolumn{1}{c|}{4} & \multicolumn{1}{c|}{5} & \multicolumn{1}{c|}{3} & \multicolumn{1}{c|}{4} & \multicolumn{1}{c|}{6} & \multicolumn{1}{c|}{5} & 5 \\ \cline{2-11} 
                 & 2 & \multicolumn{1}{c|}{5} & \multicolumn{1}{c|}{5} & \multicolumn{1}{c|}{4} & \multicolumn{1}{c|}{5} & \multicolumn{1}{c|}{5} & \multicolumn{1}{c|}{5} & \multicolumn{1}{c|}{5} & \multicolumn{1}{c|}{5} & 5 \\ \cline{2-11} 
                 & 3 & \multicolumn{1}{c|}{3} & \multicolumn{1}{c|}{5} & \multicolumn{1}{c|}{5} & \multicolumn{1}{c|}{5} & \multicolumn{1}{c|}{6} & \multicolumn{1}{c|}{4} & \multicolumn{1}{c|}{6} & \multicolumn{1}{c|}{6} & 4 \\ \cline{2-11} 
                 & 4 & \multicolumn{1}{c|}{5} & \multicolumn{1}{c|}{5} & \multicolumn{1}{c|}{5} & \multicolumn{1}{c|}{4} & \multicolumn{1}{c|}{4} & \multicolumn{1}{c|}{5} & \multicolumn{1}{c|}{4} & \multicolumn{1}{c|}{6} & 6 \\ \hline
                \end{tabular}%
        }
\end{table}

\begin{table}[]
        \caption{The correlation coefficients of methods by the probability of $rand>1$ astrocytes phagocytose synapses, the cosine filter simulation includes the same frequency of the upstream and downstream brain regions}
        \resizebox{\textwidth}{!}{
        \label{table13}
        \begin{tabular}{|c|c|c|c|c|c|c|}
        \hline
        Method                  & \begin{tabular}[c]{@{}c@{}}NCHLOI\\    \\ LTM\end{tabular} & \begin{tabular}[c]{@{}c@{}}NHLOI\\    \\ LTM\end{tabular} & \begin{tabular}[c]{@{}c@{}}AHLOI\\    \\ LTM\end{tabular} & \begin{tabular}[c]{@{}c@{}}NCHLOI\\    \\ LTM\&STM\end{tabular} & \begin{tabular}[c]{@{}c@{}}NHLOI\\    \\ LTM\& STM\end{tabular} & \begin{tabular}[c]{@{}c@{}}AHLOI\\    \\ LTM\& STM\end{tabular} \\ \hline
        Correlation coefficient & 0.9449                                                     & 0.9550                                                    & 0.9113                                                    & 0.9403                                                          & 0.9451                                                          & 0.9200                                                          \\ \hline
        \end{tabular}
        }
\end{table}

\begin{table}[]
        \caption{The synaptic strength of methods by the probability of $rand>1$ astrocytes phagocytose synapses, the cosine filter simulation includes the low frequency of the upstream and high frequency downstream brain regions}
        \resizebox{\textwidth}{!}{
        \label{table14}
        \begin{tabular}{|c|c|lllllllll|}
                \hline
                \multirow{2}{*}{Method} & \multirow{2}{*}{Neuron types} & \multicolumn{9}{c|}{Synaptic strength} \\ \cline{3-11} 
                 &  & \multicolumn{1}{c|}{1} & \multicolumn{1}{c|}{2} & \multicolumn{1}{c|}{3} & \multicolumn{1}{c|}{4} & \multicolumn{1}{c|}{5} & \multicolumn{1}{c|}{6} & \multicolumn{1}{c|}{7} & \multicolumn{1}{c|}{8} & \multicolumn{1}{c|}{9} \\ \hline
                \multirow{4}{*}{NCHLOILTM} & 1 & \multicolumn{1}{l|}{3} & \multicolumn{1}{l|}{5} & \multicolumn{1}{l|}{6} & \multicolumn{1}{l|}{6} & \multicolumn{1}{l|}{4} & \multicolumn{1}{l|}{6} & \multicolumn{1}{l|}{6} & \multicolumn{1}{l|}{4} & 4 \\ \cline{2-11} 
                 & 2 & \multicolumn{1}{l|}{2} & \multicolumn{1}{l|}{4} & \multicolumn{1}{l|}{5} & \multicolumn{1}{l|}{6} & \multicolumn{1}{l|}{6} & \multicolumn{1}{l|}{4} & \multicolumn{1}{l|}{5} & \multicolumn{1}{l|}{6} & 6 \\ \cline{2-11} 
                 & 3 & \multicolumn{1}{l|}{2} & \multicolumn{1}{l|}{6} & \multicolumn{1}{l|}{6} & \multicolumn{1}{l|}{5} & \multicolumn{1}{l|}{4} & \multicolumn{1}{l|}{4} & \multicolumn{1}{l|}{6} & \multicolumn{1}{l|}{6} & 5 \\ \cline{2-11} 
                 & 4 & \multicolumn{1}{l|}{2} & \multicolumn{1}{l|}{6} & \multicolumn{1}{l|}{5} & \multicolumn{1}{l|}{3} & \multicolumn{1}{l|}{5} & \multicolumn{1}{l|}{6} & \multicolumn{1}{l|}{5} & \multicolumn{1}{l|}{6} & 6 \\ \hline
                \multirow{4}{*}{NHLOILTM} & 1 & \multicolumn{1}{l|}{3} & \multicolumn{1}{l|}{4} & \multicolumn{1}{l|}{5} & \multicolumn{1}{l|}{6} & \multicolumn{1}{l|}{6} & \multicolumn{1}{l|}{6} & \multicolumn{1}{l|}{5} & \multicolumn{1}{l|}{5} & 4 \\ \cline{2-11} 
                 & 2 & \multicolumn{1}{l|}{3} & \multicolumn{1}{l|}{4} & \multicolumn{1}{l|}{5} & \multicolumn{1}{l|}{6} & \multicolumn{1}{l|}{6} & \multicolumn{1}{l|}{5} & \multicolumn{1}{l|}{6} & \multicolumn{1}{l|}{4} & 5 \\ \cline{2-11} 
                 & 3 & \multicolumn{1}{l|}{5} & \multicolumn{1}{l|}{6} & \multicolumn{1}{l|}{5} & \multicolumn{1}{l|}{5} & \multicolumn{1}{l|}{4} & \multicolumn{1}{l|}{3} & \multicolumn{1}{l|}{5} & \multicolumn{1}{l|}{6} & 5 \\ \cline{2-11} 
                 & 4 & \multicolumn{1}{l|}{4} & \multicolumn{1}{l|}{4} & \multicolumn{1}{l|}{4} & \multicolumn{1}{l|}{5} & \multicolumn{1}{l|}{5} & \multicolumn{1}{l|}{6} & \multicolumn{1}{l|}{6} & \multicolumn{1}{l|}{5} & 5 \\ \hline
                \multirow{4}{*}{AHLOILTM} & 1 & \multicolumn{1}{l|}{6} & \multicolumn{1}{l|}{5} & \multicolumn{1}{l|}{4} & \multicolumn{1}{l|}{6} & \multicolumn{1}{l|}{4} & \multicolumn{1}{l|}{3} & \multicolumn{1}{l|}{5} & \multicolumn{1}{l|}{6} & 5 \\ \cline{2-11} 
                 & 2 & \multicolumn{1}{l|}{4} & \multicolumn{1}{l|}{5} & \multicolumn{1}{l|}{5} & \multicolumn{1}{l|}{6} & \multicolumn{1}{l|}{6} & \multicolumn{1}{l|}{3} & \multicolumn{1}{l|}{5} & \multicolumn{1}{l|}{4} & 6 \\ \cline{2-11} 
                 & 3 & \multicolumn{1}{l|}{7} & \multicolumn{1}{l|}{6} & \multicolumn{1}{l|}{5} & \multicolumn{1}{l|}{4} & \multicolumn{1}{l|}{4} & \multicolumn{1}{l|}{6} & \multicolumn{1}{l|}{6} & \multicolumn{1}{l|}{3} & 3 \\ \cline{2-11} 
                 & 4 & \multicolumn{1}{l|}{5} & \multicolumn{1}{l|}{5} & \multicolumn{1}{l|}{3} & \multicolumn{1}{l|}{4} & \multicolumn{1}{l|}{6} & \multicolumn{1}{l|}{6} & \multicolumn{1}{l|}{6} & \multicolumn{1}{l|}{5} & 4 \\ \hline
                \multirow{4}{*}{NCHLOILTM \& STM} & 1 & \multicolumn{1}{l|}{3} & \multicolumn{1}{l|}{5} & \multicolumn{1}{l|}{4} & \multicolumn{1}{l|}{4} & \multicolumn{1}{l|}{6} & \multicolumn{1}{l|}{6} & \multicolumn{1}{l|}{5} & \multicolumn{1}{l|}{5} & 6 \\ \cline{2-11} 
                 & 2 & \multicolumn{1}{l|}{4} & \multicolumn{1}{l|}{4} & \multicolumn{1}{l|}{3} & \multicolumn{1}{l|}{6} & \multicolumn{1}{l|}{7} & \multicolumn{1}{l|}{4} & \multicolumn{1}{l|}{4} & \multicolumn{1}{l|}{7} & 5 \\ \cline{2-11} 
                 & 3 & \multicolumn{1}{l|}{5} & \multicolumn{1}{l|}{4} & \multicolumn{1}{l|}{4} & \multicolumn{1}{l|}{6} & \multicolumn{1}{l|}{5} & \multicolumn{1}{l|}{5} & \multicolumn{1}{l|}{5} & \multicolumn{1}{l|}{5} & 5 \\ \cline{2-11} 
                 & 4 & \multicolumn{1}{l|}{3} & \multicolumn{1}{l|}{4} & \multicolumn{1}{l|}{6} & \multicolumn{1}{l|}{4} & \multicolumn{1}{l|}{5} & \multicolumn{1}{l|}{4} & \multicolumn{1}{l|}{5} & \multicolumn{1}{l|}{6} & 7 \\ \hline
                \multirow{4}{*}{NHLOILTM \& STM} & 1 & \multicolumn{1}{l|}{4} & \multicolumn{1}{l|}{5} & \multicolumn{1}{l|}{5} & \multicolumn{1}{l|}{5} & \multicolumn{1}{l|}{3} & \multicolumn{1}{l|}{5} & \multicolumn{1}{l|}{6} & \multicolumn{1}{l|}{5} & 6 \\ \cline{2-11} 
                 & 2 & \multicolumn{1}{l|}{2} & \multicolumn{1}{l|}{3} & \multicolumn{1}{l|}{5} & \multicolumn{1}{l|}{5} & \multicolumn{1}{l|}{5} & \multicolumn{1}{l|}{4} & \multicolumn{1}{l|}{6} & \multicolumn{1}{l|}{6} & 8 \\ \cline{2-11} 
                 & 3 & \multicolumn{1}{l|}{3} & \multicolumn{1}{l|}{4} & \multicolumn{1}{l|}{4} & \multicolumn{1}{l|}{7} & \multicolumn{1}{l|}{4} & \multicolumn{1}{l|}{4} & \multicolumn{1}{l|}{6} & \multicolumn{1}{l|}{5} & 7 \\ \cline{2-11} 
                 & 4 & \multicolumn{1}{l|}{2} & \multicolumn{1}{l|}{5} & \multicolumn{1}{l|}{4} & \multicolumn{1}{l|}{5} & \multicolumn{1}{l|}{4} & \multicolumn{1}{l|}{6} & \multicolumn{1}{l|}{6} & \multicolumn{1}{l|}{6} & 6 \\ \hline
                \multirow{4}{*}{AHLOILTM \& STM} & 1 & \multicolumn{1}{l|}{5} & \multicolumn{1}{l|}{6} & \multicolumn{1}{l|}{3} & \multicolumn{1}{l|}{6} & \multicolumn{1}{l|}{6} & \multicolumn{1}{l|}{3} & \multicolumn{1}{l|}{5} & \multicolumn{1}{l|}{6} & 4 \\ \cline{2-11} 
                 & 2 & \multicolumn{1}{l|}{6} & \multicolumn{1}{l|}{7} & \multicolumn{1}{l|}{5} & \multicolumn{1}{l|}{4} & \multicolumn{1}{l|}{4} & \multicolumn{1}{l|}{4} & \multicolumn{1}{l|}{6} & \multicolumn{1}{l|}{4} & 4 \\ \cline{2-11} 
                 & 3 & \multicolumn{1}{l|}{5} & \multicolumn{1}{l|}{5} & \multicolumn{1}{l|}{6} & \multicolumn{1}{l|}{5} & \multicolumn{1}{l|}{4} & \multicolumn{1}{l|}{5} & \multicolumn{1}{l|}{5} & \multicolumn{1}{l|}{5} & 4 \\ \cline{2-11} 
                 & 4 & \multicolumn{1}{l|}{5} & \multicolumn{1}{l|}{6} & \multicolumn{1}{l|}{4} & \multicolumn{1}{l|}{5} & \multicolumn{1}{l|}{5} & \multicolumn{1}{l|}{4} & \multicolumn{1}{l|}{6} & \multicolumn{1}{l|}{4} & 5 \\ \hline
                \end{tabular}%
        }
\end{table}

\begin{table}[]
        \caption{The correlation coefficients of methods by the probability of $rand>1$ astrocytes phagocytose synapses, the cosine filter simulation includes the low frequency of the upstream, and high frequency downstream brain regions}
        \resizebox{\textwidth}{!}{
        \label{table15}
        \begin{tabular}{|c|c|c|c|c|c|c|}
        \hline
        Method                  & \begin{tabular}[c]{@{}c@{}}NCHLOI\\    \\ LTM\end{tabular} & \begin{tabular}[c]{@{}c@{}}NHLOI\\    \\ LTM\end{tabular} & \begin{tabular}[c]{@{}c@{}}AHLOI\\    \\ LTM\end{tabular} & \begin{tabular}[c]{@{}c@{}}NCHLOI\\    \\ LTM\&STM\end{tabular} & \begin{tabular}[c]{@{}c@{}}NHLOI\\    \\ LTM\& STM\end{tabular} & \begin{tabular}[c]{@{}c@{}}AHLOI\\    \\ LTM\& STM\end{tabular} \\ \hline
        Correlation coefficient & 0.9814                                                     & 0.9891                                                    & 0.9488                                                    & 0.9661                                                          & 0.9852                                                          & 0.9396                                                          \\ \hline
        \end{tabular}
        }
\end{table}

\begin{figure}%
        \centering
        \begin{minipage}{0.8\textwidth}
                \centering
                \includegraphics[width=0.8\textwidth]{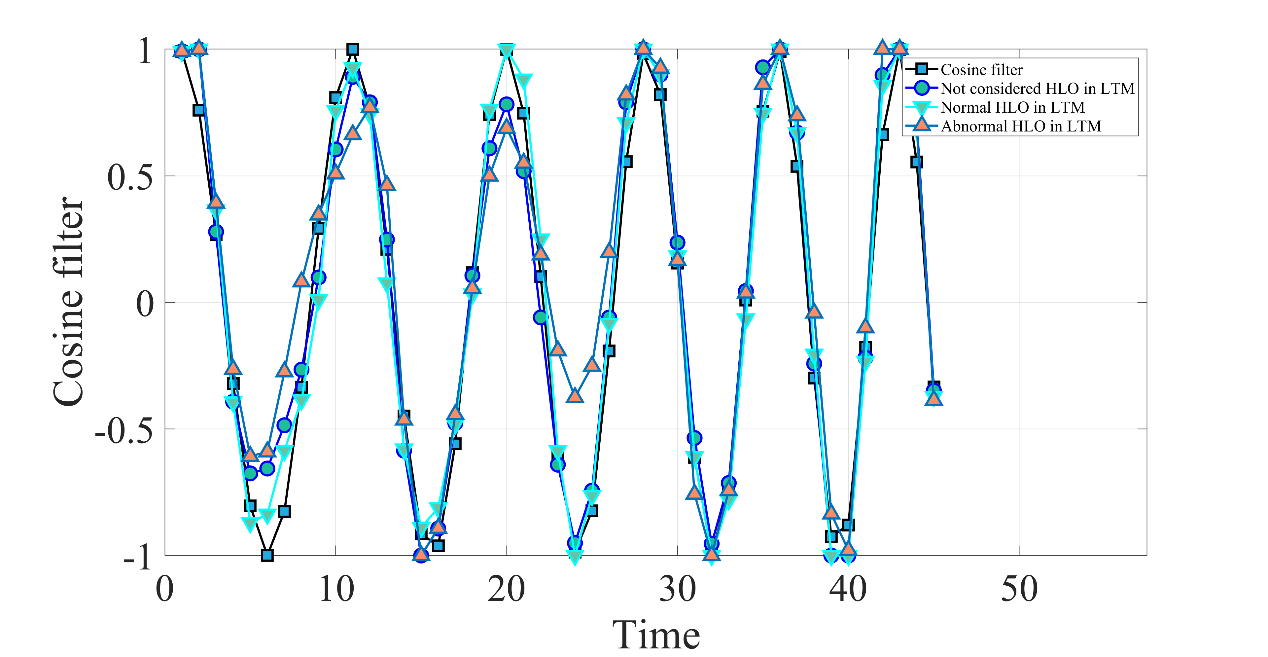}
        \end{minipage}
        \begin{minipage}{0.8\textwidth}
                \centering
                \includegraphics[width=0.8\textwidth]{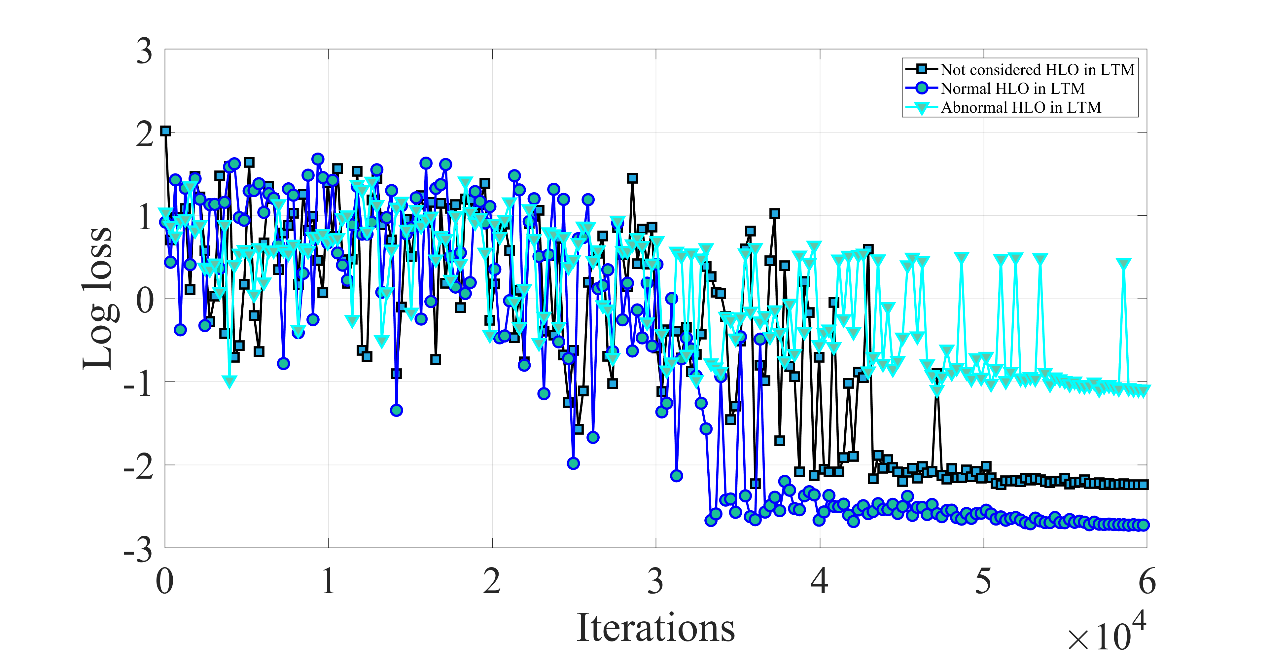}
                \caption{Only long-term memory is considered}
        \end{minipage}
        \begin{minipage}{0.8\textwidth}
                \centering
                \includegraphics[width=0.8\textwidth]{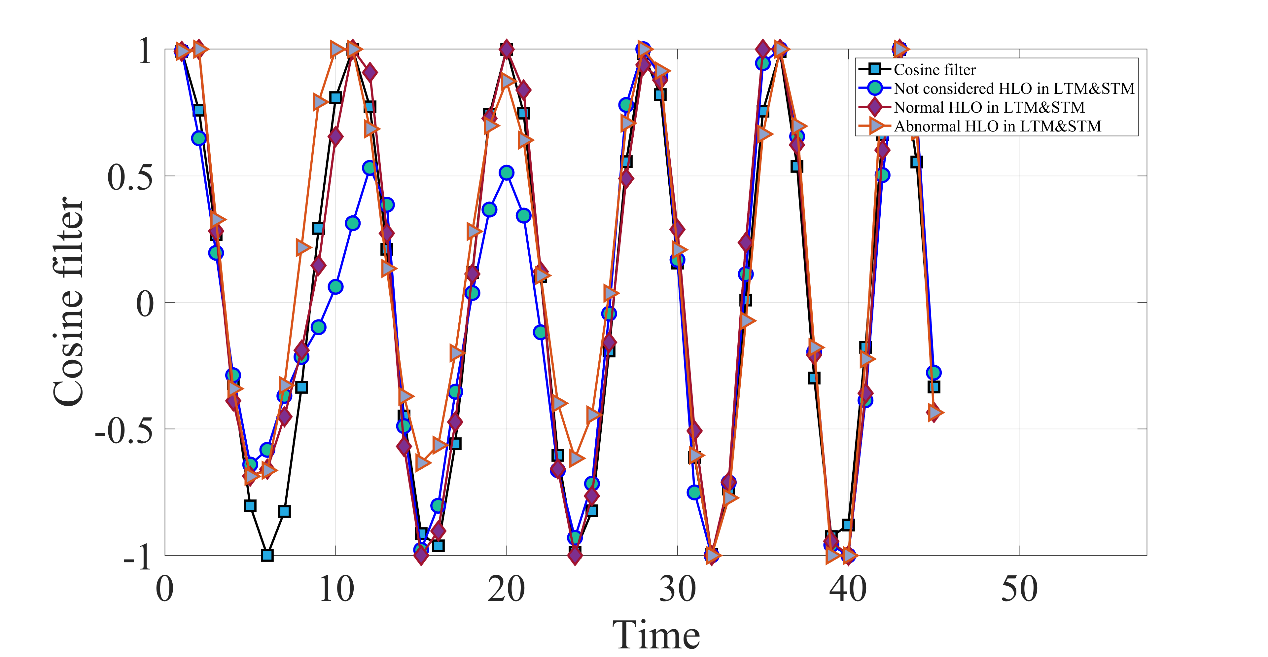}
        \end{minipage}
        \begin{minipage}{0.8\textwidth}
                \centering
                \includegraphics[width=0.8\textwidth]{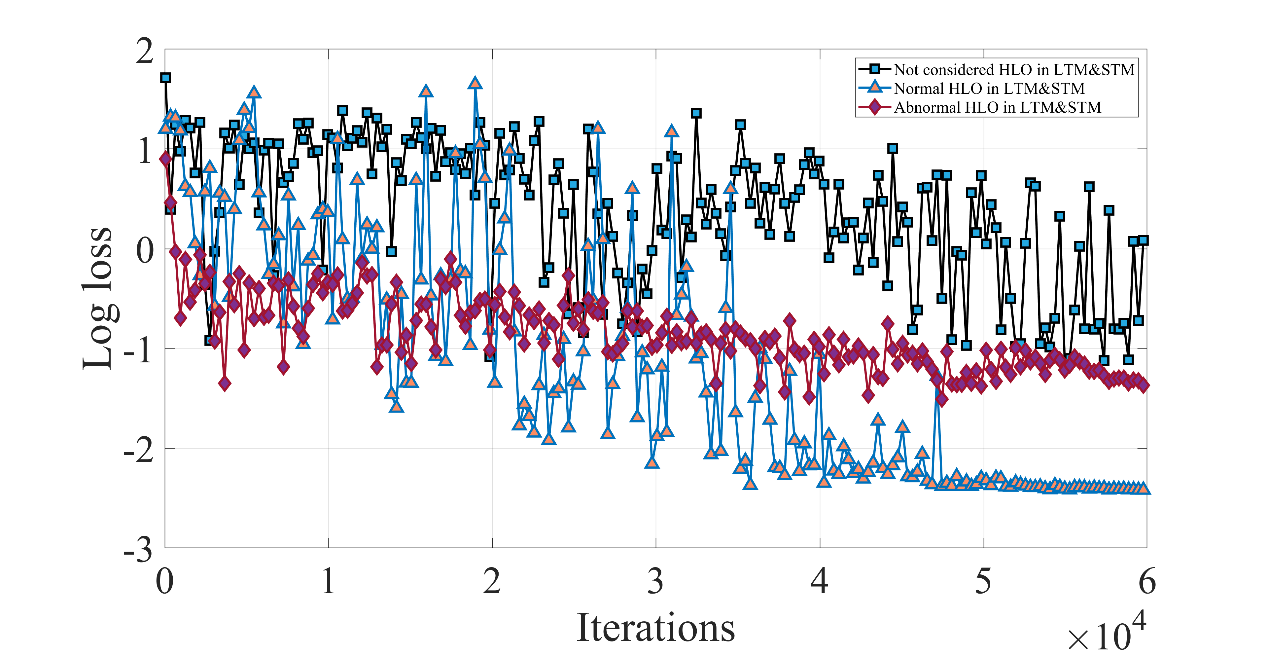}
                \caption{Both short-term and long-term memory are considered}
        \end{minipage}
\caption{The comparisons of higher lower order brain by the probability of $rand>1$ astrocytes phagocytose synapses, the cosine filter simulation includes the low frequency of the upstream, and high frequency downstream brain regions}
\label{fig23}
\end{figure}

Synaptic activity will be weakened as the synapse strengthens, and synaptic activity will be strengthened when the synapse weakens, $\tanh(\beta t)$ is taken according to the activation function of synaptic activity, and the $\beta$ value ranges from 0.9 to 1, and the smaller the $\beta$ value, the synapse is more active. Considering the low probability of glial cells phagocytose synapses, when taking $rand>0.9$, the probability of glial cells phagocytose synapses is $10\%$. The cosine filter simulation includes the same frequency of upstream and downstream brain regions, as shown in Table \ref{table16} for synaptic strength, and the correlation coefficients between the actual results and the expected results in Table \ref{table17}. Fig. \ref{fig23}: Comparison of fitting effects and the iterations of the loss function by different methods. And the cosine filter of the low frequency of the upstream brain regions and the high frequency of the downstream brain regions, as shown in Table \ref{table18}-\ref{table19} and Fig. \ref{fig25}. And the cosine filter of the high frequency of the upstream brain regions and the low frequency of the downstream brain regions, see Table \ref{table20}-\ref{table21}, Fig. \ref{fig26}, and the fitting effect is getting worse.

Why do there be more high frequencies in brain regions downstream of the normal brain? From the upstream brain regions to the downstream brain regions, the brain regions received by the signals gradually accumulate, and the signal error also gradually accumulates. If there is a low frequency in the downstream brain regions, the high frequency in the upstream brain regions will increase according to the rebalancing of synaptic strength. High frequencies lead to greater error in signal processing in upstream brain regions, and larger signal error accumulates in downstream brain regions. Even if the signal processing in the downstream brain regions is improved by the low-frequency, due to the larger signal error accumulation in the downstream brain regions, which makes the low frequencies of the downstream brain regions meaningless. The signal cognition of the upstream and the downstream brain regions will not be improved in the next iteration of the circuit. Therefore, the normal brain should be the downstream brain regions with more high frequencies, in line with the rules of Evolutionary Computation.

\begin{table}[]
        \caption{The synaptic strength of methods by the probability of $rand>1$ astrocytes phagocytose synapses, the cosine filter simulation includes the same frequency of the upstream and downstream brain regions through synaptic activity in activation function}
        \resizebox{\textwidth}{!}{
        \label{table16}
        \begin{tabular}{|c|c|lllllllll|}
                \hline
                \multirow{2}{*}{Method} & \multirow{2}{*}{Neuron types} & \multicolumn{9}{c|}{Synaptic strength} \\ \cline{3-11} 
                 &  & \multicolumn{1}{c|}{1} & \multicolumn{1}{c|}{2} & \multicolumn{1}{c|}{3} & \multicolumn{1}{c|}{4} & \multicolumn{1}{c|}{5} & \multicolumn{1}{c|}{6} & \multicolumn{1}{c|}{7} & \multicolumn{1}{c|}{8} & \multicolumn{1}{c|}{9} \\ \hline
                \multirow{4}{*}{NCHLOILTM} & 1 & \multicolumn{1}{l|}{4} & \multicolumn{1}{l|}{5} & \multicolumn{1}{l|}{5} & \multicolumn{1}{l|}{5} & \multicolumn{1}{l|}{5} & \multicolumn{1}{l|}{5} & \multicolumn{1}{l|}{5} & \multicolumn{1}{l|}{5} & 5 \\ \cline{2-11} 
                 & 2 & \multicolumn{1}{l|}{4} & \multicolumn{1}{l|}{5} & \multicolumn{1}{l|}{6} & \multicolumn{1}{l|}{3} & \multicolumn{1}{l|}{6} & \multicolumn{1}{l|}{6} & \multicolumn{1}{l|}{5} & \multicolumn{1}{l|}{3} & 6 \\ \cline{2-11} 
                 & 3 & \multicolumn{1}{l|}{5} & \multicolumn{1}{l|}{4} & \multicolumn{1}{l|}{5} & \multicolumn{1}{l|}{5} & \multicolumn{1}{l|}{5} & \multicolumn{1}{l|}{6} & \multicolumn{1}{l|}{6} & \multicolumn{1}{l|}{5} & 3 \\ \cline{2-11} 
                 & 4 & \multicolumn{1}{l|}{3} & \multicolumn{1}{l|}{6} & \multicolumn{1}{l|}{5} & \multicolumn{1}{l|}{4} & \multicolumn{1}{l|}{5} & \multicolumn{1}{l|}{6} & \multicolumn{1}{l|}{4} & \multicolumn{1}{l|}{6} & 5 \\ \hline
                \multirow{4}{*}{NHLOILTM} & 1 & \multicolumn{1}{l|}{3} & \multicolumn{1}{l|}{5} & \multicolumn{1}{l|}{5} & \multicolumn{1}{l|}{5} & \multicolumn{1}{l|}{5} & \multicolumn{1}{l|}{6} & \multicolumn{1}{l|}{3} & \multicolumn{1}{l|}{6} & 6 \\ \cline{2-11} 
                 & 2 & \multicolumn{1}{l|}{3} & \multicolumn{1}{l|}{6} & \multicolumn{1}{l|}{5} & \multicolumn{1}{l|}{5} & \multicolumn{1}{l|}{3} & \multicolumn{1}{l|}{6} & \multicolumn{1}{l|}{5} & \multicolumn{1}{l|}{5} & 6 \\ \cline{2-11} 
                 & 3 & \multicolumn{1}{l|}{2} & \multicolumn{1}{l|}{6} & \multicolumn{1}{l|}{5} & \multicolumn{1}{l|}{5} & \multicolumn{1}{l|}{4} & \multicolumn{1}{l|}{6} & \multicolumn{1}{l|}{5} & \multicolumn{1}{l|}{5} & 6 \\ \cline{2-11} 
                 & 4 & \multicolumn{1}{l|}{4} & \multicolumn{1}{l|}{6} & \multicolumn{1}{l|}{3} & \multicolumn{1}{l|}{5} & \multicolumn{1}{l|}{4} & \multicolumn{1}{l|}{6} & \multicolumn{1}{l|}{4} & \multicolumn{1}{l|}{6} & 6 \\ \hline
                \multirow{4}{*}{AHLOILTM} & 1 & \multicolumn{1}{l|}{6} & \multicolumn{1}{l|}{6} & \multicolumn{1}{l|}{4} & \multicolumn{1}{l|}{5} & \multicolumn{1}{l|}{5} & \multicolumn{1}{l|}{5} & \multicolumn{1}{l|}{4} & \multicolumn{1}{l|}{4} & 5 \\ \cline{2-11} 
                 & 2 & \multicolumn{1}{l|}{6} & \multicolumn{1}{l|}{4} & \multicolumn{1}{l|}{3} & \multicolumn{1}{l|}{4} & \multicolumn{1}{l|}{4} & \multicolumn{1}{l|}{5} & \multicolumn{1}{l|}{6} & \multicolumn{1}{l|}{6} & 6 \\ \cline{2-11} 
                 & 3 & \multicolumn{1}{l|}{6} & \multicolumn{1}{l|}{5} & \multicolumn{1}{l|}{5} & \multicolumn{1}{l|}{5} & \multicolumn{1}{l|}{5} & \multicolumn{1}{l|}{5} & \multicolumn{1}{l|}{4} & \multicolumn{1}{l|}{5} & 4 \\ \cline{2-11} 
                 & 4 & \multicolumn{1}{l|}{4} & \multicolumn{1}{l|}{3} & \multicolumn{1}{l|}{6} & \multicolumn{1}{l|}{6} & \multicolumn{1}{l|}{6} & \multicolumn{1}{l|}{4} & \multicolumn{1}{l|}{4} & \multicolumn{1}{l|}{5} & 6 \\ \hline
                \multirow{4}{*}{NCHLOILTM \& STM} & 1 & \multicolumn{1}{l|}{5} & \multicolumn{1}{l|}{5} & \multicolumn{1}{l|}{3} & \multicolumn{1}{l|}{5} & \multicolumn{1}{l|}{5} & \multicolumn{1}{l|}{5} & \multicolumn{1}{l|}{6} & \multicolumn{1}{l|}{5} & 5 \\ \cline{2-11} 
                 & 2 & \multicolumn{1}{l|}{5} & \multicolumn{1}{l|}{4} & \multicolumn{1}{l|}{5} & \multicolumn{1}{l|}{7} & \multicolumn{1}{l|}{6} & \multicolumn{1}{l|}{3} & \multicolumn{1}{l|}{6} & \multicolumn{1}{l|}{3} & 5 \\ \cline{2-11} 
                 & 3 & \multicolumn{1}{l|}{5} & \multicolumn{1}{l|}{6} & \multicolumn{1}{l|}{4} & \multicolumn{1}{l|}{4} & \multicolumn{1}{l|}{5} & \multicolumn{1}{l|}{4} & \multicolumn{1}{l|}{6} & \multicolumn{1}{l|}{5} & 5 \\ \cline{2-11} 
                 & 4 & \multicolumn{1}{l|}{4} & \multicolumn{1}{l|}{4} & \multicolumn{1}{l|}{5} & \multicolumn{1}{l|}{7} & \multicolumn{1}{l|}{4} & \multicolumn{1}{l|}{4} & \multicolumn{1}{l|}{4} & \multicolumn{1}{l|}{6} & 6 \\ \hline
                \multirow{4}{*}{NHLOILTM \& STM} & 1 & \multicolumn{1}{l|}{4} & \multicolumn{1}{l|}{5} & \multicolumn{1}{l|}{4} & \multicolumn{1}{l|}{5} & \multicolumn{1}{l|}{4} & \multicolumn{1}{l|}{5} & \multicolumn{1}{l|}{5} & \multicolumn{1}{l|}{6} & 6 \\ \cline{2-11} 
                 & 2 & \multicolumn{1}{l|}{4} & \multicolumn{1}{l|}{3} & \multicolumn{1}{l|}{5} & \multicolumn{1}{l|}{5} & \multicolumn{1}{l|}{4} & \multicolumn{1}{l|}{4} & \multicolumn{1}{l|}{7} & \multicolumn{1}{l|}{6} & 6 \\ \cline{2-11} 
                 & 3 & \multicolumn{1}{l|}{4} & \multicolumn{1}{l|}{4} & \multicolumn{1}{l|}{5} & \multicolumn{1}{l|}{4} & \multicolumn{1}{l|}{5} & \multicolumn{1}{l|}{6} & \multicolumn{1}{l|}{5} & \multicolumn{1}{l|}{4} & 7 \\ \cline{2-11} 
                 & 4 & \multicolumn{1}{l|}{4} & \multicolumn{1}{l|}{4} & \multicolumn{1}{l|}{5} & \multicolumn{1}{l|}{4} & \multicolumn{1}{l|}{5} & \multicolumn{1}{l|}{4} & \multicolumn{1}{l|}{5} & \multicolumn{1}{l|}{6} & 7 \\ \hline
                \multirow{4}{*}{AHLOILTM \& STM} & 1 & \multicolumn{1}{l|}{8} & \multicolumn{1}{l|}{4} & \multicolumn{1}{l|}{4} & \multicolumn{1}{l|}{6} & \multicolumn{1}{l|}{5} & \multicolumn{1}{l|}{4} & \multicolumn{1}{l|}{5} & \multicolumn{1}{l|}{5} & 3 \\ \cline{2-11} 
                 & 2 & \multicolumn{1}{l|}{7} & \multicolumn{1}{l|}{6} & \multicolumn{1}{l|}{6} & \multicolumn{1}{l|}{4} & \multicolumn{1}{l|}{5} & \multicolumn{1}{l|}{4} & \multicolumn{1}{l|}{4} & \multicolumn{1}{l|}{4} & 4 \\ \cline{2-11} 
                 & 3 & \multicolumn{1}{l|}{6} & \multicolumn{1}{l|}{7} & \multicolumn{1}{l|}{5} & \multicolumn{1}{l|}{6} & \multicolumn{1}{l|}{3} & \multicolumn{1}{l|}{4} & \multicolumn{1}{l|}{4} & \multicolumn{1}{l|}{5} & 4 \\ \cline{2-11} 
                 & 4 & \multicolumn{1}{l|}{6} & \multicolumn{1}{l|}{7} & \multicolumn{1}{l|}{5} & \multicolumn{1}{l|}{4} & \multicolumn{1}{l|}{4} & \multicolumn{1}{l|}{4} & \multicolumn{1}{l|}{4} & \multicolumn{1}{l|}{6} & 4 \\ \hline
                \end{tabular}%
        }
\end{table}

\begin{table}[]
        \caption{The correlation coefficients of methods by the probability of $rand>1$ astrocytes phagocytose synapses, the cosine filter simulation includes the same frequency of the upstream and downstream brain regions through synaptic activity in activation function}
        \resizebox{\textwidth}{!}{
        \label{table17}
        \begin{tabular}{|c|c|c|c|c|c|c|}
        \hline
        Method                  & \begin{tabular}[c]{@{}c@{}}NCHLOI\\    \\ LTM\end{tabular} & \begin{tabular}[c]{@{}c@{}}NHLOI\\    \\ LTM\end{tabular} & \begin{tabular}[c]{@{}c@{}}AHLOI\\    \\ LTM\end{tabular} & \begin{tabular}[c]{@{}c@{}}NCHLOI\\    \\ LTM\&STM\end{tabular} & \begin{tabular}[c]{@{}c@{}}NHLOI\\    \\ LTM\& STM\end{tabular} & \begin{tabular}[c]{@{}c@{}}AHLOI\\    \\ LTM\& STM\end{tabular} \\ \hline
        Correlation coefficient & 0.9527                                                     & 0.9544                                                    & 0.9207                                                    & 0.9369                                                          & 0.9475                                                          & 0.9180                                                          \\ \hline
        \end{tabular}
        }
\end{table}

\begin{figure}%
        \centering
        \begin{minipage}{0.8\textwidth}
                \centering
                \includegraphics[width=0.8\textwidth]{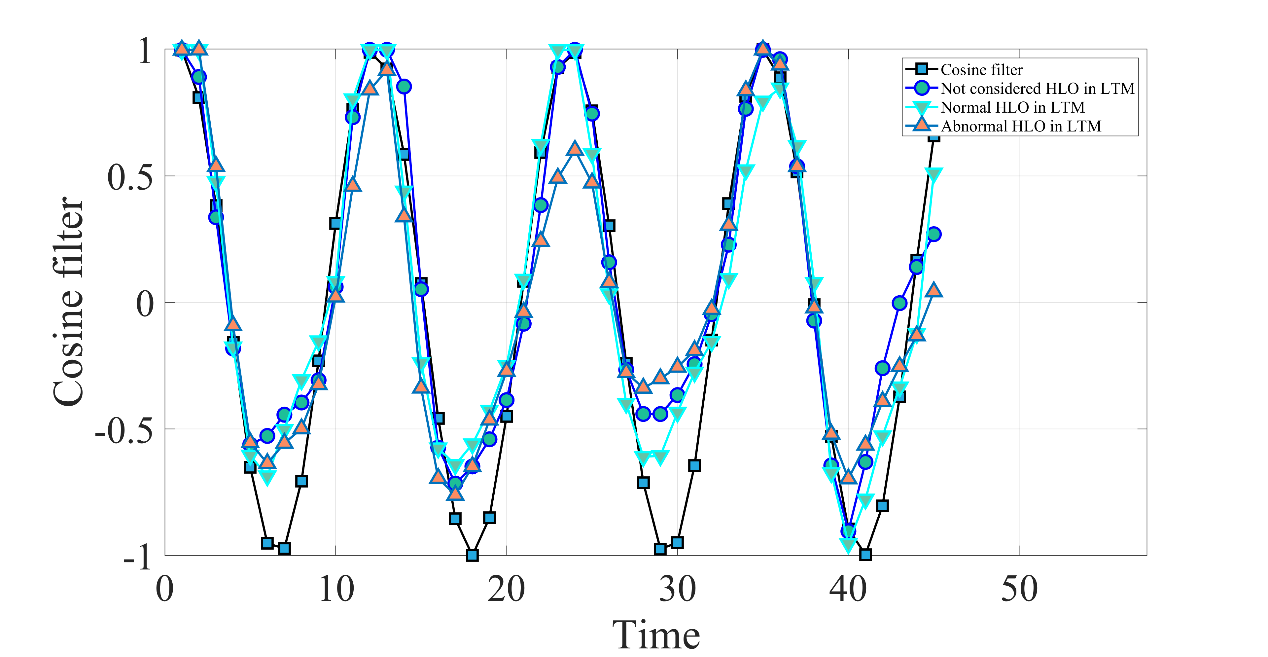}
        \end{minipage}
        \begin{minipage}{0.8\textwidth}
                \centering
                \includegraphics[width=0.8\textwidth]{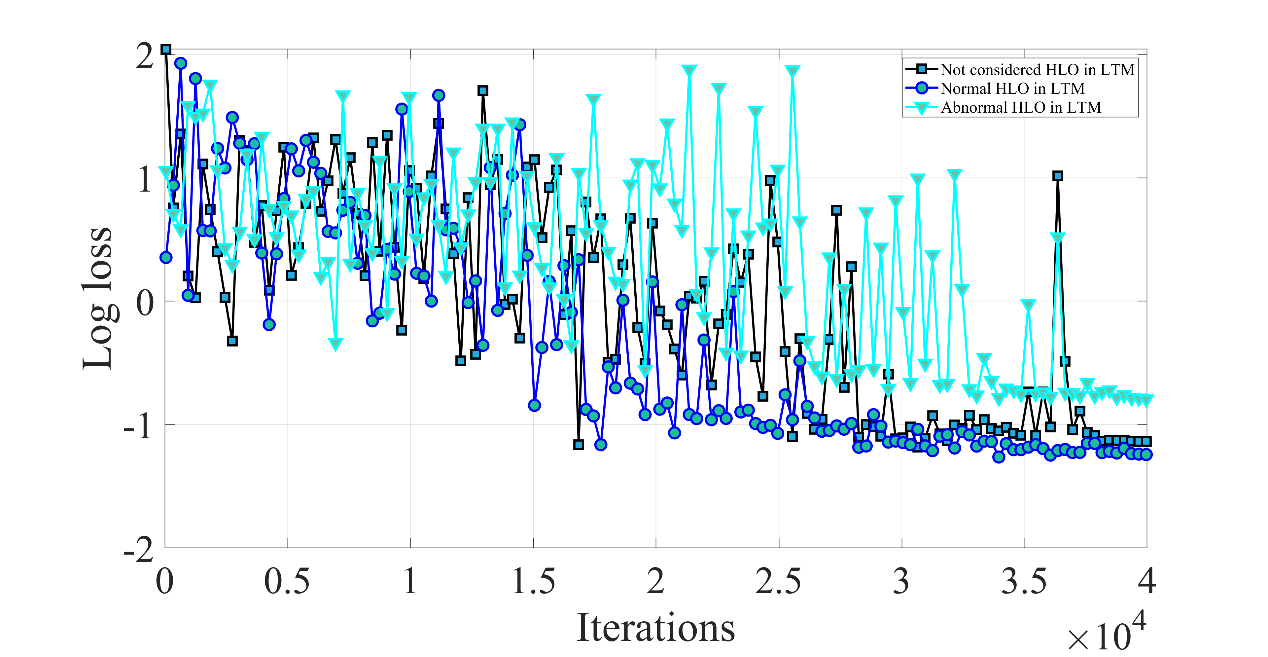}
                \caption{Only long-term memory is considered}
        \end{minipage}
        \begin{minipage}{0.8\textwidth}
                \centering
                \includegraphics[width=0.8\textwidth]{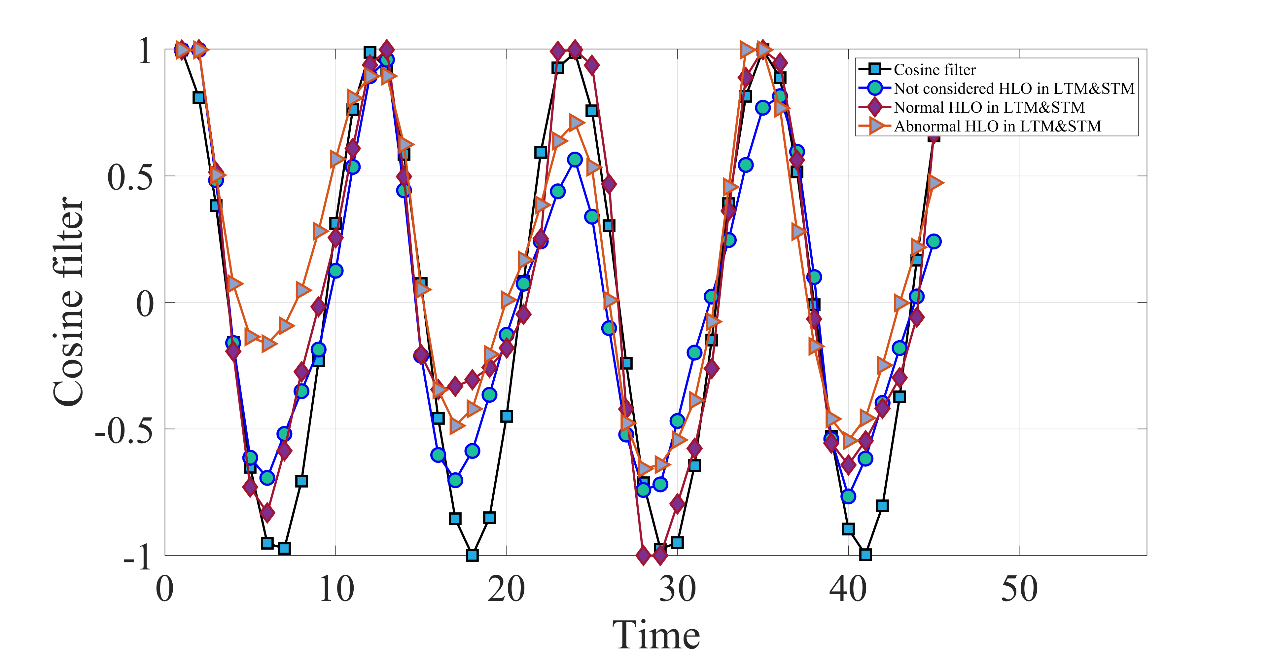}
        \end{minipage}
        \begin{minipage}{0.8\textwidth}
                \centering
                \includegraphics[width=0.8\textwidth]{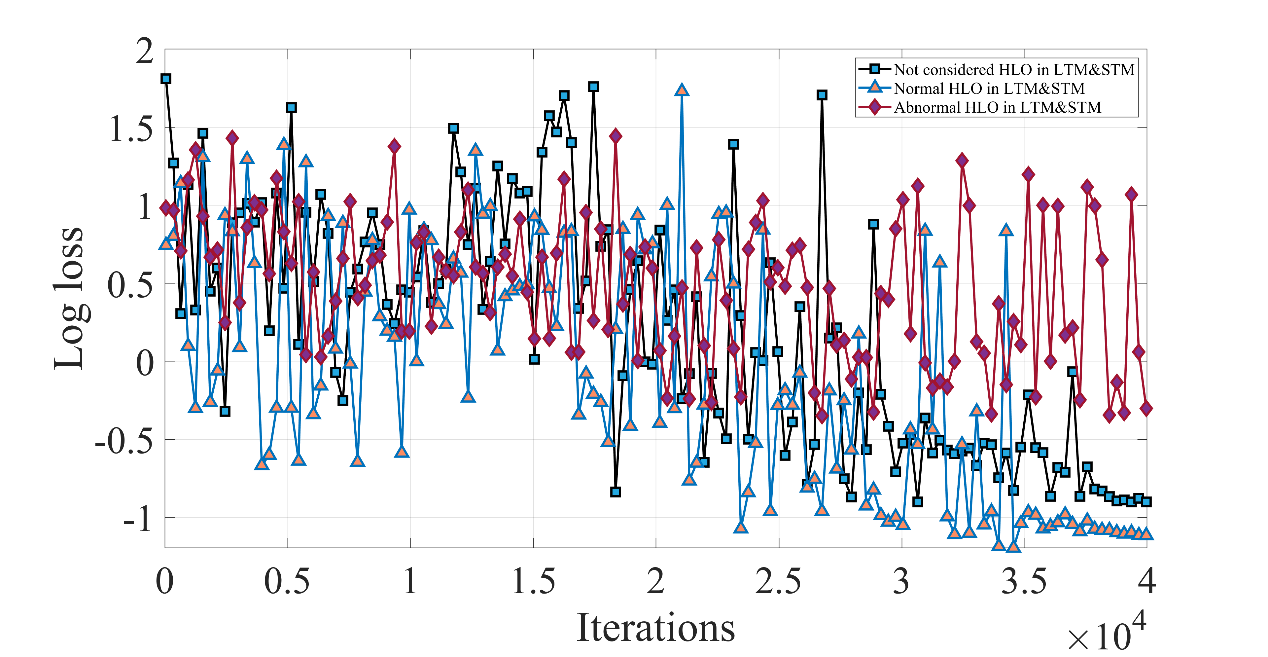}
                \caption{Both short-term and long-term memory are considered}
        \end{minipage}
\caption{The comparisons of higher lower order brain by the probability of $rand>0.9$ astrocytes phagocytose synapses, the cosine filter simulation includes the high frequency of the upstream and low frequency downstream brain regions through synaptic activity in activation function}
\label{fig24}
\end{figure}

\begin{table}[]
        \caption{The synaptic strength of methods by the probability of $rand>0.9$ astrocytes phagocytose synapses, the cosine filter simulation includes the low frequency of the upstream, and high frequency downstream brain regions through synaptic activity in activation function}
        \resizebox{\textwidth}{!}{
        \label{table18}
        \begin{tabular}{|c|c|lllllllll|}
                \hline
                \multirow{2}{*}{Method} & \multirow{2}{*}{Neuron types} & \multicolumn{9}{c|}{Synaptic strength} \\ \cline{3-11} 
                 &  & \multicolumn{1}{c|}{1} & \multicolumn{1}{c|}{2} & \multicolumn{1}{c|}{3} & \multicolumn{1}{c|}{4} & \multicolumn{1}{c|}{5} & \multicolumn{1}{c|}{6} & \multicolumn{1}{c|}{7} & \multicolumn{1}{c|}{8} & \multicolumn{1}{c|}{9} \\ \hline
                \multirow{4}{*}{NCHLOILTM} & 1 & \multicolumn{1}{l|}{5} & \multicolumn{1}{l|}{5} & \multicolumn{1}{l|}{3} & \multicolumn{1}{l|}{5} & \multicolumn{1}{l|}{4} & \multicolumn{1}{l|}{6} & \multicolumn{1}{l|}{6} & \multicolumn{1}{l|}{4} & 6 \\ \cline{2-11} 
                 & 2 & \multicolumn{1}{l|}{4} & \multicolumn{1}{l|}{5} & \multicolumn{1}{l|}{4} & \multicolumn{1}{l|}{6} & \multicolumn{1}{l|}{5} & \multicolumn{1}{l|}{5} & \multicolumn{1}{l|}{6} & \multicolumn{1}{l|}{4} & 5 \\ \cline{2-11} 
                 & 3 & \multicolumn{1}{l|}{6} & \multicolumn{1}{l|}{5} & \multicolumn{1}{l|}{5} & \multicolumn{1}{l|}{6} & \multicolumn{1}{l|}{3} & \multicolumn{1}{l|}{4} & \multicolumn{1}{l|}{6} & \multicolumn{1}{l|}{5} & 4 \\ \cline{2-11} 
                 & 4 & \multicolumn{1}{l|}{4} & \multicolumn{1}{l|}{5} & \multicolumn{1}{l|}{5} & \multicolumn{1}{l|}{5} & \multicolumn{1}{l|}{5} & \multicolumn{1}{l|}{5} & \multicolumn{1}{l|}{5} & \multicolumn{1}{l|}{4} & 6 \\ \hline
                \multirow{4}{*}{NHLOILTM} & 1 & \multicolumn{1}{l|}{5} & \multicolumn{1}{l|}{6} & \multicolumn{1}{l|}{5} & \multicolumn{1}{l|}{5} & \multicolumn{1}{l|}{3} & \multicolumn{1}{l|}{5} & \multicolumn{1}{l|}{4} & \multicolumn{1}{l|}{5} & 6 \\ \cline{2-11} 
                 & 2 & \multicolumn{1}{l|}{4} & \multicolumn{1}{l|}{4} & \multicolumn{1}{l|}{5} & \multicolumn{1}{l|}{4} & \multicolumn{1}{l|}{6} & \multicolumn{1}{l|}{5} & \multicolumn{1}{l|}{5} & \multicolumn{1}{l|}{6} & 5 \\ \cline{2-11} 
                 & 3 & \multicolumn{1}{l|}{5} & \multicolumn{1}{l|}{3} & \multicolumn{1}{l|}{5} & \multicolumn{1}{l|}{5} & \multicolumn{1}{l|}{5} & \multicolumn{1}{l|}{5} & \multicolumn{1}{l|}{6} & \multicolumn{1}{l|}{6} & 4 \\ \cline{2-11} 
                 & 4 & \multicolumn{1}{l|}{4} & \multicolumn{1}{l|}{3} & \multicolumn{1}{l|}{4} & \multicolumn{1}{l|}{6} & \multicolumn{1}{l|}{6} & \multicolumn{1}{l|}{6} & \multicolumn{1}{l|}{3} & \multicolumn{1}{l|}{6} & 6 \\ \hline
                \multirow{4}{*}{AHLOILTM} & 1 & \multicolumn{1}{l|}{5} & \multicolumn{1}{l|}{5} & \multicolumn{1}{l|}{6} & \multicolumn{1}{l|}{4} & \multicolumn{1}{l|}{5} & \multicolumn{1}{l|}{5} & \multicolumn{1}{l|}{5} & \multicolumn{1}{l|}{6} & 3 \\ \cline{2-11} 
                 & 2 & \multicolumn{1}{l|}{6} & \multicolumn{1}{l|}{3} & \multicolumn{1}{l|}{6} & \multicolumn{1}{l|}{5} & \multicolumn{1}{l|}{5} & \multicolumn{1}{l|}{3} & \multicolumn{1}{l|}{6} & \multicolumn{1}{l|}{6} & 4 \\ \cline{2-11} 
                 & 3 & \multicolumn{1}{l|}{6} & \multicolumn{1}{l|}{5} & \multicolumn{1}{l|}{5} & \multicolumn{1}{l|}{3} & \multicolumn{1}{l|}{4} & \multicolumn{1}{l|}{5} & \multicolumn{1}{l|}{5} & \multicolumn{1}{l|}{6} & 5 \\ \cline{2-11} 
                 & 4 & \multicolumn{1}{l|}{4} & \multicolumn{1}{l|}{2} & \multicolumn{1}{l|}{4} & \multicolumn{1}{l|}{7} & \multicolumn{1}{l|}{7} & \multicolumn{1}{l|}{6} & \multicolumn{1}{l|}{7} & \multicolumn{1}{l|}{5} & 2 \\ \hline
                \multirow{4}{*}{NCHLOILTM \& STM} & 1 & \multicolumn{1}{l|}{4} & \multicolumn{1}{l|}{5} & \multicolumn{1}{l|}{4} & \multicolumn{1}{l|}{6} & \multicolumn{1}{l|}{6} & \multicolumn{1}{l|}{5} & \multicolumn{1}{l|}{5} & \multicolumn{1}{l|}{4} & 5 \\ \cline{2-11} 
                 & 2 & \multicolumn{1}{l|}{5} & \multicolumn{1}{l|}{5} & \multicolumn{1}{l|}{5} & \multicolumn{1}{l|}{6} & \multicolumn{1}{l|}{4} & \multicolumn{1}{l|}{6} & \multicolumn{1}{l|}{3} & \multicolumn{1}{l|}{6} & 4 \\ \cline{2-11} 
                 & 3 & \multicolumn{1}{l|}{5} & \multicolumn{1}{l|}{5} & \multicolumn{1}{l|}{4} & \multicolumn{1}{l|}{6} & \multicolumn{1}{l|}{6} & \multicolumn{1}{l|}{6} & \multicolumn{1}{l|}{4} & \multicolumn{1}{l|}{4} & 4 \\ \cline{2-11} 
                 & 4 & \multicolumn{1}{l|}{3} & \multicolumn{1}{l|}{5} & \multicolumn{1}{l|}{4} & \multicolumn{1}{l|}{5} & \multicolumn{1}{l|}{5} & \multicolumn{1}{l|}{7} & \multicolumn{1}{l|}{6} & \multicolumn{1}{l|}{5} & 4 \\ \hline
                \multirow{4}{*}{NHLOILTM \& STM} & 1 & \multicolumn{1}{l|}{5} & \multicolumn{1}{l|}{3} & \multicolumn{1}{l|}{5} & \multicolumn{1}{l|}{4} & \multicolumn{1}{l|}{6} & \multicolumn{1}{l|}{5} & \multicolumn{1}{l|}{6} & \multicolumn{1}{l|}{5} & 5 \\ \cline{2-11} 
                 & 2 & \multicolumn{1}{l|}{4} & \multicolumn{1}{l|}{5} & \multicolumn{1}{l|}{4} & \multicolumn{1}{l|}{5} & \multicolumn{1}{l|}{5} & \multicolumn{1}{l|}{5} & \multicolumn{1}{l|}{5} & \multicolumn{1}{l|}{5} & 6 \\ \cline{2-11} 
                 & 3 & \multicolumn{1}{l|}{6} & \multicolumn{1}{l|}{4} & \multicolumn{1}{l|}{5} & \multicolumn{1}{l|}{5} & \multicolumn{1}{l|}{5} & \multicolumn{1}{l|}{4} & \multicolumn{1}{l|}{4} & \multicolumn{1}{l|}{6} & 5 \\ \cline{2-11} 
                 & 4 & \multicolumn{1}{l|}{3} & \multicolumn{1}{l|}{5} & \multicolumn{1}{l|}{3} & \multicolumn{1}{l|}{5} & \multicolumn{1}{l|}{5} & \multicolumn{1}{l|}{5} & \multicolumn{1}{l|}{6} & \multicolumn{1}{l|}{5} & 7 \\ \hline
                \multirow{4}{*}{AHLOILTM \& STM} & 1 & \multicolumn{1}{l|}{6} & \multicolumn{1}{l|}{5} & \multicolumn{1}{l|}{5} & \multicolumn{1}{l|}{4} & \multicolumn{1}{l|}{6} & \multicolumn{1}{l|}{6} & \multicolumn{1}{l|}{4} & \multicolumn{1}{l|}{5} & 3 \\ \cline{2-11} 
                 & 2 & \multicolumn{1}{l|}{6} & \multicolumn{1}{l|}{5} & \multicolumn{1}{l|}{6} & \multicolumn{1}{l|}{5} & \multicolumn{1}{l|}{5} & \multicolumn{1}{l|}{4} & \multicolumn{1}{l|}{5} & \multicolumn{1}{l|}{5} & 3 \\ \cline{2-11} 
                 & 3 & \multicolumn{1}{l|}{6} & \multicolumn{1}{l|}{5} & \multicolumn{1}{l|}{4} & \multicolumn{1}{l|}{5} & \multicolumn{1}{l|}{5} & \multicolumn{1}{l|}{4} & \multicolumn{1}{l|}{5} & \multicolumn{1}{l|}{5} & 5 \\ \cline{2-11} 
                 & 4 & \multicolumn{1}{l|}{6} & \multicolumn{1}{l|}{6} & \multicolumn{1}{l|}{4} & \multicolumn{1}{l|}{6} & \multicolumn{1}{l|}{5} & \multicolumn{1}{l|}{5} & \multicolumn{1}{l|}{5} & \multicolumn{1}{l|}{4} & 3 \\ \hline
                \end{tabular}%
        }
\end{table}

\begin{table}[]
        \caption{The correlation coefficients of methods by the probability of $rand>0.9$ astrocytes phagocytose synapses, the cosine filter simulation includes the low frequency of the upstream, and high frequency downstream brain regions through synaptic activity in activation function}
        \resizebox{\textwidth}{!}{
        \label{table19}
        \begin{tabular}{|c|c|c|c|c|c|c|}
        \hline
        Method                  & \begin{tabular}[c]{@{}c@{}}NCHLOI\\    \\ LTM\end{tabular} & \begin{tabular}[c]{@{}c@{}}NHLOI\\    \\ LTM\end{tabular} & \begin{tabular}[c]{@{}c@{}}AHLOI\\    \\ LTM\end{tabular} & \begin{tabular}[c]{@{}c@{}}NCHLOI\\    \\ LTM\&STM\end{tabular} & \begin{tabular}[c]{@{}c@{}}NHLOI\\    \\ LTM\& STM\end{tabular} & \begin{tabular}[c]{@{}c@{}}AHLOI\\    \\ LTM\& STM\end{tabular} \\ \hline
        Correlation coefficient & 0.9676                                                     & 0.9772                                                    & 0.9466                                                    & 0.9637                                                          & 0.9854                                                          & 0.9417                                                          \\ \hline
        \end{tabular}
        }
\end{table}

\begin{figure}%
        \centering
        \begin{minipage}{0.8\textwidth}
                \centering
                \includegraphics[width=0.8\textwidth]{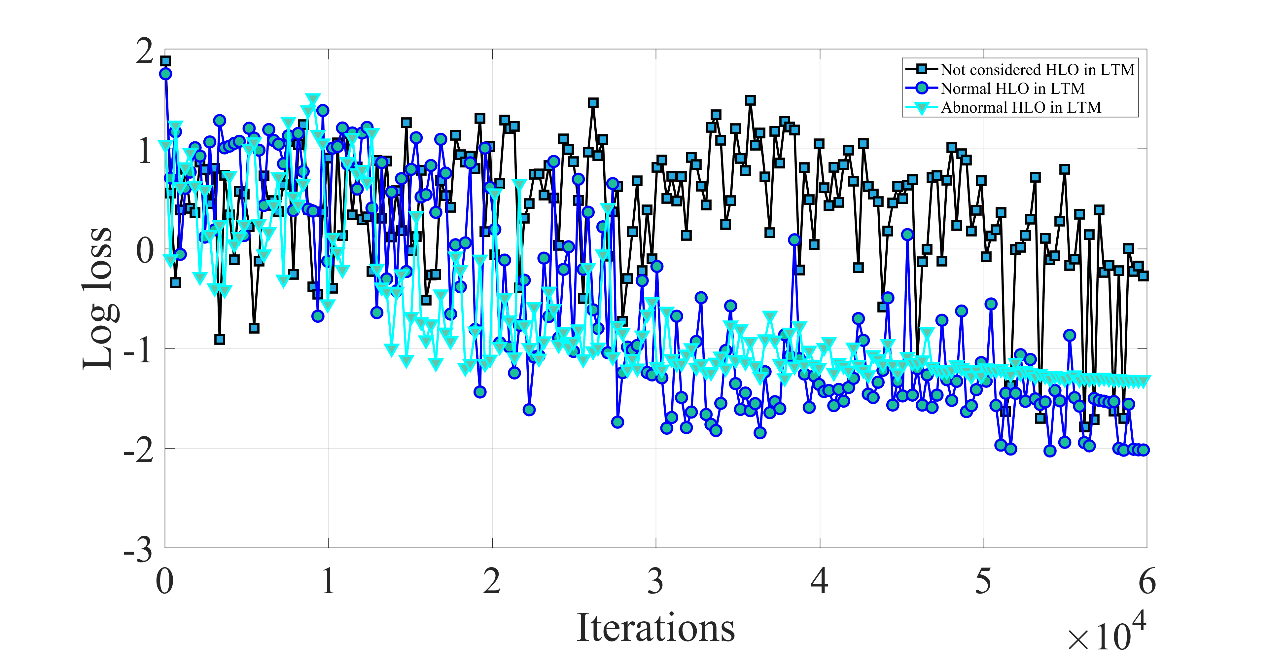}
        \end{minipage}
        \begin{minipage}{0.8\textwidth}
                \centering
                \includegraphics[width=0.8\textwidth]{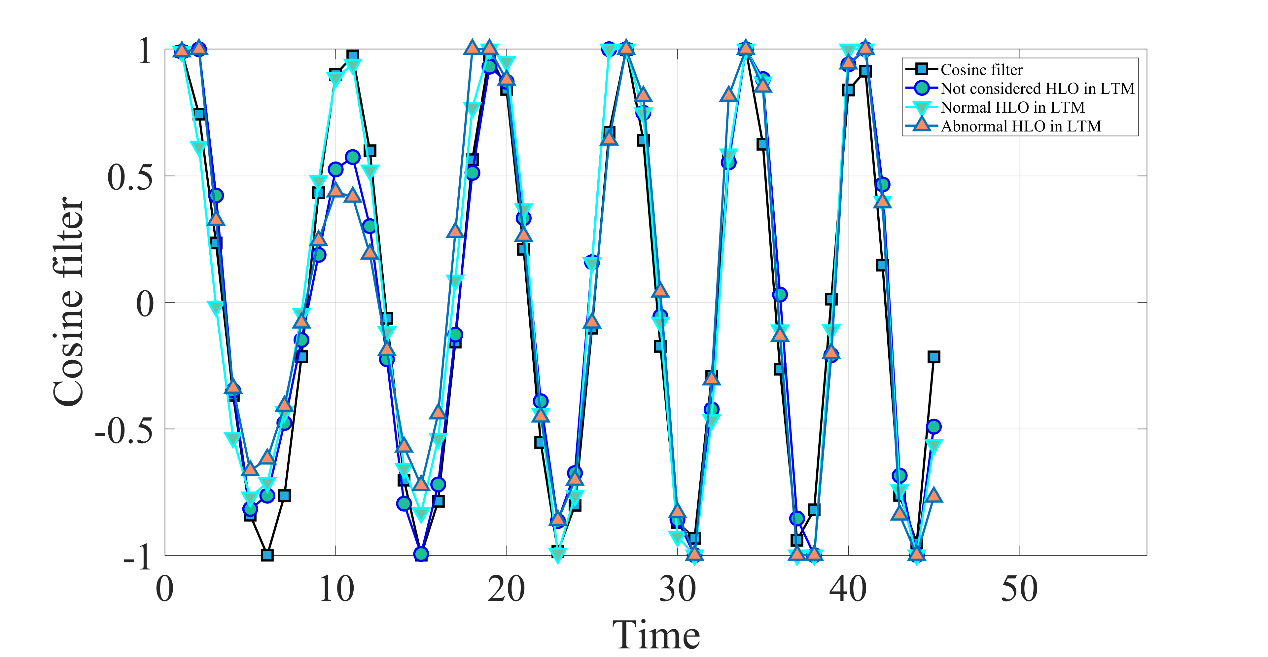}
                \caption{Only long-term memory is considered}
        \end{minipage}
        \begin{minipage}{0.8\textwidth}
                \centering
                \includegraphics[width=0.8\textwidth]{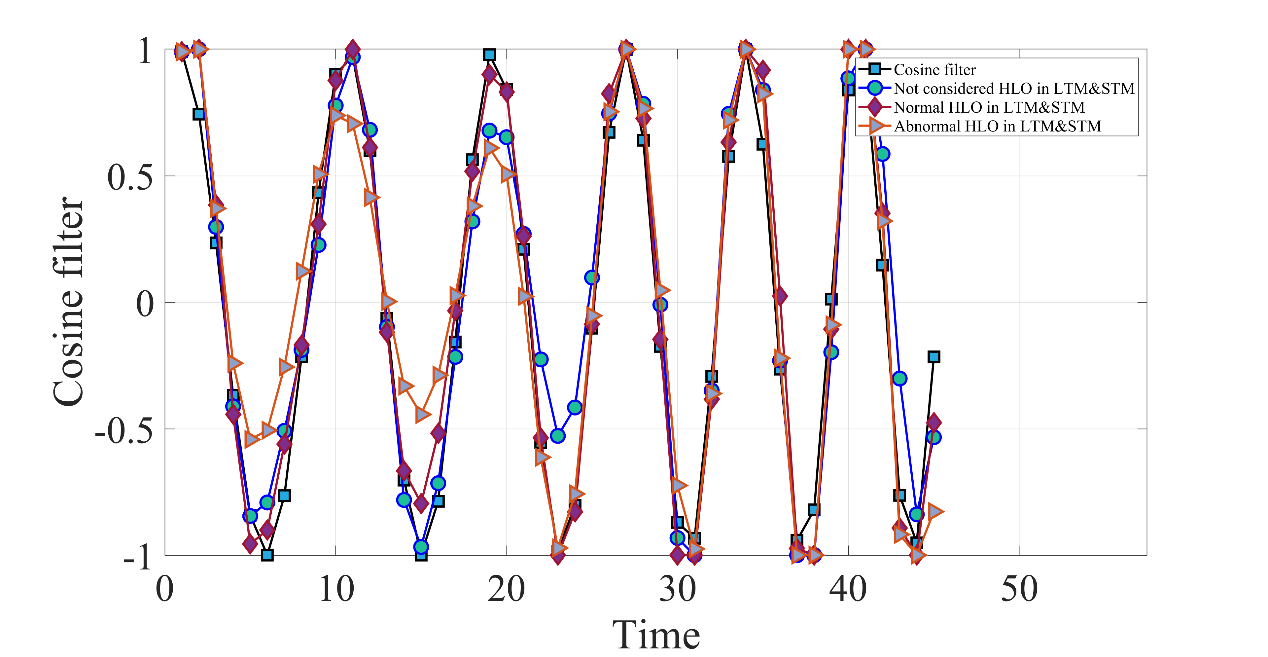}
        \end{minipage}
        \begin{minipage}{0.8\textwidth}
                \centering
                \includegraphics[width=0.8\textwidth]{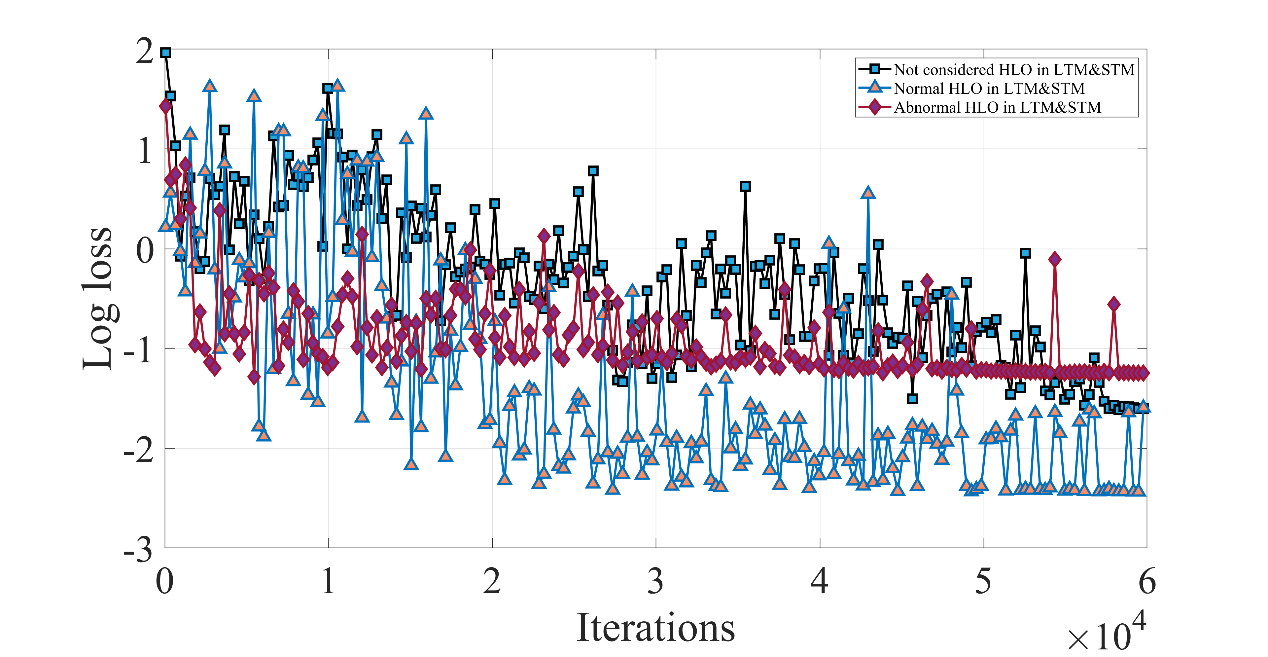}
                \caption{Both short-term and long-term memory are considered}
        \end{minipage}
\caption{The comparisons of higher lower order brain by the probability of $rand>0.9$ astrocytes phagocytose synapses, the cosine filter simulation includes the low frequency of the upstream, and high frequency downstream brain regions through synaptic activity in activation function}
\label{fig25}
\end{figure}

\begin{table}[]
        \caption{The synaptic strength of methods by the probability of $rand>0.9$ astrocytes phagocytose synapses, the cosine filter simulation includes the high frequency of the upstream, and low frequency downstream brain regions through synaptic activity in activation function}
        \resizebox{\textwidth}{!}{
        \label{table20}
        \begin{tabular}{|c|c|lllllllll|}
                \hline
                \multirow{2}{*}{Method} & \multirow{2}{*}{Neuron types} & \multicolumn{9}{c|}{Synaptic strength} \\ \cline{3-11} 
                 &  & \multicolumn{1}{c|}{1} & \multicolumn{1}{c|}{2} & \multicolumn{1}{c|}{3} & \multicolumn{1}{c|}{4} & \multicolumn{1}{c|}{5} & \multicolumn{1}{c|}{6} & \multicolumn{1}{c|}{7} & \multicolumn{1}{c|}{8} & \multicolumn{1}{c|}{9} \\ \hline
                \multirow{4}{*}{NCHLOILTM} & 1 & \multicolumn{1}{l|}{3} & \multicolumn{1}{l|}{6} & \multicolumn{1}{l|}{5} & \multicolumn{1}{l|}{5} & \multicolumn{1}{l|}{5} & \multicolumn{1}{l|}{4} & \multicolumn{1}{l|}{5} & \multicolumn{1}{l|}{6} & 5 \\ \cline{2-11} 
                 & 2 & \multicolumn{1}{l|}{4} & \multicolumn{1}{l|}{6} & \multicolumn{1}{l|}{5} & \multicolumn{1}{l|}{4} & \multicolumn{1}{l|}{4} & \multicolumn{1}{l|}{5} & \multicolumn{1}{l|}{6} & \multicolumn{1}{l|}{6} & 4 \\ \cline{2-11} 
                 & 3 & \multicolumn{1}{l|}{5} & \multicolumn{1}{l|}{5} & \multicolumn{1}{l|}{5} & \multicolumn{1}{l|}{5} & \multicolumn{1}{l|}{5} & \multicolumn{1}{l|}{4} & \multicolumn{1}{l|}{5} & \multicolumn{1}{l|}{4} & 6 \\ \cline{2-11} 
                 & 4 & \multicolumn{1}{l|}{3} & \multicolumn{1}{l|}{5} & \multicolumn{1}{l|}{5} & \multicolumn{1}{l|}{5} & \multicolumn{1}{l|}{5} & \multicolumn{1}{l|}{4} & \multicolumn{1}{l|}{6} & \multicolumn{1}{l|}{6} & 5 \\ \hline
                \multirow{4}{*}{NHLOILTM} & 1 & \multicolumn{1}{l|}{5} & \multicolumn{1}{l|}{5} & \multicolumn{1}{l|}{5} & \multicolumn{1}{l|}{5} & \multicolumn{1}{l|}{5} & \multicolumn{1}{l|}{4} & \multicolumn{1}{l|}{5} & \multicolumn{1}{l|}{5} & 5 \\ \cline{2-11} 
                 & 2 & \multicolumn{1}{l|}{5} & \multicolumn{1}{l|}{5} & \multicolumn{1}{l|}{5} & \multicolumn{1}{l|}{6} & \multicolumn{1}{l|}{4} & \multicolumn{1}{l|}{3} & \multicolumn{1}{l|}{5} & \multicolumn{1}{l|}{5} & 6 \\ \cline{2-11} 
                 & 3 & \multicolumn{1}{l|}{4} & \multicolumn{1}{l|}{4} & \multicolumn{1}{l|}{4} & \multicolumn{1}{l|}{6} & \multicolumn{1}{l|}{6} & \multicolumn{1}{l|}{4} & \multicolumn{1}{l|}{5} & \multicolumn{1}{l|}{6} & 5 \\ \cline{2-11} 
                 & 4 & \multicolumn{1}{l|}{3} & \multicolumn{1}{l|}{5} & \multicolumn{1}{l|}{5} & \multicolumn{1}{l|}{5} & \multicolumn{1}{l|}{5} & \multicolumn{1}{l|}{5} & \multicolumn{1}{l|}{5} & \multicolumn{1}{l|}{5} & 6 \\ \hline
                \multirow{4}{*}{AHLOILTM} & 1 & \multicolumn{1}{l|}{5} & \multicolumn{1}{l|}{5} & \multicolumn{1}{l|}{5} & \multicolumn{1}{l|}{4} & \multicolumn{1}{l|}{6} & \multicolumn{1}{l|}{4} & \multicolumn{1}{l|}{5} & \multicolumn{1}{l|}{6} & 4 \\ \cline{2-11} 
                 & 2 & \multicolumn{1}{l|}{4} & \multicolumn{1}{l|}{4} & \multicolumn{1}{l|}{5} & \multicolumn{1}{l|}{6} & \multicolumn{1}{l|}{6} & \multicolumn{1}{l|}{5} & \multicolumn{1}{l|}{5} & \multicolumn{1}{l|}{5} & 4 \\ \cline{2-11} 
                 & 3 & \multicolumn{1}{l|}{5} & \multicolumn{1}{l|}{6} & \multicolumn{1}{l|}{6} & \multicolumn{1}{l|}{6} & \multicolumn{1}{l|}{5} & \multicolumn{1}{l|}{6} & \multicolumn{1}{l|}{3} & \multicolumn{1}{l|}{5} & 2 \\ \cline{2-11} 
                 & 4 & \multicolumn{1}{l|}{4} & \multicolumn{1}{l|}{4} & \multicolumn{1}{l|}{5} & \multicolumn{1}{l|}{5} & \multicolumn{1}{l|}{5} & \multicolumn{1}{l|}{6} & \multicolumn{1}{l|}{6} & \multicolumn{1}{l|}{5} & 4 \\ \hline
                \multirow{4}{*}{NCHLOILTM \& STM} & 1 & \multicolumn{1}{l|}{3} & \multicolumn{1}{l|}{6} & \multicolumn{1}{l|}{5} & \multicolumn{1}{l|}{4} & \multicolumn{1}{l|}{7} & \multicolumn{1}{l|}{3} & \multicolumn{1}{l|}{6} & \multicolumn{1}{l|}{5} & 5 \\ \cline{2-11} 
                 & 2 & \multicolumn{1}{l|}{3} & \multicolumn{1}{l|}{5} & \multicolumn{1}{l|}{6} & \multicolumn{1}{l|}{5} & \multicolumn{1}{l|}{6} & \multicolumn{1}{l|}{5} & \multicolumn{1}{l|}{3} & \multicolumn{1}{l|}{5} & 6 \\ \cline{2-11} 
                 & 3 & \multicolumn{1}{l|}{4} & \multicolumn{1}{l|}{5} & \multicolumn{1}{l|}{4} & \multicolumn{1}{l|}{5} & \multicolumn{1}{l|}{5} & \multicolumn{1}{l|}{6} & \multicolumn{1}{l|}{5} & \multicolumn{1}{l|}{5} & 5 \\ \cline{2-11} 
                 & 4 & \multicolumn{1}{l|}{5} & \multicolumn{1}{l|}{3} & \multicolumn{1}{l|}{6} & \multicolumn{1}{l|}{5} & \multicolumn{1}{l|}{5} & \multicolumn{1}{l|}{7} & \multicolumn{1}{l|}{4} & \multicolumn{1}{l|}{4} & 5 \\ \hline
                \multirow{4}{*}{NHLOILTM \& STM} & 1 & \multicolumn{1}{l|}{4} & \multicolumn{1}{l|}{5} & \multicolumn{1}{l|}{5} & \multicolumn{1}{l|}{5} & \multicolumn{1}{l|}{4} & \multicolumn{1}{l|}{4} & \multicolumn{1}{l|}{5} & \multicolumn{1}{l|}{5} & 7 \\ \cline{2-11} 
                 & 2 & \multicolumn{1}{l|}{4} & \multicolumn{1}{l|}{4} & \multicolumn{1}{l|}{5} & \multicolumn{1}{l|}{4} & \multicolumn{1}{l|}{4} & \multicolumn{1}{l|}{4} & \multicolumn{1}{l|}{6} & \multicolumn{1}{l|}{6} & 7 \\ \cline{2-11} 
                 & 3 & \multicolumn{1}{l|}{2} & \multicolumn{1}{l|}{5} & \multicolumn{1}{l|}{4} & \multicolumn{1}{l|}{4} & \multicolumn{1}{l|}{7} & \multicolumn{1}{l|}{4} & \multicolumn{1}{l|}{5} & \multicolumn{1}{l|}{5} & 8 \\ \cline{2-11} 
                 & 4 & \multicolumn{1}{l|}{3} & \multicolumn{1}{l|}{5} & \multicolumn{1}{l|}{4} & \multicolumn{1}{l|}{4} & \multicolumn{1}{l|}{4} & \multicolumn{1}{l|}{5} & \multicolumn{1}{l|}{5} & \multicolumn{1}{l|}{6} & 8 \\ \hline
                \multirow{4}{*}{AHLOILTM \& STM} & 1 & \multicolumn{1}{l|}{6} & \multicolumn{1}{l|}{7} & \multicolumn{1}{l|}{4} & \multicolumn{1}{l|}{5} & \multicolumn{1}{l|}{4} & \multicolumn{1}{l|}{5} & \multicolumn{1}{l|}{4} & \multicolumn{1}{l|}{4} & 5 \\ \cline{2-11} 
                 & 2 & \multicolumn{1}{l|}{6} & \multicolumn{1}{l|}{4} & \multicolumn{1}{l|}{4} & \multicolumn{1}{l|}{6} & \multicolumn{1}{l|}{5} & \multicolumn{1}{l|}{5} & \multicolumn{1}{l|}{5} & \multicolumn{1}{l|}{4} & 5 \\ \cline{2-11} 
                 & 3 & \multicolumn{1}{l|}{8} & \multicolumn{1}{l|}{6} & \multicolumn{1}{l|}{5} & \multicolumn{1}{l|}{5} & \multicolumn{1}{l|}{4} & \multicolumn{1}{l|}{4} & \multicolumn{1}{l|}{5} & \multicolumn{1}{l|}{4} & 3 \\ \cline{2-11} 
                 & 4 & \multicolumn{1}{l|}{5} & \multicolumn{1}{l|}{5} & \multicolumn{1}{l|}{6} & \multicolumn{1}{l|}{7} & \multicolumn{1}{l|}{4} & \multicolumn{1}{l|}{5} & \multicolumn{1}{l|}{4} & \multicolumn{1}{l|}{5} & 3 \\ \hline
                \end{tabular}%
        }
\end{table}

\begin{table}[]
        \caption{The correlation coefficients of methods by the probability of $rand>0.9$ astrocytes phagocytose synapses, the cosine filter simulation includes the high frequency of the upstream, and low frequency downstream brain regions through synaptic activity in activation function}
        \resizebox{\textwidth}{!}{
        \label{table21}
        \begin{tabular}{|c|c|c|c|c|c|c|}
        \hline
        Method                  & \begin{tabular}[c]{@{}c@{}}NCHLOI\\    \\ LTM\end{tabular} & \begin{tabular}[c]{@{}c@{}}NHLOI\\    \\ LTM\end{tabular} & \begin{tabular}[c]{@{}c@{}}AHLOI\\    \\ LTM\end{tabular} & \begin{tabular}[c]{@{}c@{}}NCHLOI\\    \\ LTM\&STM\end{tabular} & \begin{tabular}[c]{@{}c@{}}NHLOI\\    \\ LTM\& STM\end{tabular} & \begin{tabular}[c]{@{}c@{}}AHLOI\\    \\ LTM\& STM\end{tabular} \\ \hline
        Correlation coefficient & 0.9449                                                     & 0.9572                                                    & 0.9266                                                    & 0.9503                                                          & 0.9653                                                          & 0.9071                                                          \\ \hline
        \end{tabular}
        }
\end{table}

\begin{figure}%
        \centering
        \begin{minipage}{0.8\textwidth}
                \centering
                \includegraphics[width=0.8\textwidth]{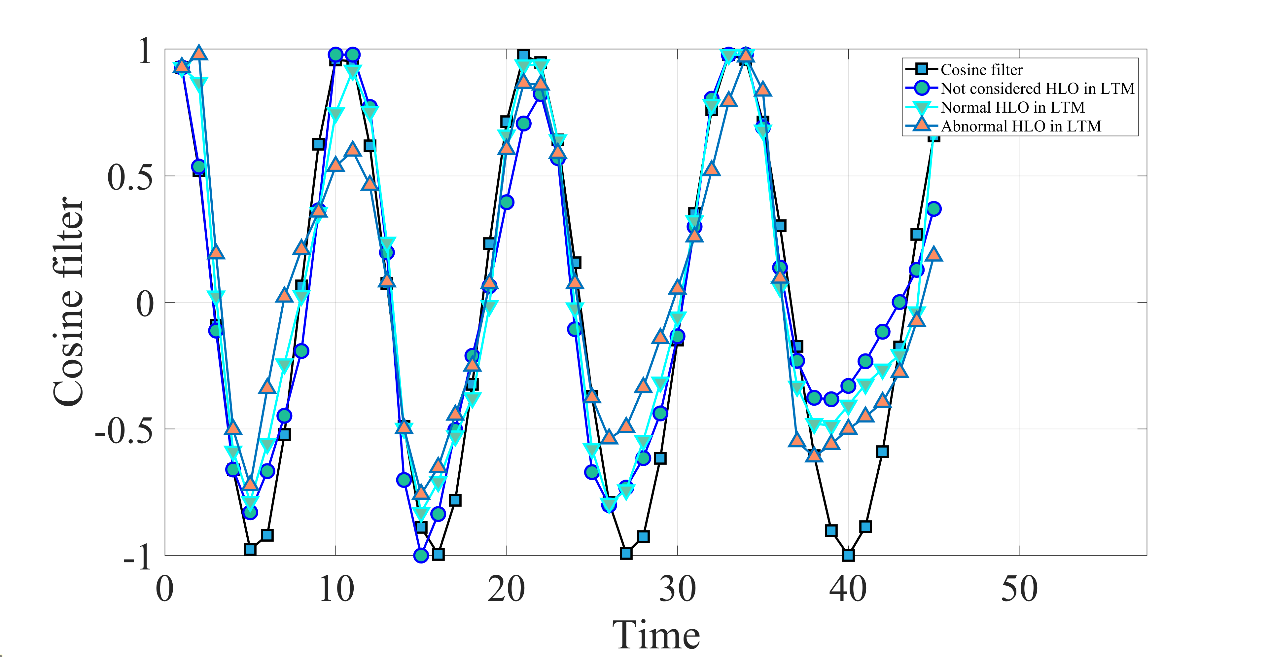}
        \end{minipage}
        \begin{minipage}{0.8\textwidth}
                \centering
                \includegraphics[width=0.8\textwidth]{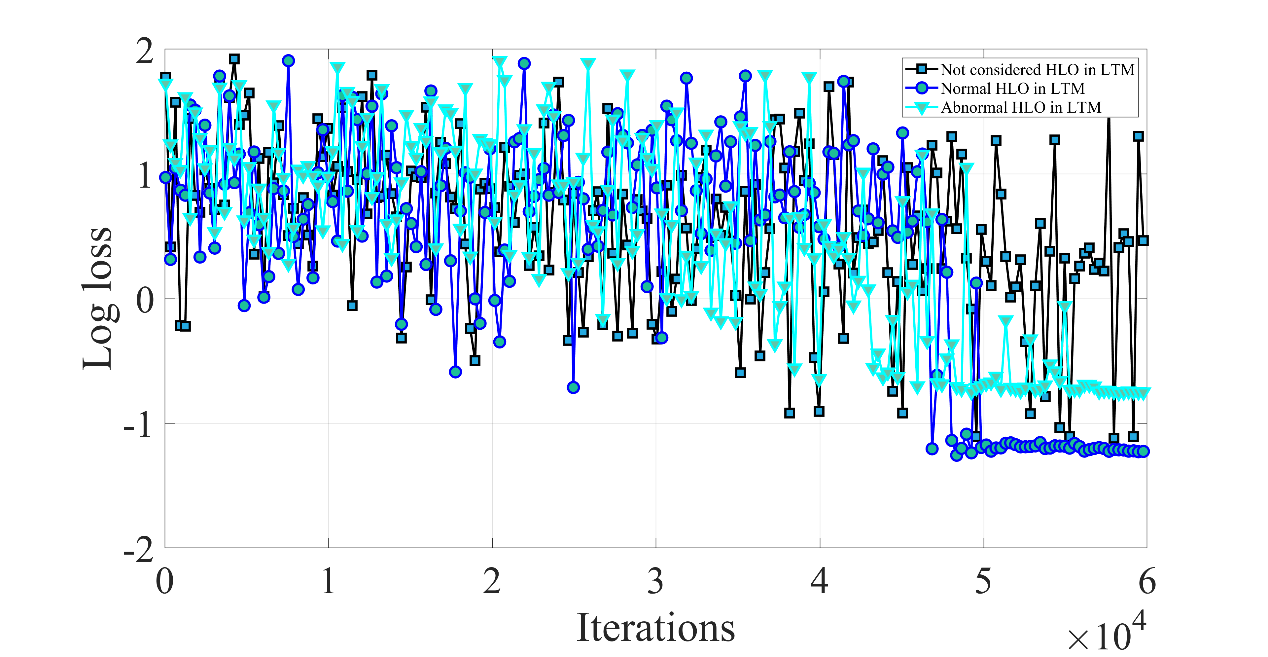}
                \caption{Only long-term memory is considered}
        \end{minipage}
        \begin{minipage}{0.8\textwidth}
                \centering
                \includegraphics[width=0.8\textwidth]{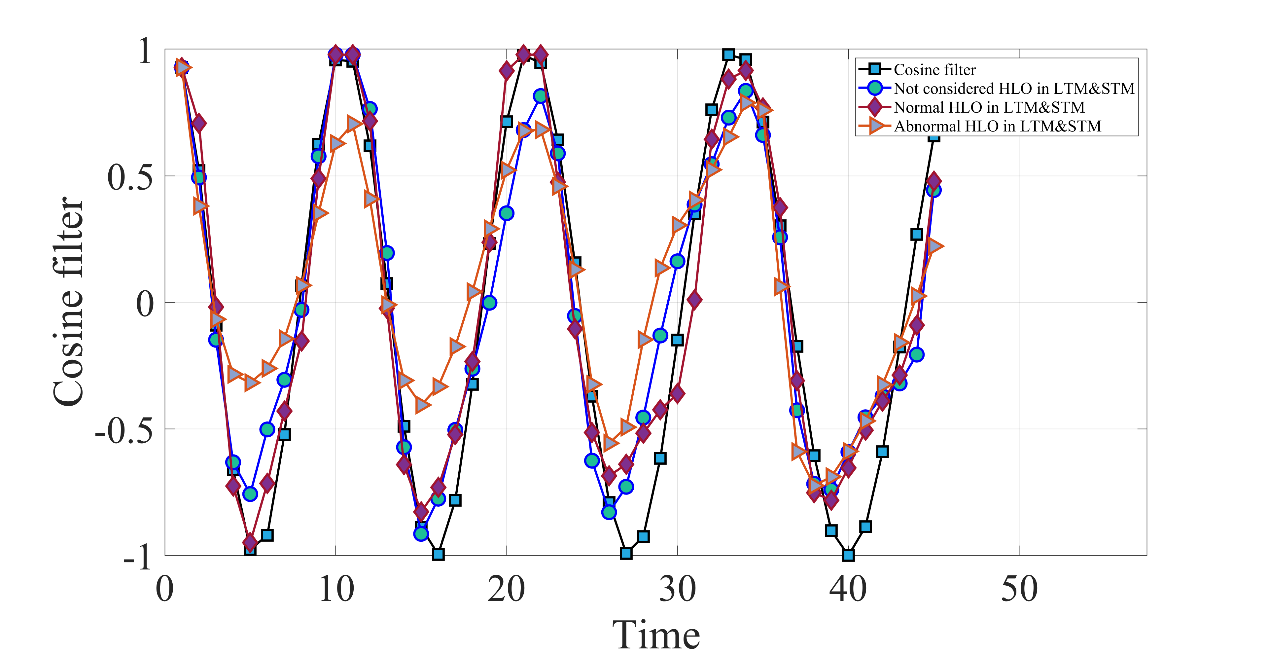}
        \end{minipage}
        \begin{minipage}{0.8\textwidth}
                \centering
                \includegraphics[width=0.8\textwidth]{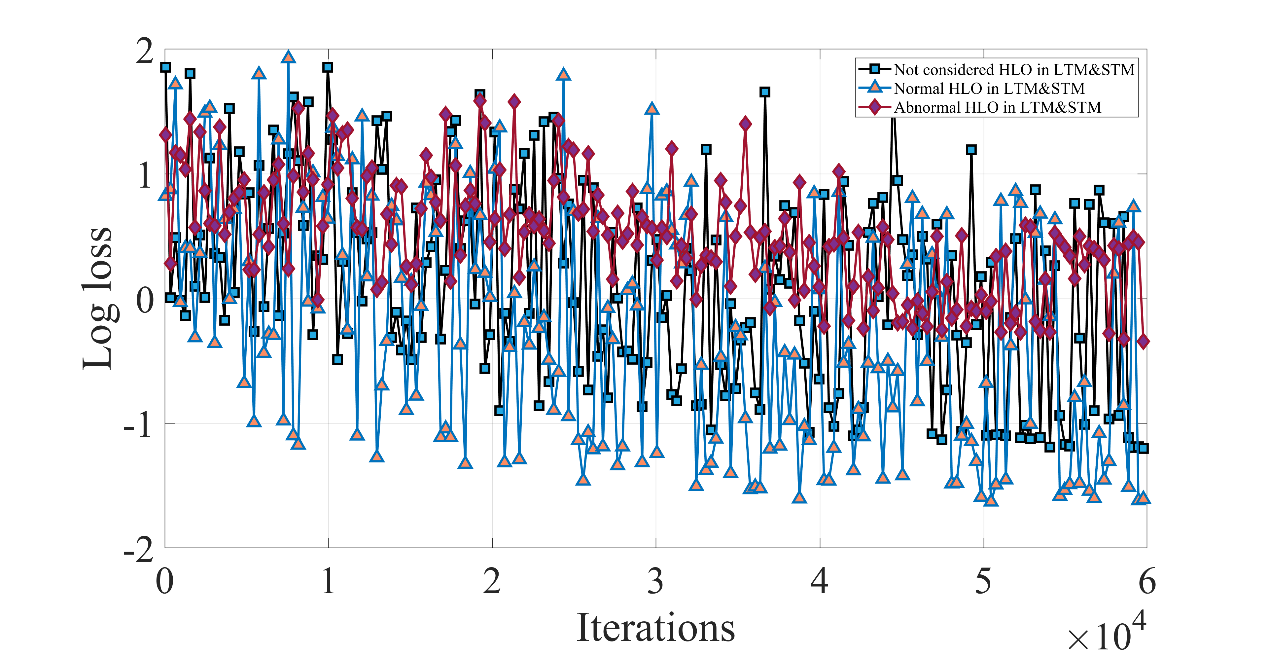}
                \caption{Both short-term and long-term memory are considered}
        \end{minipage}
\caption{The comparisons of higher lower order brain by the probability of $rand>0.9$ astrocytes phagocytose synapses, the cosine filter simulation includes the high frequency of the upstream, and low frequency downstream brain regions through synaptic activity in activation function}
\label{fig26}
\end{figure}

\section{Mechanism, Interpretability, Findings and Hypotheses}
In addition to the shared weights of synaptic connections, we proposed a new Neural Network that includes weights of synaptic ranges for Forward Propagation and Back Propagation. And it gave a unified model of brain plasticity and synapse formation, and explains the mechanism of brain. And using more simulations to compare with the experiments which RNN cannot be achieved \cite{bib14, bib15, bib16, bib17, bib18, bib19, bib20, bib21, bib33,bib22,bib34,bib36,bib37,bib38,bib40}.

The mechanism of PNN enables the appropriate synapse to change the effective range for dendrite morphogenesis at critical period. When one synapse transfers a signal from one neuron to other with a better stimulation signal, PNN will change the current synaptic effective range based on Mean Squared Error (MSE) of loss function, with reference to current and memory brain plasticity, current gradient informational and memory positive and negative gradient informational synapse formation, and vice versa for the same reason.

In terms of the infectious disease dynamics, the infection rates of primitive strains and Delta strains varies in different time ranges, and the time interval corresponding to the cure rate will not be in line with the time interval corresponding to the infection rate, nor will the cure rate correspond exactly to the synaptic effective range effect of the infection rate, due to its high dependence on the medical condition. 

Explaining in terms of synaptic competition, PNN is well-positioned to calculate the time ranges corresponding to the cure and infection rates, i.e., the synaptic effective ranges, through data processing, and then obtain the appropriate cure and infection rates, i.e., the shared weights of synaptic connections. During the emergency and non-emergency phase, i.e., critical period and closure of critical period, astrocytic effect is similar to government policies such as lockdown and mask rules, thereby affecting infection rate, cure rate and time ranges corresponding to them.

The optimization of synaptic effective range is also similar to the Residual Neural Network (ResNet)\cite{bib12}, as more layers don't necessarily perform better in training than fewer layers, and the number of layers is determined by training the synaptic effective range by gradient update with iterations.

By previous discussion's comparisons, we can summarize as follows:

PNN considers synaptic strength balance in dynamic of phagocytosing of synapses and static of constant sum of synapses length, it implements rebalancing strength between synapses\cite{bib14}. From input to output units, neurons go through hidden layers, and when the lead synaptic effective range changes at critical period, these neurons, like a school of fish, also affect the postsynaptic neurons.

Synapse formation will inhibit dendrites generation to a certain extent in experiments, by simulations synapse formation will inhibit the function of dendrites such as receiving signals then affecting cognition \cite{bib15}. The effects of phagocytosing of synapses also help to synaptic growth. Because of rebalancing of synaptic strength, the phagocytosing of synapses makes neighboring synapses easily to grow. High IQ will have more neurons and their actions process fast, but synapse formation will inhabit those. Because of the rebalancing strength from input to output units, synapse formation leads to less neurons, and longer distance between neurons causes processing slower.

The Alcino J Silva 's team suggested that memory ensembles by a retrograde mechanism \cite{bib16}. Our research gave retrograde formula in preprint arXiv:2203.11740 version 1, it used memory gradient of the best previous synaptic effective range weight to update current gradient of synaptic effective range weight, it employs the retrograde mechanism, the memory best previous gradient is positive, and current gradient is negative. Memory gradient of the best previous synaptic effective range weight may be stored in the memory engram cells. And preprint version 2, it makes more sense the best previous memory retrieval process employs a retrograde mechanism through derivation of formula. The preprint version 3 considered the memory relatively inferior gradient of synaptic effective range weight by a retrograde mechanism.

The Alvarez-Buylla's lab proposed that human hippocampal neurogenesis drops sharply in children to undetectable levels in adults \cite{bib17}. While PNN's simulations consider synapse formation, synapse formation leads to the decrease of neurogenesis, through the process from infancy to the senile phase. 

But another study claimed human hippocampal neurogenesis persists throughout aging \cite{bib18}. We guess if it may activate a new and longer circuit (a larger $\rm {l_{max}}$)  in late iteration, even if synapse formation happens then decrease of neurons in early iteration, so a new circuit may lead to neurogenesis persists and may increase neurons in senile phase.

Closing the critical period (Such as astrocytic cortex memory persistence or astrocytes phagocytose synapses at critical period) will cause neurological disorder in experiments, but worse results in PNN simulations \cite{bib19}.

The cortex memory positive and negative synapse formation formula which considers gradient information of retrograde circuit, although it has same effects like astrocytes phagocytose synapses in simulation comparing with experiments. The cortex memory persistence gradient information of retrograde circuit similar to the Enforcing Resilience in a Spring Boot, in addition to considering the memory brain plasticity of architecture. The relatively good and inferior gradient information in synapse formation of retrograde circuit like the folds of the brain.  Only considering positive (best previous gradient information) cortex memory persistence in simulation will inhibit synapse length changes with iterations. The memory of fear learning (or negative memory) can activate Synapses and observed obviously. By simulations, the PNN gave the two cases with and without negative memories. The brain plasticity was much easier to restore without negative memories, and the paper proposed $\alpha$7-nAChR-dependent astrocytic responsiveness is an integral part of the cellular substrate underlying memory persistence by impairing fear memory  \cite{bib20}.

Astrocytes phagocytose synapses will avoid the local accumulation of synapses by simulation (Lack of astrocytes phagocytose synapses causes excitatory synapses and functionally impaired synapses accumulate in experiments and lead to destruction of cognition, but local longer synapses and worse results in PNN simulations) \cite{bib21}.

The brain cortex may be the space-time accumulation of the loss function. And the thicker cortex and more diverse individuals in brain may have high IQ in simulations and experiments \cite{bib33}.

The weights of synaptic ranges' memory negative and positive gradient information (or memory relatively good or inferior gradient information) is not memory brain plasticity, it used stimulus information to update brain plasticity. The stimulus information may be stored in the memory engram cells for inhibiting and inducing synapses and updating synaptic strength \cite{bib22}.

Based on principles of Optics, the relatively good or inferior gradient information's location is near the core of lens, and signals from input to output units with smaller losses go through core of lens. So different cortexes' relatively good and inferior gradient information is stimulated by stronger signals. And the relatively good or inferior gradient information may reflect improved memory retrieval process. And paper found that 1064-nm Transcranial photobiomodulation (tPBM) applied to the right prefrontal cortex improves visual working memory capacity \cite{bib34}. The effects of PNN memory structure and tPBM longer wavelengths may be the same, powerful penetrability of signals for improving memory capacity. It means signals go through easily neighboring areas of focus on convex or concave lens.

The findings may have witnessed entanglement mediated by consciousness-related brain functions, the consciousness-related or electrophysiological signals are unknown in NMR. Those brain functions must then operate non-classically, which would mean that consciousness is non-classical \cite{bib36}. So PNN introduces exacted memory brain plasticity by quantum computing, and affects global brain, at hippocampus. At different cortices, but normal computing in details by Deep Learning, Gradient Descent Method and Statistics in gradient of synaptic connections, gradient of effective ranges, astrocytic cortex memory persistence by memory engram cells, astrocytes phagocytose synapses, different types neurons and synapses.

Priya Rajasethupathy research group shows that anteromedial thalamus selects strong memories and connects with cortex to selectively stabilize memories at remote time \cite{bib37}. Refer to formula \eqref{eq2} and \eqref{eq4}, short-term memory is consolidated and turns long-term memory. And we suggest strong and high flow short-term memory means maximum of directional derivatives of exacted memory brain plasticity is relatively good or inferior gradient of memory brain plasticity, means long-term memory.

Fast, efficient information transfer recently shown in functional magnetic resonance imaging (FMRI) with high spatial resolution, turbulence shows to offer a based way to facilitate energy and information transfer across spatiotemporal scales in brain dynamics \cite{bib38}. Memory from hippocampus to different cortices because of turbulence rather than laminar flow. Long-term memory is gradient of brain plasticity $\frac{dr}{dn}$ which is too idealistic, there is an angle $\alpha_c=\tan^{-1} \frac{\frac{dr}{dn}}{\frac{dr}{dn}}$, $\alpha_c$ will increase because of human aging, and turbulence becomes laminar flow gradually throughout aging, and smaller value tangent vector $\frac{dr}{dt}$ might impair to human intelligence.

Melanin-concentrating hormone decreases synaptic strength and modulates firing rate homeostasis. The paper's findings identify the MCH (Melanin-concentrating hormone) system as vulnerable in early AD. MCH neurons are prominently active during rapid eye movement (REM) sleep. They find that a reduction in the percentage of active MCH neurons in App\textsuperscript{NL-G-F} mice is paralleled by a decrease in the time spent in REM sleep. Homeostatic plasticity response counteracts increased excitatory drive to CA1 pyramidal neurons in App\textsuperscript{NL-G-F} mice. Melanin-concentrating hormone reverses increased excitatory drive in App\textsuperscript{NL-G-F} CA1. Paper's work suggests a model in which impaired MCH-dependent synaptic function in CA1 and perturbed REM sleep synergistically compromise neuronal homeostasis, contributing to aberrant neuronal activity in CA1\cite{bib40}. It can be explained by the possible mechanism of AD, reverse turbulence relates to analyses of MCH, that synapses that should be decreased strength become excitatory, and should be increased strength become inhibitory.

We also hypothesized that heart failure affected the hippocampus. Like a water pump, the ability to pump and release water are weakened, making the radius of the large middle artery and the large middle duct in the brain, turn to smaller, which may lead to Alzheimer's disease.

We see that formula \eqref{eq2}, $-g_r(m,k)$ are synaptic updates in the cortex, and $MW_1 \times g_r(m,k)_{best}+MW_2 \times g_r(m,k)_{worse}+MW_3\times g_r(m,k)_{better}$ are from the upstream brain regions cortical memory engrams.

For memory engrams in the cortex, because reverse turbulence conjecture makes the gradient that updates the synaptic effective range weight take a positive gradient, resulting in synaptic loss, synaptic inhibition excitation reversal, forward and backward memory loss makes the memory engrams smoothing, and neuroinflammation.

The healthy brain retrieves memory engrams from transcerebral regions through a retrograde mechanism, and the ensembles of these memory engrams through smaller upstream ducts to larger downstream ducts without needing to meet the critical conditions of turbulence.

For the memory engrams of other cortices obtained across brain regions, due to the reverse turbulence, the cortex should be retrograde to obtain the memory engrams of the upstream brain regions cortexes without obtaining causes neurons to be lost, or difficulty in obtaining at early stage of AD and causing amnesia, but obtain memory engrams in downstream brain regions cause redundant memory information leading to hallucinations, while more emotional memories in downstream brain regions, which will make people irritable and depressed.

Furthermore, reverse turbulence causes emotional memories to be transmitted retrogradely to the upstream brain regions, leading to a reduction in emotion in the downstream brain regions. Consequently, the downstream brain regions experience synapse lose activity, resulting in the hardening of the hippocampus and downstream brain regions, which accelerates brain aging.

The Heart-Brain Deep Learning model considering intestine, liver and kidney function through the reverse turbulence of brain explains that the memory engrams of the upstream brain regions cannot be retrieved; Local cortex synapse loss; Inhibition and excitation of synapse reversal in the cortex, smoothing memory engrams because of blood disturbances; Extraction of redundant memory engrams in downstream brain regions caused hallucinations, irritability, depression; Neuroinflammation by lacking immune cells, high probability accompanied by atherosclerosis leading to systolic hypertension, high probability accompanied by cerebral infarction, high probability accompanied by heart failure; Gradual atrophy and hardening of the hippocampus, brain aging, circadian rhythm disorders, and cognitive impairment occurs as a result.

The flow turbulent of immune cells through the ducts between brain regions, reverse turbulence of immune cells can bring the opposite stimulating effect to microglia, which affect the direction of memory engrams because of the opposite stimulation, the reverse of memory engrams leads to memory loss and cognitive decline. According to our model, clearance of immune cells weakens microglial activation, it does help with delayed memory loss and cognitive decline, and cognitive search is changed from the correct negative gradient to the false positive gradient, but the cognitive search becomes a random search based on early memory engrams. Compared to clearing immune cells, it is also a good idea to possibly improve the turbulent flow of immune cells crossing between different brain regions.

The Reynolds number distinguishes whether the fluid flow is laminar or turbulent. The Reynolds number formula $Re=\rho v d/\mu $, where $v$, $\rho $ and $\mu $ are the flow velocity, density and viscosity coefficient of the fluid, respectively, and $d$ is a characteristic length. For example, if fluid flows through a circular pipe, $d$ is the equivalent diameter of the pipe. If moderate exercise promotes arterial dilation and increase $d$, increases blood flow and increases $v$, prevents blood viscosity from decreasing $\mu $. Moderate aerobic exercise increases the Raynaud number so that laminar flow between the ducts from downstream brain regions to upstream brain regions becomes turbulent, which can prevent Alzheimer's disease. 

In addition, according to the above analysis of Intestine-Liver-Kidney-Heart-Brain model for Alzheimer's disease, we should consider drugs and foods that Sleep/Wake Regulation, drugs and foods that restore liver function, drugs and foods that restore kidney function, foods and drugs that improve intestinal probiotics, drugs and foods that prevent systolic hypertension caused by atherosclerosis, drugs and foods that prevent cerebral infarction, drugs and foods that prevent heart failure, drugs and foods that prevent depression, drugs and foods to treat aging, drugs and foods that promote brain metabolism and aerobic exercise, which help remove garbage and toxins from cerebral arteries and ducts, which may be effective for early Alzheimer's disease.

We may be able to design Alzheimer's disease drugs based on this.

The 4 findings were as follows:

1. Astrocytic cortex memory persistence also inhibits local synaptic accumulation and excitation, and the model inspires experiments.

2. But the thickest cortex and the most diverse individuals in brain may have low IQ in simulation, so the PNN can guide the experiments, and experiments are special situations of models. And our guess is that more advanced cortexes near the core of brain.

3. It may be the process of astrocytes phagocytose synapses is driven by positive and negative memories of brain plasticity, based on $r(m,k)=r(m,k)+rand\times[MW_1 \times R+MW_2 \times Q_N+MW_3 \times Q_P-r(m,k)]+P$  in formula \eqref{eq4}, because of the positive or negative value of $Q_P$ or $Q_N$.

4. For PNN, the role of long-term memory is more pronounced than short-term memory.

The 4 hypotheses were as follows:

1. Negative or positive emotion and cognition reflect non-classical exacted memories brains plasticity by quantum mechanics, are short-term hippocampal memories. The wave function is high-frequency. And wave function of exacted memories brains plasticity shows kinetic energy because of high-speed particles, produce at hippocampus. By PNN, we propose these appeared barriers from hippocampus to different cortexes, wave function will lead to exponential decay, and wave function of high-frequency turns to low-frequency. At last, long-term stimulus information stored in memory engram cells of different cortexes. And long-term memories exhibit potential energy releases;

2. Barriers may relate to astrocytes;

3. Directional derivative of short-term memory will flow to the different cortices and stored in memory engram cells. High flow or maximum of directional derivatives of short-term memory means gradient of short-term memory will turn to long-term memory.

4. In PNN simulations, ensembles of long-term memory produced once each iteration, so it occurs once at strong probability. but retrieval processes of short-term memory happened once by many iterations, and occurrences are poor probability.

\section{Conclusion}
\begin{itemize}
        \item[1] Based on the RNN architecture, we accomplished the model construction, formula derivation and algorithm testing for PNN. We elucidated the mechanism of PNN based on the latest research on synaptic strength rebalance, and also grounded our study on the basis of findings of the research, which suggested that synapse formation is important for competition in dendrite morphogenesis. The effects of astrocyte impact on brain plasticity and synapse formation is an important mechanism of our Neural Network at critical period or the closure of critical period. In addition to formula \eqref{eq1} and \eqref{eq2}, which is derived from Back Propagation and features the synaptic shared connection weight and the effective range weight respectively. We also managed to figure out synaptic competition by current gradient informational and memory positive and negative gradient informational synapse formation at critical period through formula \eqref{eq2}, synaptic strength rebalance through formula \eqref{eq2} and \eqref{eq4}, and current and memory brain plasticity at critical period through formula \eqref{eq4}. Formula \eqref{eq4} through short-term memory, but Formula \eqref{eq2} through long-term memory.However, our Neural Network is no longer confined to the simple physical concept of synaptic strength rebalance - it treats synapses as a group of intelligent agents to achieve synaptic strength rebalance. The Gradient Descent Method to update the synaptic effective range weights by long-term memory. The quantum computing to update the synaptic effective range weights by short-term memory. Short-term memory travels through the barriers of hippocampus and different cortices and turns into long-term memory.
        \item[2]  The memory positive and negative gradient informational synapse formation needs to consider the forgotten memory-astrocytic synapse formation cortex memory persistence factor. Memory brain plasticity involves plus or minus disturbance- astrocytes phagocytose synapses factor, thus enabling dynamic synaptic strength balance. The effect of astrocyte made local synaptic effective range remain in an appropriate length at critical period. We give decreasing weights of factors at critical period. We try to deduce the mechanism of failure in brain plasticity by model at the closure of critical period in details by contrasting with previous studies. The simulation in Tables similar to that failure to the closure of critical period results in neurodevelopmental disorders\cite{bib19}. The findings of our study proved meaningful, which revealed that much like astrocytes phagocytose synapses factor. The cortex memory persistence factor obtained by the simulation of the mathematical model also inhibits the local accumulation of synapses \cite{bib21}.
        \item[3] Test results of the 3 scenarios all concern the memory best previous solution. All the PNNs with gradient-optimized synaptic effective range fared better than those with random synaptic effective range, which in turn outperformed PNNs with constant synaptic effective range. Constant synaptic effective range is achieved when the minimum synaptic effective range grows to coincide with the maximum synaptic effective range, which usually takes place in the cognition of the senile phase. The iteration of CRPNN is difficult to convergence due to a lack of plasticity and diversity, and RRPNN demonstrates slow convergence in testing.
        \item[4] We believe our work is of profound significance for the development of NN research. Furthermore, PNN considers the memory positive and negative gradient informational synapse formation, and brain plasticity and synapse formation change architecture of NN is a new method of Deep Learning. We can obtain better results in Tables of correlation coefficients respectively, when ORPNN contains both astrocytic synapse formation cortex memory persistence factor and astrocytes phagocytose synapses factor at critical period. Both the memory best previous gradient information, the relatively good gradient information and the relatively inferior gradient information are considered, which increases the activity of the synaptic effective range with iteration changes and thereby improves the simulation results. The relatively good and inferior gradient information in synapse formation of retrograde circuit like the folds of the brain \cite{bib16}. And we discussed the relationship of human intelligence and cortical thickness, individual differences in brain by simulations and experiments, and PNN can guide the experiments, and experiments are special situations of models\cite{bib33}. Our model PNN fits very well with the memory engram cells that strengthened synaptic strength \cite{bib22}. The effects of PNN's memory structure and longer wavelengths of tPBM may be the same, powerful penetrability of signals for improving memory capacity \cite{bib34}.
        \item[5] In addition to the shared weights of synaptic connections, Forward Propagation and Back Propagation also include weights of synaptic ranges. We developed the concept of weights for NN and divided the weights into categories, namely weights of synaptic shared connection and effective range. The latter reflects synaptic competition, strength rebalance, current and memory brain plasticity and current gradient informational and memory negative and positive gradient informational synapse formation at critical period.  From input to output units, they go through hidden layers, and when the lead synaptic effective range changes at critical period, these neurons, like a school of fish, also affect the postsynaptic neurons.
        \item[6] We also attempted to simulate PNN cognition process from infancy to the senile phase, and by analyzing relationship between different synapse formation and correlation coefficients. We may reasonably assume that the minimum synaptic effective range increases and synapse population decreases for synapse formation of senile phase.
        \item[7] In the next step of our study, we took the real pandemic data into consideration, and PNN may well be used to process the pandemic data of the original strain and the variant Delta strain, in an effort to optimize the time interval of the synaptic effective range, and then different connection weights will be applied for cure and infection rates. We will calculate the connection weights by PNN, and the synaptic effective range, based on which we then formulate contingency plans for different period.
        \item[8] By testing, we made a similar observation to that of Dr. Luo's team through PNN: synapse formation to a certain extent is detrimental to dendrites, and synapse formation leads to a reduction in changes of the synaptic effective range, therefore disrupting diversity and plasticity, including results considering cortex memory persistence factor and  astrocytes phagocytose synapses factor separately.
        \item[9] The synaptic competition mechanism NN is suitable for RNN architecture as well as BP, and our next step is to conduct research on the implementation of synaptic competition, strength rebalance, memory generation, memory consolidation, memory loss and critical period by BP.
        \item[10] In simulation, if a same type of shared connection weight changes, all the weights of effective ranges may be consequently updated for better results.
        \item[11] To be honest, the effects of astrocyte affect on brain plasticity and synapse formation was inspired by Particle Swarm Optimization (PSO) \cite{bib31} and consolidation of concrete, and critical period of brain plasticity really depends on inertia weight of PSO \cite{bib32}. Only synapse formation affects brain cognition refers biological tests of neuroscience \cite{bib15}, quantum computer of brain \cite{bib36}  and memory consolidation \cite{bib37}, but also is relevant to the FEM. Synaptic strength rebalance is actually inspired by the conservation of energy. The question we proposed is whether the promotion of computational neuroscience and brain cognition was achieved by model construction, formula derivation or algorithm testing, and whether we can reduce the number of animal experiments and their suffering through the guidance of model simulation planning. We resorted to the Artificial Neural Network (ANN), Evolutionary Computation and other numerical methods for hypotheses, possible explanations and rules, rather than only biological experiments.
        \item[12] In formula \eqref{eq2}, the PNN gradient information considering both positive and negative memories outperforms the case where only positive memory is considered, and the synaptic effective range of the former is more active with iteration changes, yielding better optimization results.
        \item[13] We simplified the PNN to only consider the update of astrocytes phagocytose synapses rather than the gradient update, and decided whether to cancel the update of astrocytes phagocytose synapses based on the correlation coefficient corresponding to the time range between the current and previous generation, each variable and synaptic position corresponding to one time range. The model considering whether or not to cancel the phagocytic update which yields better simulation results than model not processing, and the synaptic effective range is more active with iterations.
        \item[14] Particle Swarm Optimization (PSO) considers the global optimal solution and the best previous solution to update the velocity of particles-2 parameters, and refers PNN can also introduce the relatively good and the relatively inferior solution to update the velocity of particle-4 parameters.
        \item[15] In formula \eqref{eq2} and \eqref{eq4}, we suggest the PNN is not only a classical normal computer in synapse formation and brain plasticity, but also a non-classical quantum computer in working memory brain plasticity, in extracted relatively good or inferior memories brains plasticity will exhibits exponential decay of wave function for a while because of barriers. When you are considering one problem, whether positive or negative, the exacted memory brain plasticity reflects quantum.
        \item[16] The process of astrocytes phagocytose synapses may both have quantum computing and traditional computing, and may be the process is driven by positive and negative memories of brain plasticity, through interactions of excitatory and inhibitory transmitters.
        \item[17] The quantum entanglement between the heart and brain may be beyond absolute space-time. Non-classical working memories by quantum computing, are short-term memories, produce at hippocampus. The wave function is high-frequency, and shows kinetic energy. These appeared barriers from hippocampus to different cortexes, wave function will lead to exponential decay. At last, long-term stimulus information stored in memory engram cells of different cortices, and memories exhibit potential energy. Barriers may relate to astrocytes. Directional derivative of short-term memory will flow to the different cortices and stored in memory engram cells. High flow or maximum of directional derivatives of short-term memory means gradient of short-term memory will turn to long-term memory. From the hippocampus to different cortexes, memory flow may be considered to be the transmission of the rate of change of the architecture. From the hippocampus to the first cortex, memory may mean Jacobian matricx of brain plasticity. From the first cortex to the second cortex, memory may be the second derivative of the brain plasticity. We also hypothesized that heart failure affected the forward and backward transmission of neurotransmitters.
        \item[18] When  $\alpha_c \geq \pi$, model of memory Generation-Consolidation-Loss tries to give the mechanism of Alzheimer's disease. The simultaneous interaction of two different turbulence directions of $\alpha_c \geq \pi$ memory loss and $\alpha_c \leq \frac{\pi}{2}$  memory consolidation leads to $\beta$-amyloid plaques in brain. Another explanation might be false positive gradient of Back Propagation of memory loss, it is also indirectly proved by the results of other's experiments \cite{bib40}. We hypothesize the possible mechanism of Alzheimer's disease, and that brain $\beta$-amyloid plaques and tau protein may only be a symptom or not the main cause.
        \item[19] We try to simulate the prefrontal cortex, amygdala, and hippocampus. Along the forward direction, the fear memory gradually increases. Along the direction of Back Propagation, the optimization order increases.
        \item[20] The first research to synaptic strength rebalance was achieved in Deep Learning. The parameters of relatively inferior negative and relatively good memory were considered in Deep Learning for the first time, and the parameters also enriched the research of particle swarm optimization and genetic algorithm. Studies first proposed that memory flow in each cortex of the brain must be less than the critical value in order for laminar flow to become turbulent and spread to the upstream brain regions, the critical angle of turbulence is related to the mnemonic weight. Based on turbulence, the study gives that the flow of memory from the downstream brain regions to the upstream brain regions may be the transmission of the brain architecture rate of change, and the mnemonic architecture of the downstream first cortex is $r_m$, then the mnemonic architecture of the upstream $n$th cortex may be approximately the $n-1$st derivative of $r_m$, our memory may be a two-dimensional logarithmic spiral, in this way, the $n-1$st derivative of $r_m$ is not much different from $r_m$, synaptic connection weight or range weight is stored in a logarithmic spiral, a unique variable that angle as a logarithmic spiral. The study first analyzed that mnemonic architecture formula-logarithmic spiral and turbulent movement in brain regions result in energy loss or consolidation, and engrams are approximate in moving different brain regions. The study first showed that the upstream brain regions are more rational learning than the downstream brain regions and requires higher-order optimization, and the downstream brain regions have more negative memory, and the Deep Learning model for upstream and downstream brain regions is given. The study considered the Deep Learning model for upstream and downstream brain regions combined with reverse turbulence of downstream brain regions to explain 15 phenomena of Alzheimer's disease. The non-classical experiment with reference to the Heart-Brain interactions were studied, using wave function with exponential decay to update the Deep Learning model of the synaptic effective range firstly. This explains the dynamics cause of shaping in the geometry of the brain, related to the turbulent movement of the logarithmic spiral of the brain.
        \item[21] Both higher-order and lower-order energy decrease from the upstream brain regions to the downstream brain regions. However, the amount of energy at the upstream brain regions, the higher-order positive energy$>$the lower-order positive energy$>$the higher-order negative energy$>$the lower-order negative energy. But the opposite results at downstream brain regions. We compared normal higher-lower-order brain architecture with abnormally reverse higher-lower-order brain architecture through simulations, and the increased probability of astrocytes or microglia cells phagocytose synapses may eliminate the cognitive difference between normal and abnormal brain architecture in some extent. In this way, we suggest that giving the healthy astrocytes or microglia cells to the brains of patients with Alzheimer's disease, which may help them recover. The movement of the logarithmic spiral may control the energy parameters, the mnemonic parameters, and the synaptic activities of the different brain regions.
        \item[22] The higher-lower-order model that considers both long-term memory and short-term memory compared to that only considers long-term memory, and that both long-term memory and short-term memory enhance the spatial distribution of high-frequency and low-frequency pulses in brain regions. Consideration of the activation function of synaptic activity also enhances the spatial distribution of high-frequency and low-frequency pulses in brain regions. The higher-lower-order model are divided into normal brain, Alzheimer's brain, and higher-lower-order case is not considered. Through simulations, we conclude that the normal brain and the case without considering the higher-lower-order are that more high-frequency waves in the downstream brain regions and more low-frequency waves in the upstream brain regions, that is, the synapses in the downstream brain regions are strengthened, while the synapses in the upstream brain regions are weakened, and the synaptic strength calculations of the normal brain are more obvious. In patients with Alzheimer's disease, there are more high-frequency waves in the upstream brain regions and more low-frequency waves in the downstream brain regions.
\end{itemize}

\section*{Author contributions}

Conceptualization, J.B.T., B.Q.S., Y.X., J.Q.L, G.Q.L and C.W.; Formal analysis, J.B.T., B.Q.S., W.D.Z., S.Y.Q., L.K.C., and Y.X., Funding acquisition, C.W., and B.Q.S.; Investigation, J.B.T., C.W., B.Q.S., Y.Y.W., Y.X., J.Q.L and S.Y.Q.; Methodology and Coding, J.B.T., Supervision, B.Q.S., Y.X., G.Q.L, C.W., S.Y.Q., and W.D.Z.; Writing-original draft, J.B.T., J.X.Z., and Y.Y.W.; Writing-review \& editing, W.D.Z. B.Q.S., J.Q.L, G.Q.L, Y.Y.W, and Y.X.; All authors have read and agreed to the published version of the manuscript.

\section*{Competing interests}
The authors declare no competing interest

\bibliography{sn-bibliography}

\end{document}